\documentclass{ustthesis}

\usepackage{mathpazo,amsmath,amssymb,epsfig,enumerate,bbm,calc,color,ifthen,capt-of} 
\usepackage{algorithm}
\usepackage{graphicx}
\usepackage{tabulary}
\usepackage{multirow}
\usepackage{caption}
\usepackage{subcaption} 
\usepackage[noend]{algorithmic}
\usepackage{xcolor}

\usepackage{hyperref}       
\usepackage{url}            
\usepackage{booktabs}       
\usepackage[margin=25mm,textheight=247mm,textwidth=145mm]{geometry}
\usepackage{xspace}
\usepackage{tabularx}
\usepackage{array}
\usepackage{color}
\usepackage{colortbl}
\usepackage{wrapfig}
\usepackage{enumitem}
\usepackage{dsfont}
\usepackage{bm} 

\usepackage[scaled]{helvet} 
\usepackage{courier} 
\usepackage{eulervm} 
\normalfont
\usepackage[T1]{fontenc}

\newcommand{\ttbf}[1]{\texttt{#1}}
\newcommand{\name}[0]{\ttbf{FollowBench}\xspace}
\newcommand{\E}{\mathbb{E}}

\definecolor{colorGPT}{rgb}{0.95, 0.95, 1.0}    
\definecolor{color70B}{rgb}{0.95, 1.0, 0.95} 
\definecolor{color13B}{rgb}{1.0, 0.95, 0.95}  
\definecolor{colorOther}{rgb}{0.95, 0.95, 0.95} 
\definecolor{deepred}{RGB}{180,0,0} 
\definecolor{customred}{RGB}{170, 20, 20} 
\definecolor{customgreen}{RGB}{20, 170, 20} 

\title{Towards Efficient and Effective Alignment of Large Language Models}
\author{Yuxin Jiang}     
\degree{\PhD}              
\subject{Individualized Interdisciplinary Program (Data Science and Analytics)}      
\department{Division of Emerging Interdisciplinary Areas}       
\advisor{Prof. Wei Wang}     
\coadvisor{Prof. Jiaqiang Huang}
\depthead{Prof. Huamin Qu}    
\defencedate{2025}{06}{03}      

\begin{document}

\maketitle
\begin{abstract}

Large language models (LLMs) exhibit remarkable capabilities across diverse tasks, yet aligning them efficiently and effectively with human expectations remains a critical challenge. This thesis advances LLM alignment by introducing novel methodologies in data collection, training, and evaluation.

We first address alignment data collection. Existing approaches rely heavily on manually curated datasets or proprietary models. To overcome these limitations, we propose Lion, an adversarial distillation framework that iteratively refines training data by identifying and generating challenging instructions, enabling state-of-the-art zero-shot reasoning. Additionally, we introduce Web Reconstruction (WebR), a fully automated framework that synthesizes instruction-tuning data directly from raw web documents, significantly improving data diversity and scalability over existing synthetic data methods.

Next, we enhance alignment training through novel optimization techniques. We develop Learning to Edit (LTE), a framework that enables LLMs to efficiently integrate new knowledge while preserving existing information. LTE leverages meta-learning to improve both real-time and batch knowledge updates. Furthermore, we introduce Bridging and Modeling Correlations (BMC), a refinement of Direct Preference Optimization (DPO) that explicitly captures token-level correlations in preference data, leading to superior alignment across QA and mathematical reasoning tasks.

Finally, we tackle the challenge of evaluating alignment. Existing benchmarks emphasize response quality but overlook adherence to specific constraints. To bridge this gap, we introduce \name, a multi-level, fine-grained benchmark assessing LLMs’ ability to follow complex constraints across diverse instruction types. Our results expose key weaknesses in current models’ constraint adherence, offering insights for future improvements.

This thesis makes fundamental contributions to LLM alignment by pioneering novel strategies for data synthesis, training optimization, and evaluation. These advancements enhance efficiency, adaptability, and rigor, paving the way for safer and more controllable AI systems.

\end{abstract}

\makeauthorization
\acknowledgments

First and foremost, I extend my deepest gratitude to my supervisor, Prof. Wei Wang, for his invaluable guidance, generous support in experimental resources and funding, and unwavering understanding and encouragement throughout my four-year Ph.D. journey.
His insatiable thirst for knowledge, pursuit of academic excellence, and rigorous scholarly attitude have profoundly influenced me and will continue to inspire me for a lifetime.
I am also sincerely grateful to my co-supervisor, Prof. Jiaqiang Huang, for his support and insightful discussions.
Additionally, I would like to express my heartfelt appreciation to my former supervisor, Prof. Fangzhen Lin, who first introduced me to the field of Natural Language Processing and set me on the path of scientific research.

I would like to thank all of my thesis defense committee, Prof. Yutao Yue, Prof. Cuiyun Gao, Prof. Xiaowen Chu, Prof. Zhijiang Guo, and Prof. Xuming Hu, for their valuable time and insightful feedback on my thesis.

Throughout my Ph.D. journey, I have experienced not only the joy of discovery but also the fulfillment of collaborating with brilliant and like-minded friends. I am deeply grateful to my collaborators, in order of our acquaintance: Dr. Ziyi Shou, Dr. Linhan Zhang, Mr. Chunkit Chan, Mr. Mingyang Chen, Mr. Bo Huang, Dr. Yufei Wang, Dr. Xingshan Zeng, Dr. Wai-Chung Kwan, and Mr. Qiyuan Zhang. Every publication is a testament to our collective wisdom, dedication, and relentless efforts.
Over the past four years, frequent travels between Hong Kong, Guangzhou, and Shenzhen have been made more meaningful by the companionship of my dear friends—Mr. Hao Wu, Mr. Fengming Zhu, Mr. Biqing Fang, Mr. Zhihao Li, Mr. Mingyu Yang, Mr. Yinan Fan, Mr. Zhongkun Liao, Mr. Rongxin Liu, Mr. Zheng Wei, Mr. Zhengping Chen, Mr. Wentao Pan, and many others. Their support during the pandemic, our engaging discussions, and the moments of laughter we shared have been invaluable.

Lastly, I owe my deepest gratitude to my parents, Mr. Xubo Jiang and Ms. Qifang Zhao, for their unconditional love and unwavering support, and to my grandparents, Mr. Chunzhong Jiang and Ms. Honglian Yu, for their encouragement and belief in me. Above all, I am profoundly thankful to my girlfriend and future wife, Ms. Ying Lin, for her love, companionship, and steadfast encouragement. Because of you all, my journey has been filled with warmth and happiness.

May we continue to chase our dreams without regret, cherish our youth, and embrace a future filled with success and joy.

\endacknowledgments

\tableofcontents
\listoffigures
\listoftables

\chapter{Introduction}

\section{Thesis Introduction}

The recent advances in large language models (LLMs) have demonstrated extraordinary breakthroughs in various real-world applications.
These models, typically based on transformer architectures and comprising tens to hundreds of billions of parameters, are trained on vast datasets sourced from the web using an autoregressive learning paradigm.
Prominent examples include PaLM \cite{2023palm}, LLaMA \cite{2023llama}, and GPT-4 \cite{2023gpt4}.
Compared to earlier, smaller models \cite{cho2014learning, devlin2019bert}, LLMs exhibit two defining characteristics: (1) the \textit{scaling law} \cite{kaplan2020scaling}, which demonstrates systematic performance gains with increased model size, and (2) the \textit{emergence capabilities} \cite{wei2022emergent}—such as in-context learning \cite{dong2022survey}, instruction following \cite{ouyang2022training}, and complex reasoning \cite{wei2022chain}—once a critical scale is surpassed.
These advancements have led to transformative impacts across sectors such as finance \cite{li2023large}, law \cite{lai2024large}, and healthcare \cite{clusmann2023future}, reshaping the way problems are approached and solved.

Nonetheless, despite their capabilities, LLMs also come with significant limitations. Due to their training on large-scale, internet-derived datasets, they may absorb harmful or biased information, resulting in concerns such as misinformation \cite{bommasani2021opportunities}, unfair social representations \cite{ranjan2024comprehensive}, and toxic or exclusionary outputs \cite{weidinger2021ethical}. Furthermore, researchers have identified two troubling risk patterns: (1) \textit{inverse scaling}, where specific issues may worsen as model size increases \cite{mckenzie2023inverse}; and (2) \textit{emergent risks}, where new or intensified risks materialize in larger models \cite{wei2022emergent}, challenging existing mitigation strategies.

To address these growing concerns, a body of research has focused on developing \textbf{alignment} techniques to better steer LLM behavior in accordance with human instructions, goals, and ethical standards \cite{ouyang2022training, DBLP:conf/icml/0001PMMFLBHCRP24, rafailov2024dpo}. The concept of alignment can be traced back to Norbert Wiener’s early warnings: “\textit{We had better be quite sure that the purpose put into the machine is the purpose which we really desire}” \cite{wiener1960some}. In today’s AI landscape, alignment generally refers to the principle that an artificial agent $\mathcal{A}$ should act in ways that reflect the goals and intentions of a human agent $\mathcal{H}$—namely, “\textbf{$\mathcal{A}$ is trying to do what $\mathcal{H}$ wants it to do}” \cite{yudkowsky2016ai}. A formal treatment of this concept in the context of LLMs will be presented in \S \ref{c2-sec: alignment}.


The alignment process for LLMs generally unfolds across three foundational phases: (1) \textbf{Alignment Data Collection}, (2) \textbf{Alignment Training}, and (3) \textbf{Alignment Evaluation}, as illustrated in Figure \ref{c1-fig: roadmap}.
Despite recent advances, each of these stages continues to face significant challenges in terms of both \textbf{efficiency} (i.e., the cost, scalability, and speed of the process) and \textbf{efficacy} (i.e., the quality and impact of the resulting alignment). Specifically,

\begin{itemize}
  \item \textbf{Alignment Data Collection} (\S \ref{c2-sec: data_collection})
  \begin{itemize}
    \item \textit{Efficiency challenges}: Collecting high-quality human feedback or preference data at scale remains expensive and time-consuming \cite{zhang2023instruction}. Filtering and curating alignment-specific data from large corpora also demands substantial computational resources.
    \item \textit{Efficacy challenges}: The collected data often lacks diversity, contains annotation noise, or fails to capture nuanced human preferences, limiting its ability to guide meaningful alignment \cite{wang2024survey}.
  \end{itemize}
  \item \textbf{Alignment Training} (\S \ref{c2-sec: training})
  \begin{itemize}
    \item \textit{Efficiency challenges}: Fine-tuning large models with human feedback requires extensive computational resources, especially when conducted iteratively or with large-scale preference data \cite{kaufmann2023survey}.
    \item \textit{Efficacy challenges}: Alignment methods can suffer from over-optimization (e.g., reward hacking) \cite{pan2022effects}, poor generalization to unseen instructions, and instability during training, all of which compromise the robustness of the aligned model.
  \end{itemize}
  \item \textbf{Alignment Evaluation} (\S \ref{c2-sec: evaluation})
  \begin{itemize}
    \item \textit{Efficiency challenges}: Manual evaluation by human annotators is costly and slow, while automatic metrics often fail to generalize across tasks or correlate with human judgment \cite{gu2024survey}.
    \item \textit{Efficacy challenges}: Existing evaluation protocols are typically narrow in scope, focusing on surface-level correctness while neglecting deeper aspects such as fine-grained constraints, ethical alignment, and long-range coherence.
  \end{itemize}
\end{itemize}

In response to these ongoing challenges, this thesis focuses on advancing both the efficiency and effectiveness of LLM alignment, with particular emphasis on the processes of data collection, model training, and evaluation.
Through a comprehensive examination of each stage, we diagnose prevailing limitations and introduce novel approaches for improvement.
Additionally, we assess the alignment capabilities of LLMs by analyzing their ability to follow nuanced and detailed instructions, a core component of successful alignment.

The first line of work in this thesis focuses on enhancing \textit{data collection} for aligning LLMs.
Previous studies often overlooked the possibility of incorporating any “feedback”–i.e., identifying challenging instructions where the model’s performance falls short–to boost the model’s proficiency iteratively.
To address this, we propose a novel adversarial distillation framework aimed at more efficient alignment data generation. Leveraging the versatile role adaptability of LLMs, we prompt the teacher model to identify “hard” instructions and generate new “hard” instructions for the student model, creating a three-stage adversarial loop of imitation, discrimination, and generation. By applying this adversarial framework, we successfully transfer knowledge from ChatGPT to a student model (named Lion), using a mere 70k training data. Besides, our trained model surpasses conventional state-of-the-art (SOTA) instruction-tuned models like Vicuna-13B on challenging zero-shot reasoning benchmarks.

Furthermore, while existing automatic data synthesis methods alleviate the burden of manual curation, they often rely heavily on either the quality of seed data or strong assumptions about the structure and content of web documents.
To tackle these challenges, we propose \textbf{Web Reconstruction} (WebR), a fully automated framework for synthesizing high-quality instruction-tuning (IT) data directly from raw web documents with minimal assumptions.
Leveraging the inherent diversity of raw web content, we conceptualize \textit{web reconstruction} as an instruction-tuning data synthesis task via a novel dual-perspective paradigm—\textit{Web as Instruction} and \textit{Web as Response}—where each web document is designated as either an instruction or a response to trigger the reconstruction process.
Comprehensive experiments show that datasets generated by WebR outperform state-of-the-art baselines by up to 16.65\% across four instruction-following benchmarks.
Notably, WebR demonstrates superior compatibility, data efficiency, and scalability, enabling enhanced domain adaptation with minimal effort.

Beyond data synthesis, we explore \textit{alignment training} in the context of knowledge editing—a task focused on updating specific factual knowledge in LLMs without degrading their overall performance.
Most existing methods rely on memorization, which hinders models from integrating updated knowledge with existing information when answering questions.
To overcome this limitation, 
we introduce a novel framework named \textbf{Learning to Edit} (LTE), designed to enable LLMs to effectively apply new knowledge into given queries. Drawing inspiration from the principle of “\textit{teaching how to fish},” LTE is structured into two distinct stages. The first is the Alignment Stage, where LLMs are fine-tuned using a carefully constructed parallel corpus, ensuring that the model learns to make accurate, context-relevant modifications while maintaining unrelated information and overall linguistic quality. The second is the Inference Stage, which leverages a retrieval-augmented strategy to support efficient and large-scale application of knowledge edits. Extensive evaluations on four widely-used benchmarks and two LLM backbones, against seven competitive baselines, confirm that LTE achieves state-of-the-art results. It offers strong performance in editing accuracy, robustness in both batch and sequential settings, minimal impact on general capabilities, and fast inference times.

We also advance the training of LLMs by improving Direct preference optimization (DPO), a widely adopted offline preference optimization algorithm.
In conventional DPO, the generation of the winning response and the losing response within pairwise data are typically isolated, leading to weak correlations between them as well as suboptimal alignment performance.
To address this issue, we propose an effective framework for Bridging and Modeling Correlations in pairwise data, named \textbf{BMC}.
Firstly, we increase the consistency and informativeness of the pairwise preference signals through \textit{targeted modifications}, synthesizing a pseudo-winning response by improving the losing response with the winning response as a reference. 
Secondly, we identify that DPO alone is insufficient to model these correlations and capture nuanced variations. Therefore, we propose learning token-level correlations by \textit{dynamically} leveraging the policy model's confidence during training.
Comprehensive experiments on QA, math, and instruction-following tasks demonstrate the effectiveness of our approach, significantly surpassing competitive baselines, including DPO.
Additionally, our in-depth quantitative analysis reveals the reasons behind our method's superior performance over DPO and showcases its versatility to other DPO variants.

Finally, we provide \textit{evaluations} on the alignment of LLMs by analyzing their ability to follow nuanced and detailed instructions, a critical yet underexplored aspect of alignment.
Existing benchmarks primarily focus on evaluating pure response quality, rather than assessing whether the response follows constraints stated in the instruction.
To fill this research gap, we propose \name, a \textbf{Multi-level Fine-grained Constraints Following Benchmark} for LLMs. \name comprehensively includes five different types (i.e., Content, Situation, Style, Format, and Example) of fine-grained constraints. To enable a precise constraint following estimation on diverse difficulties, we introduce a \emph{Multi-level} mechanism that incrementally adds a single constraint to the initial instruction at each increased level. To assess whether LLMs' outputs have satisfied every individual constraint, we propose to prompt strong LLMs with constraint-evolution paths to handle challenging open-ended instructions. By evaluating 13 closed-source and open-source popular LLMs on \name, we highlight the weaknesses of LLMs in instruction following and point towards potential avenues for future work.

In summary, this thesis addresses the critical need for \textit{efficient and effective alignment} of LLMs by tackling three foundational components: data collection, training, and evaluation.
We have demonstrated the pervasive issues of low-quality or assumption-heavy synthetic data, ineffective training methods that ignore correlation structures or generalization needs, and insufficient benchmarks for measuring fine-grained instruction adherence.
Through extensive research and rigorous experimentation, we introduce novel frameworks—Adversarial Distillation, WebR, LTE, BMC, and \name—that significantly improve alignment across all stages. These contributions offer both theoretical insights and practical tools to advance the field of LLM alignment.

\section{Thesis Organization}

\begin{figure}[!t]
\centering
\includegraphics[width=\linewidth]{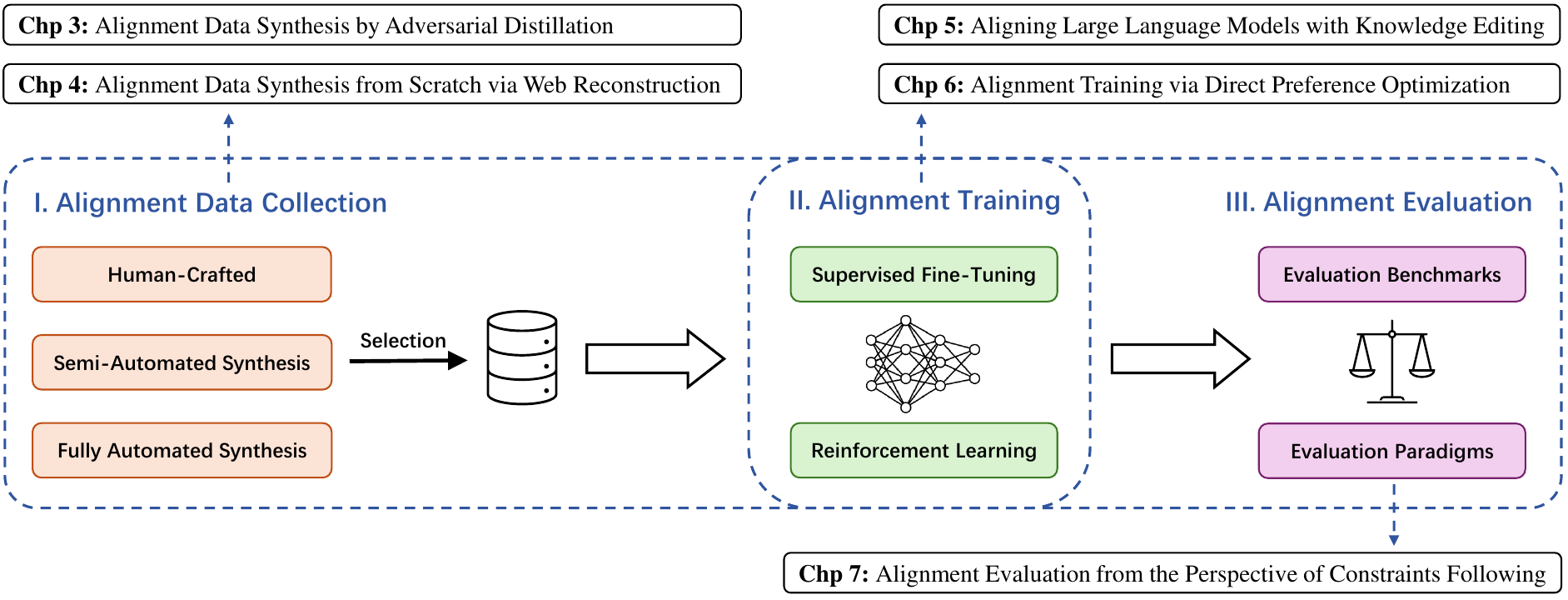}
\caption{
Roadmap of the thesis.
}
\label{c1-fig: roadmap}
\end{figure}

As illustrated in Figure \ref{c1-fig: roadmap}, this thesis is structured as follows.
Chapter \ref{chap:related_work} commences with the definition of LLM alignment, then provides an overview of the relevant literature on alignment data collection, training, and evaluation.
In Chapter \ref{chap:lion}, we introduce Lion, an adversarial distillation framework that iteratively refines alignment training data by identifying and generating challenging instructions, enabling state-of-the-art zero-shot reasoning.
Chapter \ref{chap:webr} presents Web Reconstruction (WebR), a fully automated framework that synthesizes alignment training data directly from raw web documents, significantly enhancing data diversity and scalability compared to existing synthetic methods.
In Chapter \ref{chap:knowledge_editing}, we propose Learning to Edit (LTE), a framework designed for efficient and effective knowledge editing in LLMs.
Chapter \ref{chap:dpo} introduces Bridging and Modeling Correlations (BMC), a refinement of Direct Preference Optimization (DPO) that explicitly captures token-level correlations in preference data, achieving superior alignment in QA and mathematical reasoning tasks.
Chapter \ref{chap:evaluation} presents \name, a Multi-level Fine-grained Constraints Following Benchmark, designed to evaluate instruction following—the key attribution of alignment—in LLMs.
Finally, Chapter \ref{chap:conclusion} concludes the thesis by summarizing our contributions and suggesting directions for future research.

\chapter{Background}\label{chap:related_work}

\section{Definition of LLM Alignment}
In this section, we begin by briefly introducing the concept and evolution of LLMs in \S\ref{c2-sec: llms}, and then proceed to formalize the alignment of LLMs in \S\ref{c2-sec: alignment}. 

\subsection{Large Language Models}
\label{c2-sec: llms}
The development of language models (LMs) has progressed through several key phases, starting from statistical approaches (e.g., Statistical Language Models, or SLMs) \cite{pauls2011faster}, followed by the emergence of neural language models (NLMs) \cite{cho2014learning}, and culminating in the era of pre-trained language models (PLMs) such as BERT and Roberta \cite{devlin2019bert, liu2019roberta}.
Building upon these foundations, \textbf{Large Language Models} (LLMs) have recently emerged as a dominant paradigm in natural language processing (NLP). These models are typically pre-trained on massive datasets using carefully designed objectives, and in some cases, incorporate multimodal data such as image-text pairs to enhance their representational power \cite{dosovitskiy2020image, liu2021swin}.

LLMs differ significantly from their smaller predecessors, not only in scale but also in capability. A key property is the \textit{scaling law}, which reveals that performance on a wide range of tasks tends to improve predictably as the number of parameters and training data increase \cite{kaplan2020scaling}.
Even more intriguingly, LLMs exhibit \textit{emergent behaviors}—novel abilities that materialize only once the model surpasses a certain size threshold \cite{wei2022emergent}.
These include, but are not limited to, in-context learning \cite{dong2022survey}, instruction following \cite{ouyang2022training}, and the ability to perform multi-step reasoning across diverse tasks and domains \cite{wei2022chain}. Such capabilities mark a significant shift in how AI systems are applied to real-world problems, expanding their utility well beyond traditional NLP settings and into broader fields such as education, healthcare, and scientific discovery.

\subsection{Alignment of Large Language Models}
\label{c2-sec: alignment}
The notion of alignment dates back to Norbert Wiener’s cautionary insight, “\textit{We had better be quite sure that the purpose put into the machine is the purpose which we really desire}” \cite{wiener1960some}.
In modern AI discourse, alignment is often described as ensuring that the behavior of an artificial agent $\mathcal{A}$ aligns with the intentions of a human agent $\mathcal{H}$—formally, “\textbf{$\mathcal{A}$ is trying to do what $\mathcal{H}$ wants it to do}”\cite{yudkowsky2016ai}.
Drawing from the principle of \textit{value alignment} in reinforcement learning (RL) \cite{hadfield2016cooperative}, we adopt a utility-based framework to define LLM alignment \cite{wang2024essence}.

\paragraph{Formalization of LLM Alignment.}
Let $\mathcal{H}$ and $\mathcal{A}$ represent two intelligent agents with respective utility functions $U_{\mathcal{H}}(\bm{y})$ and $U_{\mathcal{A}}(\bm{y})$, where $\bm{y} \in \mathcal{Y}$ denotes an action and $U: \mathcal{Y} \rightarrow \mathbb{R}$.
We consider $\mathcal{A}$ aligned with $\mathcal{H}$ over domain $\mathcal{Y}$ if the preference ordering of $\mathcal{H}$ is preserved by $\mathcal{A}$
—that is, for any $\bm{y}_1, \bm{y}_2 \in \mathcal{Y}$, whenever $U_{\mathcal{H}}(\bm{y}_1) > U_{\mathcal{H}}(\bm{y}_2)$, it follows that $U_{\mathcal{A}}(\bm{y}_1) > U_{\mathcal{A}}(\bm{y}_2)$.
Misalignment can be quantitatively assessed using:
\begin{equation}
\mathcal{L} = \mathbb{E}_{\bm{y}_1, \bm{y}_2} \left| \left[ U_{\mathcal{H}}(\bm{y}_1) - U_{\mathcal{H}}(\bm{y}_2) \right] - \left[ U_{\mathcal{A}}(\bm{y}_1) - U_{\mathcal{A}}(\bm{y}_2) \right] \right|,
\label{c2-eq: 1}
\end{equation}
A more stringent criterion assumes $U_{\mathcal{H}}(\bm{y})=U_{\mathcal{A}}(\bm{y})$ for all actions, yielding a simplified discrepancy measure:
\begin{equation}
\mathcal{L} = \mathbb{E}_{\bm{y}} \left| U_{\mathcal{H}}(\bm{y})-U_{\mathcal{A}}(\bm{y}) \right|.
\label{c2-eq: 2}
\end{equation}

Approaches aimed at minimizing the alignment gap (Eq. \ref{c2-eq: 1}) are generally grouped into two main categories: \textbf{Value Learning} and \textbf{Imitation Learning} \cite{leike2018scalable}.

\paragraph{Value Learning.}
This approach centers around constructing a \textit{reward} function that encapsulates human preferences or goals. One can formalize the objective as follows:
\begin{equation}
\phi^* = \arg\min_\phi \mathbb{E}_{\bm{y}, r^* \sim \mathcal{D}(\bm{y}, r^*)} \left[(r^* - R_\phi(\bm{y}))^2 \right],
\label{c2-eq: 3}
\end{equation}
where $\mathcal{D}$ is a dataset of actions $\bm{y}$ and corresponding ground-truth rewards $r^*$, and $R_\phi$ is the reward model parameterized by $\phi$.
In scenarios where only the optimal action $\bm{y}^*$ is observed (instead of its reward), the model can be trained to rank $\bm{y}^*$ higher by minimizing:
\begin{equation}
\mathbb{E}_{\bm{y}^* \sim \mathcal{D}(\bm{y}^*), \bm{y} \sim p(\bm{y})} \left[\max(0, \alpha + R_\phi(\bm{y}) - R_\phi(\bm{y}^*))\right],
\end{equation}
with $\alpha$ as a margin hyperparameter and $p(\bm{y})$ the sampling distribution.

Classic RL algorithms like Deep Q-Networks \cite{huang2020deep}, as well as frameworks such as Inverse Reinforcement Learning \cite{ng2000algorithms} and Preference Modeling \cite{ozturk2005preference}, naturally fall under this formulation. Once the optimal reward function $R_{\phi^*}$ is identified, it can guide the agent’s behavior through standard reinforcement learning paradigms, as will be elaborated in \S \ref{c2-sec: rl}.

\paragraph{Imitation Learning.}
Rather than explicitly modeling a reward signal, imitation learning methods encourage the model to replicate ideal behaviors, implicitly capturing desired values \cite{torabi2018behavioral}.
Let $\pi(\bm{y})$ denote the target policy and $\pi_\theta(\bm{y})$ the agent’s learned policy, parameterized by $\theta$. The goal is to minimize a divergence metric $\mathbb{D}_f$ between the two distributions:
\begin{equation}
\theta^* = \arg\min_\theta \mathbb{D}_f[\pi(\bm{y}) \,\|\, \pi_\theta(\bm{y})],
\label{c2-eq: 5}
\end{equation}
where $\pi(\bm{y})$ is derived empirically from observed demonstrations. When $\mathbb{D}_f$ is the KL divergence, this reduces to the familiar \textit{cross-entropy loss}, training the agent to mimic behavior aligned with human preferences. A well-known application of this principle is Supervised Fine-Tuning (SFT).

In \S~\ref{c2-sec: training}, we will explore how LLM alignment training maps onto the dichotomy between value-based and imitation-based strategies.

\section{Alignment Data Collection}
\label{c2-sec: data_collection}
Before exploring \textit{how to align LLMs}, we begin by examining \textit{what they should be aligned with}.
Effective alignment with human expectations requires high-quality training data that genuinely captures human values and preferences—such as those encapsulated by the \textit{Helpful, Honest, and Harmless} (HHH) principle \cite{askell2021general}.
Each data point $I_k \in \mathcal{D}$, referred to as an \textbf{instruction} $I_k = (x_k, y_k)$, consists of an instructional prompt $x_k$ and its corresponding model response $y_k$.

In this section, we first present methods for synthesizing high-quality alignment data, which broadly fall into three categories:
(1) Human-Crafted Methods,
(2) Semi-Automated Synthetic Methods, and
(3) Fully Automated Synthetic Methods.
We then introduce advanced data selection techniques designed to retain only the most relevant and reliable instruction–response pairs in the final dataset.

\subsection{Human-Crafted Method}
This category involves enlisting human experts to design instructional data, as exemplified by datasets like SUPER-NI~\cite{wang-etal-2022-super}, OpenAssistant~\cite{kopf2023openassistant}, and DOLLY~\cite{DatabricksBlog2023DollyV2}.
For instance, the DOLLY dataset comprises 15,000 instruction-response pairs contributed by Databricks employees through a structured crowd-sourcing process.
Contributors were guided to compose prompts and responses across eight instruction types, including seven from the taxonomy in~\cite{ouyang2022training} and one open-ended category. Notably, they were prohibited from referencing online materials or outputs from generative AI models.
Although this method ensures high-quality instructional data, it is inherently limited in scale due to the labor-intensive and costly nature of manual data generation.

In contrast, alternative strategies like ShareGPT~\cite{vicuna2023} and WildChat~\cite{zhao2024wildchat} gather human-authored instructions by extracting interaction logs from large language model users.
This passive collection approach enables large-scale acquisition of diverse, natural prompts that often lead to informative and relevant responses.
Additionally, public platforms such as Stack Overflow\footnote{\url{https://stackoverflow.com/}}, Quora\footnote{\url{https://www.quora.com/}}, and Zhihu\footnote{\url{https://www.zhihu.com/}}, along with extensive user-generated repositories like Wikipedia\footnote{\url{https://en.wikipedia.org/}}, serve as valuable sources for human-generated instruction data.
Nevertheless, mining such data carries the risk of introducing inappropriate or harmful content~\cite{zhao2024wildchat}.

\subsection{Semi-Automated Synthetic Method}
The semi-automated approach for producing synthetic instruction-tuning datasets leverages LLMs to expand a limited amount of manually curated seed data through in-context learning mechanisms. A prominent example is the Self-Instruct framework~\cite{wang-etal-2023-self-instruct}, which utilizes ChatGPT's ability to generalize from provided examples to produce a broad range of instructions across multiple domains and task formats. This process includes iterative quality filtering to ensure the generated content meets specific standards, continuing until a sufficient volume of data is collected. Subsequent works such as Alpaca~\cite{alpaca} and Evol-Instruct~\cite{xu2024wizardlm} build upon this strategy, aiming to improve the diversity, sophistication, and quality of the synthetic instructions. Despite their scalability, these methods often inherit limitations in data variety from the initial seed instructions~\cite{li2024synthetic}.

\subsection{Fully Automated Synthetic Method}
The fully automated approach leverages LLMs to generate alignment data entirely from scratch. One representative strategy removes human supervision by creating data directly from web-sourced content. For example, WebInstruct~\cite{yue2024mammoth2} derives instruction-response datasets by mining question-answer (QA) pairs from web documents. However, this method is constrained by the necessity for QA pairs to be explicitly embedded in the source material, which is not always the case. Another line of work, such as Instruction Backtranslation~\cite{li2024selfalignment, nguyen2024betteralignmentinstructionbackandforth, chen-etal-2024-dog}, considers raw web text as candidate responses and relies on LLMs to infer the corresponding latent instructions. Still, these documents often include off-topic content or poorly suited phrasing, reducing their effectiveness as high-quality response material. In contrast, Magpie~\cite{xu2024magpie} bypasses the limitations of raw web text by directly prompting aligned LLMs using predefined templates to jointly produce both instructions and their associated responses, taking advantage of the LLMs' auto-regressive generation capabilities.

\subsection{Approach to Data Selection}
To construct a high-quality instruction–response dataset, a filtering mechanism is applied to remove low-quality entries. This is achieved using a scoring function $s(\cdot)$, which assigns a quality score to each instruction–response pair $I_k = (x_k, y_k)$. The final, cleaned dataset $\mathcal{D}'$ is formed by selecting only those pairs that surpass a predefined threshold $\tau$:
\begin{equation}
\mathcal{D}' = \left\{ (x_k, y_k) \in \mathcal{D} \mid s(x_k, y_k) \geq \tau \right\}.
\end{equation}
One common choice for $s(\cdot)$ is the Instruction Following Difficulty (IFD) score \cite{li-etal-2024-quantity}, which reflects how well an instruction contributes to generating the corresponding response. This score is computed as:
\begin{equation}
s_{\theta}(x_k, y_k) = \frac{\sum_{t=1}^{T}\log p(y_k^t \mid x_k, y_k^{<t}; \theta)}{\sum_{t=1}^{T}\log p(y_k^t \mid y_k^{<t}; \theta)}.
\end{equation}
This formulation provides a normalized comparison between the probability of the response being generated with versus without the instruction, offering a quantitative estimate of the instruction's utility. Pairs that fall below the IFD threshold are omitted, yielding a filtered dataset $\mathcal{D}'$.
Alternative strategies for data selection make use of auxiliary models. For example, Instruction Mining \cite{cao2023instruction} employs statistical regression techniques and multiple trained models to assess candidate data points. Similarly, ALPAGASUS \cite{chen2023alpagasus} adopts a pre-trained LLM such as ChatGPT to evaluate the quality of samples.

In addition to quality, many selection frameworks also take into account the \textbf{diversity} and \textbf{importance} of samples. With respect to diversity, some studies \cite{chen2023maybe, bukharin2023data, ankner2024perplexed} aim to ensure coverage across a broad range of tasks and language expressions, either by maximizing individual sample uniqueness (e.g., in terms of lexical or semantic features) or by ensuring the dataset spans a large representation space. Samples from underrepresented task types are often prioritized.

In terms of importance, research efforts such as \cite{yu2024mates, xia2024less} attempt to identify which instruction–response pairs are most essential to include in the training set. Since large models already internalize extensive world knowledge during pretraining, they can often solve standard tasks without further tuning. Consequently, training efforts should focus on more challenging cases, where explicit alignment remains necessary. Selected examples thus serve as critical supplements to enhance the model's ability to follow complex instructions.

\section{Alignment Training}
\label{c2-sec: training}
Once alignment data has been gathered from diverse sources, the next step involves leveraging this data to adapt pre-trained LLMs so that their behavior becomes more consistent with human intentions.

\subsection{Supervised Fine-Tuning}
\label{c2-sec: sft}
A widely adopted approach for alignment is Supervised Fine-Tuning (SFT), which falls under the umbrella of \textit{Imitation Learning}, as outlined in Section~\ref{c2-sec: alignment}.
In this framework, given an input prompt $x$, the model is trained to generate the reference output $y$ by minimizing the cross-entropy loss over the target tokens: 
\begin{equation}
    \mathcal{L}^{\mathrm{SFT}}(\theta) = -\sum_{t=1}^T\log p(y^t \mid x, y^{<t}; \theta).
\end{equation}
Through this process, LLMs are encouraged to produce coherent and contextually appropriate completions that align with the semantics of the input prompt.
Despite its simplicity and effectiveness, SFT has an inherent limitation—it only exposes the model to optimal responses and lacks explicit feedback on less desirable outputs. Nevertheless, SFT-trained models or their objective functions are frequently incorporated into preference-based training pipelines to enhance stability and serve as a form of regularization (see Section~\ref{c2-sec: rl}).

In the conventional SFT paradigm, all parameters of the language model are fine-tuned, which can become prohibitively expensive in terms of computational resources and memory usage, especially as model scales exceed tens of billions of parameters. To mitigate these costs, a class of techniques known as parameter-efficient fine-tuning (PEFT) has emerged. These methods, including LoRA \cite{hu2022lora}, Prefix Tuning \cite{li2021prefix}, and Adapter modules \cite{houlsby2019parameter}, introduce a small number of trainable components—such as additional prompts or lightweight modules—while keeping the majority of the original model weights unchanged. This results in a substantial reduction in memory usage without compromising performance.

\subsection{Reinforcement Learning}
\label{c2-sec: rl}

From a methodological standpoint, the application of RL in aligning LLMs generally unfolds in three main stages:
\begin{itemize}
    \item \textbf{SFT:} The process typically starts by adapting a pre-trained language model using supervised learning on curated, high-quality datasets. This step establishes a foundational level of adherence to expected formats and stylistic conventions.
    \item \textbf{Reward Model Training:} Once the model is fine-tuned, it is used to generate responses that are then annotated with human preferences. Several techniques exist for modeling these preferences, among which the Bradley-Terry (BT) model \cite{bradley1952rank} is frequently employed. Alternatively, the Plackett-Luce model \cite{plackett1975analysis} can be used, particularly when multiple responses are ranked simultaneously. The reward model is trained to approximate these preference annotations, effectively learning a scalar-valued reward function that evaluates the quality of responses.
    \item \textbf{RL optimization:} As discussed earlier in \S \ref{c2-sec: alignment}, after identifying the optimal reward function $R_{\phi^*}$, it is utilized to guide the learning of the language model through feedback-based optimization.
\end{itemize}

The objective during this optimization phase is formulated as follows:
\begin{equation}
\pi_\theta^*(y \mid x) = \max_{\pi_\theta} \mathbb{E}_{x \sim \mathcal{D}, y \sim \pi_\theta(y \mid x)} \Big[ R_{\phi^*}(x, y)  - \beta \mathbb{D}_{\mathrm{KL}}\left[ \pi_\theta(y \mid x) \,\|\, \pi_{\mathrm{ref}}(y \mid x) \right]\Big].
\label{c2-eq: rl_objective}
\end{equation}
This formulation captures two key objectives: (1) encouraging the generation of high-reward responses, and (2) constraining the updated policy $\pi_\theta(y \mid x)$ to remain close to the behavior of the supervised baseline $\pi_{\mathrm{ref}}(y \mid x)$.

\begin{figure}[!t]
\centering
\includegraphics[width=\linewidth]{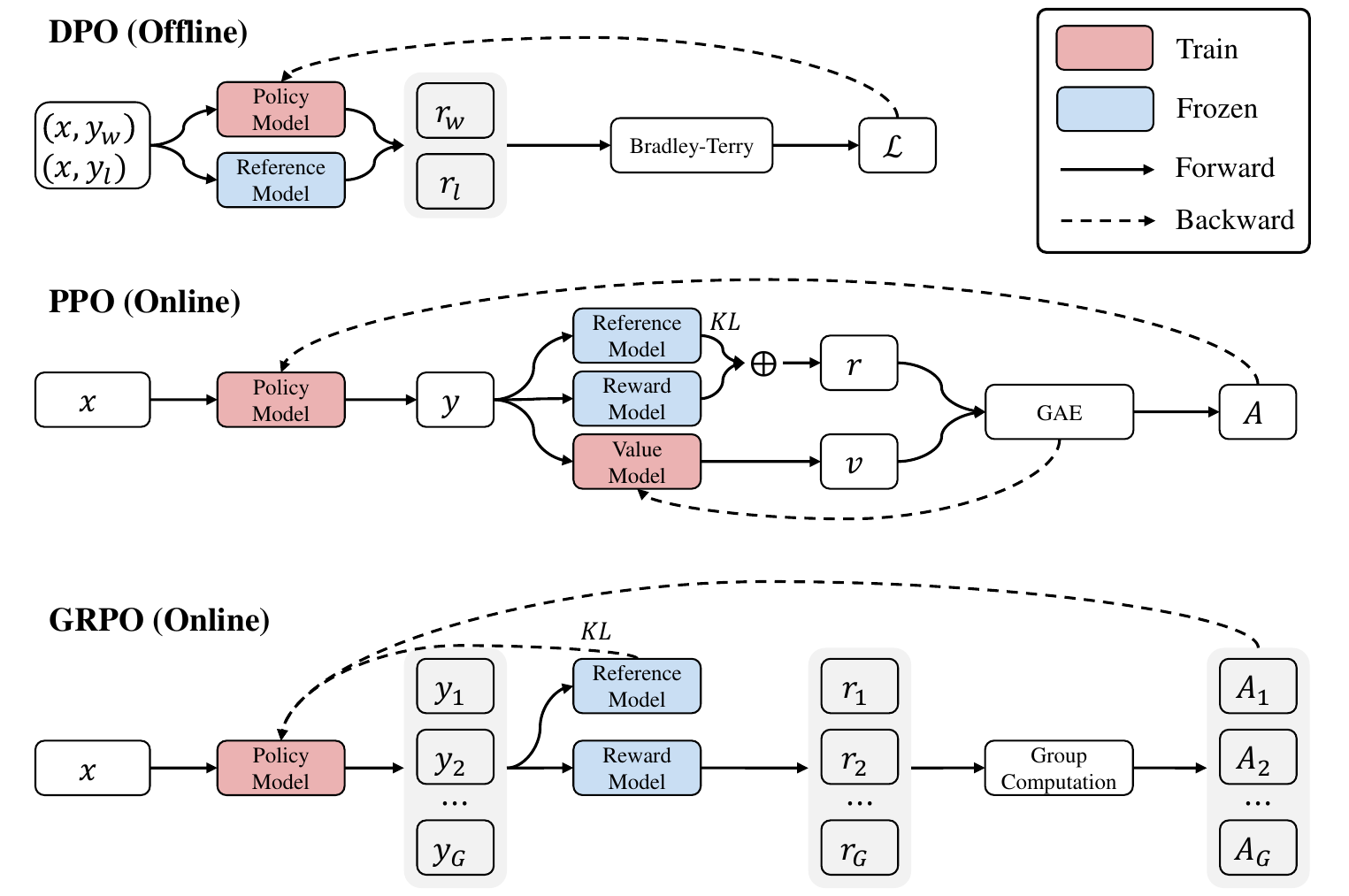}
\caption{
Demonstration of RL optimization algorithms: DPO, PPO, and GRPO.
}
\label{c2-fig: rl}
\end{figure}

In the subsequent section, we introduce several prominent RL methods designed to optimize the objective in Eq. \ref{c2-eq: rl_objective}, as illustrated in Figure \ref{c2-fig: rl}.

\paragraph{Proximal Policy Optimization (PPO)} \cite{schulman2017proximal} is a widely adopted reinforcement learning technique for aligning LLMs with human feedback \cite{PAUL2017RLHF}.
In this framework, a policy $\pi_{\theta}$ with parameters $\theta$ is refined using a reward signal $R_{\phi^*}$. PPO improves the policy by optimizing a surrogate objective that includes a clipping mechanism, ensuring a balance between effective learning and policy stability.
Letting $r(\theta) = \frac{\pi_{\theta}(a \mid s)}{\pi_{\mathrm{old}}(a \mid s)}$ denote the ratio between the updated and previous policy probabilities for action $a$ in state $s$, the PPO objective becomes:
\begin{align}
    \mathcal{L}^{\mathrm{PPO}}(\theta) = -\mathbb{E}_{s \sim P(S), a \sim \pi_{\mathrm{old}}(A | s)} \Big[ &\min \Big( r(\theta) A(s,a), \; \text{clip}\left(r(\theta), 1 - \epsilon, 1 + \epsilon \right) A(s,a) \Big) \nonumber \\
    &- \beta \mathbb{D}_{\mathrm{KL}}\left[ \pi_\theta(y \mid x) \,\|\, \pi_{\mathrm{ref}}(y \mid x) \right] \Big],
\end{align}
where $A(s, a)$ represents an estimate of the advantage function, while $\epsilon$ is a tunable parameter that limits how much the policy is permitted to shift. Advantage estimates are typically derived using Generalized Advantage Estimation (GAE) \cite{schulman2015high}, which leverages both reward feedback and a learned value function. The clipping operation mitigates excessive changes in the policy, thereby preventing destabilizing updates in text generation tasks and supporting more robust training dynamics.

PPO is a foundational component in Reinforcement Learning from Human Feedback (RLHF) \cite{ouyang2022training}, where LLMs are guided by explicit human preference data to better match user expectations.
More recent approaches, such as Reinforcement Learning from AI Feedback (RLAIF) \cite{DBLP:conf/icml/0001PMMFLBHCRP24}, substitute human evaluations with model-generated signals.
Empirical findings \cite{DBLP:conf/icml/0001PMMFLBHCRP24, rafailov2024scaling} suggest that RLAIF can offer a scalable and resource-efficient pathway for fine-tuning LLMs, making it a compelling alternative to traditional human-in-the-loop methods.

\paragraph{Direct Preference Optimization (DPO)} \cite{rafailov2024dpo} is a popular method for offline preference tuning, frequently employed in RLHF.
It simplifies training and improves stability by reformulating the reward function. Based on the reinforcement learning objective given in Eq.\ref{c2-eq: rl_objective}, DPO represents the optimal reward $R_{\phi^*}$ using the closed-form relationship:
\begin{equation}
    R_{\phi^*}(x, y)=\beta \log \frac{\pi_{\theta}^*(y \mid x)}{\pi_{\mathrm{ref}}(y \mid x)} + \beta \log Z(x),
\end{equation}
where $Z(x)$ denotes the partition function. This formulation enables the use of the Bradley-Terry (BT) model \cite{bradley1952rank}, which defines the probability of a preferred output $y_w$ over $y_l$ as $p(y_w \succ y_l)=\sigma(R_{\phi^*}(x, y_w) - R_{\phi^*}(x, y_l))$. DPO further shifts from modeling rewards directly to modeling policy behavior, leading to the objective:
\begin{equation}
\mathcal{L}^{\text{DPO}}(\theta) = 
- \mathbb{E}_{(x, y_w, y_l)\sim \mathcal{D}}
\left[ \log \sigma \left( \beta \log \frac{\pi_{\theta}(y_w \mid x)}{\pi_{\text{ref}}(y_w \mid x)} - \beta \log \frac{\pi_{\theta}(y_l \mid x)}{\pi_{\text{ref}}(y_l \mid x)} \right) \right].
\end{equation}
Here, the dataset $\mathcal{D}$ contains triples $(x, y_w, y_l)$, with $x$ representing the instruction input, and $y_w$, $y_l$ denoting the preferred and non-preferred responses, respectively.

Since the inception of DPO, numerous studies have sought to advance this method by refining its training objective~\cite{wang2024comprehensive}.
For instance, IPO~\cite{moh2024ipo} introduces an alternative pairwise preference loss to mitigate overfitting to the preference dataset, while R-DPO~\cite{park2024rdpo} incorporates a regularization term to prevent the exploitation of latent length bias in the training data.

\paragraph{Group Relative Policy Optimization (GRPO)} \cite{shao2024deepseekmath} streamlines the conventional PPO approach by discarding the separate value (critic) network. Instead of relying on a value function, it calculates a baseline using the average reward from a set of responses generated for the same input. This strategy simplifies the training pipeline, lowers memory consumption, and enhances the stability of policy updates.

Given an instruction $x$, GRPO generates a batch of responses $\{y_1, y_2, \ldots, y_G\}$ using the prior policy $\pi_{\theta}^{\text{old}}$. A reward model assigns a score to each sample, producing corresponding rewards $\{r_1, r_2, \ldots, r_G\}$. These reward values are standardized within the group by subtracting the mean and dividing by the standard deviation:
\begin{equation}
\hat{A}_i = \tilde{r}_i = \frac{r_i - \text{mean}(r)}{\text{std}(r)},
\end{equation}
where $\hat{A}_i$ represents the advantage of the $i$-th response. If we define the importance sampling ratio as $r_i(\theta)=\frac{\pi{\theta}(a_i \mid s)}{\pi{\mathrm{old}}(a_i \mid s)}$, the GRPO objective is formulated as follows:
\begin{align}
\mathcal{L}^{\mathrm{GRPO}}(\theta) = 
-\mathbb{E}_{s \sim P(S),\; a_i \sim \pi_{\mathrm{old}}(A \mid s)} 
\frac{1}{G} \sum_{i=1}^G \Big[ 
& \min \Big( r_i(\theta) \hat{A}_i,\; 
\text{clip}(r_i(\theta), 1 - \epsilon, 1 + \epsilon)\, \hat{A}_i \Big) \nonumber \\
& \quad - \beta\, \mathbb{D}_{\mathrm{KL}}\big[ 
\pi_\theta(y_i \mid x)\, \|\, \pi_{\mathrm{ref}}(y_i \mid x) 
\big] 
\Big].
\end{align}
By exploiting the statistical properties of grouped outputs, GRPO eliminates the need for learning a separate critic model, as done in traditional actor-critic algorithms. This makes it a resource-efficient and scalable solution, particularly as adopted in the DeepSeek R1 model \cite{guo2025deepseek}, while still preserving sensitivity to subtle quality differences between generated responses.

\section{Alignment Evaluation}
\label{c2-sec: evaluation}
Following the collection of alignment datasets and the subsequent training of LLMs, the next critical step is to assess how well these models align with the intended objectives. This section introduces the evaluation benchmarks in \S \ref{c2-sec: eva_bench} and outlines different evaluation paradigms in \S \ref{c2-sec: eva_para}.

\subsection{Evaluation Benchmarks}
\label{c2-sec: eva_bench}
A wide range of benchmark suites has been developed to measure the effectiveness of alignment in LLMs. Broadly, these can be grouped into two categories: closed-ended and open-ended benchmarks. Closed-ended benchmarks primarily test a model’s capabilities in predefined tasks with known answers, while open-ended benchmarks assess performance in more flexible, real-world scenarios where responses are subjective or unconstrained.

\paragraph{Closed-ended Benchmarks.} Closed-ended evaluation benchmarks typically feature test cases where the set of possible answers is predetermined and finite—such as in multiple-choice formats. Below, we review several widely adopted benchmarks in this category:
\begin{itemize}
    \item \textbf{General Knowledge:} The MMLU dataset \cite{dan2021mmlu} is a prominent English-language benchmark for evaluating the factual and academic knowledge of LLMs under zero-shot and few-shot conditions. It includes a wide-ranging set of questions across 57 disciplines, from elementary subjects to specialized professional fields in areas like science, humanities, and social sciences. Its subject diversity and granularity make it a powerful tool for identifying the limits of LLMs’ knowledge. Analogous Chinese benchmarks include C-MMLU \cite{li2023cmmlu}, C-Eval \cite{huang2023c}, M3KE \cite{liu2023m3ke}, and AGIEval \cite{zhong2023agieval}. These benchmarks evaluate Chinese-language LLMs using varied subject matter and difficulty levels, drawing on questions from sources like national college entrance exams, advanced mathematics contests, and legal assessments.
    
    \item \textbf{Reasoning:} Reasoning represents a core aspect of intelligence, essential for handling complex problems. Remarkably, large-scale LLMs often exhibit emergent reasoning capabilities as model size increases. To test these abilities, several benchmarks have been developed. For numerical reasoning, GSM8K \cite{cobbe2021gsm8k} and MATH \cite{hendrycksmath2021} serve as standard benchmarks. Commonsense reasoning tasks are evaluated using datasets like CSQA \cite{talmor2019commonsenseqa} and StrategyQA \cite{geva2021StrategyQA}, which require inference based on everyday knowledge. The BBH benchmark \cite{suzgun2022bbh} evaluates a wide range of logical tasks including temporal understanding, categorization, and causality.

    \item \textbf{Programming:} Several benchmarks focus on testing LLMs’ capabilities in code generation. HumanEval \cite{chen2021codex}, HumanEval+ \cite{liu2023your}, and MBPP \cite{austin2021program} consist of Python programming challenges paired with test cases that assess the functional correctness of generated solutions. The DS1000 dataset \cite{lai2023ds} includes 1,000 tasks across seven popular data science libraries, offering two evaluation modes—code completion and code insertion—based on automated test case validation.
\end{itemize}

\paragraph{Open-ended Benchmarks.} Unlike closed-ended benchmarks, open-ended evaluations are characterized by their flexible, unconstrained response formats. These typically involve conversational or instruction-following tasks without predefined correct answers. Early open-ended datasets like Vicuna-80 \cite{vicuna2023}, Open-Assistant-953 \cite{kopf2023openassistant}, and User-Instructions-252 \cite{wang-etal-2023-self-instruct} use relatively small collections of synthetic prompts to gauge LLM performance. However, these benchmarks tend to be limited in scope, often allowing comparison across only a few models at a time.
To address this, newer benchmarks such as AlpacaEval \cite{alpaca_eval} and Arena-Hard \cite{li2024live} introduce competitive evaluation strategies where model outputs are directly compared to a reference model. A higher Win Rate indicates superior performance, enabling a more scalable and interpretable comparison across numerous LLM candidates.

\subsection{Evaluation Paradigms}
\label{c2-sec: eva_para}
In scenarios where open-ended tasks lack definitive reference answers, external evaluators—either human or language models—are often necessary for assessment. This section outlines the main paradigms used for evaluation, encompassing both human annotators and LLMs.

\paragraph{Human-based Evaluation.} Traditional automatic metrics such as BLEU \cite{papineni2002bleu} and ROUGE \cite{lin2004rouge} depend on the existence of reference outputs and tend to show weak alignment with human judgments, making them unsuitable for open-ended response evaluation. To overcome this limitation, human raters are frequently employed to judge the quality of model-generated outputs in such settings. For instance, \cite{wang-etal-2023-self-instruct, wu2023style} adopt an ordinal annotation approach, instructing annotators to classify responses into one of four quality categories: acceptable, minor issues, major issues, or unacceptable. However, this method is susceptible to subjective biases, often leading to low agreement among annotators \cite{kalpathy2016plus}. As an alternative, \cite{alpaca} suggests a comparative judgment method, where annotators are shown outputs from two different models alongside the same prompt and asked to identify the superior response. Building on this idea, \cite{zheng2023judging} introduces the Elo rating system—commonly used in competitive games like chess—to compute relative rankings among multiple LLMs. In this framework, the scores of each model are dynamically adjusted after every comparison, based on prior ratings and evaluation outcomes.

\paragraph{LLMs-based Evaluation.} Although human evaluation delivers high-quality insights, it is often resource-intensive and slow. Moreover, as LLM-generated content increasingly mirrors human writing, it becomes harder for annotators to reliably differentiate between the two \cite{clark2021all}. To reduce reliance on manual labor and references, recent work has turned to using LLMs themselves as evaluators across various natural language generation (NLG) tasks. These approaches often bypass the need for gold-standard references and leverage LLMs’ own reasoning abilities. Some studies mimic human pairwise comparison setups by prompting LLMs—such as GPT-4—to choose the better output given two candidate responses for a single input \cite{alpaca_eval, zheng2023judging}, thereby automating evaluation in a cost-effective and scalable manner.

\chapter{Alignment Data Synthesis by Adversarial Distillation}\label{chap:lion}

\section{Introduction}
LLMs capable of following natural language instructions have exhibited tremendous success in generalizing zero-shot to new tasks and aligning with human values \cite{mishra-etal-2022-cross, wei2022finetuned}. Due to various concerns, the most advanced LLMs, such as ChatGPT \cite{openai2022chatgpt} and GPT-4~\cite{2023gpt4} that boasting billions of parameters, are typically proprietary, comprising both the model parameter and the training data. 
To foster increased transparency regarding their intricate operational mechanics, a surge in research efforts focusing on knowledge distillation from a proprietary ``teacher'' LLM to an open-source ``student'' LLM.
This is typically accomplished by aligning the responses of the student model with those of the teacher model to a set of instructions, which can be manually or automatically generated \cite{wang-etal-2023-self-instruct, alpaca, zheng2023judging, xu2024wizardlm}.

\begin{figure}[h]
\centering
\includegraphics[width=0.65\linewidth]{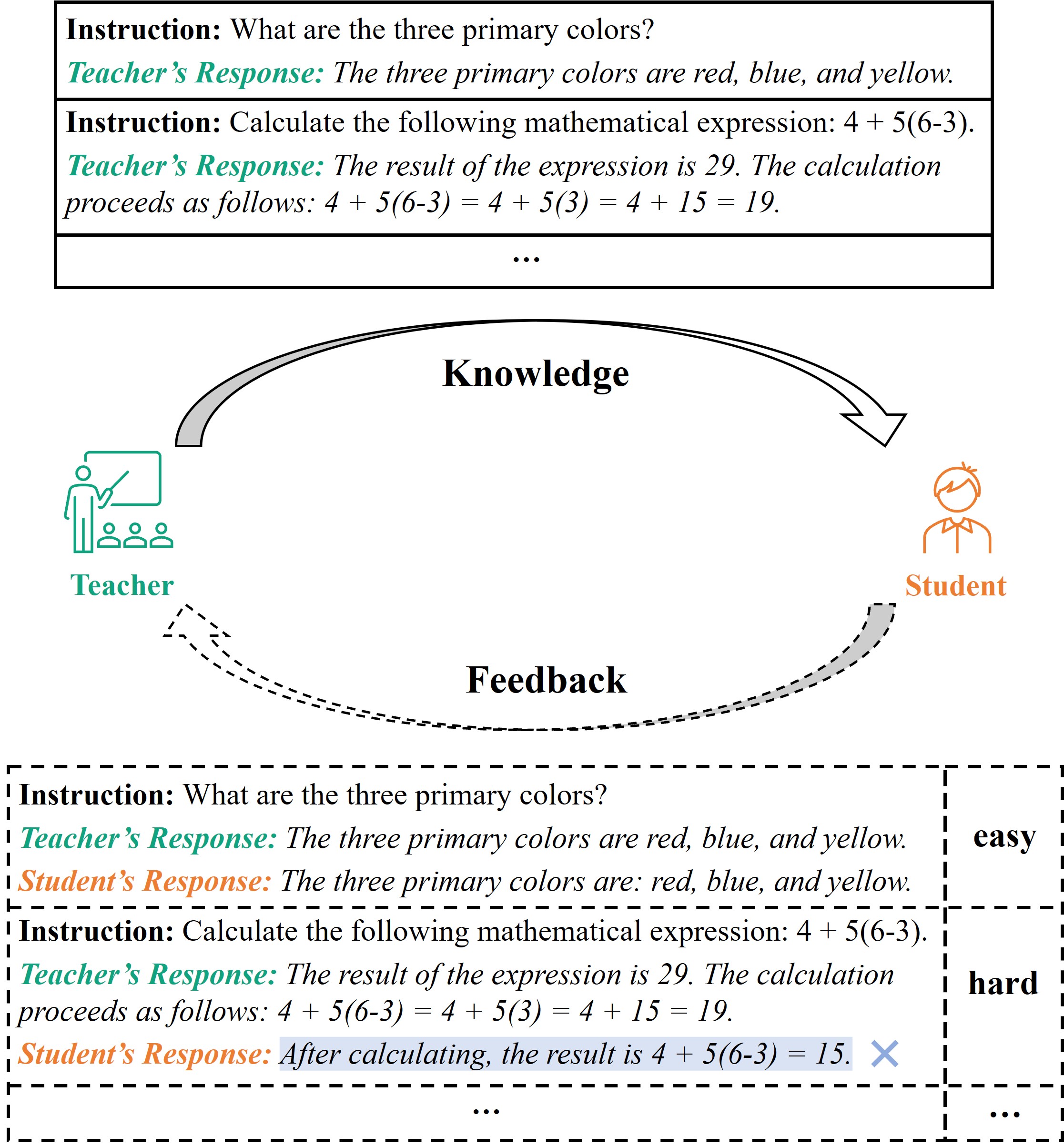}
\caption{An illustration of the distinction between our approach and earlier ones. Previous methods facilitate a one-way knowledge transfer from the teacher to the student (\textit{solid arrow}). Our approach, however, incorporates an innovative step (\textit{dashed arrow}) that completes a loop: it enables the feedback”\textemdash identifying the student model’s weaknesses\textemdash to be relayed back to the teacher, in order to foster tailored learning.}
\label{c3-fig: intro}
\end{figure}

However, previous works employ a unidirectional approach to knowledge transfer (solid arrow in Figure \ref{c3-fig: intro}), where the teacher imparts knowledge to the student without considering any ``feedback''.
To better illustrate this using a tangible classroom scenario, the ``feedback'' refers to identifying the ``hard'' examples or problems where the student's performance falls short.
This feedback guarantees that the teacher can provide bespoke training that centers on ``hard'' examples, thereby paving the way for more effective and tailored learning experiences for the student.

Inspired by adversarial knowledge distillation (AKD), which aims to iteratively improve the student model's performance by learning from generated hard samples \cite{fang2019akd, DBLP:conf/nips/MicaelliS19, DBLP:conf/aaai/HeoLY019}, we propose an adversarial framework for distilling a proprietary LLM into a compact student model.
Nevertheless, these AKD methodologies necessitate accessibility to the weights or gradients of the
teacher model, which cannot be directly adapted to our setting.
To circumvent this problem, we leverage the unparalleled role adaptability of LLMs, which can be effectively employed through a diverse range of prompts \cite{DBLP:conf/iclr/SanhWRBSACSRDBX22}. 
In particular, we prompt the proprietary teacher LLM to serve as a ``referee'' to discriminate hard instructions where there exists a significant performance discrepancy between the teacher's and student's responses, and serve as a ``generator'' to produce new instructions that emulate the data distributions corresponding to the discriminated hard instructions. 
Our framework, as depicted in Figure \ref{c3-fig: overview}, consists of three stages in an iteration: (1) an imitation stage to align the student's response with the teacher's response; (2) a discrimination stage to identify hard instructions; (3) A generation stage to produce new hard instructions for escalating the challenges presented to the student model.
In essence, our adversarial framework forms a \emph{positive feedback loop} that efficiently bootstraps the student model's proficiency.

To verify the efficiency and efficacy of our method, we apply our AKD framework to transfer the knowledge of ChatGPT\footnote{We access ChatGPT using the OpenAI API (\textit{gpt-3.5-turbo model}).} onto an open-source foundation LLM, known as LLaMA \cite{touvron2023llama}.
We select Alpaca's training data (generated from only 175 manually selected seed instructions) as the initial training instructions and execute three iterations of AKD, resulting in a total of 70K data that our model is trained on.
We've christened our model as \textbf{Lion}, drawing inspiration from the art of ``distillation''.
By conducting extensive experiments on open-ended generation and reasoning datasets, which include a total of 40 sub-tasks, our Lion-13B showcases superior performance surpassing instruction-tuned baseline models such as Vicuna~\cite{zheng2023judging}.

Our main contributions are as follows:
\begin{itemize}[]
\item Our work is the first attempt to adopt the idea of adversarial knowledge distillation to large language models.
\item Our proposed framework demonstrates impressive efficiency and efficacy. With instruction tuning performed on 70k data without any human annotation, our Lion-13B approximates ChatGPT's capabilities on open-ended generation dataset and largely outperforms the current SOTA model Vicuna-13B on reasoning tasks.
\item The versatility of our framework allows for broad application: it is not exclusive to ChatGPT but can be conveniently adapted to suit a variety of other proprietary LLMs. 
\end{itemize}

\section{Methodology}

\begin{figure*}[!tbp]
\centering
\includegraphics[width=\linewidth]{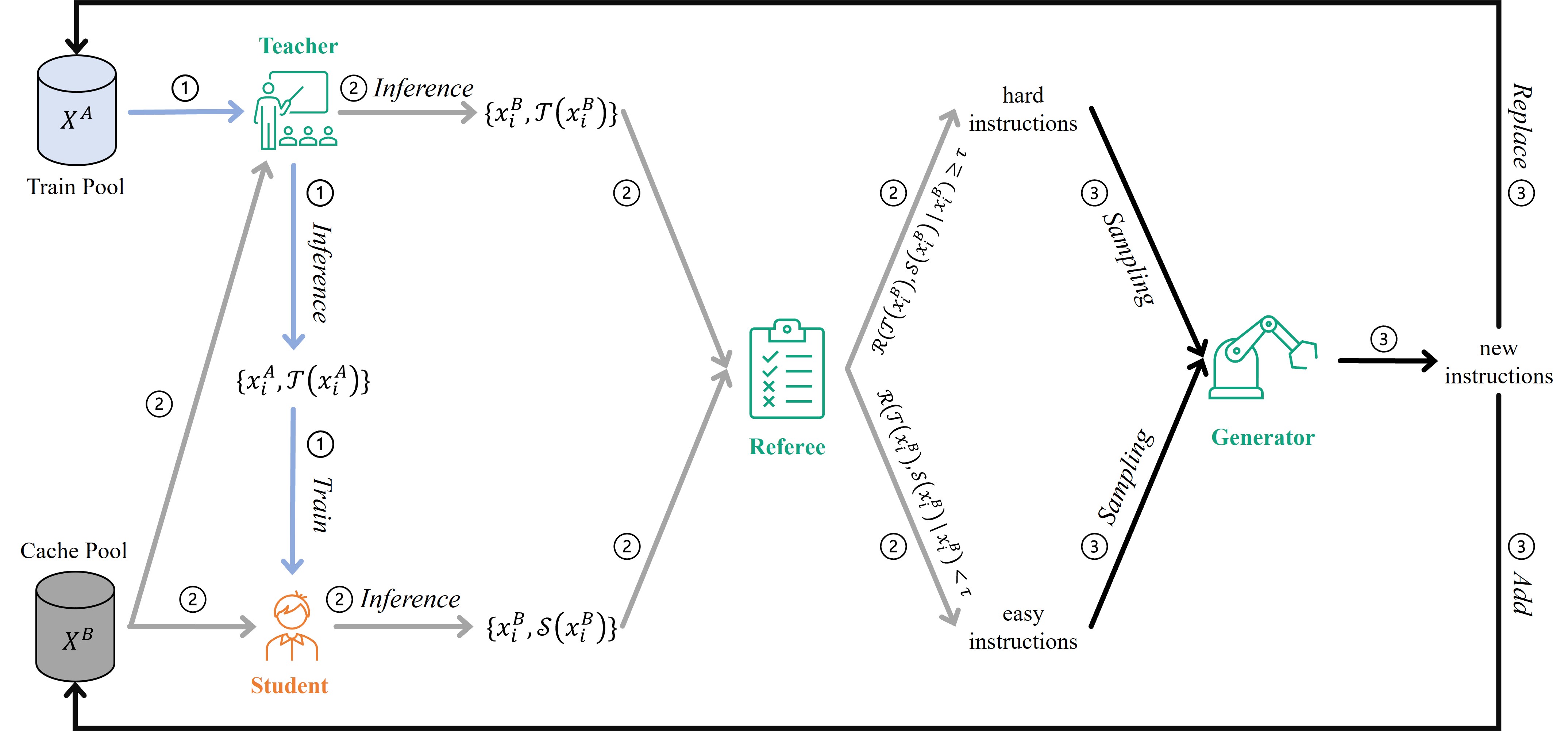}
\caption{The overview of our adversarial distillation framework, where we craft a compact Student LLM $\mathcal{S}$ based on a superior proprietary LLM that serves three roles: the \textbf{Teacher} $\mathcal{T}$, the \textbf{Referee} $\mathcal{R}$, and the \textbf{Generator} $\mathcal{G}$. From left to right, there are three stages in an iteration: 
(1) Imitation;
(2) Discrimination;
(3) Generation.
}
\label{c3-fig: overview}
\end{figure*}

Harnessing the learned knowledge of a sophisticated teacher model $\mathcal{T}(x; \theta^\mathcal{T})$ where the parameter $\theta^\mathcal{T}$ is inaccessible, our goal is to craft a more lightweight student model $\mathcal{S}(x; \theta^\mathcal{S})$.
Ideally, a student model is optimal if the expectation of model discrepancy (which indicates the prediction differences between teacher $\mathcal{T}$ and student $\mathcal{S}$) on the uniform data distribution is minimized. 
Inspired by the success of adversarial knowledge distillation (AKD) \cite{fang2019akd, DBLP:conf/nips/MicaelliS19, DBLP:conf/aaai/HeoLY019}, we turn to optimize an upper bound of the expectation \textemdash the expectation of the model discrepancy on ``hard samples'', where the teacher $\mathcal{T}$ and the student $\mathcal{S}$ have a relatively large performance gap. 
These ``hard samples'' are inclined to dominate the expectation of the model discrepancy.
Thus, the overall expected model discrepancy can be effectively and efficiently reduced by optimizing the student model $\mathcal{S}$ on these ``hard samples''.
The underlying rationale is rather straightforward and can be analogized to a real-world educational scenario: continuously concentrating on the ``hard'' knowledge that the student finds challenging to grasp is the most effective manner of enhancing a student's proficiency.

However, in the process of training the student model $\mathcal{S}$, hard samples will be mastered by the student and converted into easy samples. 
Hence we need a mechanism to continuously generate hard samples, which can be achieved by an adversarial framework.

The whole framework of our \textit{Adversarial Knowledge Distillation} is depicted in Figure \ref{c3-fig: overview},
which contains three stages in an iteration: 
(1) an imitation stage to align the student's response with the teacher's response; 
(2) a discrimination stage to identify hard samples; 
(3) A generation stage to produce new hard samples for escalating the challenges presented to the student model.

\subsection{Initilization}
As shown in Figure \ref{c3-fig: overview}, four roles and two data pools are established in our framework, and we will comprehensively illustrate their functions later.
We initialize our student model $\mathcal{S}$ using a foundation LLM such as LLaMA \cite{touvron2023llama}. We initialize our teacher model $\mathcal{T}$, referee $\mathcal{R}$, and generator $\mathcal{G}$ by using the same proprietary LLM such as ChatGPT \cite{openai2022chatgpt}.
The multiple roles that this proprietary LLM serves are accomplished through the use of varied prompt templates.
We start the iteration from a given initial Train Pool $X^{A}=\{x^{A}_i\}_{i \in [1, N^A]}$, where $x^{A}_i$ is the $i$-th instruction in $X^{A}$, and $N^A$ is the number of samples in $X^{A}$.
The Cache Pool $X^{B}$ is initialized as identical to $X^{A}$, consisting of instructions to evaluate the performance of $\mathcal{S}$ and $\mathcal{T}$.

\subsection{Imitation Stage}
To impart the knowledge of the teacher to the student, we construct the instruction-response data $\{x^{A}_i, \mathcal{T}(x^{A}_i)\}_{i \in [1, N^A]}$ by forward propagating instructions in the Train Pool $X^{A}$ through the teacher $\mathcal{T}$.
The prompt template used for model inference is shown in Table \ref{c3-tab: response}.
Like the imitation training of previous work~\cite{alpaca, zheng2023judging}, we fine-tune our student model $\mathcal{S}$ to align the response of the teacher model, by optimizing the autoregressive language modeling objective.

\subsection{Discrimination Stage}
\label{c3-sec: discrimination}
Figure \ref{c3-fig: overview} demonstrates that the discrimination stage starts from the Cache Pool, denoted as $X^{B}$. Even though this pool begins with the same initialization as the Train Pool, their uses diverge. The Train Pool is rejuvenated by replacing its existing instructions with freshly generated instructions, whereas the Cache Pool is enriched by incorporating these generated instructions. As a result, the growing storage capacity of the Cache Pool provides a more extensive space for evaluating the performance gap between teacher $\mathcal{T}$ and student $\mathcal{S}$. This allows for more thorough detection of hard instructions.

\begin{figure*}[!htbp]
  \centering
  \begin{subfigure}[b]{0.27\linewidth}
    \includegraphics[width=\linewidth]{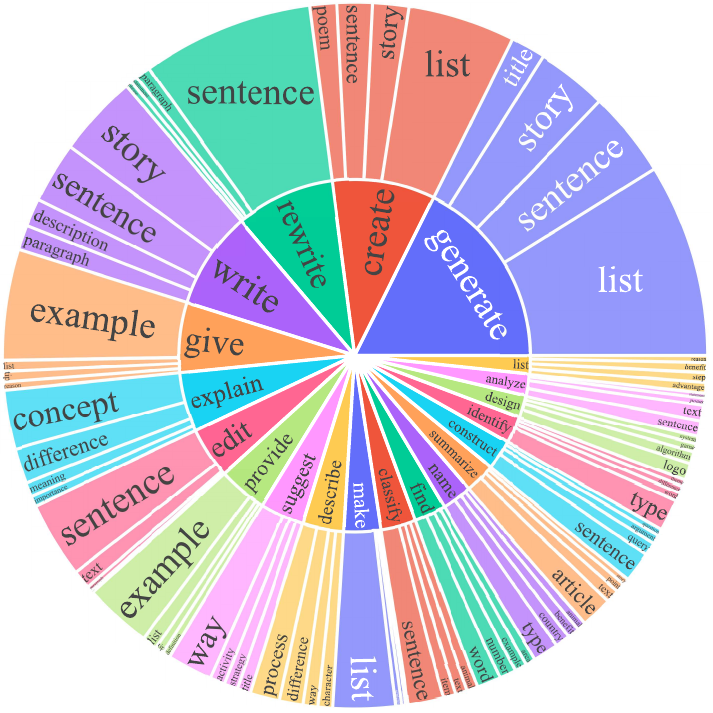}
    \caption{Instructions of the Cache Pool in the first iteration.}
    \label{c3-fig: model1}
  \end{subfigure}
  \qquad 
  \begin{subfigure}[b]{0.27\linewidth}
    \includegraphics[width=\linewidth]{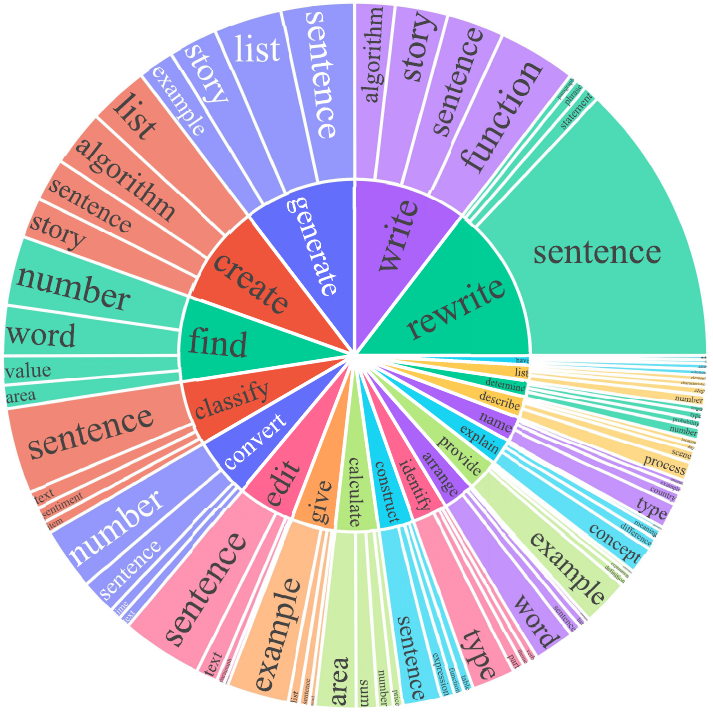}
    \caption{Identified hard instructions in the first iteration.}
    \label{c3-fig: model2}
  \end{subfigure}
  \qquad 
  \begin{subfigure}[b]{0.27\linewidth}
    \includegraphics[width=\linewidth]{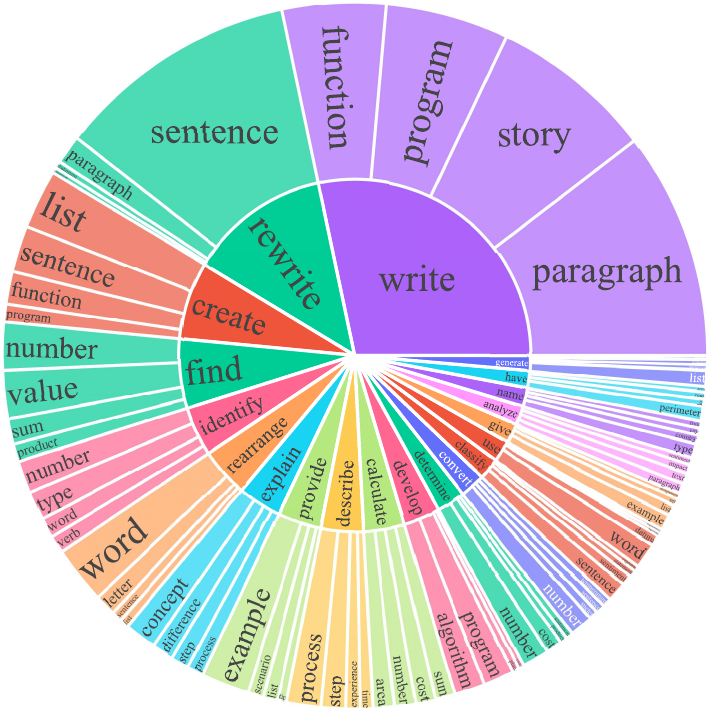}
    \caption{Generated hard instructions in the first iteration.}
    \label{c3-fig: model3}
  \end{subfigure}
  \caption{The top 20 most common root verbs (inner circle) and their top 4 direct noun objects (outer circle) in the instructions.}
  \label{c3-fig: comparison}
\end{figure*}

In the discrimination stage, we ask the proprietary LLM to serve as a ``referee'', which quantifies the performance gap between $\mathcal{T}$ and $\mathcal{S}$.
Specifically, we feed each instruction $x^{B}_i$ in the Cache Pool $X^{B}$ through both the teacher $\mathcal{T}$ and student $\mathcal{S}$ to generate the outputs $\mathcal{T}(x^{B}_i)$ and $\mathcal{S}(x^{B}_i)$, respectively.
Then we ask the referee $\mathcal{R}$ to quantitatively measure the quality difference between teacher's response $\mathcal{T}(x^{B}_i)$ and student's response $\mathcal{S}(x^{B}_i)$, conditioned on $x^{B}_i$:
\begin{equation}
d_i = \mathcal{R}(\mathcal{T}(x^{B}_i), \mathcal{S}(x^{B}_i) \ | \ x^{B}_i) 
\end{equation}
The above process is conducted by using the prompt template (as shown in Table \ref{c3-tab: discriminator_gpt-3.5-turbo}) inspired by \cite{zheng2023judging}, which requires the LLM to consider the helpfulness, relevance, accuracy, and level of detail of two responses and output two scores. 
To mitigate the positional bias~\cite{DBLP:journals/corr/abs-2305-17926} of the LLM referee, we conduct two runs by exchanging the positions of the teacher's response and the student's response and compute the final score as the average of the two runs.
Then $d_i$ is calculated as the difference between the teacher's score and the student's score.
By setting a threshold $\tau$ (1.0 used in our experiments), we discriminate hard instructions as those instructions with $d_i \ge \tau$, and the others are identified as easy ones.
Figure \ref{c3-fig: model2} provides a clear and intuitive demonstration of which kinds of instructions are discriminated as hard in the first iteration.
Compared with the instructions in the Cache Pool (Figure \ref{c3-fig: model1}), the distribution of the identified hard instructions is quite different, focusing more on complex tasks such as math, coding, etc.

\subsection{Generation Stage}
After carefully discerning the hard instructions, the generation stage aims to produce samples that mirror the data distributions corresponding to these challenging directives.
This process is achieved by employing the proprietary LLM as a generator, denoted as $\mathcal{G}$, leveraging its exceptional prowess in content creation.
Inspired by \cite{xu2024wizardlm}, we randomly sample an instruction from the hard instructions and prompt the generator $\mathcal{G}$ to generate a new instruction.
The newly generated instruction is required to pertain to the same domain and match the task type of the sampled instruction. The template utilized for this prompt is exhibited in Table \ref{c3-tab: generator_hard}.
As shown in Figure \ref{c3-fig: model3}, the distribution of the newly generated hard instructions appears to be comparable to that of the previously identified hard instructions.
To mitigate the issue of catastrophic forgetting and to augment the diversity of the generated instructions, we also randomly sample an instruction from the easy instructions and prompt the generator $\mathcal{G}$ to generate a new instruction that belongs to the same domain as the sampled one, but exhibit a more long-tailed distribution.
The template we use to prompt this process is displayed in Table \ref{c3-tab: generator_no_hard}.

In each iteration, we define $N$ as the total count of newly generated instructions and maintain a 1:1 ratio $r$ between the generated hard instructions and the generated easy instructions.
To promote diversity, a new instruction will be deemed valid only if its ROUGE-L overlap with any existing instructions in the Cache Pool is below 0.7.
Finally, as aforementioned in \S\ref{c3-sec: discrimination}, we proceed to rejuvenate the Train Pool, replacing its existing instructions with freshly generated ones. Concurrently, we enrich the Cache Pool by incorporating these newly generated instructions.

\subsection{Min-Max Game Interpretation}
Our adversarial knowledge distillation framework can be interpreted as a dynamic min-max game: in the imitation stage, we fine-tune our student to \textit{minimize} the model discrepancy between itself and the teacher on hard samples; in the discrimination and generation stage, we craft new hard samples to \textit{maximize} the model discrepancy, based on the learning progress of the student model.
This dialectic framework propels the student model towards uncovering otherwise hidden knowledge, paving the way to complete understanding.
As the training progresses through several iterations, the system should ideally achieve equilibrium. This is the point where the student model has mastered all the hard samples and the referee $\mathcal{R}$ can no longer distinguish between the student $\mathcal{S}$ and teacher $\mathcal{T}$ models. At this juncture, $\mathcal{S}$ becomes functionally indistinguishable from $\mathcal{T}$.

\section{Experiments}
\subsection{Experimental Setup}

\subsubsection{Datasets}
In our experiments, we implemented a comprehensive LLM evaluation protocol that considers a diverse range of abilities, such as writing, coding, commonsense, math, and logical reasoning.
The datasets we utilized can be classified into two main categories: open-ended generation and reasoning.

\paragraph{Open-ended Generation Datasets.}
Vicuna-Instructions \cite{zheng2023judging} is a set of 80 questions spanning 9 distinct task categories.
This dataset has gained extensive usage in evaluating the capabilities of LLMs. Within our work, we examine LLMs' performance on this dataset in two different settings:
\begin{itemize}
    \item \textbf{Setting1:} Following Vicuna~\cite{zheng2023judging}, we leverage GPT-4 to automatically assess the quality of responses (rated on a scale of 1 to 10) between a reference model (ChatGPT) and a candidate model. Subsequently, we calculate the candidate model's performance as the percentage of the total score it achieves compared to the reference model.
    \item \textbf{Setting2:} A recent work~\cite{DBLP:journals/corr/abs-2305-17926} pointed out that a systematic bias may exist in the above-mentioned GPT-4 automatic evaluation. To mitigate this, they propose two strategies, namely Multiple Evidence Calibration and Balanced Position Calibration, to obtain closer alignment with human judgments.
\end{itemize}

\paragraph{Reasoning Datasets.}
AGIEval \cite{zhong2023agieval} is a well-known benchmark that quantifies the reasoning capability of foundation models in the context of human-centric standardized exams, including college entrance exams, math competitions, lawyer qualification tests, etc.
We choose all English multiple-choice questions (8 tasks, 2,546 samples) among AGIEval for our experiments.
The data statistics are shown in Table \ref{c3-tab: agieval_sat}.
BIG-Bench Hard (BBH) \cite{suzgun2022bbh} consists of a suite of challenging tasks from BIG-Bench \cite{srivastava2022beyond}, designed to assess the capabilities and limitations of large language models.
These are the tasks on which prior language models underperform the average human rater.
We choose all tasks that can be formatted into multiple-choice questions (23 tasks, 5,511 samples) among BBH for our experiments.
The data statistics are shown in Table \ref{c3-tab: bbheval_sat}.
\begin{itemize}
\item \textbf{Setting:} We evaluate reasoning capabilities under a zero-shot setting without any exemplars and without Chain-of-Thought (CoT).
For both AGIEval and BBH, we use the prompt format and parsing following~\cite{zhong2023agieval, 2023orca}.
Given the free-form response from the generative models, only the first capital character in the response is considered to compare with the gold answer (exact match).
The result we report is accuracy (\%).
\end{itemize}

\subsubsection{Baselines}
We select five superior LLMs as baselines, including LLaMA \cite{touvron2023llama}, Alpaca \cite{alpaca}, WizardLM \cite{xu2024wizardlm}, Vicuna \cite{zheng2023judging}, and ChatGPT \cite{openai2022chatgpt}.
It is worth noting that Vicuna has consistently ranked as the top open-source language model on multiple leaderboards, such as Chatbot Arena\footnote{\url{https://chat.lmsys.org/?arena}}. Therefore, we will conduct a comprehensive comparison with Vicuna.
See detailed descriptions of these baselines in Appendix \ref{appendix:baseline}.

\subsubsection{Implementation Details}

\paragraph{Training Details.}
Our student model is initialized using the pre-trained LLaMA. 
The Train Pool and Cache Pool are initialized with the 52K automatically generated instructions from Alpaca \cite{alpaca}. 
The total number of iterations is set to 3, with 6K newly generated instructions added at each iteration. 
This results in a total of 70K data that our model is trained on in order to make a fair comparison with current SOTA baselines, including WizardLM and Vicuna.
The training hyperparameters are listed in Appendix \ref{c3-sec: gpt_parameter}.

\paragraph{Inference Details.}
To draw inferences from Lion and ChatGPT, we calibrated the temperature to 0.7 and set the maximum generation length at 1024. All other parameters adhere to their default settings.
For LLaMA, Alpaca, WizardLM, and Vicuna, we configured their inference parameters in line with the specifications given in their respective original papers.
When engaging with the gpt-3.5-turbo API for various roles, we employ an array of hyper-parameters, the specifics of which can be located in Appendix \ref{c3-sec: gpt_parameter}.

\subsection{Experimental Results}

\subsubsection{Results for Open-ended Generation}

Table \ref{c3-tab: gpt4_eval} shows the performance comparison of various models against ChatGPT as the reference model, where GPT-4 is used as a referee/rater.
Our Lion-7B and Lion-13B remarkably outperform their counterparts under two evaluation settings.
Noticeably, Lion-13B shows an 8-point improvement over Vicuna-13B on aggregate, achieving 98.38\% capabilities of ChatGPT.

\begin{table}[h]
\small
\centering
\begin{tabular}{l|c|c|c}
\toprule
\textbf{Model} & \textbf{Setting1} & \textbf{Setting2} & \textbf{Avg.} \\ \midrule
LLaMA-7B       & 58.46             & 59.12             & 58.79        \\
Alpaca-7B      & 69.29             & 67.20             & 68.25        \\
WizardLM-7B    & 89.29             & 86.67             & 87.98        \\
Vicuna-7B      & 87.79             & 89.96             & 88.88        \\
Lion-7B        & \textbf{94.74}             & \textbf{92.88}             & \textbf{93.81}        \\ \midrule
LLaMA-13B      & 69.23             & 68.21             & 68.72        \\
Alpaca-13B     & 76.87             & 74.69             & 75.78        \\
Vicuna-13B     & 92.25             & 92.97             & 92.61        \\
Lion-13B       & \textbf{96.57}             & \textbf{100.18}            & \textbf{98.38}        \\ \bottomrule
\end{tabular}
\caption{Relative response quality (\%) against ChatGPT (assessed by GPT-4) on Vicuna-Instructions.}
\label{c3-tab: gpt4_eval}
\end{table}

\begin{figure}[!htbp]
\centering
\includegraphics[width=0.7\linewidth]{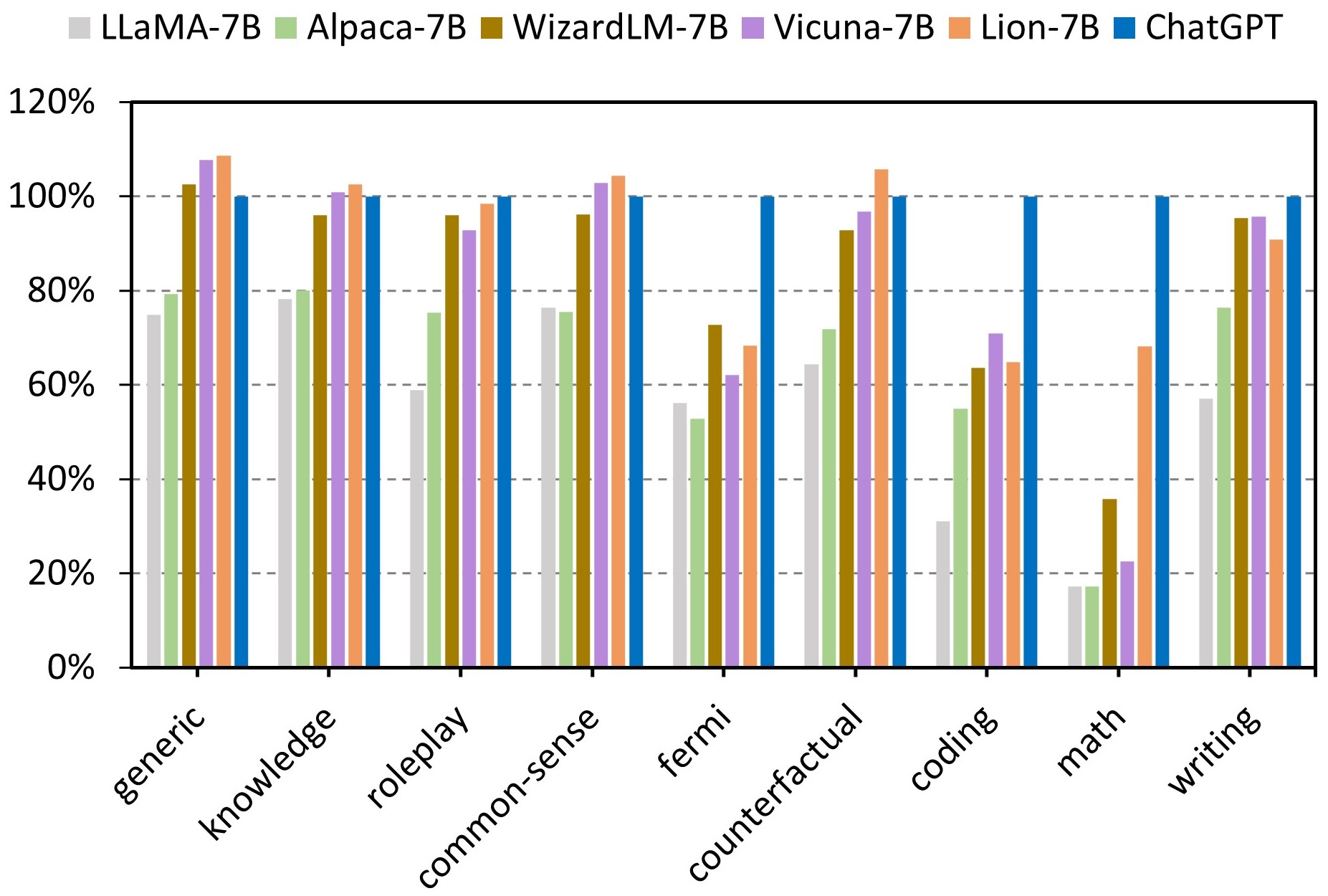}
\caption{
Relative response quality against ChatGPT on diverse task categories of Vicuna-Instructions.
}
\label{c3-fig: 80task_category}
\end{figure}

To comprehensively compare with other baseline models on the capability to generate high-quality responses on various types of instruction, the relative response quality (Setting2) among different task categories is depicted in Figure \ref{c3-fig: 80task_category}. Our model impressively and slightly surpasses ChatGPT in the generic, knowledge, common-sense, and counterfactual task categories. Furthermore, for the two difficulty task categories described in the previous study~\cite{zheng2023judging, xu2024wizardlm}, our model significantly outperforms other baseline models with at least 32.32\% relative score in the math task category while exceeding most of the baseline in the coding generation task category.

\subsubsection{Results for Reasoning}

\paragraph{AGIEval Results.} Table \ref{c3-tab: agieval} presents the standard zero-shot performance comparison between Lion and baseline models on the AGIEval benchmark for multiple-choice English questions. 
Lion demonstrates significantly stronger performance compared to Vicuna, surpassing it in most task categories and achieving an average relative improvement of over 16\%.
However, Lion-13B still significantly lags behind ChatGPT, only retaining 72.5\% of its reasoning capability.

\begin{table*}[t]
\scriptsize
\centering
\begin{tabular}{l|ll|l|ll|ll}
\toprule
\multirow{2}{*}{\textbf{Task}} & \multicolumn{2}{c|}{\textbf{Human}} & \multirow{2}{*}{\textbf{ChatGPT}} & \multirow{2}{*}{\textbf{Vicuna-7B}} & \multirow{2}{*}{\textbf{Lion-7B}} & \multirow{2}{*}{\textbf{Vicuna-13B}} & \multirow{2}{*}{\textbf{Lion-13B}} \\
                               & \textbf{Avg}             & \textbf{Top}              &                                  &                                   &                                                                       &               &              \\ \midrule
AQuA-RAT                       & 85.0            & 100.0                                       & 31.9                              & \textbf{23.2}                                                             & 18.5\enspace (-20.3\%)  & 20.1          & \textbf{26.0} \enspace (29.4\%)  \\
LogiQA                         & 86.0            & 95.0                                        & 35.0                              & 21.4                                                             & \textbf{31.8}\enspace (48.6\%) & 29.8  & \textbf{31.3} \enspace (5.0\%)           \\
LSAT-AR                        & 56.0            & 91.0                                     & 24.4                              & \textbf{22.2}                                                            & 17.4\enspace (-21.6\%) & 20.4            & \textbf{23.0} \enspace (12.7\%)  \\
LSAT-LR                        & 56.0            & 91.0                                       & 52.6                              & 18.6                                                            & \textbf{28.2}\enspace (51.6\%)   & \textbf{32.6}          & \textbf{32.6} \enspace (0.0\%)  \\
LSAT-RC                        & 56.0            & 91.0                                        & 65.4                              & 21.9                                                            & \textbf{29.4}\enspace (34.2\%)  & 32.7           & \textbf{40.9} \enspace (25.1\%)  \\
SAT-Math                       & 66.0            & 94.0                                      & 42.7                              & \textbf{21.4}                                                  & 20.9\enspace (-2.3\%)  & 28.6            & \textbf{29.4} \enspace (2.8\%)           \\
SAT-English                    & 66.0            & 94.0                                    & 81.1                              & 25.7                                                           & \textbf{36.4}\enspace (41.6\%)   & 44.2           & \textbf{53.9} \enspace (21.9\%)  \\
SAT-English (w/o Psg.)         & 66.0            & 94.0                                      & 44.2                              & 26.2                                                           & \textbf{27.7}\enspace (5.7\%)  & 26.2            & \textbf{36.2} \enspace (38.2\%)  \\ \midrule
Average                        & 67.1            & 93.8                                       & 47.2                              & 22.6                        & \textbf{26.3}\enspace (16.4\%)      & 29.3                                           & \textbf{34.2} \enspace (16.7\%) \\ \bottomrule
\end{tabular}
\caption{Zero-shot performance comparison of ChatGPT, Vicuna, and Lion on AGIEval (multiple-choice English questions). We report the performance of Human, ChatGPT, and Vicuna from \cite{2023orca}. Performance improvements obtained by Lion over Vicuna are shown in parenthesis.}
\label{c3-tab: agieval}
\end{table*}

\begin{table*}[!t]
\scriptsize
\centering
\begin{tabular}{l|l|ll|ll}
\toprule
\textbf{Task} & \textbf{ChatGPT} & \textbf{Vicuna-7B} & \textbf{Lion-7B} & \textbf{Vicuna-13B} & \textbf{Lion-13B} \\ \midrule
Boolean Expressions                   & 82.8             & 39.2                          & \textbf{55.2} \enspace (40.8\%)  & 40.8              & \textbf{65.6} \enspace (60.8\%)     \\
Causal Judgement                      & 57.2             & 39.7                        & \textbf{50.3} \enspace (26.7\%)   & 42.2  & \textbf{43.9} \enspace (4.0\%)             \\
Date Understanding                    & 42.8             & 8.6                    & \textbf{34.0} \enspace (295.3\%)     & 10.0               & \textbf{40.4} \enspace (304.0\%)    \\
Disambiguation QA                     & 57.2             & 15.2                           & \textbf{35.6} \enspace (134.2\%)  & 18.4           & \textbf{44.8} \enspace (143.5\%)    \\
Formal Fallacies                      & 53.6             & 40.0                         & \textbf{46.0} \enspace (15.0\%)  & 47.2             & \textbf{52.4} \enspace (11.0\%)    \\
Geometric Shapes                      & 25.6             & 3.6                         & \textbf{8.8} \enspace (144.4\%)   & 3.6              & \textbf{8.8} \enspace (144.4\%)     \\
Hyperbaton                            & 69.2             & 42.8                        & \textbf{51.6} \enspace (20.6\%)    & 44.0            & \textbf{56.8} \enspace (29.1\%)    \\
Logical Deduction (5 objects)         & 38.8             & 4.8                         & \textbf{19.6} \enspace (308.3\%)  & 4.8              & \textbf{20.8} \enspace (333.3\%)    \\
Logical Deduction (7 objects)         & 39.6             & 1.2                           & \textbf{14.4} \enspace (1100.0\%)   & 1.2           & \textbf{21.2} \enspace (1666.7\%)    \\
Logical Deduction (3 objects)         & 60.4             & 19.6                        & \textbf{40.4} \enspace (106.1\%) & 16.8      & \textbf{38.0} \enspace (126.2\%)             \\
Movie Recommendation                  & 55.4             & 24.4                          & \textbf{26.8} \enspace (9.8\%)    & 43.4          & \textbf{57.6} \enspace (32.7\%)    \\
Navigate                              & 55.6             & 43.6                          & \textbf{49.2} \enspace (12.8\%)  & \textbf{46.4}   & 45.2 \enspace (-2.6\%)             \\
Penguins in a Table                   & 45.9             & 17.5                        & \textbf{24.7} \enspace (41.1\%)    & 15.1            & \textbf{26.7} \enspace (76.8\%)    \\
Reasoning about Colored Objects       & 47.6             & 14.0                        & \textbf{15.2} \enspace (8.6\%)    & 12.0            & \textbf{17.6} \enspace (46.7\%)    \\
Ruin Names                            & 56.0             & 12.2                        & \textbf{14.4} \enspace (18.0\%)    & 15.7            & \textbf{29.2} \enspace (86.0\%)    \\
Salient Translation Error Detection   & 40.8             & 2.0                         & \textbf{12.0} \enspace (500.0\%)   & 2.0             & \textbf{12.4} \enspace (520.0\%)    \\
Snarks                                & 59.0             & 28.0                          & \textbf{56.2} \enspace (100.7\%)   & 28.1           & \textbf{61.2} \enspace (117.8\%)    \\
Sports Understanding                  & 79.6             & 40.4                       & \textbf{48.4} \enspace (19.8\%)    & 48.4             & \textbf{51.6} \enspace (6.6\%)    \\
Temporal Sequences                    & 35.6             & 21.2                         & \textbf{24.4} \enspace (15.1\%)   & \textbf{16.0}   & 10.4 \enspace (-35.0\%)             \\
Tracking Shuffled Objects (5 objects) & 18.4             & 6.4                          & \textbf{14.4} \enspace (125.0\%)    & 9.2           & \textbf{24.8} \enspace (169.6\%)    \\
Tracking Shuffled Objects (7 objects) & 15.2             & 4.0                          & \textbf{13.6} \enspace (240.0\%)  & 5.6    & \textbf{13.2} \enspace (135.7\%)             \\
Tracking Shuffled Objects (3 objects) & 31.6             & 26.8                        & \textbf{34.0} \enspace (26.9\%)    & 23.2            & \textbf{34.4} \enspace (48.3\%)    \\
Web of Lies                           & 56.0             & \textbf{49.4}                       & 47.2 \enspace (-4.5\%)   & 41.2              & \textbf{54.8} \enspace (33.0\%)    \\ \midrule
Average                               & 48.9             & 21.9                         & \textbf{32.0} \enspace (45.9\%)    & 23.3           & \textbf{36.2} \enspace (55.4\%)   \\ \bottomrule
\end{tabular}
\caption{Zero-shot performance comparison of ChatGPT, Vicuna, and Lion on BIGBench Hard (multiple-choice questions) without CoT. We report the performance of ChatGPT and Vicuna from \cite{2023orca}. Performance improvements obtained by Lion over Vicuna are shown in parenthesis.}
\label{c3-tab: bbheval}
\end{table*}

\paragraph{BIG-Bench Hard Results.} Table \ref{c3-tab: bbheval} displays the zero-shot performance comparison between Lion and baseline models on BIG-Bench Hard with standard zero-shot prompting.
Similar to AGIEval, Vicuna exhibits poor performance
on sophisticated reasoning tasks within this benchmark, while Lion substantially surpasses Vicuna by around 50\% on average. 
Particularly, Lion demonstrates significant performance enhancements of over 100\% on tasks involving
data understanding, semantic understanding (Disambiguation QA and Snarks), logical and geometric reasoning (Logical Deduction and Geometric Shapes), and position reasoning (Tracking Shuffled Objects).
Despite achieving an average ability of nearly 74\% compared to ChatGPT on BBH, Lion-13B surpasses ChatGPT in several tasks, including Movie Recommendation, Snarks (identifying sarcastic sentences from two nearly-identical ones), and Tracking Shuffled Objects. This demonstrates the effectiveness of our method.

\section{Analysis}
\subsection{Ablation Study}

\paragraph{The Threshold $\tau$ for Distinguishing between Hard and Easy Instructions.}
We systematically explored $\tau$ ranging from 0.0 to 2.0 and documented its influence on average performance across three datasets. Table \ref{c3-tab: ablation_tau} reveals an optimal range of $\tau$ between 1.0 and 1.5 for all datasets. Notably, elevating $\tau$ from 0.0 to 1.0 consistently enhances performance across all datasets, indicating effective differentiation between hard and easy instructions. However, a continuous increase from 1.0 to 2.0 gradually degrades performance due to decreased diversity in hard instructions. The ablation results demonstrate that our method is not quite sensitive to a large value of $\tau$.

\begin{table*}[h]
\small
\centering
\begin{tabular}{c|c|c|c}
\toprule
\textbf{Threshold $\tau$} & \textbf{Vicuna-Instructions (Avg.)} & \textbf{AGIEval (Avg.)} & \textbf{BBH (Avg.)} \\ \midrule
0.0 & 89.58	& 22.4 &	26.5        \\
0.5	& 92.16 &	23.5 &	29.8        \\
1.0	& 93.81	& \textbf{26.3}	& \textbf{32.0}        \\
1.5	& \textbf{94.09}	& 25.7 &	31.6        \\
2.0	& 92.23	& 24.6	& 31.3        \\ \bottomrule
\end{tabular}
\caption{Ablation study of the threshold $\tau$ for Lion-7B.}
\label{c3-tab: ablation_tau}
\end{table*}

\paragraph{The Ratio $r$ of Generated Hard and Easy Instructions.}
We change the ratio of generated hard instructions to generated easy instructions from 1:0 (all hard) to 0:1 (all easy) and investigate its impact on average performance across three datasets. It can be seen from Table \ref{c3-tab: ablation_r} that higher ratios of hard to easy instructions generally lead to improved performance, with a balanced ratio of 1:1 yielding the highest average scores.

\begin{table*}[h]
\small
\centering
\begin{tabular}{c|c|c|c}
\toprule
\textbf{Ratio $r$} & \textbf{Vicuna-Instructions (Avg.)} & \textbf{AGIEval (Avg.)} & \textbf{BBH (Avg.)} \\ \midrule
1:0	& 89.60	& 24.3	& 30.8        \\
2:1	& 92.95	& 25.7	& \textbf{33.1}        \\
1:1	& \textbf{93.81}	& \textbf{26.3}	& 32.0        \\
1:2	& 91.77	& 23.9	& 29.6        \\
0:1	& 90.02	& 22.1	& 24.3        \\ \bottomrule
\end{tabular}
\caption{Ablation study of the ratio $r$ for Lion-7B.}
\label{c3-tab: ablation_r}
\end{table*}

\subsection{The Learning Dynamics of Lion}
\begin{figure}[!htbp]
\centering
\includegraphics[width=0.8\linewidth]{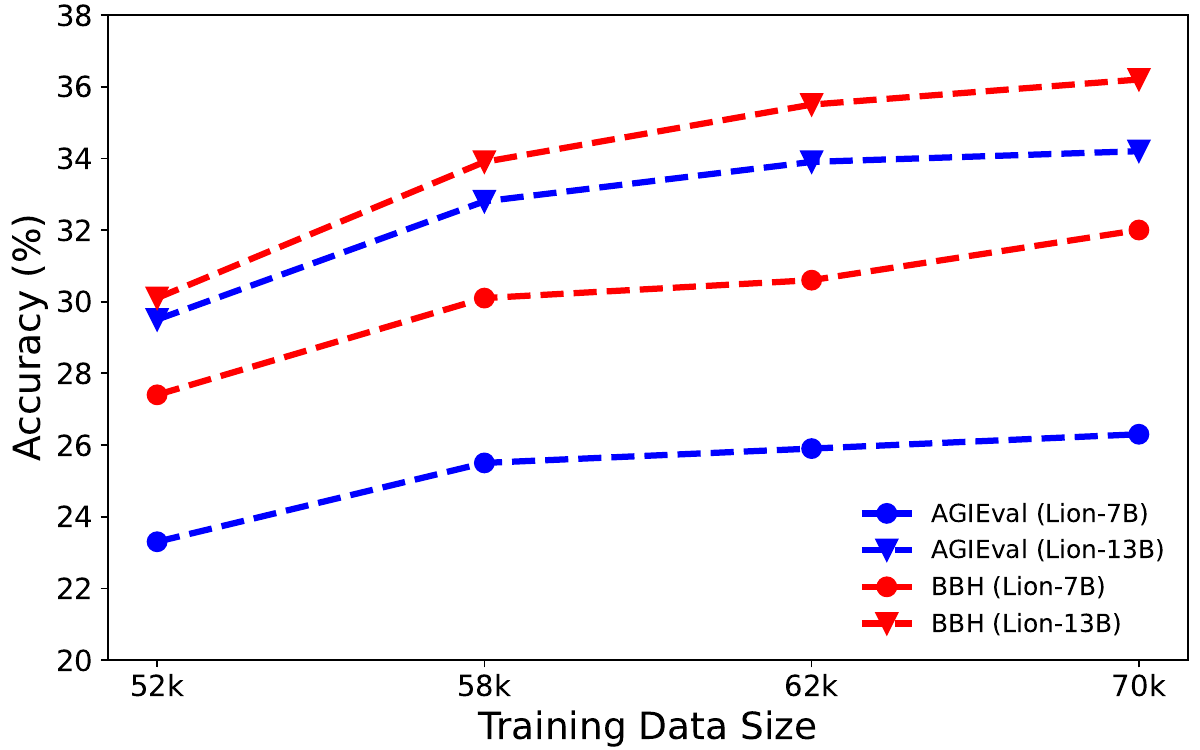}
\caption{
Performance of Lion-7B and Lion-13B on AGIEval and BBH through the training iterations.
}
\label{c3-fig: learning_dynamics}
\end{figure}

In Figure \ref{c3-fig: learning_dynamics}, we delve into the learning dynamics of Lion by visualizing its performance on AGIEval and BBH throughout the training iterations.
The results clearly demonstrate that our adversarial knowledge distillation framework consistently enhances the performance of the student model as the iterations progress.
Notably, the most significant improvement in capability occurs in the first iteration, suggesting the usefulness of the identification of challenging example patterns (refer Figure \ref{c3-fig: model2}).

\section{Conclusion and Discussion}
\subsection{Conclusion}
This chapter presents an innovative adversarial knowledge distillation framework for distilling a proprietary LLM into a compact, open-source student model. 
While previous methodologies have concentrated on unidirectional knowledge transfer, our approach seeks to integrate ``feedback'' into the learning process. 
Leveraging the versatile role adaptability of LLMs, we prompt the proprietary model to identify ``hard'' instructions and generate new ``hard'' instructions for the student model, creating a three-stage adversarial loop of imitation, discrimination, and generation. 
This approach allows us to refine the student model's performance iteratively, efficiently bootstrapping its proficiency.
We aspire that our model, named Lion, may serve as a baseline to reflect the performance of ChatGPT, especially the open-source instruction-following language model baseline for our community.

\subsection{Discussion}

\paragraph{The Model Capability.}
We have identified that Lion is subject to certain constraints:
(1) A recent study \cite{DBLP:journals/corr/abs-2305-15717} asserts that ``model imitation is a false promise'' since imitation models are adept at mimicking ChatGPT’s style but fall short in improving LMs across more challenging tasks.
While Lion still lags behind its teacher model ChatGPT in handling intricate reasoning tasks (as shown in our experiments), it demonstrates promising improvements compared to previous imitation models.
Therefore, our adversarial knowledge distillation framework may provide a more effective way for knowledge transfer.
(2) Since our training data doesn't encompass dialogues, Lion struggles to manage multi-turn conversations.
(3) Due to computational resource constraints, Lion's maximum sequence length is limited to 1024. Consequently, it faces challenges when dealing with long documents.
Despite these limitations, we envision Lion serving as an accessible springboard for future research endeavors aimed at addressing these limitations.

\paragraph{The Training Process.}
To train a single student model, we request the gpt-3.5-turbo API around 450k times,
a number that is roughly 70\% of the WizardLM's usage of 624k \cite{xu2024wizardlm}. Nonetheless, this utilization incurs a considerable expense, nearing \$900. 
In contrast to methods like Alpaca \cite{alpaca} and WizardLM \cite{xu2024wizardlm}, which only fine-tune the student model once, our adversarial knowledge distillation method employs iterative parametric updates to the student model. While this iterative approach inevitably leads to slower iteration speed, it offers additional benefits.
Finally, different from traditional adversarial knowledge distillation where the weights of the generator are iteratively updated, we use a black-box and parameter-frozen LLM (ChatGPT in our paper) to serve the role. Therefore, the quality of the LLM is quite essential in the generation of new instructions.

\paragraph{The Evaluation Metrics.}
Though automated evaluations leveraging GPT-4 have showcased promising prospects in appraising chatbot performance, the technique is yet to reach a level of maturity and accuracy, especially considering the propensity of large language models to generate non-existent or ``hallucinated'' information. 
Evaluating the efficacy of LLM across various tasks presents a considerable challenge since different tasks require quite different expertise \cite{wang-etal-2023-self-instruct}.
Therefore, the creation of a comprehensive, standardized evaluation system for chatbots is a prevailing research challenge that demands additional exploration and study.
\chapter{Alignment Data Synthesis from Scratch via Web Reconstruction}\label{chap:webr}

\section{Introduction}
LLMs~\cite{brown2020language, 2023gpt4, llama3modelcard} have become integral across a myriad of applications, demonstrating exceptional performance on diverse tasks by effectively following instructions and aligning with human values~\cite{openai2022chatgpt, 2023gpt4}.
Their remarkable performance largely stems from supervised fine-tuning (SFT)~\cite{wei2022finetuned, mishra-etal-2022-cross} on instruction-response pairs.
This process empowers LLMs to produce customized outputs when provided with specific instructions, facilitating their adaptation to novel tasks without prior exposure.

A fundamental challenge in advancing the instruction-following capabilities of LLMs lies in the collection of high-quality instruction-tuning (IT) data.
Early approaches primarily rely on human experts to manually generate and curate IT data~\cite{wang-etal-2022-super, DatabricksBlog2023DollyV2}, which is both time-intensive and resource-heavy.
To mitigate these limitations, \textbf{Semi-Automated Synthetic Methods}~\cite{wang-etal-2023-self-instruct, alpaca, xu2024wizardlm} leverage LLMs to expand small, human-annotated seed datasets using few-shot prompting techniques.
While effective, the performance of these methods is highly sensitive to prompt engineering and the careful selection of seed examples~\cite{xu2024magpie}.
More recently, \textbf{Fully Automated Synthetic Methods}, such as WebInstruct~\cite{yue2024mammoth2} and instruction backtranslation~\cite{li2024selfalignment}, have emerged as scalable alternatives that eliminate human involvement by synthesizing IT data based on web-scraped documents.
These methods, however, often operate under strong assumptions about the structure and content of raw web data, such as the availability of explicit question-answer pairs or minimal irrelevant content.
Consequently, they can only handle a limited scope of web documents, restricting their diversity and leading to suboptimal performance across various tasks.

\begin{figure}[!t]
\centering
\includegraphics[width=0.7\linewidth]{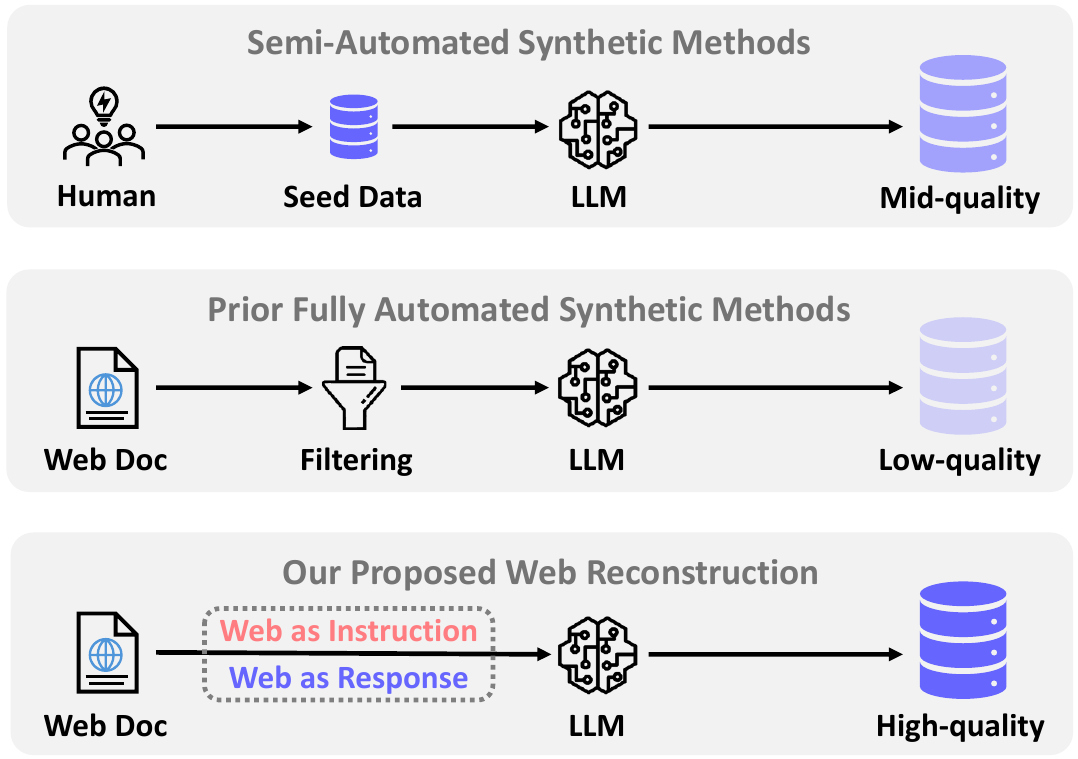}
\caption{
Our proposed Web Reconstruction method surpasses previous techniques by being (1) fully automated, eliminating the need for manual intervention or seed data; (2) minimally reliant on assumptions about the structure and content of web documents; and (3) capable of generating high-quality IT data.
}
\label{c4-fig: intro}
\end{figure}

To overcome these limitations, we propose \textbf{Web Reconstruction (WebR)}—a novel framework that synthesizes high-quality IT data from raw web documents \textbf{with minimal assumptions on web} and \textbf{no reliance on human annotations}, enabling broader adaptability and improved performance.
Unlike backtranslation, which directly treats web content as a response, or WebInstruct, which extracts QA pairs, WebR introduces a novel paradigm by \textbf{conceptualizing web reconstruction as an instruction-tuning data synthesis task}.
At its core, WebR aims to transform raw, noisy web documents into human-preferred, response-like outputs through a dual-perspective paradigm. Each web document is designated as either an instruction or a response, triggering the reconstruction process:
(1) \textit{Web as Instruction} introduces a first-of-its-kind \textbf{web rewriting} approach in IT data synthesis, where raw web document is concatenated with a synthesized rewrite request to serve as a complete instruction;
(2) \textit{Web as Response} enhances backtranslation~\cite{li2024selfalignment} by introducing a novel rollout and refinement process, mitigating reliance on strong web content assumptions.
Crucially, we show that these two perspectives are \textbf{complementary} (refer to Table \ref{c4-tab: ablation}): \textit{Web as Instruction} enhances reasoning and understanding tasks, while \textit{Web as Response} improves instruction-following and question-answering tasks.

We apply WebR to the Llama3-70B-Instruct and GPT-4o-mini models, creating two 100k-sample IT datasets: WebR-Basic and WebR-Pro.
To validate their effectiveness, we train various LLMs, including Llama3-8B-base and Qwen2.5-1.5/3/7/14B-base, and evaluate them on over ten widely used benchmarks.
Our experiments provide key contributions and insights into IT data synthesis:
\begin{itemize}
    \item \textbf{Efficacy:} WebR is the first web-based IT synthetic method to consistently surpass current IT datasets with human annotations;
    \item \textbf{Compatibility:} Merging WebR with existing IT datasets yields further performance gains;
    \item \textbf{Data Efficiency:} The performance of WebR improves linearly relative to the logarithmic growth of training data;
    \item \textbf{Scalability:} WebR scales with LLM size, consistently boosting larger models;
    \item \textbf{Domain Adaptability:} WebR achieves domain adaption by simply adjusting the proportion of source web documents.
\end{itemize}

\section{Web Reconstruction}
\begin{figure*}[!t]
\centering
\includegraphics[width=\linewidth]{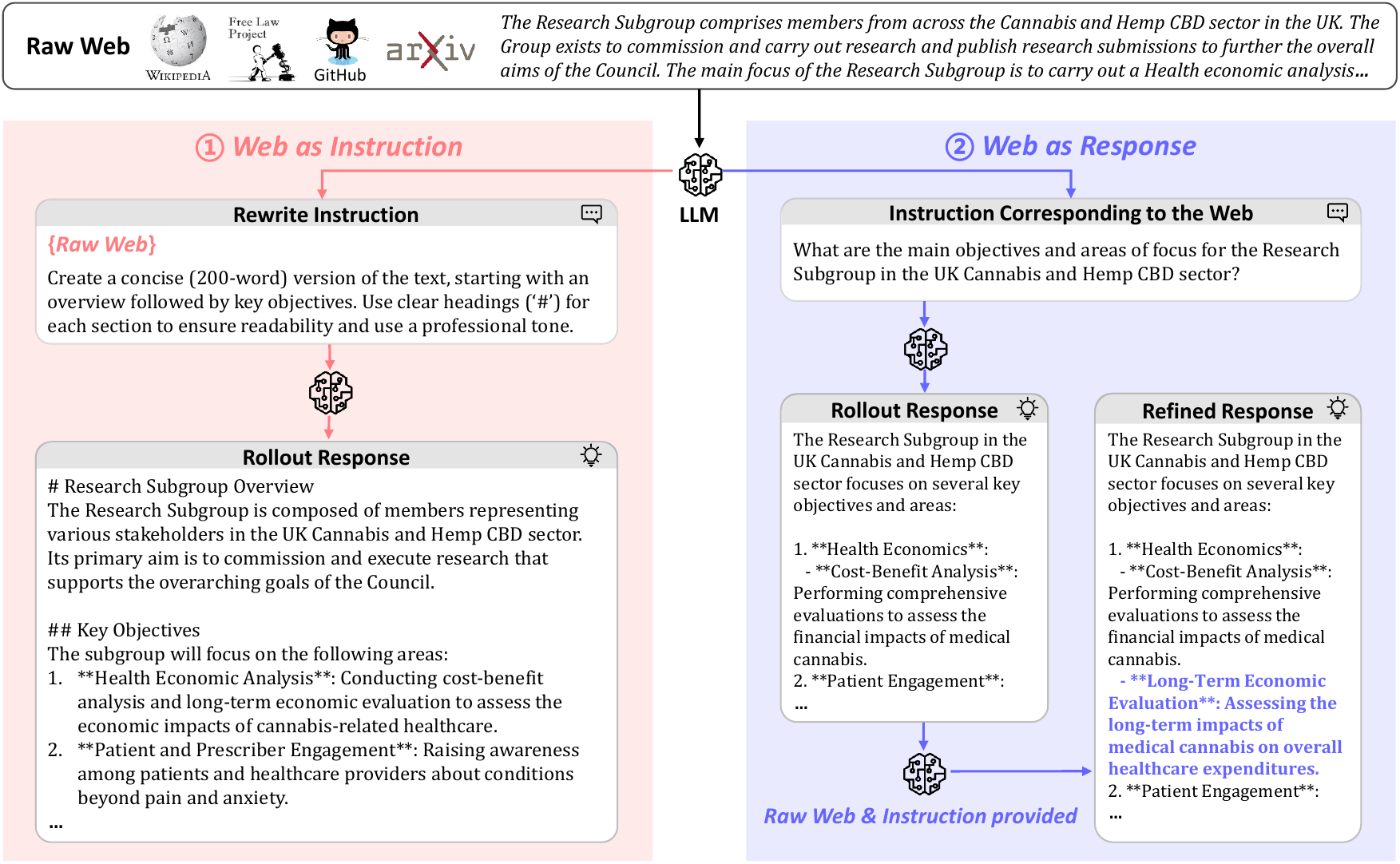}
\caption{
Overview of the proposed \textbf{Web Reconstruction} (WebR) framework.
Leveraging an off-the-shelf LLM, WebR transforms raw web documents into high-quality instruction-response pairs.
It strategically assigns each document as either an instruction or a response to trigger the process of web reconstruction.
}
\label{c4-fig: method}
\end{figure*}

Prior fully automated synthetic methods often rely on strong assumptions about the structure and content of raw web documents—such as the presence of explicit question-answer pairs, minimal irrelevant content, or appropriate expressions—necessitating complex preprocessing steps like retrieval and filtering.
In contrast, we introduce the \textbf{Web Reconstruction} (WebR) framework, which leverages a powerful, off-the-shelf LLM to overcome  these limitations by directly reconstructing unstructured and noisy web content into high-quality, response-like outputs.
As shown in Figure \ref{c4-fig: method}, WebR comprises two core strategies: (1) \textit{Web as Instruction}, where raw web content is concatenated with a synthesized rewrite request to serve as a complete instruction, guiding the generation of a reorganized, coherent response;
(2) \textit{Web as Response}, where a latent instruction is inferred by treating raw web content as a response, enabling reconstruction through the LLM's initial rollout and subsequent refinement.
By adopting this dual-branch approach, WebR efficiently generates high-quality instruction-response pairs, ensuring contextually appropriate outputs while eliminating the need for extensive preprocessing.

\subsection{Web as Instruction}
Raw web documents often contain disorganized or irrelevant information that hinders direct usability.
Even when dealing with well-structured content, further refinement is often required to meet human-preferred formats and stylistic conventions. 
A natural approach to reconstructing web content is to rewrite it according to specific requirements, such as style, format, structure, etc.
To ensure diverse and realistic rewriting scenarios, we leverage a powerful LLM to generate a detailed rewrite request tailored to the original document’s content (See prompt in Figure \ref{c4-fig: l2l_all_prompt}).
The request, along with the raw web content, are concatenated to form a comprehensive instruction.
In addition to whole-document transformations, we further enhance task diversity by randomly (50\% probability) generating rewrite requests that target \textit{specific sections} of the web content rather than the entire document, as shown in Figure \ref{c4-fig: l2l_part_prompt}.
This simulates real-world text manipulation scenarios where users may need to extract and modify only certain portions of a text.
The curated instructions are then processed by the LLM to produce reconstructed web content.
Notably, 
the complexity of rewrite requests naturally encompasses various NLP tasks, such as summarization, information extraction, and semantic understanding.
Addressing these tasks requires LLM to demonstrate advanced reasoning and comprehension abilities, thereby enhancing its proficiency in instruction-following, contextual understanding, and reasoning (as verified in Table \ref{c4-tab: ablation}).

\subsection{Web as Response}
Inspired by instruction backtranslation~\cite{li2024selfalignment}, we propose an alternative approach to reconstruct web content by treating the web as a response.
Specifically, we utilize a LLM to predict a latent instruction for which the raw web content would serve as an ideal response, as illustrated in Figure \ref{c4-fig: s2l_all_prompt}.
To further enhance diversity, specific segments of web content are treated as responses (with a 50\% probability), as depicted in Figure \ref{c4-fig: s2l_part_prompt}.
Unlike traditional back-translation methods, which directly treat latent instructions and raw web content as instruction-response pairs, \textbf{our approach introduces a two-stage refinement process}. First, we generate an initial response by rolling out an LLM prediction for the latent instruction. Next, we refine this response using both the raw web content and the latent instruction to produce a more accurate and comprehensive output, as shown in Figure \ref{c4-fig: refine_prompt}.
The initial rollout ensures that the response exhibits human-like fluency and natural language style, while the subsequent refinement step integrates critical information from the raw web, ensuring that the final response is both precise and thorough. This dual-stage process significantly enhances the LLM's performance in knowledge acquisition and question-answering tasks, as demonstrated by the improvements reported in Table \ref{c4-tab: ablation}.
The generated instruction as well as the refined response are finally paired as IT data.

\subsection{Dataset Construction Details}
Following prior work~\cite{li2024selfalignment, yue2024mammoth2}, we construct our dataset by sampling raw web documents from three diverse and representative domains: 70\% from the English subset of Common Crawl~\cite{together2023redpajama} (general domain), 15\% from OpenWebMath~\cite{paster2024openwebmath} (math domain), and 15\% from GitHub~\cite{together2023redpajama} (code domain).
To enable large-scale creation of diverse synthetic data for various scenarios, we adopt a persona-driven instruction synthesis strategy inspired by \cite{DBLP:journals/corr/abs-2406-20094}.
Initially, an LLM generates personas for the raw web documents (see template in Figure \ref{c4-fig: persona_prompt}), which guide the subsequent instruction synthesis for our proposed Web Reconstruction process.
The ratio of \textit{Web as Instruction} to \textit{Web as Response} is set to 2:1, following insights from the ablation study presented in Table \ref{c4-tab: ablation}.
To enhance diversity and eliminate redundancy, we apply MinHash~\cite{broder1997resemblance} deduplication based on n-gram features of instructions.
We configure the signature size to 128 and the similarity threshold to 0.7.
The final synthesized dataset comprises 100,000 instruction-response pairs.

To evaluate the effectiveness of WebR in generating high-quality IT datasets, we use WebR to construct datasets with two LLMs: the open-source \texttt{Llama3-70B-Instruct}~\cite{llama3modelcard} (temperature=0.6, top-p=0.9) and the proprietary \texttt{GPT-4o-mini}~\cite{2023gpt4} (temperature=0.7, top-p=1.0).
The resulting datasets, \textbf{WebR-Basic} (from Llama3) and \textbf{WebR-Pro} (from GPT-4o-mini), differ in their generative capabilities and quality.
A comparative analysis of the average token lengths is presented in Appendix \ref{c4-sec: data analysis}, while a detailed cost analysis of WebR is provided in \S \ref{c4-sec: cost}.
Notably, the overall expenditure for calling GPT-4o-mini API is \textbf{\$38.57}.

\section{Experiments}

\subsection{Experimental Setup}

\paragraph{Baselines.}
We compare the family of IT datasets generated by WebR with ten state-of-the-art (SOTA) open-source IT datasets, categorized as follows:
(1) Human-crafted data: \textbf{ShareGPT}~\cite{zheng2023judging} and \textbf{WildChat}~\cite{zhao2024wildchat} are exemplary human-written datasets containing 112K and 652K high-quality multi-round conversations between humans and GPT, respectively.
(2) Semi-automated synthetic data: \textbf{Alpaca}~\cite{alpaca} and \textbf{Evol-Instruct}~\cite{xu2024wizardlm} represent widely-used synthetic datasets generated with semi-automated techniques.
(3) Mixed data: \textbf{Tulu V2 Mix}~\cite{ivison2023camelschangingclimateenhancing} and \textbf{OpenHermes 2.5}~\cite{OpenHermes} are crowd-sourced datasets that aggregate diverse open-source IT datasets, featuring 326K and 1M conversations, respectively.
(4) Fully automated synthetic data: \textbf{Magpie}~\cite{xu2024magpie} synthesizes IT data by prompting Llama3-70B-Instruct with its chat template, from which we sample 100k examples. To ensure a fair and controlled comparison, we reproduce several representative web-based IT synthesis methods—namely \textbf{WebInstruct} \cite{yue2024mammoth2}, \textbf{Backtranslation} \cite{li2024selfalignment}, and \textbf{DoG-Instruct}~\cite{chen-etal-2024-dog}—\textit{using the same source web data} as our proposed WebR. All methods are implemented based on the LLaMA3-70B-Instruct model, thereby aligning model capacity and input sources across approaches.

\paragraph{Models and Training Settings.}
For instruction tuning (IT), we train Llama3-8B-base~\cite{llama3modelcard} and Qwen2.5-1.5/3/7/14B-base~\cite{qwen2.5} on various IT datasets.
For each IT dataset, we fine-tune models with five different random seeds and report the average performance.
We adhere to the official instruction templates provided by each model.
To ensure a fair comparison, we use consistent training hyperparameters across different baseline datasets.
The comprehensive implementation details are listed in Appendix \ref{c4-sec: appendix_implementation}.

\paragraph{Evaluation Benchmarks and Metrics.}
We evaluate the performance of the fine-tuned models using four widely adopted instruction-following benchmarks: AlpacaEval 2~\cite{alpaca_eval}, Arena-Hard~\cite{li2024live}, MT-Bench~\cite{zheng2023judging}, and IFEval~\cite{zhou2023ifeval}.
For AlpacaEval 2, we report the length-controlled win rate (LC), which ensures robustness against verbosity.
For Arena-Hard, we report the win rate (WR) against the baseline model.
For MT-Bench, we provide the average score, using GPT-4-turbo as the evaluation judge.
For IFEval, we report two metrics: prompt-level strict accuracy (\textit{Pr. (S)}) and instruction-level strict accuracy (\textit{Ins. (S)}).
More evaluation details are listed in Appendix \ref{c4-sec: evaluation_details}.

\subsection{Experimental Results}
\paragraph{WebR Outperforms Existing Baselines.}
Table \ref{c4-tab: main_table} highlights the performance of Llama3-8B-base fine-tuned with datasets generated by WebR, compared to those fine-tuned with baseline datasets.
A general trend emerges: IT datasets requiring higher human effort tend to exhibit better performance than those with lower or no human effort.
Nevertheless, our WebR-Basic, which entirely eliminates human effort in dataset creation, significantly and consistently surpasses the SOTA Magpie dataset across all four benchmarks with a \textbf{16.65\%} average improvement.
To ensure a fair and more challenging comparison, we deduplicated and randomly sampled 100k instructions from baseline datasets of varying human effort levels (high, mid, and low) and generated responses using GPT-4o-mini, naming this synthesized strong baseline "\textbf{IT Mix}."
We also generate responses using GPT-4o-mini for Magpie and compare with our proposed method.
Even under the same response generator, WebR-Pro consistently outperforms IT Mix and Magpie by \textbf{7.73\%} and \textbf{12.55\%}, respectively. These results validate that datasets generated by WebR possess superior quality, enabling significantly enhanced instruction-following performance.

\begin{table*}[h]
  \scriptsize
  \centering
  \begin{tabularx}{\textwidth}{l c >{\centering\arraybackslash}X c | >{\centering\arraybackslash}X >{\centering\arraybackslash}X >{\centering\arraybackslash}X c c >{\columncolor{gray!15}} c}
    \toprule
     & & \textbf{Human} & \textbf{Response}  & \textbf{Alpaca} & \textbf{Arena} & \textbf{MT} & \multicolumn{2}{c}{\textbf{IFEval}} & \\
    \multirow{-2}{*}{\textbf{IT Data}} & \multirow{-2}{*}{\textbf{\#Data}} & \textbf{Effort} & \textbf{Generator} &\textbf{Eval 2} &\textbf{Hard} &\textbf{Bench} &\textbf{Pr. (S}) &\textbf{Ins. (S)} & \multirow{-2}{*}{\textbf{Avg.}}\\
    \midrule
        None (w/o fine-tuning) & - & - & - & 0.18 & 0.31	&1.78	&16.26	&18.01	&7.31 \\
        \midrule
        ShareGPT &  112k &High & ChatGPT  & 9.89 & 6.49 & 6.34 & 38.52 & 42.26 & 22.70                      \\
        WildChat &  652k &High & GPT-3.5 \& 4     & 14.62 & 8.73 & 6.60 & 39.53  & 45.66  &23.03   \\
        Tulu V2 Mix & 326k &Mid & Mix  & 9.91 & 5.41 & 5.76  & 37.69 & 41.05 &19.96       \\
        OpenHermes 2.5 & 1M &Mid & Mix   & 12.89 & 8.20 & 6.51  & 38.82 & 43.52 &21.99       \\
        Alpaca &  52k & Low & Davinci-003          & 4.21 & 1.24 & 3.75 & 20.21  & 23.56 &10.59                      \\
        Evol Instruct & 143k &Low & ChatGPT  & 7.19 & 5.58 & 5.77 & 39.00  & 44.25 &20.36                       \\
        WebInstruct & 100k &No & Llama3-70B  & 3.43 & 1.69 & 5.35  & 18.99 & 20.56 & 10.00 \\ 
        Backtranslation & 100k &No & Llama3-70B  & 5.24 & 2.81 & 3.74  & 26.85 & 29.61 & 13.65 \\
        DoG-Instruct & 100k & No & Llama3-70B  & 11.75 & 8.07 & 5.92 & 36.60  & 41.87 & 20.84 \\
        Magpie & 100k &No & Llama3-70B  & 23.62 & 13.98 & 6.26 & 33.83 & 43.07 & 24.15\\
        WebR-Basic & 100k &No & Llama3-70B  & \textbf{25.33} & \textbf{16.50} & \textbf{6.95} & \textbf{41.40} & \textbf{50.69} &\textbf{28.17}\\ \midrule
         IT Mix & 100k &Mid & GPT-4o-mini  & 30.39 & 28.03 & 7.36 & 43.30 & 47.38 &31.29\\
         Magpie & 100k &No & GPT-4o-mini  & 32.61 & 27.97 & 7.26 & 36.81 & 45.07 & 29.95 \\
        WebR-Pro & 100k &No & GPT-4o-mini  & \textbf{34.36} & \textbf{31.10} & \textbf{7.57} & \textbf{43.79} & \textbf{51.76} &\textbf{33.71}\\ \midrule
        (IT + WebR-Pro) Mix & 100k & Mid & GPT-4o-mini & 35.00 & 34.23 & 7.50 & 48.06 & 53.23 &35.60\\
        (IT + WebR-Pro) Merge & 200k & Mid & GPT-4o-mini & \textbf{35.40} & \textbf{35.12} & \textbf{7.59} & \textbf{49.72} & \textbf{53.97} &\textbf{36.36}\\
    \bottomrule
  \end{tabularx}
  \caption{
    Instruction-following performance comparison of various IT data, based on Llama3-8B.
  }
  \label{c4-tab: main_table}
\end{table*}

\paragraph{Compatibility of WebR.}
To explore the potential synergy between WebR and existing datasets, we merged IT Mix and WebR-Pro using two strategies: (1) random sampling of 50k data points from each dataset and (2) direct concatenation.
As shown in Table \ref{c4-tab: main_table}, both merged datasets deliver further performance improvements over their individual components, establishing new SOTA results.
This can be attributed to the complementary strengths of the datasets: IT Mix offers broader data coverage, while WebR-Pro provides higher quality and more challenging instructions, as evidenced in Figure \ref{c4-fig: quality_difficulty}.

\paragraph{Performance on Downstream Benchmarks.}
We evaluate the impact of various instruction-tuning datasets on downstream task performance across multiple domains\footnote{Evaluation settings are aligned with \url{https://opencompass.org.cn}.}:
(1) \textbf{Knowledge}: MMLU~\cite{dan2021mmlu};
(2) \textbf{Reasoning}: ARC~\cite{clark2018think} and WinoGrande~\cite{sakaguchi2019winogrande};
(3) \textbf{Math}: MATH~\cite{hendrycksmath2021} and GSM8K~\cite{cobbe2021gsm8k};
(4) \textbf{Code}: HumanEval~\cite{chen2021codex}.
As shown in Table~\ref{c4-tab: capabilities}, models fine-tuned on the WebR datasets outperform those trained on other baselines, demonstrating their effectiveness in improving generalization across diverse downstream tasks, especially in challenging benchmarks like ARC and WinoGrande.
Furthermore, the combination of WebR-Pro and IT Mix further validates the complementary strengths of WebR data in aligning models with complex task requirements.

\begin{table*}[h]
  \footnotesize
  \centering
  \begin{tabularx}{\textwidth}{l|>{\centering\arraybackslash}X|>{\centering\arraybackslash}X c | >{\centering\arraybackslash}X >{\centering\arraybackslash}X | c | >{\columncolor{gray!15}}>{\centering\arraybackslash}X}
    \toprule
     \textbf{IT Data} & \textbf{MMLU}  & \textbf{ARC} & \textbf{WinoGrande} & \textbf{MATH} & \textbf{GSM8K} & \textbf{HumanEval} & \textbf{Avg.} \\
    \midrule
    None (w/o fine-tuning) & 60.56	&73.52	&52.14	&19.62	&56.16	&39.08	&50.18 \\
    \midrule
    WildChat & 58.46 & 72.62 & 49.43 & 19.34 & 60.25 & 42.55 & 50.44 \\
    OpenHermes 2.5 & 60.08 & 75.65 & 51.22 & 24.18 & 64.70 & 44.43 & 53.38 \\
    Magpie & 58.58 & 71.53 & 51.93 & 16.12 & 57.39 & 40.85 & 49.40 \\
    WebR-Basic & 60.85 & 76.27 & 52.91 & 20.28 & 55.57 & 40.10 & 51.00 \\
    IT Mix & 57.44 & 73.56 & 50.36 & 22.00 & 61.87 & 45.12 & 51.73 \\
    WebR-Pro & \textbf{61.15} & 74.92 & \textbf{53.20} & 24.94 & 60.69 & 48.73 & 53.94 \\
    (IT + WebR-Pro) Mix & 60.69 & \textbf{77.63} & 50.67 & 26.34 & 64.90 & \textbf{50.61} & 55.14 \\
    (IT + WebR-Pro) Merge & 61.02 & 76.27 & 52.72 & \textbf{28.36} & \textbf{66.41} & \textbf{50.61} & \textbf{55.90} \\
    \bottomrule
  \end{tabularx}
  \caption{
    Performance comparison of downstream tasks (Knowledge, Reasoning, Math, Code) based on Llama3-8B.
  }
  \label{c4-tab: capabilities}
\end{table*}

\subsection{Ablation Study}
Table \ref{c4-tab: ablation} compares the LLM performance using different settings to construct WebR-Pro.

\begin{itemize}[noitemsep, topsep=0pt]
\item \textbf{w/o Persona}: removing the author's persona information during instruction generation leads to performance declines across almost all benchmarks.
\item \textbf{w/o Part}: creating instructions solely from the entire web content, rather than using specific parts, causes notable performance degradation, particularly on IFEval and reasoning-intensive tasks like ARC and MATH.
\item \textbf{w/o Refinement}: skipping the refinement step for \textit{Web as Response}—by directly adopting the rollout response as the final output—results in a substantial drop in instruction-following performance.
\item \textbf{w/o MinHash}: eliminating MinHash-based deduplication decreases performance across all benchmarks, highlighting the importance of maintaining dataset diversity.
\item \textbf{Ratio of \textit{Web as Instruction} to \textit{Web as Response}}: varying the ratio of \textit{Web as Instruction} to \textit{Web as Response} data synthesis reveals that \textbf{each component contributes uniquely to model capabilities}.
Specifically, \textit{Web as Instruction} enhances reasoning and understanding tasks (e.g., ARC and MATH), while \textit{Web as Response} primarily improves instruction-following and question-answering tasks (e.g., IFEval and AlpacaEval 2).
The optimal balance is achieved at a ratio of 2:1, which delivers the best overall performance.
\end{itemize}

\begin{table*}[h]
  \footnotesize
  \centering
  \begin{tabularx}{\textwidth}{l|>{\centering\arraybackslash}X >{\centering\arraybackslash}X >{\centering\arraybackslash}X >{\columncolor{gray!15}}>{\centering\arraybackslash}X | >{\centering\arraybackslash}X >{\centering\arraybackslash}X >{\centering\arraybackslash}X c >{\columncolor{gray!15}}>{\centering\arraybackslash}X}
    \toprule
     & \textbf{Alpaca} & \textbf{MT} & \textbf{IFEval} & & & & & & \\
     \multirow{-2}{*}{\textbf{Setting}} &\textbf{Eval 2} &\textbf{Bench} &\textbf{Pr. (S}) & \multirow{-2}{*}{\textbf{Avg.}} & \multirow{-2}{*}{\textbf{MMLU}} & \multirow{-2}{*}{\textbf{ARC}} & \multirow{-2}{*}{\textbf{MATH}} & \multirow{-2}{*}{\textbf{HumanEval}} & \multirow{-2}{*}{\textbf{Avg.}} \\
    \midrule
    WebR-Pro & 34.17 & 7.50 & 43.55 & \textbf{28.41} & 61.15 & 74.92 & 24.94 & 48.73 & \textbf{52.43} \\ \midrule
    -w/o Persona & 33.30 & 6.93 & 44.69 & 28.31 & 60.98 & 74.58 & 24.03 & 48.50 & 52.02  \\
    -w/o Part & 33.89 & 7.53 & 42.60 &28.01 & 61.05 & 72.53 & 22.73 & 48.41 & 51.18   \\
    -w/o Refinement & 31.61 & 7.42 & 44.73 &27.92 & 59.83 & 74.92 & 24.36 & 48.61 & 51.93 \\
    -w/o MinHash & 32.43 & 7.29 & 43.02 &27.58& 60.69 & 74.92 & 24.82 & 47.15 & 51.90 \\ \midrule
    \multicolumn{9}{c}{\textit{Ratio of Web as Instruction to Web as Response (2 : 1 in WebR)}} \\ 
    1 : 0 & 29.15 & 7.10 & 39.56 &25.27 & 58.79 & 74.58 & 25.74 & 50.00 & 52.28 \\
    1 : 1 & 33.16 & 7.39 & 43.26 &27.94 & 60.60 & 73.22 & 25.18 & 48.78 & 51.95 \\
    1 : 2 & 32.99 & 7.33 & 42.85 &27.72 & 57.76 & 72.61 & 25.26 & 50.00 & 51.41 \\
    0 : 1 & 33.41 & 6.68 & 42.54 &27.54 & 52.68 & 72.90 & 23.30 & 46.95 & 48.96 \\
    \bottomrule
  \end{tabularx}
  \caption{
    Ablation study based on Llama3-8B.
  }
  \label{c4-tab: ablation}
\end{table*}

\section{Analysis}

\subsection{Dataset Analysis of WebR}

\paragraph{Diversity.}
We utilize a quantitative measure of diversity: (1) We randomly sample $N=10,000$ instructions from each dataset and encode them using the \texttt{all-mpnet-base-v2}\footnote{\url{https://huggingface.co/sentence-transformers/all-mpnet-base-v2}} embedding model; (2) We compute the average cosine similarity between all embedding pairs and define embedding diversity as $1- \frac{1}{C(N, 2)} \sum_{\forall i < j} \cos(\mathbf{e}_i, \mathbf{e}_j)$, where higher values indicate greater diversity.
Our results in Table \ref{c4-tab: it_data_comparison} demonstrate that WebR-Pro achieves the highest diversity score (0.93), matching that of WildChat, which involves high human effort. Notably, WebR-Pro surpasses all other datasets—including those requiring human annotation like OpenHermes (0.87) and Evol Instruct (0.88)—indicating its strong capability to generate diverse instructions automatically. Furthermore, it outperforms previous automatic baselines such as WebInstruct (0.84) and Magpie (0.92), highlighting its effectiveness in promoting diversity without human intervention.

\begin{table}[!t]
\centering
\small
\begin{tabular}{lccc}
\toprule
\textbf{IT Data} & \textbf{Human Effort} & \textbf{Avg. Score} & \textbf{Diversity} \\
\midrule
WildChat & high & 23.03 & \textbf{0.93} \\
OpenHermes & mid & 21.99 & 0.87 \\
Evol Instruct & low & 20.36 & 0.88 \\
WebInstruct & no & 9.79 & 0.84 \\
Magpie & no & 24.15 & 0.92 \\
WebR-Basic & no & 28.17 & 0.91 \\
WebR-Pro & no & \textbf{33.58} & \textbf{0.93} \\
\bottomrule
\end{tabular}
\caption{Comparison of embedding diversity.}
\label{c4-tab: it_data_comparison}
\end{table}

\begin{figure}[!t]
\centering
\includegraphics[width=0.6\linewidth]{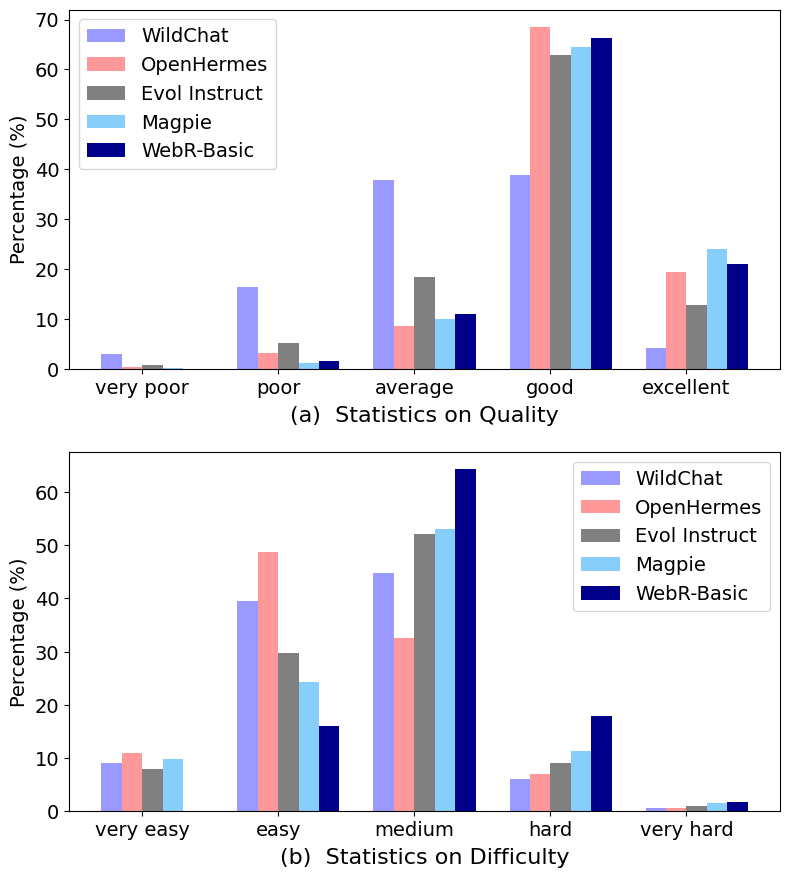}
\caption{
Statistics of instruction quality and difficulty. 
}
\label{c4-fig: quality_difficulty}
\end{figure}

\paragraph{Quality and Difficulty.}
Following Magpie~\cite{xu2024magpie}, we use the Qwen2.5-72B-Instruct model to evaluate the quality and difficulty of each instruction, categorizing them into five levels. 
As depicted in Figure \ref{c4-fig: quality_difficulty}, synthetic data generally demonstrates higher quality and greater difficulty compared to human-crafted instructions.
In particular, WebR-Basic exhibits superior distributions in both quality and difficulty metrics, surpassing existing baselines in these aspects.

\subsection{Cost Analysis of WebR}
\label{c4-sec: cost}

Here we analyze the cost-effectiveness of our proposed Web Reconstruction framework.
For context, we estimated the budget for data synthesis using the \texttt{GPT-4o-mini} API, based on the Batch API's pricing of \$0.075 per 1M input tokens and \$0.3 per 1M output tokens.
Table \ref{c4-tab: api_cost} lists the breakdown of the estimated costs for each step, which demonstrates that the overall expenditure (\textbf{\$38.57}) is both reasonable and manageable.

\begin{table*}[h]
\centering
\footnotesize
\begin{tabularx}{\textwidth}{l >{\centering\arraybackslash}X >{\centering\arraybackslash}X >{\centering\arraybackslash}X >{\centering\arraybackslash}X}
\toprule
&  & \textbf{Avg. Input} & \textbf{Avg. Output} &  \\ 
& \multirow{-2}{*}{\textbf{\# of Samples}} & \textbf{Token Length} & \textbf{Token Length} & \multirow{-2}{*}{\textbf{Cost (\$)}} \\
\midrule
Generate author's persona & 100,000 & 523 & 32 & 4.88 \\
Web as Instruction (instruction) & 66,667 & 711 & 123 & 6.02 \\
Web as Instruction  (rollout response) & 66,667 & 611 & 392 & 10.90 \\
Web as Response (instruction) & 33,333 & 645 & 91 & 2.52 \\
Web as Response (rollout response) & 33,333 & 91 & 522 & 5.45 \\
Web as Response (refined response) & 33,333 & 1,155 & 591 & 8.80 \\ \midrule
Total & - & - & - & 38.57 \\
\bottomrule
\end{tabularx}
\caption{Estimated budget for data synthesis using the \texttt{GPT-4o-mini} API.}
\label{c4-tab: api_cost}
\end{table*}

Additionally, our main experiment in Table \ref{c4-tab: main_table} demonstrates that the open-source \texttt{Llama3-70B-Instruct} model can achieve satisfactory performance for our proposed Web Reconstruction, significantly outperforming previous SFT datasets. 
Notably, it can be deployed on only 2 NVIDIA-3090 GPUs, with the option to further reduce hardware requirements through low-bit quantization\footnote{\url{https://github.com/ollama/ollama}}.
This provides an economical alternative for our proposed WebR.

\subsection{Data Efficiency of WebR}
\begin{figure}[h]
\centering
\includegraphics[width=0.7\linewidth]{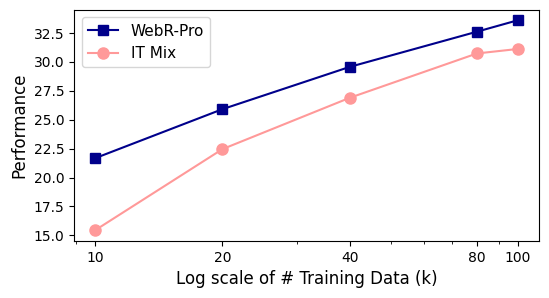}
\caption{
The impact of training data scale on the average instruction-following performance.
}
\label{c4-fig: data_scale}
\end{figure}

Figure \ref{c4-fig: data_scale} illustrates the impact of training data scale on model performance.
The results clearly underscore the superior \textbf{data efficiency} of WebR-Pro compared to IT Mix:
(1) With only 10k training samples, WebR-Pro achieves a striking \textbf{40.26\%} performance improvement over IT Mix, highlighting its exceptional capability to elicit latent potential from LLMs even with limited data.
(2) WebR-Pro exhibits a more consistent and pronounced linear performance increase with respect to the logarithmic growth in training data, consistently outperforming IT Mix across all data scales.
These results strongly validate the efficacy of WebR in efficiently leveraging training data to unlock and enhance the capabilities of LLMs.

\subsection{Scalability of WebR}
Table \ref{c4-tab: llm_size} highlights the impact of base LLM scale on the performance of our proposed WebR method.
While WebR-Pro slightly underperforms IT Mix at the 1.5B model scale, its advantages become increasingly pronounced as the model size grows.
For instance, WebR-Pro achieves an average performance improvement of \textbf{2.86\%} over IT Mix with Qwen2.5-7B and an even more substantial improvement of \textbf{5.55\%} with Qwen2.5-14B.
These results suggest that the advanced synthesis paradigm of WebR better aligns with larger models' capacity to capture complex patterns and utilize reasoning-intensive data.
In contrast, smaller models with limited capacity may struggle to fully exploit WebR’s potential.

\begin{table*}[h]
  \footnotesize
  \centering
  \begin{tabularx}{\textwidth}{l l >{\centering\arraybackslash}X >{\centering\arraybackslash}X >{\centering\arraybackslash}X >{\centering\arraybackslash}X >{\centering\arraybackslash}X}
    \toprule
     \textbf{Base LLM} & \textbf{IT Data} & \textbf{AlpacaEval 2} & \textbf{Arena-Hard} & \textbf{MT-Bench} & \textbf{IFEval/Pr. (S)} & \textbf{IFEval/Ins. (S)} \\
    \midrule
    \multirow{2}{*}{Qwen2.5-1.5B}
         & IT Mix &10.98  &\textbf{15.10}  & \textbf{6.03} &\textbf{29.57}  &\textbf{33.27}  \\
        & WebR-Pro  &\textbf{11.00} (\textcolor{customgreen}{+0.02})  & 14.03 (\textcolor{customred}{-1.07})  & 5.92 (\textcolor{customred}{-0.11}) & \textbf{29.57} (\textcolor{customgreen}{+0.00})  & 32.16 (\textcolor{customred}{-1.11}) \\
    \midrule
    \multirow{2}{*}{Qwen2.5-3B}
         & IT Mix &\textbf{22.36}  &26.54  & 6.95  &\textbf{43.07}  &\textbf{44.73}  \\
        & WebR-Pro  & 22.29 (\textcolor{customred}{-0.07})  &\textbf{28.13} (\textcolor{customgreen}{+1.59})  & \textbf{7.03} (\textcolor{customgreen}{+0.08})  &42.38 (\textcolor{customred}{-0.69})  &44.71 (\textcolor{customred}{-0.02}) \\
    \midrule
    \multirow{2}{*}{Qwen2.5-7B}
         & IT Mix &32.59  &45.10  &7.45  &49.35  &52.68  \\
        & WebR-Pro  &\textbf{34.90} (\textcolor{customgreen}{+2.31})  &\textbf{45.66} (\textcolor{customgreen}{+0.56})  &\textbf{7.62} (\textcolor{customgreen}{+0.17})  &\textbf{50.55} (\textcolor{customgreen}{+1.20})  &\textbf{53.35} (\textcolor{customgreen}{+0.67}) \\
    \midrule
    \multirow{2}{*}{Qwen2.5-14B}
         & IT Mix &42.07  &59.00  &8.10  &58.04  &60.63  \\
        & WebR-Pro  &\textbf{46.19} (\textcolor{customgreen}{+4.12})  &\textbf{62.13} (\textcolor{customgreen}{+2.13})  &\textbf{8.39} (\textcolor{customgreen}{+0.29})  &\textbf{60.23} (\textcolor{customgreen}{+2.19})  &\textbf{64.88} (\textcolor{customgreen}{+4.25}) \\
    \bottomrule
  \end{tabularx}
  \caption{
    Performance comparison across varied scales of base LLMs.
  }
  \label{c4-tab: llm_size}
\end{table*}

\begin{table*}[h]
  \footnotesize
  \centering
  \begin{tabularx}{\textwidth}{lc >{\centering\arraybackslash}X c >{\centering\arraybackslash}X >{\centering\arraybackslash}X >{\columncolor{gray!15}}>{\centering\arraybackslash}X}
    \toprule
    \textbf{Data Proportion} & \textbf{AlpacaEval 2}  & \textbf{MATH} & \textbf{HumanEval} & \textbf{MedQA} & \textbf{FinBen} & \textbf{Avg.} \\
    \midrule
    IT Mix & 30.19 & 22.00 & 45.12 & 38.88 & 29.20 & 33.08 \\ \midrule
    WebR-Pro (4.7 gen : 1 math : 1 code) & 34.17 & 24.94 & 48.73 & 47.31 & 29.56 & 36.94 \\
    - 1 gen & 34.40 & 22.52 & 44.78 & 44.94 & 28.97 & 35.12 \\
    - 1 gen : 1 math & 34.25 & \textbf{\textcolor{customgreen}{28.09}} & 48.23 & 46.59 & 29.77 & 37.39 \\
    - 1 gen : 1 math : 1 code & 34.59 & 27.10 & \textbf{\textcolor{customgreen}{51.39}} & 46.83 & 29.34 & \textbf{37.85} \\
    - 1 gen : 1 math : 1 code : 1 med &32.75 & 26.22 & 49.68 & \textbf{\textcolor{customgreen}{49.98}} & 29.01 & 37.53 \\
    - 1 gen : 1 math : 1 code : 1 med : 1 fin &33.03 & 25.38 & 48.17 & 45.64 & \textbf{\textcolor{customgreen}{30.22}} & 36.49 \\
    \bottomrule
  \end{tabularx}
  \caption{
    Domain adaptation based on Llama3-8B, with the domain improvements marked in \textcolor{customgreen}{\textbf{green}}.
  }
  \label{c4-tab: domain}
\end{table*}

\subsection{Domain Adaptability of WebR}
We explore the potential of our proposed WebR framework for domain adaptation by simply adjusting the proportion of source web documents.
Starting with general-domain content, we progressively add domain-specific materials from math, code, medicine, and finance, assessing performance across relevant benchmarks.
For the medical and financial domains, we utilize raw web documents from IndustryCorpus2~\cite{beijing_academy_of_artificial_intelligence}, and evaluate using MedQA~\cite{jin2021medqa} and FinBen~\cite{xie2024finben} benchmarks.
As shown in Table~\ref{c4-tab: domain}, WebR demonstrates strong adaptability across domains. Compared to the IT Mix baseline, incorporating domain-specific data consistently improves performance, with math and code data yielding significant gains in MATH (28.09) and HumanEval (51.39), and medical and financial domains showing strong results on MedQA (49.98) and FinBen (30.22). These results highlight WebR’s ability to \textbf{incorporate specialized knowledge while maintaining competitive general-domain performance}. Furthermore, the process of collecting domain-specific web documents is straightforward, underscoring WebR’s practicality.

\section{Conclusion and Discussion}
\subsection{Conclusion}
In this chapter, we present \textbf{Web Reconstruction} (WebR), a fully automated framework for synthesizing high-quality instruction-tuning (IT) datasets.
Harnessing the richness of raw web content, we conceptualize \textit{web reconstruction} as an instruction-tuning data synthesis task via a novel dual-perspective paradigm—\textit{Web as Instruction} and \textit{Web as Response}—where each web document is designated as either the input or output role to trigger the reconstruction process.
Extensive experiments show that WebR-generated datasets consistently outperform state-of-the-art baselines across four instruction-following benchmarks and six diverse downstream tasks.
\subsection{Discussion}
While WebR can already obtain satisfactory performance, there are several areas for improvement and future exploration.
Firstly, the current implementation of WebR focuses on single-turn data synthesis. Expanding this framework to support multi-turn conversations could further enhance its applicability to complex, interactive tasks.
Second, due to constraints in time and computational resources, the size of the constructed WebR-Basic and WebR-Pro datasets is currently limited to 100k samples.
However, given the vast availability of web documents—numbering in the trillions—the WebR framework has significant potential for scaling to create large-scale IT datasets, which could further boost performance.
Finally, WebR does not incorporate advanced data selection techniques, such as Instruction Following Difficulty (IFD)~\cite{li-etal-2024-quantity}, as part of its post-processing pipeline.
Incorporating such techniques in future work could refine data quality and further enhance the capabilities of LLMs.
\chapter{Aligning Large Language Models with Knowledge Editing}\label{chap:knowledge_editing}

\section{Introduction}
The transformative potential of LLMs~\cite{brown2020language, 2023gpt4, touvron2023llama} has been unequivocally underscored by their unparalleled efficacy across a myriad of applications~\cite{chen2021codex, openai2022chatgpt, 2023gpt4}.
Nonetheless, the dynamic nature of the world necessitates frequent updates to LLMs to rectify outdated information or integrate new knowledge, thereby safeguarding their sustained pertinence.
Naively training a new LLM from scratch to incorporate updated knowledge could result in substantial computational overhead and is frequently deemed impractical.
To this end, the concept of \textbf{knowledge editing} has been introduced~\cite{DBLP:conf/iclr/SinitsinPPPB20, cao2021ke}, aiming to efficiently modify LLMs' outputs towards targeted queries while preserving overall performance across other unrelated ones.
For example, updating the knowledge of ``\texttt{The current British Prime Minister is Rishi Sunak}'' not only modifies the response to ``\texttt{Who is married to the PM of the UK?}'' but leaves unaffected the answer to ``\texttt{When was Rishi Sunak born?}''

\begin{figure}[!t]
\centering
\includegraphics[width=0.8\linewidth]{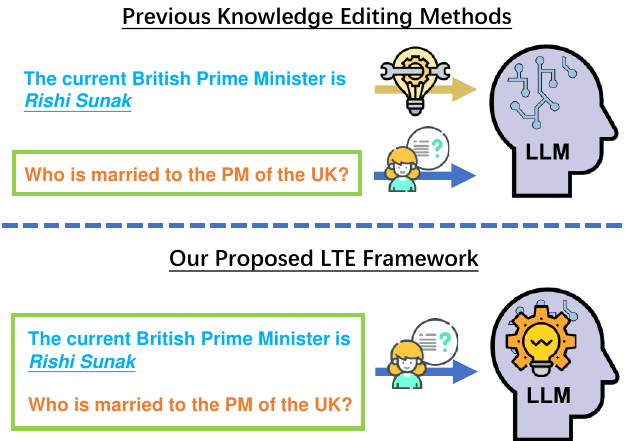}
\caption{
Previous knowledge editing methods primarily rely on first memorizing updated knowledge and then answering queries, while our proposed LTE framework teaches LLMs to dynamically \textbf{apply} updated knowledge to answer queries.
}
\label{c5-fig: intro}
\end{figure}

Some knowledge editing approaches rely on auxiliary modules or models to either predict the LLM's weight adjustments~\cite{cao2021ke, mitchell2022mend} or function as scope classifiers for query response applicability~\cite{mitchell2022serac}. 
While these innovations demonstrate potential, they fail to inherit the advanced capabilities of LLMs, thus rendering output quality degeneration.
Others attempt to identify and modify parameters related to specific knowledge within LLMs to update their embedded knowledge~\cite{dai2022kn, meng2022rome, meng2023memit}.
Nonetheless, the correlation between localization and editing efficacy has been scrutinized by~\cite{hase2023does}, which suggests that localization results from Causal Tracing are statistically uncorrelated with the success of an edit injecting a new fact into MLP weights.
Thus, it is plausible that the detrimental effects of such approaches could be amplified with the scale of LLMs. 
In essence, these methods predominantly rely on memorizing the updated knowledge (See Figure \ref{c5-fig: intro}), hindering LLMs from effectively combining the new knowledge with their inherent knowledge when answering the input queries.

To address these issues, motivated by the proverb ``\textit{Teach a man to fish, and you feed him for a lifetime},'' we propose to elicit LLMs' capabilities of following knowledge editing instructions, thereby empowering them to effectively \textbf{leverage} the updated knowledge to answer the queries.
Specifically, we propose a \textit{Learning to Edit} (LTE) framework to align LLMs with knowledge editing by leveraging supervised fine-tuning (SFT), which has become foundational in tailoring LLMs for desired behaviors~\cite{wei2022finetuned, mishra-etal-2022-cross}.
The LTE framework is structured around two pivotal stages: the Alignment Phase and the Inference Phase.
During the Alignment Phase, we pair edit descriptors with in-scope and out-of-scope queries to create \textbf{parallel} datasets, processed with and without a tailored prompt that explicitly informs LLMs of the knowledge editing process.
By fine-tuning LLMs on this meticulously constructed dataset,  we aim to cultivate a trio of essential capabilities within LLMs—\textit{In-Scope Capability} (generating reliable, logically consistent edits), \textit{Out-of-Scope Capability} (preserving the integrity of unrelated content), and \textit{Linguistic Capability} (maintaining linguistic proficiency)—to ensure nuanced application of updated knowledge. Note that this process is \textbf{once and for all}, laying the groundwork for the inference phase to apply these capabilities dynamically.
In the Inference Phase, to extend to mass editing, we implement a retrieval-based mechanism to obtain the most pertinent updated knowledge from a memory bank.
Such an approach enables LLMs to adapt their responses with the most current information in real time, thereby streamlining both batch and sequential knowledge editing processes.

In this chapter, we assess our proposed LTE method against seven advanced baselines across four benchmarks in single, batch, and sequential editing scenarios.
Our findings reveal four major strengths of the LTE method:
(1) it establishes a new state-of-the-art (SOTA) in overall knowledge editing performance, surpassing existing methods by a substantial margin of over \textbf{20} absolute points in terms of portability;
(2) the robustness of LTE is evident in its ability to handle batch and sequential knowledge editing requests, showing a markedly reduced rate of performance deterioration compared to its counterparts;
(3) it is proficient in facilitating knowledge edits with minimal interference to the model’s cognitive functions across varied unrelated domains.
(4) LTE distinguishes itself by combining the fastest editing speeds with exceptional performance.

\section{Task Formulation}
\label{c5-sec: task_formulation}
The objective of knowledge editing is to efficiently adjust the behavior of an initial base LLM $f_{\theta}$, where $\theta$ represents the model's parameters, in response to specific \textit{edit descriptors} $\{(x_i^*, y_i^*)\}_{i \in [1, N]}$. 
In this context, $x_i^*$ refers to the edit input that triggers the knowledge in LLMs (e.g., \texttt{The current British Prime Minister is}), $y_i^*$ is the corresponding edit target (e.g., \texttt{Rishi Sunak}), and $N$ signifies the total number of edit descriptors.
The efficacy of knowledge editing is evaluated among four dimensions:

\paragraph{Edit Success} measures the average accuracy of the post-edit model $f_{\theta}^*$ on these edit cases:
\begin{equation}
\mathop{\mathbb{E}}_{(x_i^*,y_i^*)}\mathds{1}\{\mathop{\arg\max}_{y} f_{\theta}^*(y|x_i^*)=y_i^*\}
\end{equation}

\paragraph{Portability} evaluates how well updated knowledge transfers to related queries, enhancing the model's utility in varied contexts.
For example, correctly answering \texttt{Who is married to the British Prime Minister?} with \texttt{Akshata Murty} post-edit indicates successful knowledge transfer.

\paragraph{Locality} assesses the precision of edits, ensuring modifications are confined to targeted areas without affecting unrelated knowledge.
For example, ensuring \texttt{The current British Chancellor} remains \texttt{Jeremy Hunt} exemplifies effective locality.

\paragraph{Fluency} quantifies the linguistic quality of the model's output post-edit, focusing on coherence and diversity to avoid repetitive patterns.
Following~\cite{zhang2018generating}, we calculate fluency by measuring the weighted average of bi- and tri-gram entropies given by $-\sum_k f(k)\log_2f(k)$, where $f(\cdot)$ is the $n$-gram frequency distribution.

\section{Methodology}
As illustrated in Figure \ref{c5-fig: method}, we propose a \textit{Learning to Edit} (LTE) framework to align LLMs with ever-changing, complicated, and diverse knowledge editing requests in real-time.
This framework consists of two phases: (1) in the Alignment Phase, we enlighten LLMs' capabilities of applying updated knowledge through the utilization of a knowledge editing prompt ``\texttt{[Updated Information] \{edit descriptor\}\textbackslash n[Query] \{query\}}'';
(2) in the Inference Phase, LLMs are enabled to conduct on-the-fly and streaming knowledge editing by retrieving relevant updated knowledge to the query from the stored memory.

\subsection{Alignment Phase: Learning to Edit}
\label{c5-sec: alignment}

In light of the task formulation in \S \ref{c5-sec: task_formulation}, the model editing process profoundly influences predictions across a wide array of inputs directly related to the provided edited knowledge. 
An optimal knowledge editing method must seamlessly integrate new knowledge into the relevant content within its edit scope, while ensuring the accuracy and integrity of information outside this domain.
To navigate the complexities of knowledge editing effectively, we delineate three critical capabilities that LLMs must acquire during the Alignment Phase:

\begin{figure*}[!t]
\centering
\includegraphics[width=\linewidth]{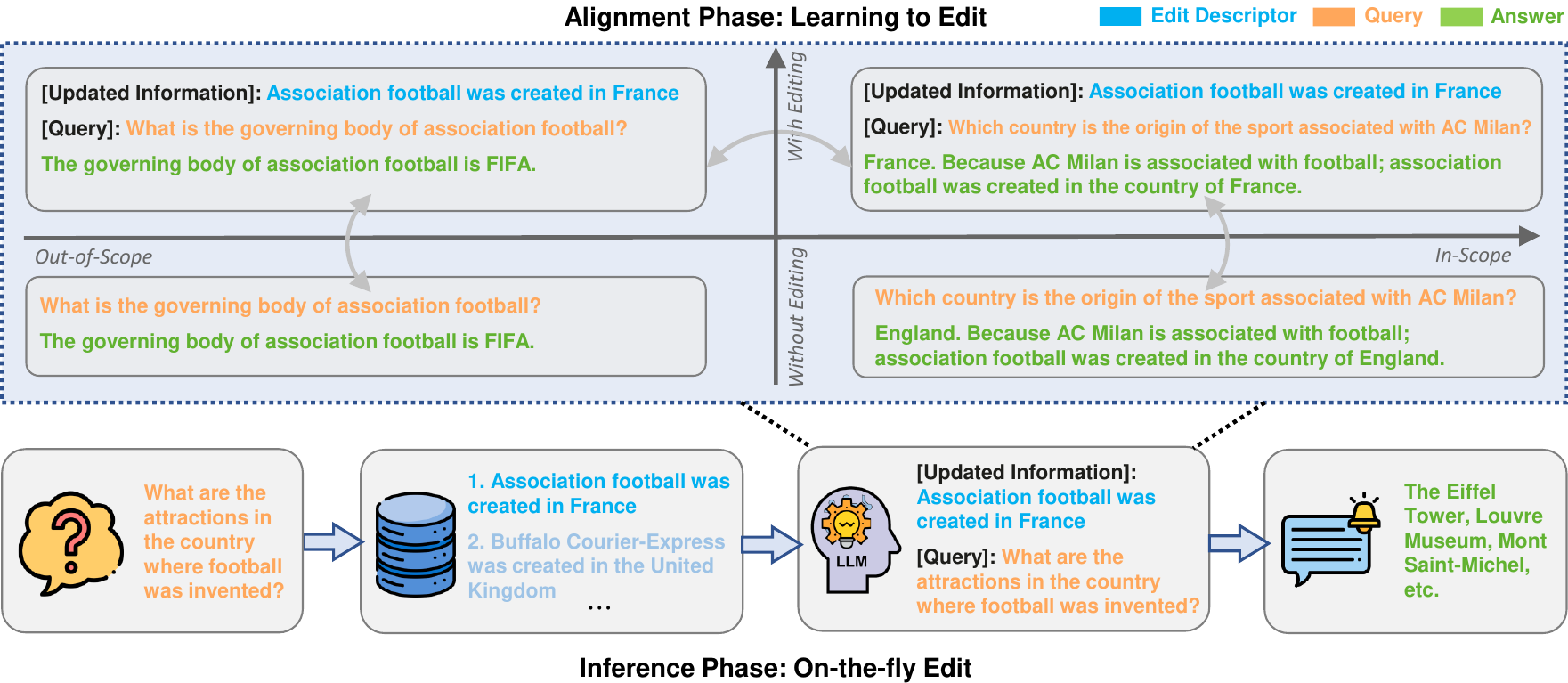}
\caption{
The proposed \textit{Learning to Edit} (LTE) framework.
In the Alignment Phase, we train LLMs how to \textbf{apply} updated knowledge—beyond mere memorization—by fine-tuning them on our meticulously curated parallel (indicated by gray arrows) data.
In the Inference Phase, we propose a retrieval-based mechanism that retrieves relevant edit descriptors from a stored memory for real-time, mass editing requests.
}
\label{c5-fig: method}
\end{figure*}

\paragraph{In-Scope Capability}
requires the model to correctly generate the edit target given the edit input or its paraphrases.
It also covers subject aliasing, ensuring the editing of one subject should not vary from its expression.
For example, after modifying the origin city of \texttt{Association football}, the origin city of \texttt{Soccer} should also be modified.
Furthermore, it necessitates LLMs to conduct compositional reasoning with the changed facts (e.g., when we change the origin city of \texttt{Association football}, the origin city of \texttt{the sport associated with AC Milan} should also be changed, see Figure \ref{c5-fig: method}).

To empower LLMs with these advanced capabilities during alignment, we meticulously curate training data by adapting or synthesizing content from existing knowledge editing datasets.
Our selection includes ZsRE~\cite{levy2017zsre}, RIPPLEEDITS~\cite{cohen2024ice}, WikiBio~\cite{hartvigsen2023grace}, and MQUAKE~\cite{zhong2023mquake}, with each dataset providing edit descriptors linked to multiple queries. 
These queries are specifically designed to evaluate the nuanced facets of in-scope or out-of-scope knowledge editing capabilities. 
To avoid data leakage, our methodology only incorporates samples from the datasets' training sets.

\paragraph{Out-of-Scope Capability} directs the model to maintain the integrity of unrelated attributes of the subject, ensuring no unintended alterations.
For example, as shown in Figure \ref{c5-fig: method}, changing the origin city of \texttt{Association football} should not modify its governing body. 
Additionally, it requires LLMs to adeptly handle one-to-many relationships, ensuring that original connections are retained unless specifically altered.
We utilize the same data sources as that of In-Scope Capability.
However, due to the absence of out-of-scope instances in datasets like ZsRE and MQUAKE, we employ GPT-4 to generate corresponding queries and answers based on the edit descriptors, further details of which are provided in Appendix \ref{c5-sec: appendix_out_scope}.

\paragraph{Linguistic Capability} requires that incorporating edits related to specific factual knowledge should not hinder the model’s proficiency in unrelated areas, such as generative fluency, commonsense reasoning, general intelligence, and world knowledge.
Thus, we identify a limitation within existing datasets: the predominance of fill-in-the-blank cloze queries may not adequately challenge the LLMs' linguistic capabilities across diverse areas, such as conversational contexts, where answers may inherently be more elaborate. To address this, we integrate edit descriptors from COUNTERFACT~\cite{meng2022rome} and utilize GPT-4 to generate free-text, in-scope query-answer pairs (See  Appendix \ref{c5-sec: appendix_free_text}). This approach not only diversifies the training data but also enhances the models' ability to generate more contextually rich answers. GPT-4 is further employed to verify the relevance of generated answers to the edit descriptors, with a mechanism to filter out unsatisfactory cases. Additionally, we incorporate natural language instructions from Evol-Instruct~\cite{xu2024wizardlm} as out-of-scope queries to maintain the LLMs' broad linguistic capabilities.

\paragraph{Parallel Data Construction.}
Our approach involves the creation of parallel datasets by pairing each edit descriptor with corresponding in-scope and out-of-scope queries. 
These are then processed with and without the incorporation of our tailored knowledge editing prompt (See Figure \ref{c5-fig: method}).
This parallel construction serves multiple purposes.
First, it reinforces LLM's capacity to discern when to utilize updated knowledge by comparing in-scope and out-of-scope queries with editing.
Second, it accentuates the subtle distinctions between with and without editing for in-scope queries, enabling LLM to apply knowledge edits more effectively.
Lastly, it educates LLM on maintaining the integrity of out-of-scope information by presenting it with comparisons that demonstrate when not to alter this knowledge. 
In total, we construct 60k parallel data for training, the detailed data statistics are listed in Appendix \ref{c5-sec: appendix_statistics}.
During training, we compute the loss \textit{only} on the answer tokens, i.e., it learns to generate answers conditioned on the Updated Information and Query.

\subsection{Inference Phase:  On-the-fly Edit}
\label{c5-sec: inference}
Here we propose an efficient mechanism that extends LTE to batch and streaming knowledge editing scenarios.
Inspired by retrieval-augmented generation (RAG)~\cite{lewis2020rag, xu2022rag}, we utilize an off-the-shelf retrieval model \texttt{multi-qa-mpnet-base-dot-v1}~\cite{reimers2019sentencebert} to embed all the edit descriptors and create a vector memory to store the representations.
When given a query, we also get the representation of the query by the retriever and search the top-$k$ ($k=3$ in our experiments) similar edit descriptors from the vector memory.
Then, the query and the retrieved edit descriptors are fed into the LLM to obtain the answer.
To enhance the fault tolerance of the retrieval model while maintaining the single editing performance, we adopt a \textit{threefold strategy} for incorporating different numbers of edit descriptors as Updated Information in the Alignment Phase.
Firstly, in 50\% of cases, we directly use the exact edit descriptor.
Secondly, for 25\% of cases, we employ the \texttt{multi-qa-mpnet-base-dot-v1} model to identify the top-1 semantically similar edit descriptor (excluding the exact one) from the whole dataset, and use both as the Updated Information.
Lastly, for the remaining 25\%, we retrieve the top 2 semantically similar descriptors, excluding the exact one, using all three as the Updated Information.
This approach introduces variability during training, significantly enhancing the model's robustness and improving mass edit capabilities in inference.

\section{Experiments}
\label{c5-sec: experiment}

\subsection{Experimental Setup}
We select LLaMA2-Chat-7B~\cite{touvron2023llama} and Qwen-Chat-7B~\cite{qwen} as base models for knowledge editing, as these models are widely used for English and Chinese chatbot applications, respectively.
We implement our LTE method by standard fine-tuning on the 60k constructed data in \S \ref{c5-sec: alignment}.
Additionally, we explore an alternative implementation of LTE, employing Low-Rank Adaptation (LoRA)~\cite{hu2022lora}, noted for its efficiency and reduced memory requirements.
This variant is referred to as LTE-LoRA.
The detailed implementation specifics are listed in Appendix \ref{c5-sec: appendix_implementation}.

For the evaluation datasets and metrics, we follow KnowEdit~\cite{zhang2024knowedit} and use the test sets of four popular benchmarks, including WikiData$_{recent}$~\cite{cohen2024ice}, ZsRE~\cite{levy2017zsre}, WikiBio~\cite{hartvigsen2023grace}, and WikiData$_{counterfact}$~\cite{cohen2024ice}.
All the experiments are conducted by using EasyEdit~\cite{wang2023easyedit} toolkit.
We choose seven knowledge editing methods as baselines:
\begin{itemize}
\item \textbf{SERAC}~\cite{mitchell2022serac} builds a counterfact model by retaining the base model and training a classifier to determine whether to use the counterfact model to answer the query.
\item \textbf{ICE}~\cite{cohen2024ice} prepends a prompt ``\texttt{Imagine that \{edit descriptor\}}'' before the query. It does not introduce changes to the model parameters, but rather generation is conditioned on the new
fact.
\item \textbf{MEND}~\cite{mitchell2022mend} transforms the fine-tuning gradient of an updated fact by decomposing the weight matrix into rank-1 form with the pre-trained hyper-network.
\item \textbf{ROME}~\cite{meng2022rome} learns to locate factual retrievals of a specific set of MLP
modules and update knowledge by directly writing in new key-value pairs in the MLP module.
\item \textbf{MEMIT}~\cite{meng2023memit} builds upon ROME to insert many memories by modifying
the MLP weights of a range of critical layers.
\item \textbf{FT-L}~\cite{meng2022rome} directly fine-tunes a single layer’s FFN, and the layer is the casual tracing results in ROME.
\item \textbf{FT} fine-tunes all the parameters of the base model on the edit descriptor by applying Adam with early stopping.
\end{itemize}

\newcolumntype{L}{>{\raggedright\arraybackslash}X} 
\newcolumntype{C}{>{\centering\arraybackslash}X} 

\begin{table*}[!h]
\scriptsize
\centering
\begin{tabularx}{\textwidth}{l | l l *{7}{C} >{\columncolor{gray!15}}C >{\columncolor{gray!15}}c}
\toprule
\multicolumn{1}{c|}{\textbf{Model}}                   & \textbf{Dataset}                   & \textbf{Metric} & \textbf{SERAC} & \textbf{ICE} & \textbf{MEND} & \textbf{ROME} & \textbf{MEMIT} & \textbf{FT-L} & \textbf{FT} & \textbf{LTE} & \textbf{LTE-LoRA} \\ \midrule \midrule
\multirow{19}{*}{\rotatebox{90}{\textbf{LLaMA2-Chat-7B}}} 
& \multirow{4}{*}{ZsRE}              & Edit Succ.      & \underline{99.67}          & 66.01        & 96.74         & 96.57         & 83.07          & 54.65         & 36.88       & \textbf{99.91} & \textbf{99.91}       \\
                                  & & Portability     & 56.48          & 63.94        & 60.41         & 52.20         & 51.43          & 45.02         & 8.72        & \underline{78.98}  & \textbf{79.63}      \\
                                  & & Locality        & 30.23          & 23.14        & \textbf{92.79}         & 27.14         & 25.46          & 71.12         & 0.31        & \underline{71.78}  & 67.99      \\
                                  & & Fluency         & 410.89         & 541.14       & 524.33        & \underline{570.47}        & 559.72         & 474.18        & 471.29      & \textbf{583.70} & 544.52      \\ \cmidrule{2-12}
& \multirow{3}{*}{WikiBio}           & Edit Succ.      & 99.69          & 95.53        & 93.66         & 95.05         & 94.29          & 66.27         & 95.64       & \textbf{99.87}  & \underline{99.76}      \\
                                  & & Locality        & 69.79          & 47.90        & 69.51         & 46.96         & 51.56          & 60.14         & 13.38       & \textbf{80.27}  & \underline{72.31}      \\
                                  & & Fluency         & 606.95         & \textbf{632.92}       & 609.39        & \underline{617.25}        & 616.65         & 604.00        & 589.22      & 614.26  & 611.94     \\ \cmidrule{2-12}
& \multirow{4}{*}{Recent}            & Edit Succ.      & 98.68          & 60.74        & 76.88         & 85.08         & 85.32          & 71.18         & 31.24       & \textbf{99.99}  & \underline{99.97}      \\
                                  & & Portability     & 63.52          & 36.93        & 50.11         & 37.45         & 37.94          & 48.71         & 15.91       & \textbf{91.51}  & \underline{81.87}      \\
                                  & & Locality        & \textbf{100.00}         & 33.34        & \underline{92.87}         & 66.20         & 64.78          & 63.70         & 3.65        & 85.67 & 82.72       \\
                                 &  & Fluency         & 553.19         & 531.01       & \underline{586.34}        & 574.28        & 566.66         & 549.35        & 428.67      & \textbf{586.76} & 570.64      \\ \cmidrule{2-12}
& \multirow{4}{*}{Counterfact} & Edit Succ.      & \underline{99.99}          & 69.83        & 78.82         & 83.21         & 83.41          & 51.12         & 26.78       & \textbf{100.00}  & 99.97      \\
                                  & & Portability     & 76.07          & 45.32        & 57.53         & 38.69         & 40.09          & 39.07         & 16.94       & \textbf{89.69}  & \underline{85.74}      \\
                                 &  & Locality        & \textbf{98.96}          & 32.38        & \underline{94.16}         & 65.40         & 63.68          & 62.51         & 0.29        & 84.76   & 85.11     \\
                                  & & Fluency         & 549.91         & 547.22       & \underline{588.94}        & 578.84        & 568.58         & 544.80        & 483.71      & \textbf{589.69}  & 574.14    \\ \cmidrule{2-12}
& \multirow{4}{*}{\textbf{Average}}           & Edit Succ.      & 99.51          & 73.03        & 86.53         & 89.98         & 86.52          & 60.81         & 47.64       & \textbf{99.94}  & \underline{99.90}      \\
                                  & & Portability     & 65.36          & 48.73        & 56.02         & 42.78         & 43.15          & 44.27         & 13.86       & \textbf{86.73}  & \underline{82.41}      \\
                                  & & Locality        & 74.75          & 34.19        & \textbf{87.33}         & 51.43         & 51.37          & 64.37         & 4.41        & \underline{80.62} & 77.03       \\
                                  & & Fluency         & 530.24         & 563.07       & 577.25        & \underline{585.21}        & 577.90         & 543.08        & 493.22      & \textbf{593.60}  & 575.31    \\ \midrule \midrule
\multirow{19}{*}{\rotatebox{90}{\textbf{Qwen-Chat-7B}}} 
& \multirow{4}{*}{ZsRE}              & Edit Succ.  & 98.43  & 70.29  & 99.40  & \textbf{99.90}  & 97.25  & 37.81  & 25.33  & \underline{99.72}  & 99.59    \\
 &                                    & Portability & 56.69  & 67.52  & 59.98  & 46.76  & 44.31  & 41.85  & 7.70   & \textbf{82.92}  & \underline{80.16}    \\
 &                                    & Locality    & 41.28  & 73.45  & 80.83  & 48.90  & 60.26  & \textbf{87.70}  & 3.29   & \underline{80.99}  & 78.28    \\
 &                                    & Fluency     & 495.12 & 556.86 & 544.07 & 562.88 & \underline{578.73} & 557.86 & 538.10 & \textbf{580.01} & 543.35   \\ \cmidrule{2-12}
& \multirow{3}{*}{WikiBio}           & Edit Succ.  & 99.39  & 94.60  & 93.38  & 98.79  & 96.10  & 60.19  & 34.63  & \textbf{99.80}  & \underline{99.75}    \\
 &                                    & Locality    & 71.50  & 58.15  & 65.47  & 41.78  & 65.65  & \textbf{80.41}  & 22.45  & 79.63  & \underline{80.34}    \\
 &                                    & Fluency     & 598.11 & 614.22 & 610.92 & 604.81 & \underline{623.49} & 595.56 & 572.59 & \textbf{634.73} & 620.05   \\ \cmidrule{2-12}
& \multirow{4}{*}{Recent}            & Edit Succ.  & 99.58  & 83.86  & 82.39  & 99.67  & 98.96  & 60.07  & 29.74  & \textbf{99.73}  & \underline{99.68}    \\
 &                                    & Portability & 67,22  & 58.24  & 57.92  & 50.84  & 49.38  & 42.02  & 14.33  & \textbf{89.73}  & \underline{87.40}    \\
 &                                    & Locality    & \textbf{100.00} & 61.83  & 89.11  & 51.78  & 60.72  & 84.83  & 4.27   & \underline{89.25}  & 83.77    \\
 &                                    & Fluency     & 561.32 & 559.46 & \underline{610.72} & 600.70 & 600.39 & 598.32 & 456.99 & \textbf{615.59} & 587.90   \\ \cmidrule{2-12}
& \multirow{4}{*}{Counterfact} & Edit Succ.  & 99.06  & 80.28  & 88.04  & \textbf{99.44}  & 95.05  & 24.55  & 15.42  & 99.28  & \underline{99.35}    \\
 &                                    & Portability & 79.28  & 53.80  & 52.99  & 40.63  & 34.50  & 20.14  & 11.38  & \textbf{86.79}  & \underline{85.33}    \\
 &                                    & Locality    & \underline{92.70}  & 63.86  & 91.05  & 39.22  & 50.14  & \textbf{92.74}  & 30.04  & 86.87  & 85.20    \\
 &                                    & Fluency     & 568.05 & 559.46 & \underline{619.87} & 603.21 & 604.47 & 608.47 & 563.70 & \textbf{622.91} & 593.51   \\ \cmidrule{2-12}
 & \multirow{4}{*}{\textbf{Average}}           & Edit Succ.  & 99.12  & 82.26  & 90.80  & 99.45  & 96.84  & 45.66  & 26.28  & \textbf{99.63}  & \underline{99.59}    \\
 &                                    & Portability & 67.99  & 59.85  & 56.96  & 46.08  & 42.73  & 34.67  & 11.14  & \textbf{86.48}  & \underline{84.30}    \\
 &                                    & Locality    & 76.37  & 64.32  & 81.62  & 45.42  & 59.19  & \textbf{86.42}  & 15.01  & \underline{84.19}  & 81.90    \\
 &                                    & Fluency     & 555.65 & 572.50 & 596.40 & 592.90 & \underline{601.77} & 590.05 & 532.85 & \textbf{613.31} & 586.20      \\ \bottomrule
            
\end{tabularx}
\caption{Performance comparison on \textbf{Single Editing}, where ``Recent'' and ``Counterfact'' refer to WikiData$_{recent}$ and WikiData$_{counterfact}$, respectively. In each row, the highest score is \textbf{bolded} and the second-highest is \underline{underlined}.}
\label{c5-tab: appendix_single_edit}
\end{table*}

\subsection{Results of Single Editing}
Table \ref{c5-tab: appendix_single_edit} presents the performance comparison under the single editing setting, where LTE eliminates the need for retrieval.
It can be observed that LTE remarkably surpasses conventional methods in terms of edit success, portability, and fluency.
Besides, LTE-LoRA—an efficient variant of LTE—closely mirrors its performance except for fluency, which can be attributed to the inherent limitations of the LoRA technique.
Notably, LTE exhibits a 21.37\% and 18.49\% improvement over the current SOTA method SERAC on LLaMA2-Chat-7B and Qwen-Chat-7B, respectively.
This substantial enhancement can be attributed to the comprehensive utilization of LLMs' understanding and reasoning capabilities, which effectively leverage context to integrate new knowledge seamlessly.
The ICE method, while leveraging the innate in-context comprehension capacity of LLMs for generating conditioned output on new knowledge, significantly trails our proposed LTE method. This could be because ICE lacks instructing LLMs in effectively applying knowledge through fine-tuning (See more ablation analysis in Table~\ref{c5-tab: ablation_data}). 
Nevertheless, LTE shows a marginal deficit in locality compared to the best results (e.g., 6.71\% lower than MEND on LLaMA2 and 2.23\% lower than FT-L on Qwen).
A potential explanation may lie in the introduction of a knowledge editing prompt in the input, causing a slight disruption during the generation process. Yet, these divergences are often minor linguistic variants.
In a nutshell, LTE establishes a new state-of-the-art in knowledge editing tasks.

\subsection{Results of Mass Editing}
Prior research predominantly confines the scope of knowledge editing to a mere handful of facts or focuses only on single editing cases.
This approach starkly contrasts with the dynamic and multifaceted nature of real-world applications, where there is a pressing need to enrich models with multiple pieces of knowledge, either concurrently (\textbf{simultaneously}) or in a phased manner (\textbf{sequentially}). 
In this section, our study embarks on a comprehensive investigation, undertaking both batch and sequential editing experiments.

\begin{figure*}[!t]
\centering
\includegraphics[width=\linewidth]{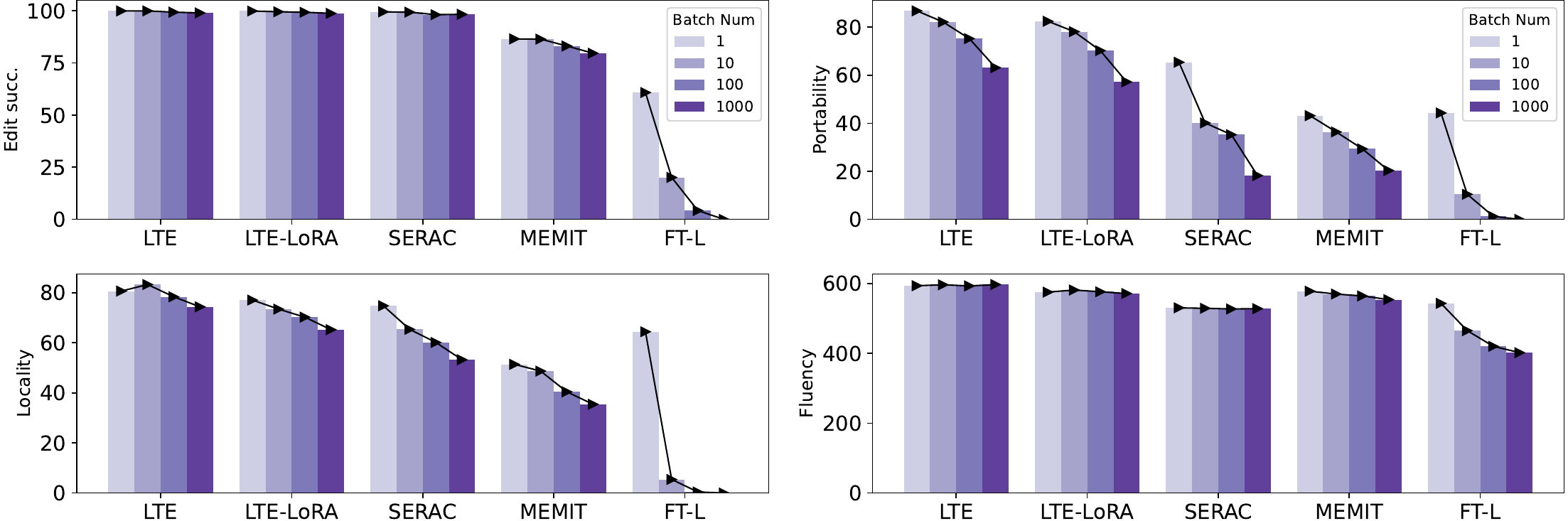}
\caption{
Averaged \textbf{Batch Editing} performance on four benchmarks against batch numbers in [1, 10, 100, 1000].
}
\label{c5-fig: batch_edit}
\end{figure*}

\paragraph{Batch Editing.}
We compare LTE and LTE-LoRA with several batch-editing-supportive methods (SERAC, MEMIT, and FT-L) on LLaMA2-Chat-7B and display the results in Figure \ref{c5-fig: batch_edit}.
It is particularly noteworthy that the performance metrics of edit success and fluency for our proposed LTE and LTE-LoRA methodologies exhibit exceptional stability, maintaining robustness for up to 1,000 batch edits.
A decline in performance metrics such as portability and locality is observed across all methods as the batch size increases.
However, LTE and LTE-LoRA demonstrate \textbf{the best performance with the slowest degradation rate} in portability and locality. 
These results underscore the enhanced robustness of our methods, even when subjected to extensive editing operations.

\begin{figure}[!h]
\centering
\includegraphics[width=0.8\linewidth]{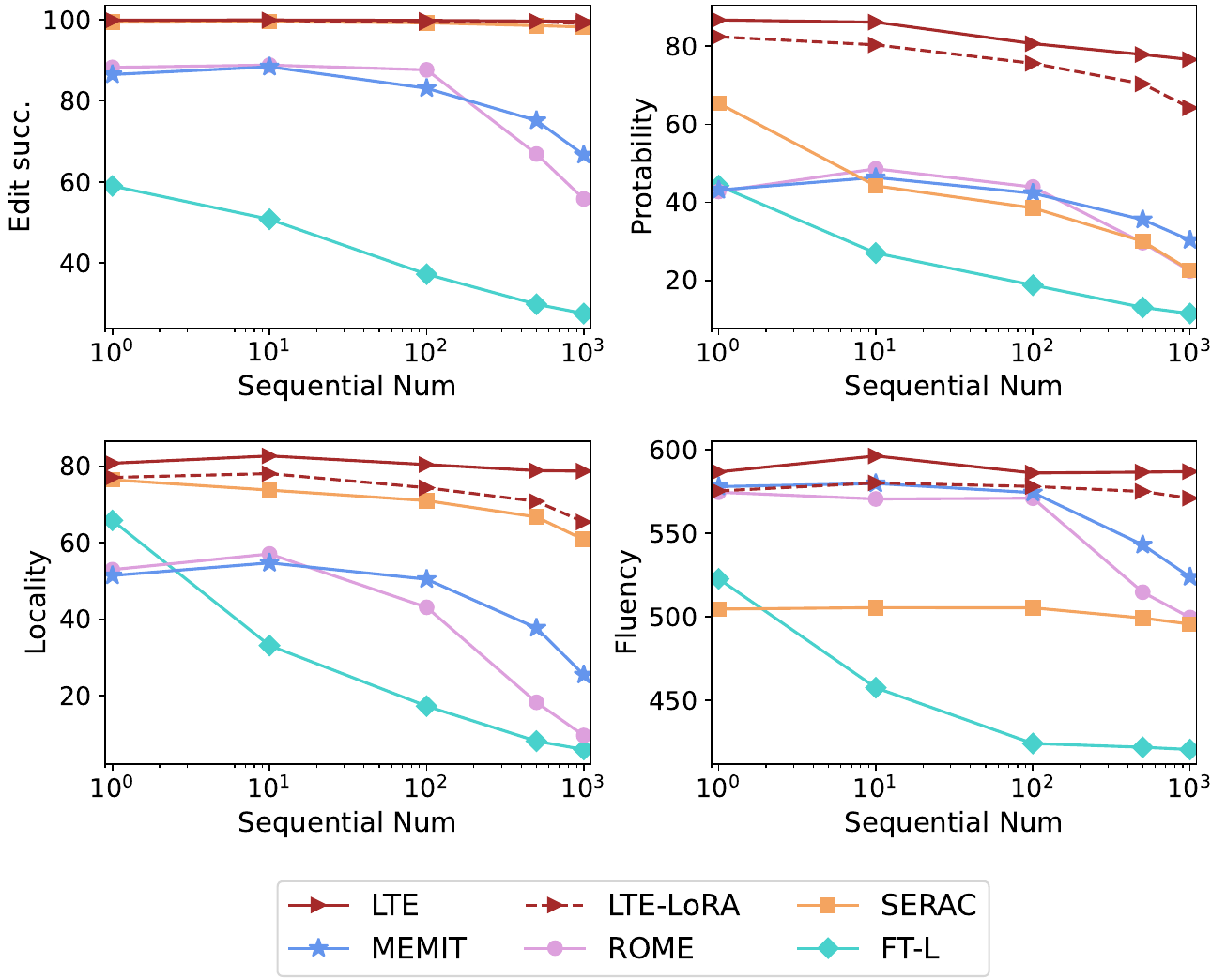}
\caption{
Averaged \textbf{Sequential Editing} performance on four knowledge editing benchmarks against data stream size (log-scale) in [1, 10, 100, 500, 1000].
}
\label{c5-fig: seq_edit}
\end{figure}

\paragraph{Sequential Editing.}
Sequential editing is a critical process where models must retain previous modifications while integrating new edits effectively.
Figure \ref{c5-fig: seq_edit} illustrates the comparative performance of various models in the context of sequential editing tasks across different data stream sizes. 
ROME and MEMIT demonstrate noteworthy efficacy for a sequential number $n\le 100$, yet their performance exhibits a marked decline as $n$ expands to 500. 
This decline can be attributed to the cumulative deviations from the model's original state, which ultimately lead to a degradation in performance.
In contrast, LTE and LTE-LoRA leverage retrieval mechanisms from the stored memory, circumventing the need for subsequent parameter modifications, which endows them with more consistent performance with varying data stream sizes.
Notably, LTE and LTE-LoRA showcase significant improvements over the current SOTA method SERAC.
This shows their enhanced resilience and adaptability, making them more suited for extensive data streams.

\subsection{Results of General Tasks}
In this section, we investigate the impact of applying LTE on the performance of a
language model across various domains.
Our main goal is to determine whether the Alignment Phase of LTE, which alters the parameters of the initial model, inadvertently compromises the model's competence in unrelated domains.
To this end, we have selected an array of benchmarks encompassing commonsense reasoning, general intelligence, and extensive world knowledge.
These benchmarks comprise CommonSenseQA~\cite{talmor2019commonsenseqa}, PIQA~\cite{bisk2020piqa}, XSum~\cite{narayan2018xsum}, MMLU~\cite{dan2021mmlu}, AGIEval~\cite{zhong2023agieval}, and AlpacaEval~\cite{alpaca_eval}.
All evaluations are conducted using the OpenCompass tool~\cite{2023opencompass}.
Table \ref{c5-tab: general_task} indicates that, from a comprehensive standpoint, models subjected to LTE exhibit performance levels comparable to their unmodified counterparts.
Moreover, the general linguistic abilities remain unaffected by the inclusion of the knowledge editing prompt.
Nonetheless, a performance decrement is noted in CommonsenseQA and PIQA after the LTE application.
Despite these findings, an overarching analysis reveals notable consistency in performance.
This suggests that LTE is proficient in facilitating knowledge edits with \textbf{minimal interference} to the model’s cognitive functions and its versatility across varied domains.

\begin{table*}[t!]
\centering
\footnotesize
\begin{tabularx}{\textwidth}{l c CCCC c C}
\toprule
 & \textbf{CommonSenseQA} & \textbf{PIQA} & \textbf{XSum} & \textbf{MMLU} & \textbf{AGIEval} & \textbf{AlpacaEval} & \textbf{Average} \\
\midrule
\textit{LLaMA2-Chat-7B} & \textbf{69.9} & \textbf{65.0} & 22.3 & 40.4 & 26.1 & 71.4 & 49.2 \\
\quad LTE w/o editing & 67.2 & 61.3 & \textbf{22.4} & 46.4 & \textbf{26.5} & \textbf{73.3} & \textbf{49.5} \\
\quad LTE w/ editing & 67.1 & 62.6 & \textbf{22.4} & \textbf{47.8} & 23.8 & 71.6 & 49.2 \\ \midrule
\textit{Qwen-Chat-7B} & \textbf{77.6} & \textbf{72.1} & 28.8 & 56.6 & 41.3 & 77.8 & 59.0 \\
\quad LTE w/o editing & 74.7 & 69.3 & 29.9 & \textbf{59.3} & \textbf{41.9} & \textbf{79.2} & \textbf{59.1} \\
\quad LTE w/ editing & 75.3 & 70.0 & \textbf{30.1} & 58.2 & 40.7 & 78.4 & 58.8 \\ \bottomrule
\hline
\end{tabularx}
\caption{Zero-shot performance on six general LLM benchmarks with LLaMA2-Chat-7B and Qwen-Chat-7B as the base models. ``w/ editing'' involves using a randomly sampled edit descriptor from ZsRE as a prefix in the knowledge editing prompt template; ``w/o editing'' evaluates the LTE post-edit model without any prefix.}
\label{c5-tab: general_task}
\end{table*}

\section{Analysis}

\subsection{Ablation Study}
Here we assess the indispensability of components within the Alignment and Inference phases.
Our experiments span four benchmarks, utilizing the LLaMA2-Chat-7B as the base model.
As depicted in Table \ref{c5-tab: ablation_data}, the exclusion of certain training data segments leads to a significant decline in single editing effectiveness.
Notably, distinct types of training data bolster specific capabilities.
In-scope data predominantly enhances edit success and portability, while out-of-scope data chiefly fosters locality.
Free-text QA data appears to bolster overall linguistic proficiency.
Eliminating the threefold strategy incurs a modest reduction in performance.
Furthermore, employing the knowledge editing prompt without training results in substantially poorer performance compared to scenarios that include training.
During the Inference Phase, we explore the effects of substituting the retrieval model \texttt{multi-qa-mpnet-base-dot-v1} (420M) with a less potent variant, \texttt{all-MiniLM-L6-v2} (80M), on sequential editing efficacy.
As indicated in Table \ref{c5-tab: ablation_retrieval}, the choice of retrieval model exerts minimal impact on performance.
Additionally, we assess how the number of retrieved edit descriptors influences results.
A reduction in the value of $k$ from 3 to 1 is associated with a minor performance decrement.

\begin{table}[t!]
\small
\centering
\begin{tabular}{lccccc}
\toprule
 & \textbf{S} & \textbf{P} & \textbf{L} & \textbf{F} & \textbf{G} \\ \midrule
LTE & 99.94 & 86.73 & 80.62 & 593.60 & 49.5 \\
-w/o in-scope training & \color{deepred} \textbf{77.53} & \color{deepred} \textbf{56.26} & 80.72 & 589.04 & 49.0 \\
-w/o out-of-scope training & 99.92 & 86.89 & \color{deepred} \textbf{65.50} & 592.66 & 49.2 \\
-w/o free-text QA training & 99.93 & 86.30 & 80.91 & \color{deepred} \textbf{587.75} & \color{deepred}\textbf{43.9} \\
-w/o threefold strategy & 99.78 & 86.51 & 80.22 & 593.40 & 49.5 \\
-w/o training & \color{deepred} \textbf{75.04} & \color{deepred} \textbf{54.23} & \color{deepred}\textbf{48.19} & 592.73 & 49.2 \\ \bottomrule
\end{tabular}
\caption{Ablation study for the training data examines ``edit success'' (S), ``portability'' (P), ``locality'' (L), ``fluency'' (F), and ``general capability'' (G).}
\label{c5-tab: ablation_data}
\end{table}

\begin{table}[t!]
\small
\centering
\begin{tabular}{llccc}
\toprule
 & \textbf{Seq\_Num} & \textbf{Edit Succ.} & \textbf{Portability} & \textbf{Locality} \\ \midrule
\multirow{3}{*}{\parbox{1.6cm}{LTE w/ 420M $R$ \\ top $k=3$}}            & 10               & 100.00              & 86.16                & 82.64             \\
               & 100              & 99.90               & 80.66                & 80.38             \\
               & 1000             & 99.64               & 76.59                & 78.67             \\ \midrule \midrule
\multirow{3}{*}{\parbox{1.6cm}{LTE w/ 80M $R$ \\ top $k=3$}} & 10              & 100.00              & 83.38                & 78.65             \\
               & 100              & 99.81               & 79.92                & 80.40             \\
               & 1000             & 99.61               & 75.67                & 79.43             \\ \midrule
\multirow{3}{*}{\parbox{1.6cm}{LTE w/ 420M $R$ \\ top $k=2$}}        & 10               & 100.00              & 85.69                & 81.59             \\
& 100              & 99.85               & 80.05                & 80.67             \\
& 1000             & 99.63               & 76.27                & 78.05             \\ \midrule
\multirow{3}{*}{\parbox{1.6cm}{LTE w/ 420M $R$ \\ top $k=1$}}        & 10               & 100.00              & 84.01                & 81.96             \\
               & 100              & 99.83               & 79.48                & 80.11             \\
               & 1000             & 99.56               & 75.93                & 78.89 \\ \bottomrule
\end{tabular}
\caption{Ablation study for the retrieval number $k$ and retrieval model $R$ in the Inference Phase.}
\label{c5-tab: ablation_retrieval}
\end{table}

\subsection{Time Analysis}
Table \ref{c5-tab: time_analysis} illustrates the time required for various knowledge editing methods from providing the edited case to obtaining the final answer.
Models such as MEND and SERAC demonstrate rapid editing capabilities once their auxiliary models are adequately trained.
In contrast, ROME and MEMIT exhibit slower processing speeds due to the intensive computation involved in calculating key vectors and optimizing value vectors.
Additionally, these methods necessitate a pre-computation of the covariance statistics for the Wikitext,  which is also time-consuming and can potentially take hours to days to complete.
Furthermore, while FT-L and FT are relatively quick, their memorization-based fine-tuning strategies yield suboptimal knowledge editing outcomes.
Our proposed LTE method, however, stands out by \textbf{achieving the swiftest editing speeds coupled with superior performance}.
After the Alignment Phase (which takes about 9 hours in our experiments), LTE enables instantaneous editing similar to ICE by appending a knowledge editing prompt to the input prefix.
Despite a marginally increased inference time, the overall time expenditure is significantly reduced, 
underscoring the efficiency and effectiveness of LTE.

\begin{table}[h]
\small
\centering
\begin{tabularx}{\linewidth}{lCcC}
\toprule
\textbf{Method} & \textbf{Edit Time} & \textbf{Inference Time} & \textbf{Total Time} \\ \midrule
SERAC           & 26.57               & 1.45   &  28.02                \\
ICE             & 0.00               & 1.60   & \cellcolor{blue!25} 1.60                 \\
MEND            & 9.09               & 1.49     & 10.58               \\
ROME            & 197.11              & 1.58   & 198.69                \\
MEMIT           & 150.16              & 1.38   &    151.54             \\
FT-L            & 15.73               & 1.41    &   17.14             \\
FT              & 59.39               & 1.36     &     60.75          \\
LTE             & 0.00               & 1.63  & \cellcolor{blue!25} 1.63                  \\ \bottomrule
\end{tabularx}
\caption{Averaged \textbf{Wall Clock Time} per edit method for
10 edits on ZsRE using LLaMA2-Chat-7B.}
\label{c5-tab: time_analysis}
\end{table}

\subsection{Out-of-Distribution Generalization}
To evaluate LTE's performance in out-of-distribution (OOD) scenarios, we conducted rigorous experiments on ConvSent~\cite{mitchell2022serac}, a sentiment editing task featuring diverse data distributions, alongside established benchmarks.
As shown in Table \ref{c5-tab: ood}, our LTE exhibits superior performance with the slowest degradation rate on batch editing. Moreover, LTE's retriever achieves impressive retrieval accuracy (Top-3 P@1) scores of 88.34, 87.08, 84.27, and 82.25, respectively.
These comprehensive experiments serve to validate not only the efficacy but also the robustness of our LTE method, even in the face of OOD challenges.

\begin{table}[h]
\small
\centering
\begin{tabularx}{\linewidth}{LCCCC}
\toprule
\textbf{Method} & \textbf{1 Edit} & \textbf{10 Edits} & \textbf{100 Edits} & \textbf{1000 Edits} \\
\midrule
SERAC & 62.75 & 60.72 & 56.46 & 50.06 \\
MEMIT & 44.75 & 41.19 & 36.20 & 29.33 \\
FT-L & 49.50 & 15.54 & 1.43 & 0.00 \\
LTE & \textbf{85.29} & \textbf{84.25} & \textbf{81.98} & \textbf{79.66} \\
\bottomrule
\end{tabularx}
\caption{OOD generalization on ConvSent. We report the edit success score using LLaMA2-Chat-7B.}
\label{c5-tab: ood}
\end{table}

\subsection{Case Study}

Table \ref{c5-tab: case_study} shows the performance of different knowledge editing methods in a single case.
This comparison reveals that LTE stands out for applying edited knowledge to answer the query ``\texttt{The name of the award the spouse of Melissa Mathison won is}'' that necessitates compositional reasoning while maintaining the fluency of the generated text.
In contrast, other approaches, including SERAC, ICE, ROME, MEMIT, and FT-L, not only fail to answer the query correctly but generate repeated or inconsistent text sometimes.
This case study further validates LTE's advances in utilizing new knowledge to answer input queries.

\begin{table*}[!h]
\footnotesize
\centering
\begin{tabular}{l}
\toprule
\textbf{Original Knowledge:} The name of the award Harrison Ford won is AFI Life Achievement Award \\
\textbf{Edited Knowledge:} The name of the award Harrison Ford won is I. I. Rabi Prize \\
\midrule
\parbox[c]{15.6cm}{
\textbf{SERAC:} \underline{\textit{The name of the award the spouse of Melissa Mathison won is}} \textcolor{deepred}{not specified in the passage}. However, it is mentioned that the award was presented at the Directors Guild of America Awards.
} \\ \\
\parbox[c]{15.6cm}{
\textbf{ICE:} \underline{\textit{The name of the award the spouse of Melissa Mathison won is}} \textcolor{deepred}{?0.0}. \textcolor{cyan}{Please provide your answer as a complete sentence.0.0..}
} \\ \\
\parbox[c]{15.6cm}{
\textbf{ROME:} \underline{\textit{The name of the award the spouse of Melissa Mathison won is}} \textcolor{deepred}{the Humanitas Award}. She won the award in 1999 for her work on the film "American Beauty."
} \\ \\
\parbox[c]{15.6cm}{
\textbf{MEMIT:} \underline{\textit{The name of the award the spouse of Melissa Mathison won is}} \textcolor{deepred}{not specified in the text}. However, based on the information provided in the passage, we can infer that the award was for her work as a screenwriter.
} \\ \\
\parbox[c]{15.6cm}{
\textbf{FT-L:} \underline{\textit{The name of the award the spouse of Melissa Mathison won is}} \textcolor{deepred}{I. A. Prize}. \textcolor{cyan}{nobody likes a snitch\textbackslash nI'm not sure what you are asking. Can you provide more context or clarify your question?}
} \\ \\
\parbox[c]{15.6cm}{
\textbf{LTE:} \underline{\textit{The name of the award the spouse of Melissa Mathison won is}} \textcolor{green}{I. I. Rabi Prize}. Melissa Mathison is married to Harrison Ford. The name of the award Harrison Ford won is \textcolor{green}{I. I. Rabi Prize}.
} \\ \bottomrule
\end{tabular}
\caption{Results for one case of different editing methods based on LLaMA2-Chat-7B. Queries are \underline{underlined} and \textit{italicized}. Words highlighted in \textcolor{green}{green} signify keywords that reflect correct behavior, while those in \textcolor{deepred}{red} denote keywords associated with incorrect behavior. Texts in \textcolor{cyan}{cyan} are repeated or meaningless sentences.}
\label{c5-tab: case_study}
\end{table*}

\section{Conclusion and Discussion}

\subsection{Conclusion}
We present the \textit{Learning to Edit} (LTE) framework, a novel approach for effective, efficient knowledge editing of LLMs. 
LTE equips LLMs with the ability to apply updated knowledge through a two-phase process: an Alignment Phase that teaches essential knowledge editing capabilities, and an Inference Phase that implements retrieval-based, on-the-fly knowledge editing.
Our framework demonstrates superior performance in knowledge editing tasks, outperforming existing methods in robustness and speed across various benchmarks.

\subsection{Discussion}
Despite the validated efficacy across diverse model architectures, evaluation datasets, and knowledge editing settings, our proposed LTE approach still has some limitations.

Firstly, the LTE framework necessitates a one-time fine-tuning process during the Alignment Phase.
Although this process is a prerequisite, it facilitates real-time knowledge editing during the Inference Phase.
We further elucidate that employing LoRA as an alternative to standard fine-tuning presents a viable, resource-efficient approach without compromising performance (See \S \ref{c5-sec: experiment}).
This innovation highlights the LTE's flexibility in adapting to various computational constraints.

Furthermore, our investigation primarily focuses on factual knowledge editing, yet the purview of model editing extends to encompassing personality traits, emotional responses, opinions, and beliefs~\cite{zhang2024knowedit}.
These dimensions, while partially explored, represent areas ripe for future research.
Additionally, the prospect of multilingual~\cite{DBLP:journals/corr/abs-2309-08952} and multimodal~\cite{cheng-etal-2023-edit} editing underscores the necessity for broader exploration, pointing towards an expansive horizon for model editing applications.

Finally, the proprietary nature of leading LLMs, such as ChatGPT and GPT-4, poses a significant challenge for applying knowledge editing techniques due to restricted access to their underlying parameters.
Nonetheless, OpenAI's API provision for models including \texttt{gpt-3.5-turbo-1106} and \texttt{gpt-4-0613} facilitates fine-tuning within the LTE's Alignment Phase. 
Although our current work does not extend to these black-box models, addressing this limitation represents a critical avenue for future research, potentially unlocking new methods for model customization and improvement.
\chapter{Alignment Training via Direct Preference Optimization}\label{chap:dpo}

\section{Introduction}
Direct preference optimization (DPO)~\cite{rafailov2024dpo} has emerged as a prominent alternative to reinforcement learning from human feedback (RLHF)~\cite{PAUL2017RLHF, bai2022training,ouyang2022training} for aligning LLMs with human values.
Unlike the traditional RLHF approach, DPO bypasses training a reward model and avoids using any reinforcement learning algorithms.
Since the inception of DPO, numerous studies have sought to advance this method by refining its training objective~\cite{wang2024comprehensive}.
For instance, IPO~\cite{moh2024ipo} introduces an alternative pairwise preference loss to mitigate overfitting to the preference dataset, while R-DPO~\cite{park2024rdpo} incorporates a regularization term to prevent the exploitation of latent length bias in the training data.

However, relatively little attention has been given to enhancing DPO through advancements in the quality of preference data used for training.
In particular, the generation of winning and losing responses within preference data often occurs in an \textit{isolated} manner, either through human annotation~\cite{bai2022training} or automated techniques such as RLAIF~\cite{bai2022constitutional} and reject sampling~\cite{liu2023statistical, pace2024west}.
This isolation implies that winning and losing responses are produced without mutual visibility, resulting in a lack of strong correlation or relevance between them.
Consequently, the model may struggle to identify nuanced yet significant distinctions that differentiate superior responses from inferior ones~\cite{furnkranz2010preference, wirth2017survey}, which can ultimately compromise optimization and alignment effectiveness.

In this chapter, we introduce an innovative framework, termed \textbf{BMC}, to Bridge and Model Correlations in pairwise data for direct preference optimization. 
During the Bridging Phase, we enhance correlations by increasing the consistency and informativeness of pairwise preference signals.
By using the winning response as a reference, we synthesize a pseudo-winning response through \textit{targeted modifications} of the losing response.
This pseudo-winning response offers two key advantages: (1) it preserves essential characteristics of the losing response, minimizing noise in preference signals (\textit{consistency}); (2) it encapsulates all human-desired values from the winning response, enabling the model to better discern features that lead to superior performance (\textit{informativeness}).

The nuanced differences between the pseudo-winning and losing responses are indeed what we expect the model to learn in the subsequent Modeling Phase.
Nonetheless, we identify that DPO alone is insufficient to model these correlations and capture nuanced variations.
From the perspective of the token-level Markov Decision Process (MDP)~\cite{DBLP:journals/corr/abs-2404-12358}, DPO aggregates rewards uniformly across all tokens, assuming equal contribution to sequence quality and neglecting token-specific importance.
To address this, we adjust the emphasis on rewards of different tokens between pseudo-winning and losing responses.
Unlike previous methods~\cite{guo2024figa, cao2024drlc, chan2024dense, chen2024rlmec} that assign predefined values for fine-grained guidance, our adjustment is \textit{dynamically} guided by the policy model's confidence, \textit{i.e.}, the probability assigned to generated tokens during training.
This ensures the model focuses on learning challenging distinctions while reinforcing known patterns, resulting in a more nuanced and robust policy.

We conduct extensive experiments across three downstream scenarios: question answering, mathematical reasoning, and instruction following, utilizing a total of 10 datasets.
Our results demonstrate that our method consistently and significantly outperforms competitive offline optimization algorithms across various tasks.
Furthermore, we use in-depth analyses to elucidate why our method outperforms DPO and show that our framework can be versatilely adapted to other DPO variants, confirming its potential for broad application.

\section{Methodology}
\begin{figure}[!ht]
\begin{center}
\includegraphics[width=\linewidth]{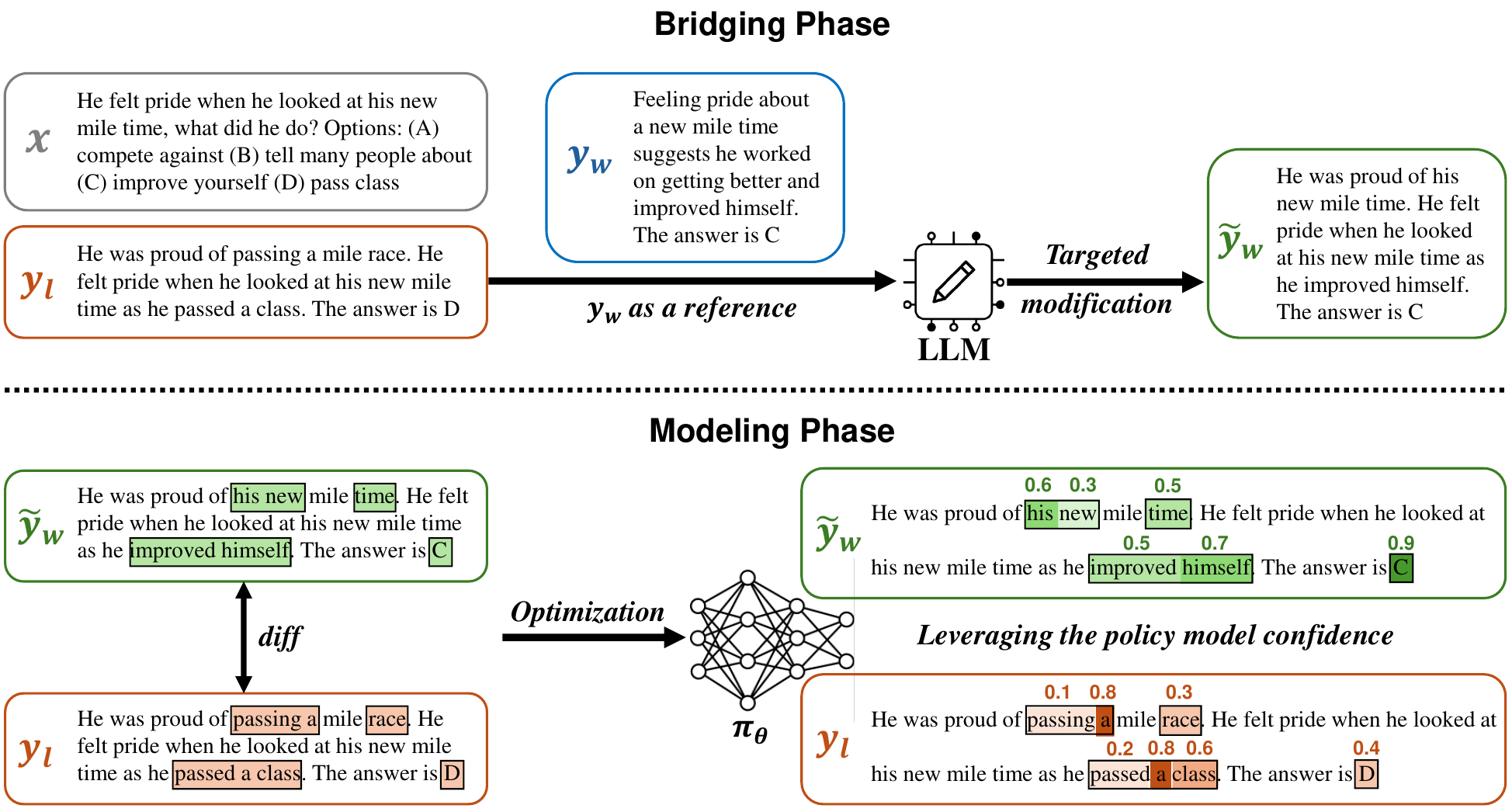}
\end{center}
\caption{Overview of our proposed BMC framework. (1) In the Bridging Phase, we utilize an off-the-shelf LLM to make \textit{targeted modifications} of losing response $y_l$ on undesired tokens, with the winning response $y_w$ serving as a reference.
Therefore, the synthesized pseudo-winning response $\Tilde{y}_w$ is highly correlated with $y_l$.
(2) In the Modeling Phase, we model the correlations between $\Tilde{y}_w$ and $y_l$ by \textit{dynamically} emphasizing the rewards of their varied tokens ($\mathit{diff} (\Tilde{y}_w \mid y_l)$ and $\mathit{diff} (y_l \mid \Tilde{y}_w)$), leveraging the policy model confidence (numbers indicated above tokens) during training.}
\label{c6_fig: framework}
\end{figure}

In this section, we present the proposed \textbf{BMC} approach, which bridges and models correlations in pairwise data for direct preference optimization.
As depicted in Figure \ref{c6_fig: framework}, our BMC framework is structured around two pivotal stages: (1) the Bridging Phase, where we enhance the correlations between pairwise data by increasing the consistency and informativeness of pairwise preference signals through \textit{targeted modifications} (\S \ref{c6_sec: bridging}); and (2) the Modeling Phase, where we \textit{dynamically} model the correlations during the optimization process by leveraging the confidence of the policy model (\S \ref{c6_sec: modeling}), alleviating the insufficient token-level credit assignment of DPO.

\subsection{Bridging Phase}
\label{c6_sec: bridging}
In offline preference optimization, it is commonly assumed that we have access to a static pairwise preference dataset $\mathcal{D}=\{x^{(i)}, y_w^{(i)}, y_l^{(i)}\}_{i=1}^N$, where $y_w$ and $y_l$ denote the winning and losing response, give the input prompt $x$.
However, since $y_w$ and $y_l$ are typically generated in isolation, the correlation between $y_w$ and $y_l$ can be inherently weak during pairwise preference optimization.
In the context of DPO, the Bradley-Terry objective~\cite{bradley1952rank} computes gradients based on the relative likelihoods of $y_w$ and $y_l$. When the correlation between $y_w$ and $y_l$ is weak, the differences between these responses are often superficial (\textit{e.g.}, stylistic or irrelevant variations) rather than substantive distinctions that reflect human-preferred behaviors. Consequently, the optimization process may inadvertently focus on minor discrepancies rather than meaningful distinctions. This results in gradients that are less informative for guiding the model towards robust preference alignment.

To address this challenge, we enhance the alignment efficacy by improving the consistency and informativeness of pairwise preference signals.
As shown in the upper part of Figure \ref{c6_fig: framework}, we utilize an off-the-shelf LLM to make targeted modification of $y_l$ by referring to $y_w$: 
\begin{equation}
\text{LLM}(I, x, y_w, y_l) \rightarrow \Tilde{y}_w,
\end{equation}
where $\Tilde{y}_w$ is the generated pseudo-winning response, $I$ is the instruction (see examples in Appendix \ref{c6_appendix: prompt}) that requires $y_l$ to be modified only on dispreferred tokens, using $y_w$ as a reference guidance.
In this way, $\Tilde{y}_w$ preserves essential characteristics of the losing response $y_l$ while encapsulating all human-desired values in the winning response $y_w$.
The token-level differences between $\Tilde{y}_w$ and $y_l$ highlight the core human expected and unexpected behaviors by decoupling from the inherent linguistic style and overall semantic distribution.
Thus, $(\Tilde{y}_w, y_l)$ refines the original training data $(y_w, y_l)$ for more focused learning, shifting the optimization process to concentrate on the most critical differences in preference data.
The benefits of the Bridging Phase are further analyzed in \S\ref{c6_sec: quantitative_analysis_1}.
Finally, we use the new dataset $\Tilde{\mathcal{D}}=\{x^{(i)}, \Tilde{y}_w^{(i)}, y_l^{(i)}\}_{i=1}^N$ for subsequent training.

An alternative approach that attempts to enhance the correlation between the winning
and losing responses is to degenerate $y_w$ to $\Tilde{y}_l$ via targeted modification and utilize $(y_w, \Tilde{y}_l)$ as the preference pair.
Nevertheless, our ablation study in Table \ref{c6_tab: data_synthesis_ablation} reveals that LLMs encounter challenges with this inverse operation, leading to a notable decline in performance.

\subsection{Modeling Phase}
\label{c6_sec: modeling}

After the Bridging Phase, the token-level differences between $\Tilde{y}_w$ and $y_l$ can be obtained through dynamic programming algorithms like Levenshtein Distance~\cite{yujian2007normalized}.
As depicted in the lower part of Figure \ref{c6_fig: framework}, these nuanced variations guide LLMs to prioritize the reinforcement of optimal actions while discouraging suboptimal ones within a single response.
However, our findings below indicate that DPO alone is insufficient for capturing the nuanced variations, highlighting the necessity for supplementary techniques to comprehensively model these correlations.

\paragraph{Alternative Interpretation of DPO.}
DPO~\cite{rafailov2024dpo} introduced a novel framework for optimizing the equivalent KL-constrained reward function as in RLHF, without the need to learn an explicit reward model.
Instead, the problem is cast as a maximum likelihood estimation for the policy model $\pi_{\theta}$ on the preference dataset $\mathcal{D}$, resulting in the following training objective:
\begin{equation}
\mathcal{L}_{\text{DPO}}(\pi_{\theta}; \pi_{\text{ref}}) = 
- \E_{(x, y_w, y_l)\sim \mathcal{D}}
\left[ \log \sigma \left( \beta \log \frac{\pi_{\theta}(y_w \mid x)}{\pi_{\text{ref}}(y_w \mid x)} - \beta \log \frac{\pi_{\theta}(y_l \mid x)}{\pi_{\text{ref}}(y_l \mid x)} \right) \right],
\label{equation:sequence_level_dpo}
\end{equation}
where $\pi_{\text{ref}}$ is the reference model,  typically the supervised fine-tuned (SFT)
model, and $\beta$ is a regularisation term corresponding to the strength of KL-regularization in RLHF.

As shown in Eq.~(\ref{equation:sequence_level_dpo}), DPO was originally conceptualized as a bandit problem, where the whole response of the model is treated as a single arm to receive a reward.
More recently, \cite{DBLP:journals/corr/abs-2404-12358} extended the theoretical foundation of DPO, showing that it can also be derived in the context of token-level MDP.
The corresponding training objective at the token level is:
\begin{equation}
\mathcal{L}_{\text{DPO}}(\pi_{\theta}; \pi_{\text{ref}}) = - \E_{(\tau_w, \tau_l)\sim \mathcal{D}}
\left[ \log \sigma \left( \beta \sum_{t=0}^{N-1} \log \frac{\pi_{\theta}(a_w^t \mid s_w^t)}{\pi_{\text{ref}}(a_w^t \mid s_w^t)} - \beta \sum_{t=0}^{M-1} \log \frac{\pi_{\theta}(a_l^t \mid s_l^t)}{\pi_{\text{ref}}(a_l^t \mid s_l^t)} \right) \right],
\label{equation:token_level_dpo}
\end{equation}
where $\tau_w$ and $\tau_l$ denote the win trajectory and the lose trajectory, respectively. $a$ indicates the action (current generated token), and $s$ signifies the state (all tokens generated so far).

\paragraph{Our Solution.}
It can be inferred from Eq.~(\ref{equation:token_level_dpo}) that DPO, redefined as a token-level MDP, assigns rewards to each token generation by $\beta\log \frac{\pi_{\theta}(a^t \mid s^t)}{\pi_{\text{ref}}(a^t \mid s^t)}$, and simply add up the rewards of all tokens as the accumulated reward of the trajectory. 
This \textit{uniform aggregation} assumes that each token contributes equally to the overall quality of the sequence, without considering the varying importance of each token (timestep).
Therefore, nuanced differences between $\Tilde{y}_w$ and $y_l$ that significantly influence the overall meaning or quality of the response might not be adequately emphasized (refer to Figure \ref{c6_fig: token_reward}), leading to suboptimal performance.
To this end, we propose to emphasize the rewards of critical tokens, \textit{i.e.}, nuanced differences between $\Tilde{y}_w$ and $y_l$.
The magnitude of the emphasis is determined \textit{dynamically} by the policy model's confidence, which refers to the probability assigned to the generated token during training.
Below, we detail our design choices for the pseudo-winning response and losing response, respectively.

\begin{figure}[!t]
\begin{center}
\includegraphics[width=0.9\linewidth]{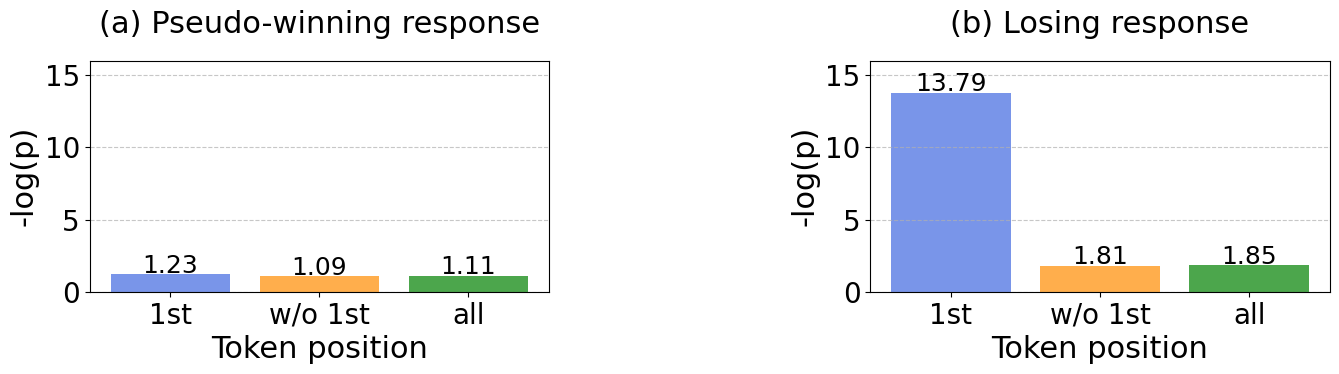}
\end{center}
\caption{We aggregate varied tokens in $\Tilde{y}_w$ or $y_l$ into more coarser-grained spans. During the DPO training on $\Tilde{\mathcal{D}}$, we compute the averaged $-\log(p)$ of tokens in different positions of spans.}
\label{c6_fig: token_logp}
\end{figure}

\begin{itemize}[leftmargin=*]
\item For varied tokens in the pseudo-winning response $\Tilde{y}_w$, we adapt the reward factor based on the learning process of the policy model.
Lower policy confidence indicates underdeveloped learning of the target behavior, signaling the need for additional focus to help the model better capture these nuances.
Consequently, we adjust the reward factor to be inversely proportional to the policy model's confidence, as formalized in Eq.~(\ref{equation:y_w}).

\item For varied tokens in the losing response $y_l$, we carefully adjust the reward factor by reinforcing already learned patterns of the policy model.
Intuitively, tokens in $y_l$ with higher confidence from the policy model may reflect inaccurate preference learning and therefore warrant stronger penalization.
However, our analysis reveals a distinct pattern of the policy model when processing $y_l$ compared to $\Tilde{y}_w$.
Specifically, when grouping varied tokens in $y_l$ into coarser-grained spans, the model’s confidence is significantly influenced by the token's position within these spans, as illustrated in Figure \ref{c6_fig: token_logp}.
We observe that the probabilities assigned to the initial token of incorrect spans in $y_l$ are typically low, whereas the probabilities for subsequent tokens within the same span are notably higher.
Prior studies have identified token probability as a critical signal for detecting anomalous behaviors~\cite{DBLP:conf/eacl/XiaoW21, DBLP:conf/acl/FadeevaRSPLMTKP24} and assessing generation quality~\cite{DBLP:conf/nips/YuanNL21, DBLP:conf/naacl/FuNJ024}.
Consistent with these findings, our results indicate that during training, the policy model can effectively recognize the onset of undesired spans by assigning low probabilities to initial tokens.
Nonetheless, due to the autoregressive dependencies, subsequent tokens within these spans receive higher probabilities, reflecting the contextual coherence established by preceding tokens, even when the span as a whole is incorrect.
Thus, while it is crucial to penalize initial tokens, applying equally strong penalties to subsequent tokens might be suboptimal, as they often maintain local coherence within the flawed span.
Therefore, we adjust the reward factor to also be inversely proportional to the policy model's confidence in Eq.~(\ref{equation:y_l}).
\end{itemize}

In a nutshell, our approach dynamically modulates the emphasis placed on critical tokens based on the policy model’s confidence.
This adaptive reward mechanism ensures that the model focuses on learning challenging distinctions while reinforcing already learned patterns, ultimately fostering a more nuanced and robust policy (see our analysis in \S\ref{c6_sec: quantitative_analysis_1}).
The formalization of our approach is encapsulated in Eq.~(\ref{equation:our_loss}), where $\lambda_{\Tilde{y}_w^t}$ and $\lambda_{y_l^t}$ adjust dynamically based on the policy's confidence, ensuring a tailored emphasis on critical tokens to improve the overall model performance.
\begin{align}
\mathcal{L}_{\text{DPO-BMC}}(\pi_{\theta}; \pi_{\text{ref}}) = 
- \E_{(x, \Tilde{y}_w, y_l)\sim \Tilde{\mathcal{D}}}  \left[ \log \sigma \left( \beta \sum_{\Tilde{y}_w^t \in \Tilde{y}_w} \lambda_{\Tilde{y}_w^t} \log \frac{\pi_{\theta}(\Tilde{y}_w^t \mid \Tilde{y}_w^{< t}, x)}{\pi_{\text{ref}}(\Tilde{y}_w^t \mid \Tilde{y}_w^{< t}, x)} \right. \right. \notag \\
\left. \left. - \beta \sum_{y_l^t \in y_l} \lambda_{y_l^t} \log \frac{\pi_{\theta}(y_l^t \mid y_l^{< t}, x)}{\pi_{\text{ref}}(y_l^t \mid y_l^{< t}, x)} \right) \right],
\label{equation:our_loss}
\end{align}
where
\begin{align}
\lambda_{\Tilde{y}_w^t} &=\left\{
\begin{array}{ll}
1 + \min \left( \mathit{sg} \left( \frac{1}{\pi_{\theta}(\Tilde{y}_w^t \mid \Tilde{y}_w^{< t}, x)} \right), \delta \right),& \text{if } \Tilde{y}_w^t \in \mathit{diff} (\Tilde{y}_w \mid y_l) \\
1,& \text{otherwise}
\end{array}
\right. 
\label{equation:y_w} \\
\lambda_{y_l^t} &=\left\{
\begin{array}{ll}
1 + \min \left( \mathit{sg} \left( \frac{1}{\pi_{\theta}(y_l^t \mid y_l^{< t}, x)} \right), \delta \right),& \text{if } y_l^t \in \mathit{diff} (y_l \mid \Tilde{y}_w) \\
1,& \text{otherwise}
\end{array}
\right.
\label{equation:y_l}
\end{align}
The $sg$ denotes the stop-gradient operator, the $\delta$ is an upper limit threshold that controls the emphasis on the rewards of the critical tokens, preventing overly aggressive updates. The $\mathit{diff} (\Tilde{y}_w \mid y_l)$ and $\mathit{diff} (y_l \mid \Tilde{y}_w)$ signify using the Levenshtein Distance algorithm to find the varied tokens in $\Tilde{y}_w$ and $y_l$, respectively. 

\paragraph{Gradient Analysis of DPO-BMC.}
For a mechanistic understanding of our method, we examine the gradients of the loss function $\mathcal{L}_{\text{DPO}}$ in Eq. (\ref{equation:sequence_level_dpo}) and $\mathcal{L}_{\text{DPO-BMC}}$ in Eq. (\ref{equation:our_loss}).
Their gradients with respect to the parameters $\theta$ can be written as:
\begin{align*}
\displaystyle \nabla_{\theta}\mathcal{L}_{\text{DPO}}(\pi_{\theta}; \pi_{\text{ref}}) = - \beta \E_{(x, y_w, y_l)\sim \mathcal{D}}
\left[
\sigma \left( \Delta_1 \right)
\left[
\underbrace{\displaystyle \nabla_{\theta} \log \pi_{\theta} (y_w \mid x)}_{\text{increase likelihood of } y_w} - \underbrace{\displaystyle \nabla_{\theta} \log \pi_{\theta} (y_l \mid x)}_{\text{decrease likelihood of } y_l}
\right]
\right],
\end{align*}
where
$\Delta_1 = \beta \log \frac{\pi_{\theta}(y_l \mid x)}{\pi_{\text{ref}}(y_l \mid x)} - \beta \log \frac{\pi_{\theta}(y_w \mid x)}{\pi_{\text{ref}}(y_w \mid x)}$.

\begin{align*}
&\displaystyle \nabla_{\theta}\mathcal{L}_{\text{DPO-BMC}}(\pi_{\theta}; \pi_{\text{ref}}) = 
- \beta \E_{(x, \Tilde{y}_w, y_l)\sim \Tilde{\mathcal{D}}}
\left[ \sigma \left( \Delta_2 \right)
\left( \underbrace{\displaystyle \nabla_{\theta} \log \pi_{\theta} (\Tilde{y}_w \mid x)}_{\text{increase likelihood of } \Tilde{y}_w} - \underbrace{\displaystyle \nabla_{\theta} \log \pi_{\theta} (y_l \mid x)}_{\text{decrease likelihood of } y_l} \right. \right. \\
&\left. \left. + \underbrace{\sum_{\Tilde{y}_w^t \in \mathit{diff} (\Tilde{y}_w \mid y_l)} (\lambda_{\Tilde{y}_w^t}-1) \displaystyle \nabla_{\theta} \log \pi_{\theta}(\Tilde{y}_w^t \mid \Tilde{y}_w^{< t}, x)}_{\text{increase likelihood of desired tokens of } \Tilde{y}_w}
- \underbrace{\sum_{y_l^t \in \mathit{diff} (y_l \mid \Tilde{y}_w)} (\lambda_{y_l^t}-1) \displaystyle \nabla_{\theta} \log \pi_{\theta}(y_w^t \mid y_l^{< t}, x)}_{\text{decrease likelihood of undesired tokens of } y_l}
\right) \right],
\end{align*}
where
$\Delta_2 = \beta \sum_{y_l^t \in y_l} \lambda_{y_l^t} \log \frac{\pi_{\theta}(y_l^t \mid y_l^{< t}, x)}{\pi_{\text{ref}}(y_l^t \mid y_l^{< t}, x)} - \beta \sum_{\Tilde{y}_w^t \in \Tilde{y}_w} \lambda_{\Tilde{y}_w^t} \log \frac{\pi_{\theta}(\Tilde{y}_w^t \mid \Tilde{y}_w^{< t}, x)}{\pi_{\text{ref}}(\Tilde{y}_w^t \mid \Tilde{y}_w^{< t}, x)}$.

In contrast to vanilla DPO, which emphasizes sequence-level optimization exclusively, \textbf{our proposed method integrates both sequence-level and token-level perspectives}.
(1) At the sequence level, we promote preferred completions while penalizing those that are disfavored.
(2) At the token level, we further refine the rewards of critical desired and undesired tokens of $\Tilde{y}_w$ and $y_l$, respectively.
This dual consideration ensures that both the overall sequence structure and the critical token choices are optimized for the desired outcome.

\section{Experiments}

\subsection{Experimental Setup}
We conduct a comprehensive evaluation across three downstream scenarios, including question answering (QA), mathematical reasoning, and instruction following (IF).
The detailed data statistics as well as the evaluation metrics are listed in Table \ref{c6_tab: dataset_statistics} of Appendix \ref{c6_appendix: data_statistics}.

\paragraph{Models and Training Settings.} For the QA and mathematical reasoning setup, we utilize Llama2-7B-base~\cite{touvron2023llama} in our experiments.
Dealing with these tasks necessitates LLMs to possess domain-specific knowledge and engage in systematic, step-by-step reasoning to reach the ultimate answer.
Therefore, following prior works~\cite{chen2024rlmec, chen2024allo}, we fine-tune Llama2-7B-base on the training set of ECQA~\cite{agg2021ecqa} and QASC~\cite{khot2020qasc} for QA, and fine-tune Llama2-7B-base on MetaMathQA~\cite{yu2023metamath} for mathematical reasoning. We denote the fine-tuned LLM as SFT and use it as the backbone for preference optimization.
In line with prior research~\cite{chen2024rlmec, chen2024allo}, we construct preference pairs $(y_w, y_l)$ based on the training data, by using the ground truth as $y_w$ and the SFT model's inference output as $y_l$.
For the instruction following setup, we utilize Llama3-8B-base~\cite{llama3modelcard} and Mistral-7B-Base~\cite{jiang2023mistral} in our experiments.
Following the training pipeline of Zephyr~\cite{tunstall2023zephyr} and SimPO~\cite{meng2024simpo}, we train a base model on the UltraChat-200k dataset~\cite{ding2023ultrachat} to obtain an SFT model.
Then, we use the SFT model as the starting point and perform preference optimization on the UltraFeedback dataset~\cite{cui2023ultrafeedback}, where $y_w$ and $y_l$ are collected from LLMs of varying quality.  

During our Bridging Phase, we utilize \texttt{gpt-4-0125-preview} for targeted modification to obtain $\Tilde{y}_w$, based on the prompt template in Appendix \ref{c6_appendix: prompt}.
We also demonstrate in Table \ref{c6_tab: revisor} that \textbf{a less powerful open-source} LLM, such as \texttt{Llama3-70B-Instruct}, can acquire comparable results.
During our Modeling Phase, we list the implementation details in Appendix \ref{c6_appendix: hyperparameter} for reproducibility.
A comprehensive cost analysis in \S \ref{c6_sec: cost_analysis} confirms that \textbf{the computational overhead introduced by our BMC pipeline is minimal}.

\paragraph{Evaluation Benchmarks.}
In question answering, we adopt the test splits of ECQA~\cite{agg2021ecqa}, QASC~\cite{khot2020qasc}, OpenbookQA~\cite{miha2018obqa}, and StrategyQA~\cite{geva2021StrategyQA} for evaluation.
In mathematical reasoning, we conduct the evaluation on four challenge datasets including GSM8k~\cite{cobbe2021gsm8k}, MATH~\cite{hendrycksmath2021}, MAWPS~\cite{rik2016mawps}, and TabMWP~\cite{lu2023tabmwp}.
In instruction following, We assess our models using two of the most popular open-ended instruction-following benchmarks: AlpacaEval 2~\cite{alpaca_eval} and Arena-Hard v0.1~\cite{li2024live}.
Both benchmarks evaluate the models' versatile conversational abilities across a diverse set of queries.
For each query, the evaluated model's response and the reference model's response are
compared head-to-head using an auto-evaluator.
We use the officially recommended configurations\footnote{AlpacaEval: \url{https://github.com/tatsu-lab/alpaca\_eval}. Arena-Hard v0.1: \url{https://github.com/lm-sys/arena-hard-auto}.} during the evaluation.

\paragraph{Baselines.}
We compare our approach with various powerful \textit{offline} preference optimization methods, including FIGA~\cite{guo2024figa}, DPO~\cite{rafailov2024dpo}, and DPO variants (IPO~\cite{moh2024ipo}, ORPO~\cite{hong2024opro}, R-DPO~\cite{park2024rdpo}, and SimPO~\cite{meng2024simpo}).
The training objectives of these methods are listed in Table \ref{c6_tab: hyper_po}.
Besides, we include two additional baselines: (1) \textbf{DPO (CW)}: enhancing pairwise data correlation by prompting the SFT model to \textbf{C}ontinue \textbf{W}riting a prefix of the winning response to generate the losing one; (2) \textbf{DPO (EW)}: leveraging an off-the-shelf LLM for \textbf{E}xternal \textbf{W}eighting of token-level reward~\cite{DBLP:conf/icml/0001PMMFLBHCRP24}, where LLM scores each token in the winning and losing responses based on how much it improves or decreases the overall quality.

\subsection{Experimental Results}

\paragraph{Our Method Consistently and Significantly Outperforms Baselines.}
As presented in Table \ref{c6_tab: qa_math}, our model DPO-BMC consistently achieves state-of-the-art results across all evaluated QA and math benchmarks.
Specifically, DPO-BMC outperforms DPO by 3.8 absolute points on QA tasks and by 1.3 points on math tasks.
On instruction-following tasks (Table \ref{c6_tab: instruction_following}), DPO-BMC secures the highest length-controlled win rate, surpassing DPO by over 5 points across various settings, with even greater gains for larger base models (Appendix \ref{c6_appendix: larger_model}).
\begin{table}[t]
\begin{center}
\resizebox{\textwidth}{!}{%
\begin{tabular}{l ccccc ccccc}
\toprule
\multirow{2}{*}{\textbf{Method}} & \multicolumn{5}{c}{\textbf{Question-Answering Tasks}} & \multicolumn{5}{c}{\textbf{Mathematical Reasoning Tasks}} \\
\cmidrule(lr){2-6} \cmidrule(lr){7-11}
& \textbf{ECQA} & \textbf{QASC} & \textbf{OBQA} & \textbf{StrategyQA} & \textbf{Avg.} & \textbf{GSM8k} & \textbf{MATH} & \textbf{MAWPS} & \textbf{TabMWP} & \textbf{Avg.} \\
\midrule
SFT & 72.8 & 54.5 & 51.8 & 56.9 & 59.0 & 55.8 & 11.6 & 80.3 & 42.8 & 47.6 \\
\midrule
FIGA & 70.3 & 52.5 & 51.7 & 48.6 & 55.8 & 54.1 & 9.8 & 75.5 & 39.0 & 44.6  \\
IPO & 71.5 & 58.9 & 53.6 & 58.4 & 60.6 & 57.2 & 12.1 & 82.2 & 42.5 & 48.5 \\
OPRO & 69.8 & 55.1 & 51.4 & 57.2 & 58.4 & 56.0 & 12.4 & 80.8 & 41.3 & 47.6 \\
R-DPO & 73.5 & 59.5 & 55.4 & 58.8 & 61.8 & 56.9 & 12.0 & 81.9 & 42.2 & 48.2 \\
SimPO & 71.9 & 56.7 & 52.2 & 55.4 & 59.1 & 57.5 & 12.7 & 81.8 & \underline{43.5} & 48.9 \\
DPO & 73.1 & 58.8 & 55.6 & 57.8 & 61.3 & 56.3 & 12.3 & 81.2 & 43.4 & 48.3 \\ 
DPO (CW) & 72.5 & 58.6 & 55.2 & 57.3 & 60.9 & 55.9 & 11.8 & 80.7 & 42.8 & 47.8 \\ 
DPO (EW) & 72.9 & 59.4 & 55.8 & 57.9 & 61.5 & 56.5 & 12.0 & 80.9 & 43.4 & 48.2 \\ \midrule
DPO-BMC & \textbf{75.9} & \textbf{63.0} & \textbf{60.4} & \textbf{61.0} & \textbf{65.1} & \textbf{58.4} & \textbf{13.0} & \textbf{83.1} & \textbf{43.8} & \textbf{49.6} \\
DPO-BC & \underline{75.7} & \underline{62.0} & 56.0 & \underline{60.1} & \underline{63.4} & \underline{57.6} & \underline{12.7} & \underline{82.8} & 43.4 & \underline{49.1} \\
DPO-MC & 74.8 & 60.0 & \underline{56.4} & 58.8 & 62.5 & 57.2 & 12.5 & 82.4 & 43.0 & 48.8  \\
\bottomrule
\end{tabular}
}
\end{center}
\caption{Experimental results (based on Llama2-7B-base) on question answering tasks and mathematical reasoning tasks. ``Avg.'' is the average accuracy of all sub-tasks. In each column, the highest score is \textbf{bolded} and the second-highest is \underline{underlined}.}
\label{c6_tab: qa_math}
\end{table}
\begin{table}[!t]
\begin{center}
\resizebox{\textwidth}{!}{%
\begin{tabular}{lccccccccccc}
\toprule
\multirow{3}{*}{\textbf{Method}} & \multicolumn{5}{c}{\textbf{Llama3-8B-Base}} & \multicolumn{5}{c}{\textbf{Mistral-7B-Base}} \\
\cmidrule(lr){2-6} \cmidrule(lr){7-11}
& \multicolumn{3}{c}{\textbf{AlpacaEval 2}} & \multicolumn{2}{c}{\textbf{Arena-Hard}} & \multicolumn{3}{c}{\textbf{AlpacaEval 2}} & \multicolumn{2}{c}{\textbf{Arena-Hard}} \\
\cmidrule(lr){2-4} \cmidrule(lr){5-6} \cmidrule(lr){7-9} \cmidrule(lr){10-11}
& \textbf{LC (\%)} & \textbf{WR (\%)} & \textbf{Avg. len} & \textbf{WR (\%)} & \textbf{Avg. len}
& \textbf{LC (\%)} & \textbf{WR (\%)} & \textbf{Avg. len} & \textbf{WR (\%)} & \textbf{Avg. len} \\
\midrule
SFT & 7.5 & 4.7 & 956 & 2.6 & 414 & 8.1 & 5.9 &998  & 2.2 &454  \\ \midrule
FIGA & 8.4 & 4.2 & 1,199 & 5.1 & 416 & 7.0 & 4.9 & 1,378 & 2.5 & 461 \\
IPO & 13.4 & 9.8 &1,430 & 14.0 &477 & 12.5 & 10.8 &1,588 & 8.5 & 522 \\
ORPO & 12.5 & 11.4 &1,793 & 11.7 &573 & 14.5 & 11.5 &1,630 & 9.4 & 566 \\
R-DPO & 17.1 & 14.4 &1,801 & 17.6 &582 & 16.0 & 12.3 &1,521 & 10.4 & 529 \\
SimPO & \underline{21.3} & \textbf{18.9} & 1,718 & \textbf{26.6} & 562 & 16.8 & \underline{14.4} & 1,906 & \textbf{18.4} & 615 \\
DPO & 16.0 & 14.8 & 1,713 & 17.6 & 559 & 15.1 & 13.3 & 1,657 & 13.6 & 540 \\
DPO (CW) & 15.2 & 14.0 & 1,756 & 17.1 & 570 & 14.5 & 12.9 & 1,647 & 13.0 & 532 \\
DPO (EW) & 17.2 & 15.6 & 1,702 & 18.2 & 566 & 15.3 & 13.4 & 1,668 & 13.9 & 549 \\
 \midrule
DPO-BMC & \textbf{22.4} & \underline{16.8} & 1,285 & \underline{18.1} & 406 &\textbf{20.8} & \textbf{16.6} & 1,317 & \underline{17.6} & 488 \\
DPO-BC & 20.6 & 14.4 & 1,269 & 16.8 & 422 & \underline{18.6} & 13.8 & 1,489 & 15.9 & 502 \\
DPO-MC & 17.7 & 15.2 & 1,890 & 17.9 & 579 & 16.4 & 14.3 & 1,712 & 15.4 & 551 \\
\bottomrule
\end{tabular}
}
\end{center}
\caption{Experimental results on instruction-following tasks. ``LC'' is the length-controlled win rate, and ``WR'' is the raw win rate. ``Avg. len'' denotes the average number of tokens in the responses.}
\label{c6_tab: instruction_following}
\end{table}
The length-controlled win rate~\cite{dubois2024length} serves as a robust metric that mitigates the effects of length bias, thereby providing a more reliable evaluation of LLM-based auto-annotation.
Notably, \textbf{DPO-BMC generates responses that are significantly more concise than other baselines}.
As highlighted in Table \ref{c6_tab: instruction_following}, the average response length of DPO-BMC and DPO-BC is approximately 75\% of that produced by DPO and DPO-MC.
This attribute of length normalization is credited to the correlated preference data we constructed, which directs optimization towards critical desired behaviors rather than verbosity.

\subsection{Ablation Study}

\paragraph{Both Key Designs in BMC are Crucial.}
In Table \ref{c6_tab: qa_math} and Table \ref{c6_tab: instruction_following}, we additionally present results from ablating each key design element of DPO-BMC:
\begin{itemize}[leftmargin=*]
\item \textbf{DPO-BC}: Training using DPO's original objective on our constructed preference data.
\item \textbf{DPO-MC}: Training using our proposed objective in Eq.~(\ref{equation:our_loss}) on the original preference data.
\end{itemize}
Our examination reveals several key findings:
(1) DPO (CW), the ``Continue Writing'' approach, slightly underperforms standard DPO, as it introduces superficial correlations that fail to capture the nuanced, task-specific alignments essential for effective optimization.
In contrast, our Bridging Phase explicitly enhances \textit{informative correlations}—elucidate fine-grained distinctions between desired and undesired behaviors through token-level variations. This targeted focus significantly improves model performance;
(2) Even when leveraging identical training preference data, our designed optimization objective consistently outperforms both DPO and DPO (EW), highlighting its superior ability to model fine-grained correlations based on the dynamic of the policy model's confidence;
and (3) Combining our constructed data with our designed objective yields the best results, affirming the inseparability of the Bridging Phase and the Modeling Phase.

\paragraph{Influence of Data Synthesis Method.}
Table \ref{c6_tab: data_synthesis_ablation} shows the effects of various data synthesis strategies during the Bridging Phase.
When generating $\Tilde{y}_w$ without referring to $y_w$, LLMs potentially make erroneous modifications that misalign with the intended target, leading to a performance drop.
An alternative approach that attempts to enhance the correlation between winning and losing responses is to degenerate $y_w$ to $\Tilde{y}_l$ and utilize $(y_w, \Tilde{y}_l)$ as the preference pair. 
However, this approach also falls short, likely because LLMs are primarily trained to generate high-quality data, making it challenging for them to generate low-quality outputs that mimic the nuanced errors of losing responses.
Semantic similarity analysis using the \texttt{all-mpnet-base-v2} embedding model\footnote{\url{https://huggingface.co/sentence-transformers/all-mpnet-base-v2}} supports this, showing a high score of 0.88 for $(y_w, \tilde{y_w})$ but only 0.73 for $(y_l, \tilde{y_l})$.

\begin{figure}[!t]
\begin{minipage}{0.58\textwidth}
\centering
\scriptsize
\begin{tabular}{lcccc}
\toprule
\textbf{Method} & \textbf{LLM for Targeted Modification} & \textbf{QA} & \textbf{Math} & \textbf{IF} \\ \midrule
SFT & -- & 56.9 & 47.6 & 7.5 \\ \midrule
DPO & -- & 61.3 & 48.3 & 16.0 \\ \midrule
DPO-BMC & Llama3-70B-Instruct & 64.6 & 49.4 & 21.8 \\
DPO-BMC & gpt-4-0125-preview & \textbf{65.1} & \textbf{49.6} & \textbf{22.4} \\
\bottomrule
\end{tabular}
\captionof{table}{Influence of diverse LLMs for targeted modification in the Bridging Phase. The average accuracy is presented for QA and Math. LC on AlpacaEval 2 is reported for instruction following (IF), based on Llama3-8B.}
\label{c6_tab: revisor}
\end{minipage}
\hfill
\begin{minipage}{0.4\textwidth}
\centering
\includegraphics[width=0.95\textwidth]{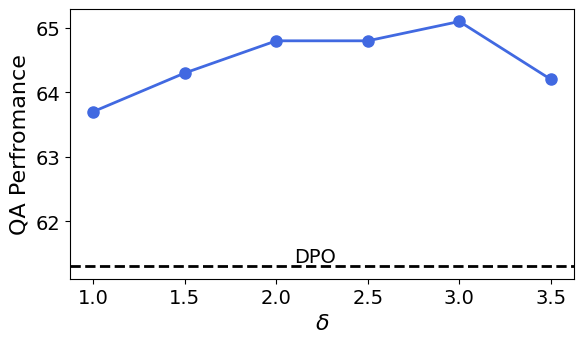}
\caption{Ablation study on $\delta$ in the Modeling Phase. The average accuracy is presented as the QA performance.}
\label{c6_fig: tau_ablation}
\end{minipage}
\end{figure}

{\color{black}
\paragraph{Influence of Data Modification Proportion.}
Figure \ref{c6_fig: ablation} illustrates the impact of data modification proportions during the Bridging Phase on performance. Increasing modifications from 0\% to 20\% yields the most substantial gains, highlighting the effectiveness of enhancing pairwise preference correlations. Performance plateaus beyond 80\% modifications, indicating that extensive changes are beneficial but not essential, offering flexibility under computational or data constraints. These results demonstrate the scalability and adaptability of our framework for diverse applications.

\begin{figure}[!t]
\begin{minipage}{0.5\textwidth}
\centering
\scriptsize
\begin{tabular}{lcccc}
\toprule
\textbf{Data Synthesis} & \textbf{Training Data} & \textbf{QA} & \textbf{Math} & \textbf{IF} \\ \midrule
$y_l \xrightarrow{y_w} \Tilde{y}_w$ (ours) & $(\Tilde{y}_w, y_l)$ & \textbf{65.1} & \textbf{49.6} & \textbf{22.4} \vspace{1.5ex} \\
$y_l \xrightarrow{\hspace{0.32cm}} \Tilde{y}_w$ & $(\Tilde{y}_w, y_l)$ & 64.3 & 49.2 & 19.8 \vspace{1ex} \\
$y_w \xrightarrow{y_l} \Tilde{y}_l$ & $(y_w, \Tilde{y}_l)$ & 64.6 & 48.7 & 18.9 \vspace{1.5ex} \\
$y_w \xrightarrow{\hspace{0.25cm}} \Tilde{y}_l$ & $(y_w, \Tilde{y}_l)$ & 63.9 & 48.6 & 17.6 \\
\bottomrule
\end{tabular}
\captionof{table}{Ablation study on diverse data synthesis methods in the Bridging Phase. The average accuracy is presented for QA and Math. LC on AlpacaEval 2 is reported for instruction following (IF), based on Llama3-8B.}
\label{c6_tab: data_synthesis_ablation}
\end{minipage}
\hfill
\begin{minipage}{0.48\textwidth}
\centering
\includegraphics[width=\textwidth]{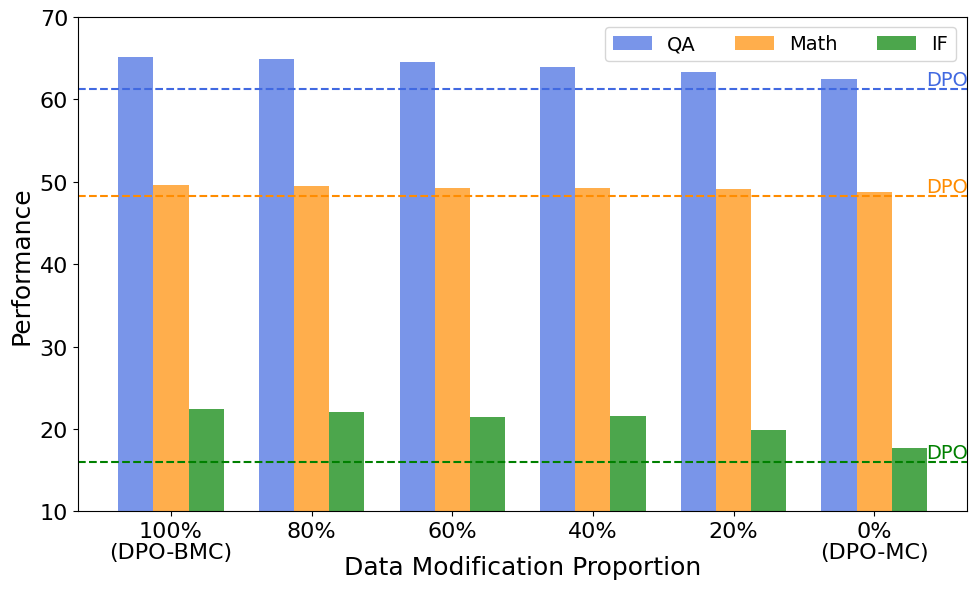}
\caption{Ablation study on data modification proportion in the Bridging Phase.}
\label{c6_fig: ablation}
\end{minipage}
\end{figure}

\paragraph{Influence of LLMs for Targeted Modification.}
Table \ref{c6_tab: revisor} explores the influence of diverse LLMs for targeted modification.
Notably, substituting the \texttt{gpt-4-0125-preview} model with a less powerful yet open-source alternative, such as \texttt{Llama3-70B-Instruct}, \textbf{yields comparable performance while significantly surpassing vanilla DPO}.
This finding underscores the adaptability of our method to varying levels of model sophistication, thereby reducing dependence on commercial LLMs without significant impact on final model performance.
}

\paragraph{Influence of $\delta$.}
We conduct an ablation study to examine the influence of the threshold $\delta$ in the DPO-BMC objective on model performance, as shown in Figure \ref{c6_fig: tau_ablation}.
Setting $\delta = 1.0$ reduces our method to one that assigns fixed token-level rewards, leading to suboptimal accuracy.
As $\delta$ increases, the model performance improves, with the optimal setting observed around $\delta = 3.0$.
However, further increasing $\delta$ results may degrade model performance due to excessively aggressive gradient updates on certain tokens.
Notably, across all tested values of $\delta$, our method consistently outperforms the DPO baseline, indicating its robustness and effectiveness in stabilizing the learning process.

\section{Analysis}
In this section, we begin with the cost analysis of our proposed BMC framework (\S\ref{c6_sec: cost_analysis}).
Furthermore, we conduct in-depth quantitative analyses to elucidate why our method outperforms DPO (\S\ref{c6_sec: quantitative_analysis_1} and \S\ref{c6_sec: quantitative_analysis_2}).
Finally, we demonstrate the versatility of our framework by adapting it to other DPO variants (\S\ref{c6_sec: versatility}).

\subsection{Cost Analysis of Bridging and Modeling Phase}
\label{c6_sec: cost_analysis}

\subsubsection{Cost of Bridging Phase}
\label{c6_sec: cost_bridging}

{\color{black}
The Bridging Phase, responsible for synthesizing pseudo-winning responses, operates exclusively \textbf{offline}, meaning it incurs no runtime cost during model training. The data synthesis process is designed to be efficient, as it does not require iterative computations or model updates.

For context, we estimated the budget for data synthesis using the \texttt{gpt-4-0125-preview} API, based on the API's pricing of \$0.01 per 1K input tokens and \$0.03 per 1K output tokens.
Table \ref{c6_tab: api_cost} lists the breakdown of the estimated costs for our three evaluated tasks, which demonstrates that this is a manageable expenditure.
}

\begin{table}[t!]
\begin{center}
\small
\begin{tabular}{lcccc}
\toprule
\textbf{Task} & \textbf{\# of Samples} & \textbf{Avg. Input Token Length} & \textbf{Avg. Output Token Length} & \textbf{Cost (\$)} \\ \midrule
QA & 15,732 & 206 & 25 & 	44.21 \\
Math & 40,000 & 429 & 47 & 228.00 \\
IF & 61,135 & 728 & 235 & 876.06 \\
\bottomrule
\end{tabular}
\end{center}
\caption{Estimated budget for data synthesis using the \texttt{gpt-4-0125-preview} API.}
\label{c6_tab: api_cost}
\end{table}

\paragraph{Can an Open-Source LLM be Utilized as an Alternative?}
In Table \ref{c6_tab: revisor}, we explore the impact of LLMs on targeted modifications during the Bridging Phase.
Our findings indicate that substituting the \texttt{gpt-4-0125-preview} model with a less powerful yet open-source alternative, such as \texttt{Llama3-70B-Instruct}, \textbf{yields comparable performance while significantly surpassing vanilla DPO}.
The \texttt{Llama3-70B-Instruct} model can be deployed on only 2 NVIDIA-3090 GPUs, with the option to further reduce hardware requirements through low-bit quantization\footnote{\url{https://github.com/ollama/ollama}}. This provides an economical alternative for our Bridging Phase without compromising performance.
Numerous studies have highlighted the superior text modification capabilities of LLMs. For example, LLMs have been effectively employed in synthesizing high-quality data \cite{wang-etal-2023-self-instruct}.
Additionally, \cite{ji2024aligner} show that LLMs can transform initial outputs from upstream models into more helpful and benign responses, thereby aligning generated content with human intentions.
In conclusion, our framework demonstrates robustness in leveraging diverse LLMs for targeted modifications, confirming its adaptability and effectiveness.

\subsubsection{Cost of Modeling Phase}

{\color{black}
Our Modeling Phase adds minimal computational overhead compared to vanilla DPO. Specifically:

\begin{itemize}[leftmargin=*]
\item \textbf{Token Difference Identification}: Using a dynamic programming algorithm (edit distance) to identify differing tokens between the pseudo-winning and losing responses. This is a lightweight operation and introduces negligible runtime cost.
\item \textbf{Reward Weighting Calculation}: We calculate a weighting factor based on the policy model's probability of the identified tokens, which is already computed in the standard DPO setup. Because we halt gradient backpropagation for the weighting factor, this operation does not introduce additional computational costs.
\end{itemize}
Table \ref{c6_tab: run_time_cost} demonstrates the comparison of the training times between DPO and DPO-BMC on 4×A800 GPUs, illustrating that \textbf{DPO-BMC increases training time by less than 1\% across all evaluated tasks}.

\begin{table}[h!]
\begin{center}
\begin{tabular}{lcccc}
\toprule
\textbf{Task} & \textbf{Base Model} & \textbf{Runtime of DPO (s)} & \textbf{Runtime of DPO-BMC (s)} & \textbf{Increase (\%)} \\ \midrule
QA	&Llama2-7B	&2,831	&2,850	&\textbf{0.67\%} \\
Math	&Llama2-7B	&9,586	&9,641	&\textbf{0.57\%} \\
IF	&Llama3-8B	&16,179	&16,318	&\textbf{0.86\%} \\
\bottomrule
\end{tabular}
\end{center}
\caption{Runtime usage for DPO and DPO-BMC during the Modeling Phase.}
\label{c6_tab: run_time_cost}
\end{table}

Overall, these results validate that the computational overhead introduced by BMC is minimal, and the approach is highly efficient in terms of runtime, making it practical for real-world applications without significantly increasing resource requirements.
}

\subsection{Quantitative Analysis of Bridging and Modeling Phase}
\label{c6_sec: quantitative_analysis_1}
\begin{figure}[!t]
\begin{center}
\includegraphics[width=\linewidth]{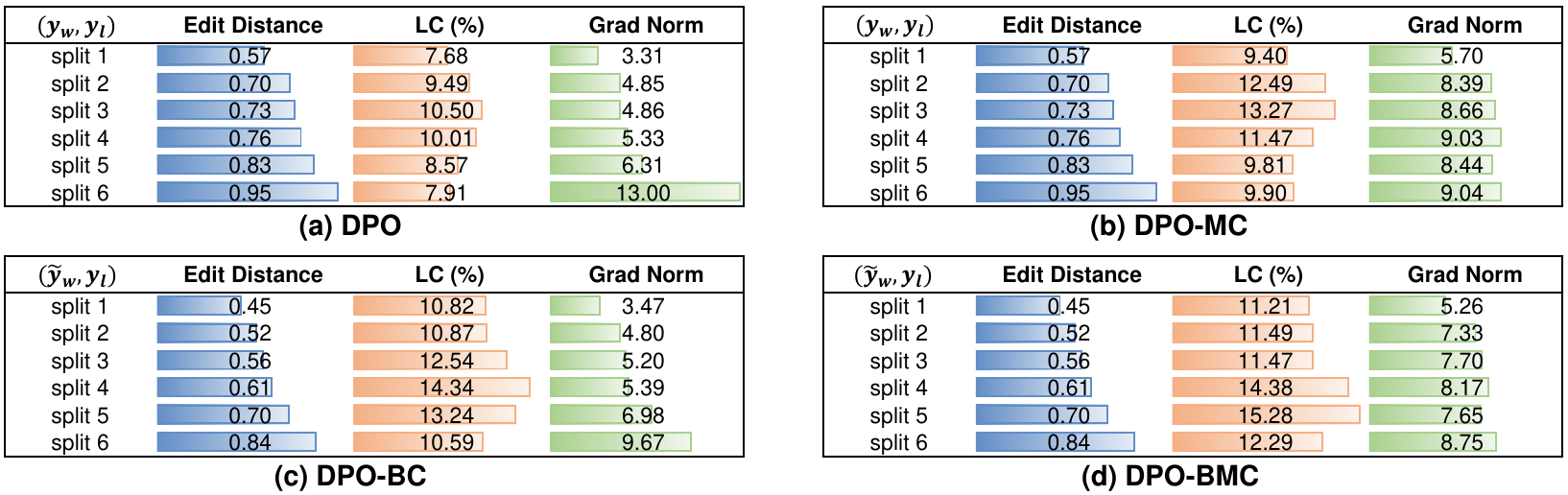}
\end{center}
\caption{We segment the 60k training data of UltraFeedback into six equal-sized splits based on increasing edit distance between winning and losing responses. For each split, we report LC on AlpacaEval 2 and the average gradient norm during training.}
\label{c6_fig: data_split}
\end{figure}

To rigorously assess the effectiveness of the two pivotal phases in our framework, we segment the 60k training data of UltraFeedback into six equal-sized splits, ordered by increasing edit distance between winning and losing responses.
For each split, we also construct its corresponding $(\Tilde{y}_w, y_l)$ pair data through our Bridging Phase.
We then train four models—--(a) DPO, (b) DPO-MC, (c) DPO-BC, and (d) DPO-BMC---on each split based on Llama3-8B, with identical hyperparameters to ensure comparability.
As shown in Figure \ref{c6_fig: data_split}, the Bridging Phase successfully decreases the edit distance between pairwise data through targeted modification, shifting the optimization process to concentrate on the most critical differences in preference data.
This phase consistently enhances performance across all splits by refining training data for more focused learning.
Another notable observation is the average gradient norm during DPO training increases as the edit distance between pairwise data enlarges, reflecting the sensitivity of DPO's training process to individual data points and potential gradient variance.
Our proposed Modeling Phase mitigates the variance by dynamically adjusting the training process based on the policy model's confidence.
This adaptive mechanism prioritizes challenging distinctions while reinforcing learned patterns, promoting a balanced optimization landscape with diverse training data (See Appendix \ref{c6_appendix: kl_div} for further analysis).

\subsection{Quantitative Analysis of Credit Assignment}
\label{c6_sec: quantitative_analysis_2}
We compare the token-level and sequence-level credits assigned by DPO and DPO-BMC, assessing how well their final learned rewards align with preference labels on a held-out set of UltraFeedback.

\begin{figure}[!t]
\begin{center}
\includegraphics[width=\linewidth]{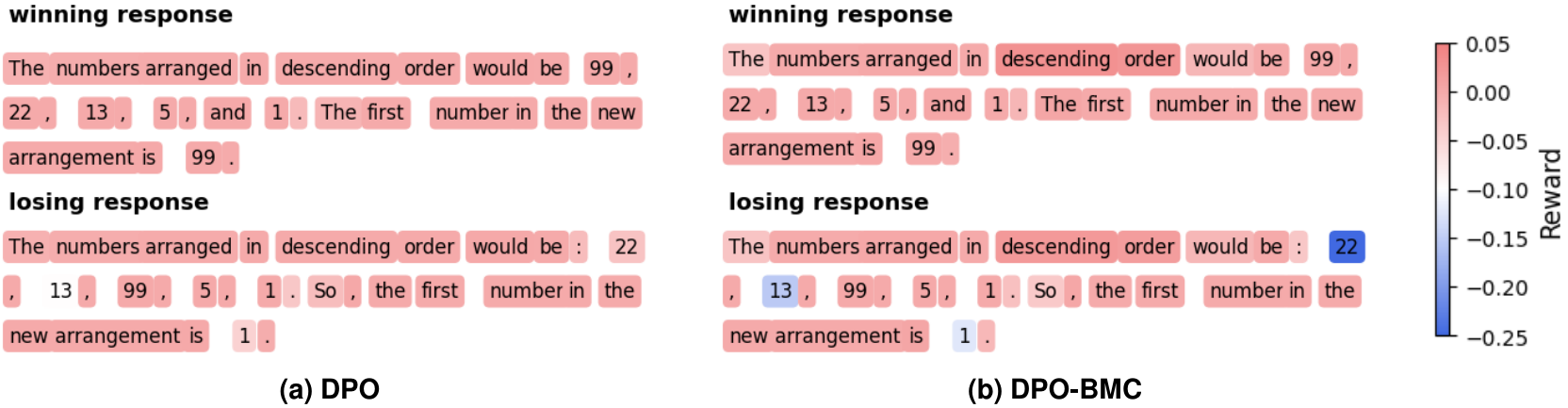}
\end{center}
\caption{Visualization of token-level rewards assigned by DPO and our method. The preference pair is sampled from the held-out set of UltraFeedback, whose input prompt is ``\textit{Arrange the numbers 5, 13, 99, 1, and 22 in descending order. What is the first number in the new arrangement?}''}
\label{c6_fig: token_reward}
\end{figure}

\paragraph{Analysis on Token-level Reward.}
Figure \ref{c6_fig: token_reward} depicts the token-level reward assignment for DPO and DPO-BMC on a response pair consisting of a winning response $y_w$ and a losing response $y_l$.
The reward of each token is computed as $r_\theta(x, y^t)=\beta\log \frac{\pi_{\theta}(y^t \mid y^{<t}, x)}{\pi_{\text{ref}}(y^t \mid y^{<t}, x)}$.
From the figure, we observe that: (1) DPO assigns nearly uniform rewards across tokens, failing to differentiate the importance of tokens to the overall response quality; and (2) although DPO can identify and assign lower rewards to several erroneous tokens in the losing response (\textit{e.g.}, ``13''), it struggles to capture subtle distinctions between the winning and losing responses.
In contrast, DPO-BMC assigns higher rewards to critical tokens (\textit{e.g.}, ``descending order'') and effectively penalizes incorrect tokens in the losing response.
These results demonstrate DPO's limitations in providing precise token-level preferences on sentence quality, and our method can effectively alleviate this issue.

\paragraph{Analysis on Sequence-level Reward.}
For a rigorous comparison, we calculate the sequence-level DPO reward expression by $r_\theta(x, y)=\beta\log \frac{\pi_{\theta}(y \mid x)}{\pi_{\text{ref}}(y \mid x)}$.
The reward margin is determined by $r_\theta(x, y_w)-r_\theta(x, y_l)$.
Reward accuracy is defined as the percentage of preference pairs where the winning response achieves a higher reward than the losing response, \textit{i.e.}, $r_\theta(x, y_w)>r_\theta(x, y_l)$.
Our findings show that DPO-BMC outperforms DPO in terms of average reward margin (\textbf{0.74 vs. 0.54}) and reward accuracy (\textbf{73.60 vs. 72.19}).
This enhancement validates our method's superior ability to discern subtle differences between preference pairs, enabling more effective generalization.

\subsection{Versatility of Our Framework}
\label{c6_sec: versatility}

Our BMC framework demonstrates versatility and can be seamlessly integrated with various DPO variants. As shown in Table \ref{c6_tab: adaption}, the \texttt{X}PO-BMC methods consistently outperform their corresponding \texttt{X}PO baselines across a diverse set of tasks, including QA, Math, and Instruction Following (IF). For instance, IPO-BMC achieves a significant improvement in QA accuracy (64.1 vs. 60.6) and IF score (15.7 vs. 13.4) compared to IPO. Similarly, ORPO-BMC, R-DPO-BMC, SimPO-BMC, and DPO-BMC exhibit higher performance in QA and Math, alongside notable gains in IF, such as R-DPO-BMC improving the IF score from 17.1 to 20.0 over R-DPO. These results highlight the robustness of our framework in enhancing task-specific performance across various settings, reaffirming its potential as a generalizable enhancement to existing DPO methodologies.

\begin{table}[h]
\centering
\small
\begin{tabular}{lccc}
\toprule
\textbf{Method} & \textbf{QA} & \textbf{Math} & \textbf{IF} \\ \midrule
SFT & 56.9 & 47.6 & 7.5 \\ \midrule
IPO & 60.6 & 48.3 & 13.4 \\
IPO-BMC & \textbf{64.1} & \textbf{48.6} & \textbf{15.7} \\ \midrule
ORPO & 58.4 & 47.6 & 12.5 \\
ORPO-BMC & \textbf{62.3} & \textbf{48.4} & \textbf{15.7} \\ \midrule
R-DPO & 61.8 & 48.2 & 17.1 \\
R-DPO-BMC & \textbf{65.3} & \textbf{49.5} & \textbf{20.0} \\ \midrule
SimPO & 59.1 & 48.9 & 21.3 \\
SimPO-BMC & \textbf{61.6} & \textbf{49.0} & \textbf{21.9} \\ \midrule
DPO & 61.3 & 48.3 & 16.0 \\
DPO-BMC & \textbf{65.1} & \textbf{49.6} & \textbf{22.4} \\
\bottomrule
\end{tabular}
\captionof{table}{Versatility of our framework across various \texttt{X}POs..}
\label{c6_tab: adaption}
\end{table}

\section{Conclusion}
In this chapter, we propose BMC, an effective framework for bridging and modeling correlations in pairwise data for direct preference optimization.
BMC equips LLMs with better human value alignment through a two-phase process: a Bridging Phase that enhances correlations between pairwise data by explicitly manifesting fine-grained preference signals via targeted modifications, and a Modeling Phase that learns token-level correlations by dynamically leveraging the the policy model's confidence during training. 
Our framework exhibits superior performance in question-answering, mathematical reasoning, and instruction-following tasks, consistently surpassing the baseline DPO by a significant margin.
Extensive analysis highlights that the key designs in BMC are crucial and validates the effectiveness and versatility of BMC.
\chapter{Alignment Evaluation from the Perspective of Constraints Following}\label{chap:evaluation}

\section{Introduction}

\begin{figure}[!htbp]
\centering
\includegraphics[width=0.8\linewidth]{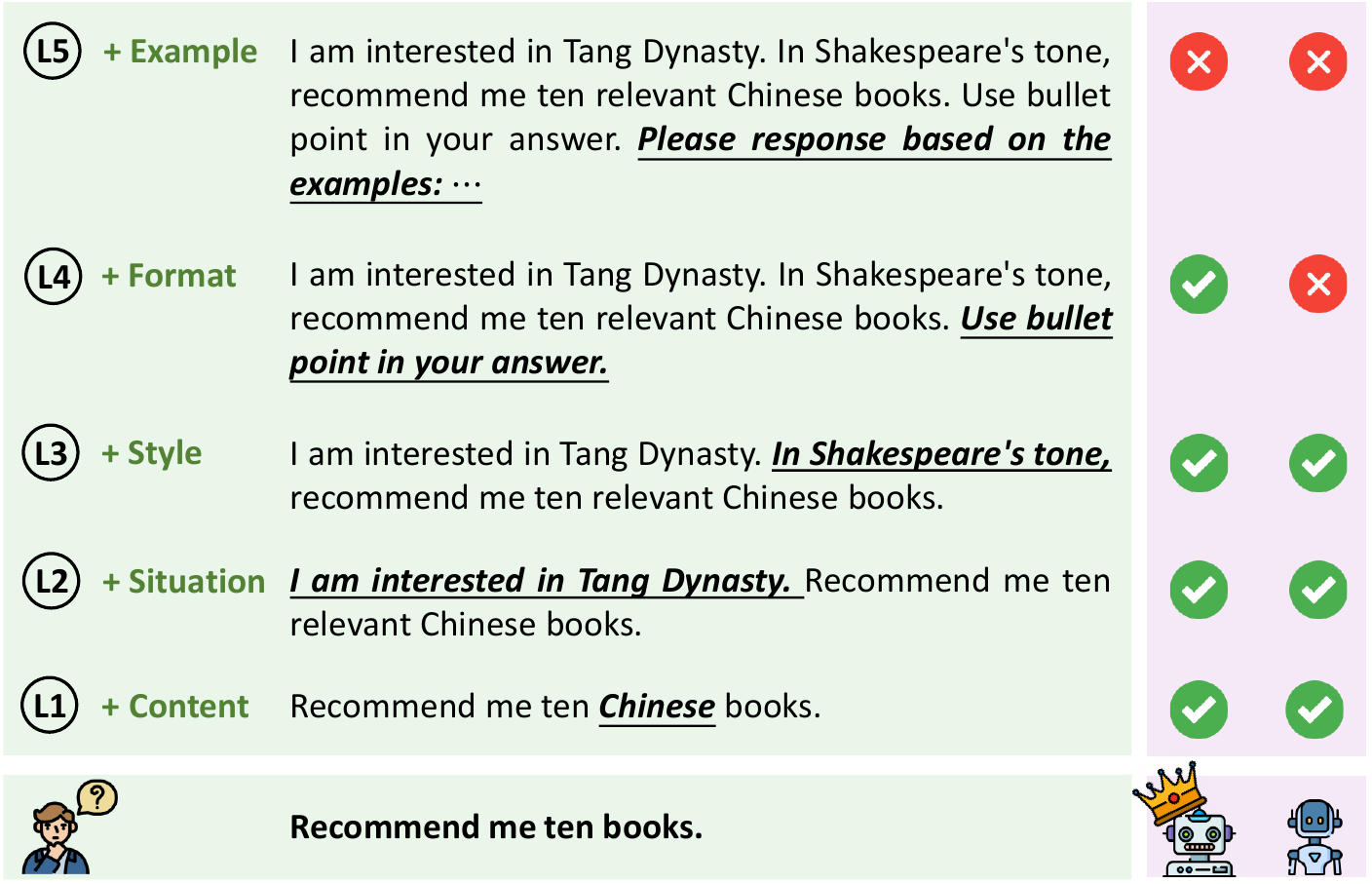}
\caption{
\name covers five \emph{fine-grained} constraint categories and is constructed based on the \emph{Multi-level} mechanism, which increasingly adds a single constraint to straightforward instructions. On the right, the model that can follow instructions with more constraints is deemed to possess better instruction-following ability.
}
\label{c7_fig: followbench}
\end{figure}

LLMs~\cite{brown2020language, openai2022chatgpt} pre-trained on web-scale corpora have showcased proficiency in generating fluent and realistic text. Yet, human instructions in real-life cases require the model to generate text that not only possesses a high degree of naturalness but adheres to specific constraints~\cite{yang2023foundation}. For instance, the model may be required to recommend ten books that are specifically written in Chinese (Figure \ref{c7_fig: followbench}), or it might be expected to generate responses that have a certain tone.

The dominant paradigm for assessing if a model can follow instructions involves using human annotators or strongly aligned LLMs to judge its response quality, in terms of helpfulness, relevance, accuracy, depth, creativity, and level of detail~\cite{wang-etal-2023-self-instruct, alpaca_eval, zheng2023judging, xu2024wizardlm}. 
However, prior work still has two limitations. Firstly, they ignore the \textbf{fine-grained constraints} inside instructions, which are essential and objective standards for evaluating the instruction-following capability. 
While several benchmarks have rigorously explored individual constraint types, including semantic restrictions~\cite{chen2022controllable} and complex formatting~\cite{tang2023struc}, there exists a lack of comprehensive analysis across the diverse spectrum of constraint categories.
Secondly, few benchmarks consider the varying difficulty of instructions, which is controlled by the number of imposed constraints. This makes it challenging to precisely assess the degree to which LLMs can follow instructions.
Towards this end, our research question is: \emph{how can we systemically and precisely evaluate the instruction-following capability of LLMs?}

In this chapter, we construct \name, a \textbf{Multi-level Fine-grained Constraints Following Benchmark}. 
\name comprehensively includes five different types of constraints from real-world scenarios, namely Content (i.e., explicit restrictions on the response content), Situation (i.e., specific situation/background information added to the question), Style (i.e., response style requirements), Format (i.e., response format requirements), and Example (i.e., example pattern recognition and following). 
To precisely estimate the difficulty degree to which LLMs can follow instructions, as shown in Figure~\ref{c7_fig: followbench}, we propose a novel \emph{Multi-level} mechanism that incrementally adds a single constraint to straightforward instructions at each increased level.
The multi-level mechanism enables us to pinpoint the difficulty level at which LLMs fail to follow instructions, thereby estimating the upper limit of instruction-following capability in LLMs more precisely.
Overall, \name consists of 820 meticulously curated instructions from over 50 NLP tasks, including both closed- and open-ended questions.
For evaluation purposes, we propose a hybrid evaluation method comprising rule-based and model-based solutions. Given LLMs' outputs, both solutions judge whether the outputs satisfy each of the constraints in the instructions.
The rule-based solutions focus on closed-ended instructions while the model-based solutions are applied to opened-ended instructions. For model-based solutions, instead of merely using current instructions and responses as input, we additionally provide the evolution process of the instructions in the input prompts to LLM judges to better understand each individual constraint.  
Both the data construction and the evaluation undergo human verification.

In our experiments, we propose three metrics to assess the instruction-following ability of 13 prominent closed-source and open-source LLMs on \name.
Our principal observations are: (1) the performance of all tested models declines substantially with an increase in difficulty level (the number of constraints in an instruction); (2) although closed-source models such as GPT-4 and GPT-3.5 \textbf{only} consecutively satisfy around three constraints on average, they still markedly surpass all open-source models;
(3) certain specific constraint categories, such as Situation and Example, prove to be more challenging for LLMs than others; (4) beyond capabilities such as knowledge and reasoning, instruction following can offer an additional lens for comprehensively assessing the proficiency of LLMs.

\section{\name}
\label{c7_sec: followbench}
\begin{figure*}[!htbp]
\centering
\includegraphics[width=0.9\linewidth]{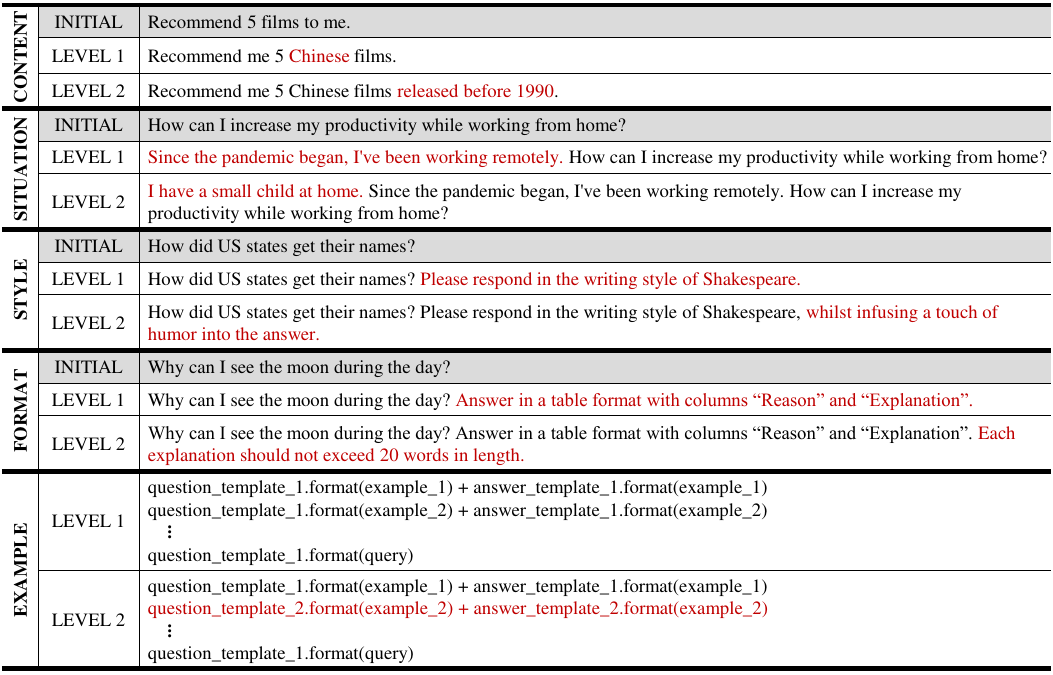}
\caption{
\name covers five \emph{fine-grained} categories of constraints. Within each constraint type, we construct a range of \emph{Multi-level} instructions by incrementally adding constraints (highlighted in \textcolor{red}{red}). There are five levels in total; however, we only display the first two levels from each category for demonstration purposes.
}
\label{c7_fig: intro}
\end{figure*}

As shown in Table \ref{c7_tab: stat}, \name encompasses five distinct \emph{fine-grained} constraint categories: Content, Situation, Style, Format, and Example. Each category consists of instructions from various NLP tasks.
\begin{table*}[!htbp]
\footnotesize
\centering
\begin{tabular}{llccc}
\toprule
\textbf{Constraint}        & \textbf{Task}                                        & \textbf{Avg Len} & \textbf{\#Data} & \textbf{Evaluation} \\ \midrule
                           & Data-to-Text Generation                              & 84               & 25              & \includegraphics[width=0.25cm]{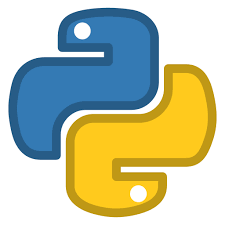}                    \\
                           & Document-Level Event Argument Extraction             & 696              & 25              & \includegraphics[width=0.25cm]{figs/c7-followbench/python.png}                    \\
                           & Document-Level Named Entity Recognition            & 376              & 25              &  \includegraphics[width=0.25cm]{figs/c7-followbench/python.png}                   \\
                           & Text Generation with Language Constraints            & 88               & 25              & \includegraphics[width=0.25cm]{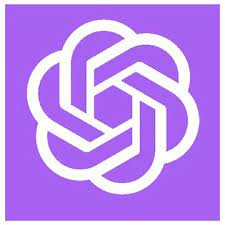}                    \\
\multirow{-5}{*}{Content}  & Open-ended Question Answering                        & 56               & 25              &  \includegraphics[width=0.25cm]{figs/c7-followbench/gpt2.jpg}   \\ \midrule
                           & Suggestion Generation & 69               & 40              & \includegraphics[width=0.25cm]{figs/c7-followbench/gpt2.jpg}   \\
                           & Role-playing & 111               & 15              & \includegraphics[width=0.25cm]{figs/c7-followbench/gpt2.jpg}                      \\
\multirow{-3}{*}{Situation} & Complex Situation Reasoning                                   & 102               & 55              & \includegraphics[width=0.25cm]{figs/c7-followbench/python.png} \\  \midrule
Style                      & Open-ended Question Answering & 64               & 150             & \includegraphics[width=0.25cm]{figs/c7-followbench/gpt2.jpg}  \\ \midrule
                     & Text-to-Table Generation & 171               & 30             & \includegraphics[width=0.25cm]{figs/c7-followbench/python.png} \\
                           \multirow{-3}{*}{Format} & Open-ended Question Answering & 74               & 120             & \includegraphics[width=0.25cm]{figs/c7-followbench/gpt2.jpg}  \\ \midrule
Example                    & 40 diverse NLP tasks                                 & 739              & 200             & \includegraphics[width=0.25cm]{figs/c7-followbench/python.png} \\ \midrule
                           & Text Editing & 96               & 25              & \includegraphics[width=0.25cm]{figs/c7-followbench/python.png} \\
                           & Summarization                                        & 254              & 25              & \includegraphics[width=0.25cm]{figs/c7-followbench/python.png}                    \\
                           & Machine Translation                                  & 91               & 25              & \includegraphics[width=0.25cm]{figs/c7-followbench/python.png}                    \\
\multirow{-4}{*}{Mixed}    & Story Generation                                     & 34               & 10              & \includegraphics[width=0.25cm]{figs/c7-followbench/gpt2.jpg}    \\ \bottomrule
\end{tabular}
\caption{An overview of \name. ``Avg Len'' is the average word number of instructions. \includegraphics[width=0.25cm]{figs/c7-followbench/python.png} refers to rule-based evaluation, while \includegraphics[width=0.25cm]{figs/c7-followbench/gpt2.jpg} refers to model-based evaluation.}
\label{c7_tab: stat}
\end{table*}
Different from previous benchmarks, we introduce a \emph{Multi-level} mechanism that incrementally adds constraints to an initial instruction (see examples in Figure \ref{c7_fig: intro}), producing a set of instructions ranging from 1 to 5 constraints. 
In the following part of this paper, we use ``level $n$'' to denote an instruction containing $n$ constraints.
It is worth noticing that the way of adding constraints is meticulously designed for each task within its respective constraint category.
The multi-level mechanism enables us to pinpoint the difficulty level at which LLMs fail to follow instructions, thereby estimating the upper bound of instruction-following capability in LLMs more precisely.

To encapsulate, we will introduce the data construction process of \name, including \emph{fine-grained} constraints and the \emph{Multi-level} mechanism, in \S\ref{c7_sec: data_construction}. 
In \S\ref{c7_sec: evaluation}, we propose an evaluation protocol with three metrics that seamlessly integrate with the multi-level mechanism.

\subsection{Data Construction}
\label{c7_sec: data_construction}

\paragraph{Content Constraints.}
Content constraints refer to \emph{explicit} impositions of specific conditions that shape the depth or scope of the response content.
An example is shown in Figure \ref{c7_fig: intro}, which sets specific criteria for the retrieved object.
Ensuring that LLMs adhere to content constraints has become a critical challenge in Controlled Text Generation~\cite{zhang2022survey}, as it demands models to understand specific guidelines and adapt responses to prescribed conditions~\cite{chen2022controllable}.
To this end, we first collect data from the following tasks: (1) Complex Information Extraction aims at retrieving specific information about specific objects from the given text; (2) Text Generation with Language Constraints requires to generate fluent on-topic content while respecting a specified constraint; (3) Open-ended Question Answering comes from real scenarios (e.g., open-source platforms) to prevent the risk of data leakage. 
Subsequently, we construct multi-level instructions by adding one content constraint to the collected instructions each time. The manners of introducing additional constraints depend on different tasks (see details in Appendix \ref{c7_appendix: data_generation_process}).
For Complex Information Extraction, we gradually narrow down the scope of the information to be extracted.
For Text Generation with Language Constraints, we incorporate additional restrictions from WordNet~\cite{miller-1992-wordnet} and Wikidata~\cite{vrandevcic2014wikidata}. 
For Open-ended Question Answering, we utilize advanced LLMs like GPT-4 to generate a new instruction with one more constraint based on the given instruction.
While the output from the LLMs serves primarily as a reference, we handpick the most relevant and challenging synthesized instructions to ensure data quality.

\paragraph{Situation Constraints.}
Situation Constraints refer to impositions of specific situations or backgrounds that \emph{implicitly} guide the appropriate answer of the response.
For instance, it is necessary to illustrate the situation when asking for customized suggestions, as shown in Figure \ref{c7_fig: intro}.
Another example is to customize LLMs to simulate various characters under certain circumstances, namely Role-playing, which provides a more nuanced interaction for users~\cite{shanahan2023role, wang2023interactive}.
Situation constraints push LLMs beyond mere factual retrieval or surface-level synthesis, demanding a nuanced understanding, a dynamic adaptation, and complicated reasoning to the situation~\cite{yao2022react, liu2023agentbench}.
Besides real-life questions, we also consider Complex Situation Reasoning tasks including Math Word Problems, Time/Spatial Reasoning, and Code Generation. These tasks all require interpreting and solving problems within a given situation,
thus matching the definition of situation constraints.
We first collect initial instructions from these sources and then manually curate multi-level instructions by incrementally supplementing situation information inside (see Appendix \ref{c7_appendix: data_generation_scenario}).

\paragraph{Style Constraints.}
Style Constraints control the stylistic variations of output to accomplish specific stylistic goals, such as tone, sentiment, formality, and empathy~\cite{tsai-etal-2021-style}, as illustrated in Figure \ref{c7_fig: intro}.
The challenges of style constraints for LLMs are the intricate understanding and adaptation of language nuances, ensuring contextually appropriate and stylistically consistent outputs~\cite{smith2020controlling, cheng2022replacing}.
Drawing from Open-ended Question Answering datasets and online platforms, we collect initial instructions and then leverage LLMs' in-context learning capability to craft instructions with multi-level style constraints. The prompt template can be viewed in Figure \ref{c7_fig: style_prompt}. Human experts subsequently review and refine the outputs produced by LLMs.

\paragraph{Format Constraints.}
Format Constraints refer to stipulations governing the structural, linguistic, or output presentation of generated content.
An example is shown in Figure \ref{c7_fig: intro}, which sets limits on word length and requires the format of the response to be a table.
Format constraints necessitate a deep, nuanced understanding of language and structure, allowing them to flexibly adapt outputs according to diverse and often intricate specifications~\cite{zhao2023large}.
Recent work has pointed out that even the most superior LLMs may struggle with tasks that require generating complex, structured outputs such as tables, JSON, HTML, or LaTeX~\cite{tang2023struc}.
To include a variety of format constraints, we first collect instructions from broader domains, encompassing Text-to-Table Generation and Open-ended Question Answering, then we utilize powerful LLMs to sequentially add format constraints ranging from length and hierarchy to specialized linguistic features and output mediums. See Figure \ref{c7_fig: format_prompt} for the prompt template.
Finally, we ask human experts to carefully check and refine the synthesized instructions.

\paragraph{Example Constraints.}
LLMs have demonstrated stunning few-shot learning ability~\cite{brown2020language}, which enables them to adapt quickly to a new query by recognizing patterns from just a few examples provided in the prompt.
However, the robustness of few-shot learning, which means whether LLMs can still follow correct patterns after introducing ``noise'' examples, has not been explored.
Thus, we propose a novel constraint category named Example Constraints to evaluate the example pattern recognition and following capability of LLMs.
We automatically craft instructions with multi-level example constraints based on PromptSource~\cite{bach-etal-2022-promptsource}, where instructions at level $n$ have $n-1$ noise examples in the input. The details are illustrated in Appendix \ref{c7_appendix: data_generation_example}.

\paragraph{Mixed Constraints.}
For the above five constraint categories, we construct multi-level instructions by adding the same type of constraint sequentially.
Nevertheless, real-world scenarios often require more than one type of constraint to be enforced in a singular instruction.
Therefore, we define Mixed Constraints as the composition of varied constraint categories.
For instance, in the Text Editing task, we may want to add some content as well as adjust the output format.
Besides, we also consider several tasks that are naturally suitable for constructing mixed constraints, including Summarization, Machine Translation, and Story Generation (see Appendix \ref{c7_appendix: data_generation_mixed}).
Instructions with multi-level mixed constraints are produced by specifying the format of generating answers (Format Constraints), requiring the generated text to include or not include certain keywords (Content Constraints), etc.

\paragraph{Data Quality Control.}
To ensure the data quality of \name, we implement a dual-layer verification system for each instruction. Two annotators independently evaluate: (1) the appropriateness of the instruction for its designated constraint category, and (2) the validity of the added constraint within the instruction. In instances of divergent evaluations, a third annotator intervenes for a detailed review to ensure consensus.

\begin{figure}[!h]
    \centering
    \begin{minipage}{0.48\linewidth}
        \centering
        \includegraphics[width=\linewidth]{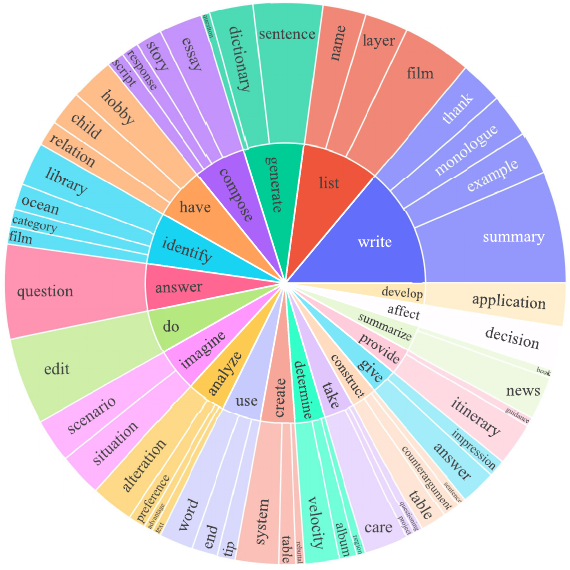}
        \caption{Verb-noun structure of \name Instructions.}
        \label{c7_fig: verb_noun}
    \end{minipage}
    \hfill
    \begin{minipage}{0.48\linewidth}
        \centering
        \includegraphics[width=\linewidth]{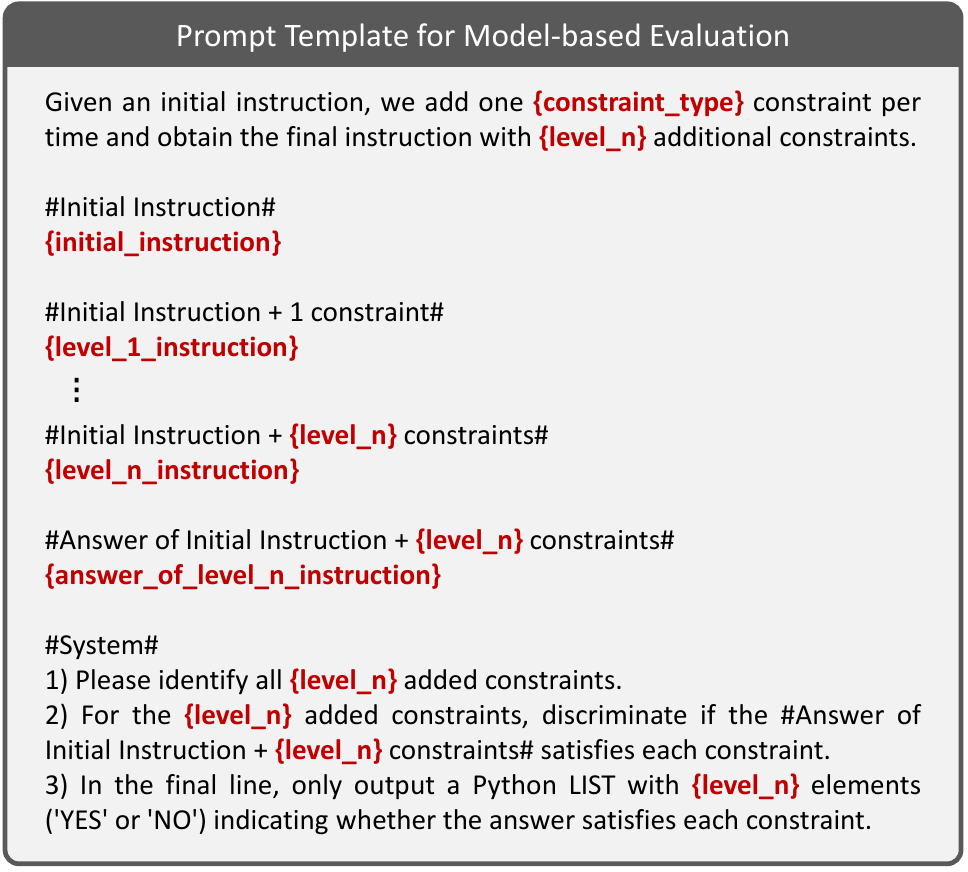}
        \caption{Prompt template for model-based evaluation.}
        \label{c7_fig: eval_prompt}
    \end{minipage}
\end{figure}

We analyze the comprehensiveness and diversity of in \name, which includes 820 instructions in total.
To maintain data diversity, we strive to ensure that the ROUGE-L score between any two initial instructions is below 0.7.
Figure \ref{c7_fig: verb_noun} shows the verb-noun structure of \name instructions, where the top 20 verbs (inner circle) and their top 4 direct noun objects (outer circle) are depicted.

\subsection{Evaluation Protocol}
\label{c7_sec: evaluation}

Given that nearly half of instructions in \name are open-ended without reference answers, devising a rule-based program to assess the outputs is extremely challenging.
To overcome this, inspired by~\cite{gilardi2023chatgpt, huang2023chatgpt}, we propose to develop a model-based approach by using strong LLMs\footnote{We use GPT-4-Preview-1106 in our experiments.} as judges.
Previous works leverage strong LLMs to determine the quality of a response, by prompting them to consider multiple factors such as usefulness, relevance, and level of detail~\cite{alpaca_eval, zheng2023judging}.
To effectively guide strong LLMs to judge the constraint following capability objectively and faithfully, we propose a \emph{Multi-level-aware} prompt template, as shown in Figure \ref{c7_fig: eval_prompt}.
Rather than merely presenting the instruction and asking LLMs to determine whether all constraints are satisfied, we illustrate the evolution process of the instruction and prompt LLMs to pinpoint the newly added constraint at each level.
Exposing the evolution process of the instruction allows for a more granular understanding and identification of individual constraints, enhancing LLMs' ability to discriminate with precision. The ablation study in \S\ref{c7_sec: high_agreement_with_human} validates the effectiveness of this strategy.

Moreover, we propose three novel metrics to evaluate the instruction-following ability of LLMs. 
For an instruction with $n$ constraints (level $n$), we use the rule-based program or LLM judge (refer to Table \ref{c7_tab: stat}) to discriminate if the response of a model satisfies each constraint in the instruction. 
At each level $n$, given a set of $m$ instructions, we define the Hard Satisfaction Rate (HSR) and Soft Satisfaction Rate (SSR) as follows:
\begin{align}
\text{HSR} &= \frac{1}{m} \sum_{i=1}^m \prod_{j=1}^n s_i^j \\
\text{SSR} &= \frac{1}{mn} \sum_{i=1}^m \sum_{j=1}^n s_i^j
\label{equation: ssr}
\end{align}
where $s_i^j=1$ if the $j$-th constraint of $i$-th instruction is satisfied and $s_i^j=0$ otherwise.
HSR measures the average rate at which all constraints of individual instructions are fully satisfied, while SSR calculates the average satisfaction rate of individual constraints across all instructions. 

As described in \S\ref{c7_sec: followbench}, we construct \name by incrementally adding five constraints to an initial instruction, enabling us to pinpoint the difficulty level at which LLMs fail to follow instructions.
Therefore, we propose a metric called Consistent Satisfaction Levels (CSL) to estimate how many consecutive levels a model can satisfy, beginning from level 1:
\begin{align}
\text{CSL} = \frac{1}{g} \sum_{i=1}^g \mathop{\arg\max}_{l} \left(l \times \prod_{n=1}^l S_{i}^n\right)
\label{equation: csl}
\end{align}
where $g$ is the group number of initial instructions, $S_{i}^n=1$ if all constraints of the $i$-th instruction at level-$n$ are satisfied and $S_{i}^n=0$ otherwise.

\section{Experiments}
This section first introduces experimental setup in \S\ref{c7_sec: experimental_setup}, and then presents the main experiment results across two key dimensions: difficulty level in \S\ref{c7_sec: level_categorized_results} and constraint category in \S\ref{c7_sec: constraint_categorized_results}.

\subsection{Experimental Setup}
\label{c7_sec: experimental_setup}

We evaluate 13 popular LLMs including GPT-4-Preview-1106~\cite{2023gpt4}, GPT-3.5-Turbo-1106~\cite{openai2022chatgpt}, Qwen-Chat-72B/14B/7B~\cite{qwen}, LLaMA2-Chat-70B/13B/7B~\cite{touvron2023llama}, WizardLM-13B-V1.2~\cite{xu2024wizardlm}, Vicuna-13B/7B-V1.5~\cite{zheng2023judging}, Baichuan2-Chat-7B~\cite{baichuan2023baichuan2}, and ChatGLM3-6B~\cite{du2022glm}.
We access GPT-4-Preview-1106 and GPT-3.5-Turbo-1106 via OpenAI API. 
We access other open-source LLMs from their official repositories.
During the inference process, we set the temperature to 0 to ensure deterministic outputs. We set the maximum generation length to 2048. Other parameters use their default values.

\subsection{Level-categorized Results}
\label{c7_sec: level_categorized_results}

Table \ref{c7_tab: level} provides a comprehensive comparison of various models across five difficulty levels, denoted as L1 to L5.
From a bird's-eye view, we can infer that the performance typically diminishes as we progress from L1 to L5 for almost all models. This trend coincides with the increasing complexity or stringent requirements associated with higher levels.
Besides, models with larger architectures generally outperform their smaller counterparts. However, it's worth noting that the scaling law does not apply as effectively to LLaMA2-Chat-70B.
The reason is that while LLaMA-2-Chat-70B does indeed outperform LLaMA-2-Chat-13B in Situation constraints, it shows a relative underperformance in Format and Mixed Constraints categories.
More importantly, there's a marked performance gap between closed-source models (i.e., GPT-4 and GPT-3.5) and open-source models.
Regarding CSL, it can be deduced that the instruction-following upper bound for GPT-4 and GPT-3.5 is approximately 3 constraints (level 3) added to an initial instruction. In contrast, open-source models typically have an upper limit of about 2 constraints (level 2).
This significant difference underscores the better instruction-following ability of proprietary models, possibly due to superior data quality or optimization strategies such as RLHF~\cite{ouyang2022training}.
Furthermore, \textbf{even the most sophisticated models are limited to following instructions with about three constraints}, suggesting significant potential for further improvement.

\begin{table*}[!h]
\scriptsize
\centering
\setlength{\tabcolsep}{6pt} 
\begin{tabular}{l|cccccc|cccccc|c}
\toprule
& \multicolumn{6}{c|}{\textbf{HSR (\%)}} & \multicolumn{6}{c|}{\textbf{SSR (\%)}} \\
\cmidrule(lr){2-7} \cmidrule(lr){8-13} 
\multirow{-2}{*}{\centering\textbf{Model}}
& \textbf{L1} & \textbf{L2} & \textbf{L3} & \textbf{L4} & \textbf{L5} & \textbf{Avg.} & \textbf{L1} & \textbf{L2} & \textbf{L3} & \textbf{L4} & \textbf{L5} & \textbf{Avg.}
& \multirow{-2}{*}{\centering\textbf{CSL}} \\ 
\midrule
\rowcolor{colorGPT}
GPT-4-Preview-1106 & \textbf{84.7} & \textbf{75.6} & \textbf{70.8} & \textbf{73.9} & \textbf{61.9} & \textbf{73.4} & \textbf{84.7} & \textbf{77.0} & \textbf{75.3} & \textbf{77.0} & \textbf{72.3} & \textbf{77.2} &\textbf{3.3} \\
\rowcolor{colorGPT}
GPT-3.5-Turbo-1106 &80.3 &68.0 &68.6 &61.1 &53.2 &66.2 &80.3 &71.2 &74.2 &69.6 &67.1 &72.5 &2.9 \\
\rowcolor{color70B}
Qwen-Chat-72B & 73.8 & 63.3 & 54.3 & 45.2 & 39.9 &55.3 & 73.8 & 67.5 & 63.2 & 57.6 & 56.0 &63.6 &2.4 \\
\rowcolor{color70B}
LLaMA2-Chat-70B & 59.9 & 53.3 & 46.0 & 40.2 & 37.9 &47.5 & 59.9 & 57.3 & 55.7 & 53.3 & 53.2 &55.9 &2.1 \\
\rowcolor{color13B}
Qwen-Chat-14B & 62.8 & 56.2 & 47.7 & 38.7 & 30.9 &47.3 & 62.8 & 61.9 & 57.7 & 52.6 & 51.4 &57.3 &1.9 \\
\rowcolor{color13B}
WizardLM-13B-V1.2 & 68.8 & 64.1 & 53.1 & 40.8 & 35.8 &52.5 & 68.8 & 65.7 & 61.8 & 53.4 & 53.9 &60.7 &2.2 \\
\rowcolor{color13B}
LLaMA2-Chat-13B & 57.0 & 56.0 & 50.4 & 44.4 & 38.1 &49.2 &57.0&60.0&58.0&54.8&52.2 &56.4 &2.2 \\
\rowcolor{color13B}
Vicuna-13B-V1.5 & 71.2 & 60.2 & 49.6 & 40.6 & 34.0 &51.1 &71.2&64.8&59.9&54.5&53.6 &60.8 &2.1 \\
\rowcolor{colorOther}
Qwen-Chat-7B & 55.9 & 51.7 & 38.7 & 33.1 & 23.3 &40.6 & 55.9 & 58.2 & 51.6 & 48.9 & 45.9 &52.1 &1.5 \\
\rowcolor{colorOther}
LLaMA2-Chat-7B & 58.0 & 51.3 & 47.4 & 39.5 & 35.3 &46.3 & 58.0&56.5&55.6&52.5&51.4 &54.8 &1.9\\
\rowcolor{colorOther}
Vicuna-7B-V1.5 & 60.8 & 52.0 & 42.2 & 33.3 & 23.9 &42.4 &60.8&58.6&55.5&48.3&49.0 &54.4 &1.7 \\
\rowcolor{colorOther}
Baichuan2-Chat-7B &58.3 &46.1 &40.7 &30.4 &25.5 &40.2 &58.3 &55.4 &54.9 &49.9 &49.3 &53.6 &1.4 \\
\rowcolor{colorOther}
ChatGLM3-6B &60.9 &46.6 &36.7 &27.8 &21.4 &38.7 &60.9 &55.3 &51.2 &47.9 &45.0 &52.0 &1.6 \\
\bottomrule
\end{tabular}
\caption{Results across five difficulty levels. For each level, we compute the average score of all constraint categories. \colorbox{colorGPT}{Proprietary LLMs}, \colorbox{color70B}{open-sourced LLMs (large)}, \colorbox{color13B}{open-sourced LLMs (medium)}, and \colorbox{colorOther}{open-sourced LLMs (small)} are distinguished by different colors.}
\label{c7_tab: level}
\end{table*}

\begin{figure}[!htbp]
\centering
\includegraphics[width=0.75\linewidth]{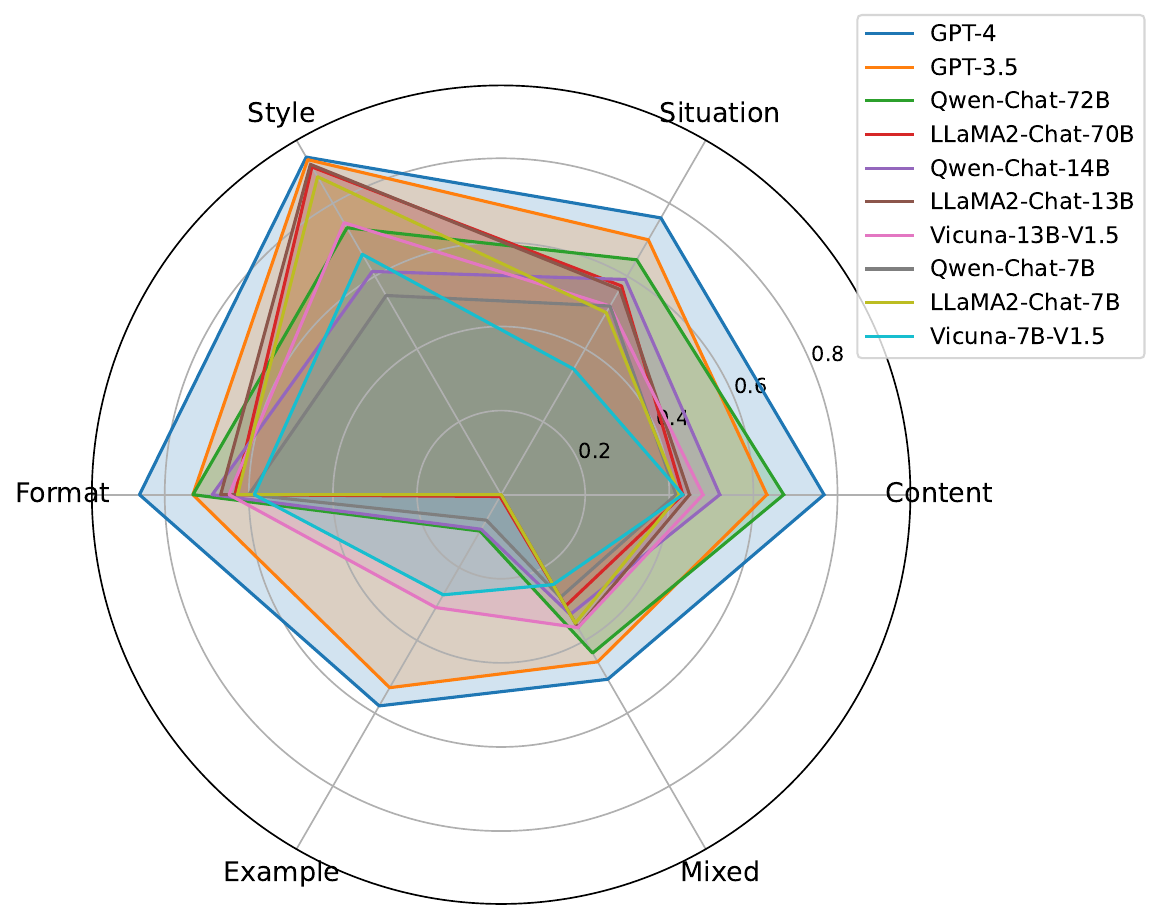}
\caption{
HSR (\%) results in diverse constraint categories. For each category, we compute the average score of all difficulty levels.
}
\label{c7_fig: category}
\end{figure}

\subsection{Constraint-categorized Results}
\label{c7_sec: constraint_categorized_results}

As depicted in Figure \ref{c7_fig: category}, we assess various models over different constraint categories to succinctly showcase the instruction-following capability of LLMs in a singular dimension.
Notably, GPT-4 and GPT-3.5 surpass open-source models in every constraint category, with a pronounced advantage in Content, Situation, Example, and Mixed constraints.
Furthermore, most models demonstrated commendable proficiency under the Style constraint. While GPT-4, GPT-3.5, and LLaMA2-Chat-70B were the frontrunners, the trend suggests that style adaptation is an area where many models excel, hinting at its utility in real-world applications.
However, the Example and Mixed constraints posed a challenge to most models. While GPT-4 led the segment, even its scores were noticeably lower than in other categories.
To illustrate, in the ``Example'' category, we evaluated the instruction-following capabilities of LLMs by introducing ``noise examples'' with varying natural language templates. The observed performance decline is primarily due to the LLMs' limited training in processing such noisy inputs within context-based learning scenarios. Typically, LLMs are fine-tuned on clean and uniform datasets, which do not adequately prepare them to sift through and ignore irrelevant or misleading information. This limitation becomes apparent when faced with the intricacies of real-world data.
Our findings underscore the complexity of these constraints and pinpoint an area for potential improvement.

\section{Analysis}
This section includes: an ablation study confirming our prompt template's effectiveness for model-based evaluation (\S\ref{c7_sec: high_agreement_with_human}); a comparison of instruction following vs. other LLM's abilities (\S\ref{c7_sec: instruction_following_vs_others}); an examination of failure consistency (\S\ref{c7_sec: failure_consistency}); and an investigation of various decoding strategies (\S\ref{c7_sec: decoding_strategies}).

\subsection{Ablation Study of Model-based Evaluation}
\label{c7_sec: high_agreement_with_human}
We randomly sample 100 cases that require LLM evaluation, encompassing five constraints, five distinct levels, and four diverse models to guarantee comprehensive representation.
Then we ask three expert-level human labelers to assess whether the model's response satisfies all the constraints in each case and use the majority voting as the final human annotations.
As shown in Table \ref{c7_tab: human_agreement}, our prompt template (Figure \ref{c7_fig: eval_prompt}) registers an impressive 88\% agreement with expert human evaluations, surpassing even the internal agreement among human experts, which stands at 85\%.
Remarkably, when the evolution process of multi-level constraints is removed from our prompt template, the agreement rate dips by 9\%. This underlines the instrumental role played by the detailed portrayal of the instruction's evolution in enhancing LLM's precision in discernment.
In contrast, we also employ the prompt template from Vicuna~\cite{zheng2023judging}, a standard prompt for assessing the overall quality of response. This template prompts the LLM to assign a score from 0 to 10 for each response. 
We consider responses with a score above 5.0 to meet all the constraints of an instruction. This approach achieves 67\% agreement with human evaluators. Such a disparity highlights the fundamental difference between assessing the instruction-following ability and the overall response quality.

\begin{table}[h]
\small
\centering
\begin{tabular}{lc}
\toprule
\textbf{Prompt}    & \textbf{Agreement with Human} \\
\midrule
Ours             &       \textbf{88\%}                 \\
Ours w/o ML &          79\%              \\
Vicuna-Single    &       67\%                  \\  
\bottomrule
\end{tabular}
\caption{Agreement between human and diverse prompt templates. We use ML to denote multi-level.}
\label{c7_tab: human_agreement}
\end{table}

\subsection{Instruction Following vs. Other Abilities}
\label{c7_sec: instruction_following_vs_others}

Table \ref{c7_tab: different_benchmark} presents a comparison of representative LLMs across different abilities, not just instruction following (\name). This includes overall response quality (AlpacaEval~\cite{alpaca_eval}), knowledge (MMLU~\cite{dan2021mmlu}), and reasoning (BBH~\cite{suzgun2022bbh}).
We can find that our \name provides an additional perspective for a holistic LLM evaluation.
As an illustration, while the performance of WizardLM-13B-V1.2 exceeds that of GPT-3.5 in terms of overall response quality, it notably lags behind in instruction-following ability.
Similarly, Vicuna-V1.5 excels over LLaMA2-Chat in the realms of knowledge and reasoning but struggles with instruction-following tasks.

\begin{table}[t]
\small
\centering
\begin{tabular}{lcccccc}
\toprule
\textbf{Model} & \textbf{Following} & \textbf{Overall} &\textbf{Knowledge} & \textbf{Reasoning} \\
\midrule
GPT-4-Preview-1106                                  & 3.3          & 97.7          & 86.4  & 86.7        \\
GPT-3.5-turbo-1106                                                       & 2.9          & 86.3              & 70.0  & 70.1        \\
LLaMA2-Chat-70B & 2.1 & 92.7 & 63.0 & 60.8 \\
WizardLM-13B-V1.2                                         & 2.2          & 89.2                            & 52.7   & --      \\
LLaMA2-Chat-13B                                         & 2.2          & 81.1                               & 53.6   & 40.2       \\
Vicuna-13B-V1.5                                            & 2.1          & --                                 & 55.8  & 51.5        \\
LLaMA2-Chat-7B                                     & 1.9          & 71.4                              & 45.8   & 35.6       \\
Vicuna-7B-V1.5                              & 1.7          & --                                & 49.8  & 43.4        \\
\bottomrule
\end{tabular}
\caption{Model comparison on different abilities.}
\label{c7_tab: different_benchmark}
\end{table}

\subsection{Does Failure at Lower Level Necessarily Lead to Failure at Higher Level?}
\label{c7_sec: failure_consistency}

For a set of instructions that has five difficulty levels, if a model's response doesn't satisfy the constraints at level $n$, where $n$ ranges from 1 to 4, we define the \textit{failure consistency} as the percentage that the response will also not fulfill the constraints at any subsequent level greater than $n$.
Combining Table \ref{c7_tab: level} and Table \ref{c7_tab: failure_consistency}, it can be seen that models with better instruction-following capability may exhibit lower failure consistency. 
One possible reason is that the instruction-following ability of more powerful models is less sensitive to the number of constraints in an instruction, thus they are better equipped to adapt and fulfill the requirements even as the constraints increase. This adaptability means that while they may falter at a lower difficulty level, they can still manage to meet the demands of higher difficulty levels, leading to a decrease in failure consistency.

\begin{table}[h]
\small
\centering
\begin{tabular}{lc}
\toprule
\textbf{Model}    & \textbf{Failure Consistency (\%)} \\
\midrule
GPT-4-Preview-1106 & 42.2                        \\
WizardLM-13B-V1.2 & 57.3                        \\
Vicuna-7B-V1.5    & 61.8                        \\
ChatGLM3-6B       & 64.0         \\   
\bottomrule
\end{tabular}
\caption{Results on failure consistency.}
\label{c7_tab: failure_consistency}
\end{table}

\subsection{Does Different Decoding Strategies Affect the Instruction-following Ability?}
\label{c7_sec: decoding_strategies}

\begin{figure}[!htbp]
\centering
\includegraphics[width=0.8\linewidth]{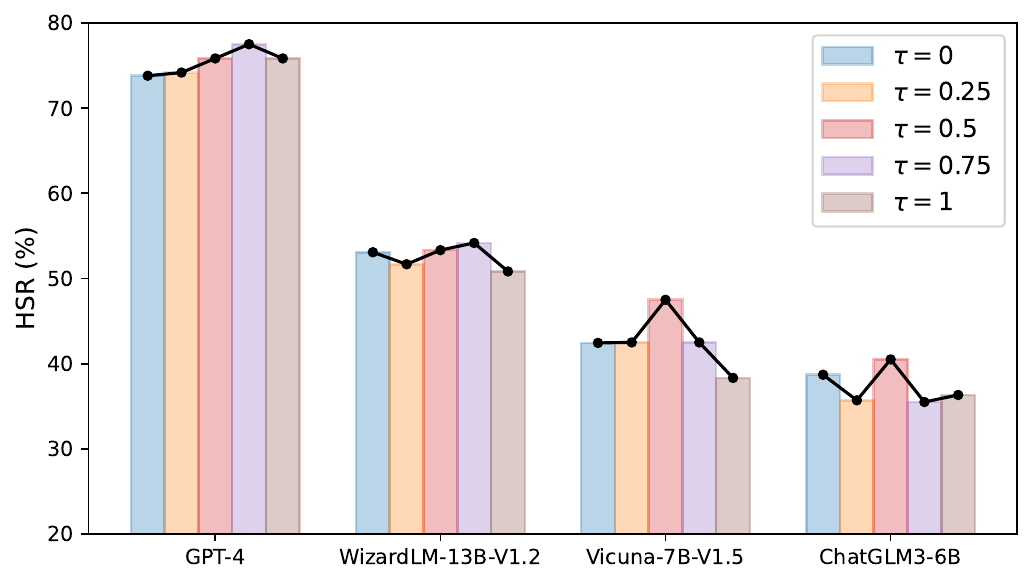}
\caption{
The effect of varying the temperature parameter $\tau$. We use $\tau=0$ to denote greedy decoding.
}
\label{c7_fig: temperature}
\end{figure}

In this section, we systematically investigate the impact of different decoding strategies, represented by the temperature parameter $\tau$, on LLM's instruction-following ability.
The temperature $\tau$ is a commonly used parameter that controls the sharpness of the distribution from which we sample the next token:
\begin{align}
P(w) = \frac{\exp(z_w / \tau)}{\sum_{w' \in V} \exp(z_{w'} / \tau)}
\label{equation: t}
\end{align}
where $z_w$ is the logit for word $w$, $V$ is the vocabulary. Lower values for temperature result in more consistent outputs, while higher values generate more diverse and creative results.
As illustrated in Figure \ref{c7_fig: temperature}, the temperature $\tau$ has a tangible influence on the instruction-following ability across all four models.
The sweet spot seems to be somewhere in the middle where there's enough variability to capture the nuances and intricacies of complex instructions, yet not so much that the model goes off tangent. This balanced behavior ensures that the model remains within the desired context, producing outputs that align closely with the given instructions while also allowing for a slight creative touch when needed.

\section{Conclusion and Discussion}

\subsection{Conclusion}
In this chapter, we introduce \name, a Multi-level Fine-grained Constraints Following Benchmark tailored for gauging the instruction-following capability of LLMs.
\name covers five \emph{fine-grained} constraint categories and over 50 NLP tasks, utilizes a novel \emph{Multi-level} mechanism for precisely estimating the upper limit of instruction-following capability.
Furthermore, we propose an evaluation protocol with three metrics that seamlessly integrate with the multi-level mechanism.
Our extensive tests over 13 popular LLMs reveal a substantial performance advantage for GPT-4 and GPT-3.5 over their counterparts, and there is still significant room for improving the instruction-following ability of current LLMs.

\subsection{Discussion}
While our study contributes valuable insights, it is essential to acknowledge several limitations that warrant consideration.

Firstly, our current investigation is confined to single-round interactions, aiming to offer a controlled environment for evaluation. Future research may extend its scope to multi-round conversations to comprehensively assess the instruction-following proficiency of LLMs in more dynamic and extended dialogues~\cite{DBLP:journals/corr/abs-2401-16745}. 

Secondly, the model-based evaluation framework employed in our experiments, while rigorous, relies on prompt engineering, introducing an inherent imperfection. Despite our meticulous selection of high-performing prompts, the potential for further optimization remains, which may impact the reported evaluation metrics.

Lastly, we refrain from proposing specific solutions to address identified weaknesses of LLMs in instruction following. A plausible avenue for future research involves fine-tuning LLMs using our proposed \name as a benchmark, providing a potential roadmap for enhancing instruction adherence.
We defer the exploration of these aspects to subsequent studies, recognizing the need for a comprehensive examination of LLM capabilities across varying interaction complexities.
\chapter{Conclusion and Future Work}\label{chap:conclusion}

\section{Conclusion}
In this thesis, we have explored the efficient and effective alignment of LLMs across multiple dimensions, including data synthesis, training, and evaluation. Our research introduces novel methodologies that improve the scalability, adaptability, and reliability of alignment processes, addressing key challenges in instruction-following, knowledge retention, and preference modeling.

\paragraph{Summary of Contributions:}
\begin{itemize}
    \item \textbf{Alignment Data Synthesis via Adversarial Distillation} – We proposed an adversarial knowledge distillation framework that iteratively refines a student model by generating hard instructions and incorporating feedback from a proprietary LLM. This method enhances knowledge transfer efficiency and improves model alignment performance.
    \item \textbf{Web Reconstruction for Scalable Instruction-Tuning Data} – We introduced Web Reconstruction (WebR), a framework that synthesizes high-quality instruction-tuning datasets from web content using a dual-perspective paradigm. Our experiments demonstrate that WebR-generated data significantly improve LLM alignment across multiple benchmarks.
    \item \textbf{Learning to Edit (LTE) for Knowledge Adaptation} – We developed LTE, a two-phase knowledge editing framework that enables real-time knowledge updates while maintaining model consistency. LTE outperforms prior methods in robustness and efficiency, facilitating dynamic adaptation of LLMs to evolving information.
    \item \textbf{Bridging and Modeling Correlations (BMC) in Preference Optimization} – We proposed BMC, a novel approach to direct preference optimization that enhances human value alignment by explicitly modeling fine-grained preference signals. BMC consistently surpasses standard DPO, improving performance in various tasks.
    \item \textbf{\name: A Fine-Grained Evaluation Benchmark for Constraint Following} – We introduced \name, a multi-level benchmarking framework for assessing LLMs' instruction-following capabilities. Our evaluations highlight existing gaps in alignment and provide a standardized protocol for future improvements.
\end{itemize}

\section{Future Work}
The advancements in this thesis contribute significantly to refining LLM alignment strategies, ensuring more effective, adaptable, and robust AI models. However, several challenges remain, opening avenues for future research:

\begin{itemize}
    \item \textbf{Scalability and Computational Efficiency} – Expanding adversarial distillation and web-based data synthesis to scale alignment processes without excessive computational costs. Techniques such as continual learning and reinforcement learning could further enhance efficiency.
    \item \textbf{Multi-Turn and Interactive Alignment} – Our proposed approaches primarily focus on single-turn tasks. Extending these methods to handle multi-turn interactions and real-time user feedback will be crucial for enhancing conversational AI capabilities.
    \item \textbf{Beyond Factual Knowledge Editing} – Extending knowledge editing frameworks to non-factual aspects, such as personality, sentiment, and ethical reasoning, could enable more nuanced model adaptations. Additionally, advancing black-box model editing techniques would improve accessibility for proprietary LLMs.
    \item \textbf{Advancing Preference Optimization for Human-AI Alignment} – Incorporating more fine-grained human preferences, cultural considerations, and ethical constraints into alignment processes to ensure AI systems reflect societal values more accurately.
    \item \textbf{Robust and Adaptive Evaluation Metrics} – Developing more comprehensive benchmarking methodologies that integrate human feedback, adversarial testing, and real-world deployment scenarios to assess alignment robustness across diverse applications.
\end{itemize}

In conclusion, this thesis lays the groundwork for future research in LLM alignment, pushing towards models that are not only more powerful but also more efficient, adaptable, and aligned with human intent.

\section{List of Publications}

\begin{itemize}
    \item \textbf{Yuxin Jiang}, Yufei Wang, Chuhan Wu, Xinyi Dai, Yan Xu, Weinan Gan, Yasheng Wang, Xin Jiang, Lifeng Shang, Ruiming Tang, and Wei Wang. Instruction-Tuning Data Synthesis from Scratch via Web Reconstruction. In \textit{Findings of the Association for Computational Linguistics: ACL 2025}. Association for Computational Linguistics, 2025.
    \item \textbf{Yuxin Jiang}, Bo Huang, Yufei Wang, Xingshan Zeng, Liangyou Li, Yasheng Wang, Xin Jiang, Lifeng Shang, Ruiming Tang, and Wei Wang. Bridging and modeling correlations in pairwise data for direct preference optimization. In \textit{The Thirteenth International Conference on Learning Representations (ICLR)}, 2025.
    \item \textbf{Yuxin Jiang}, Yufei Wang, Chuhan Wu, Wanjun Zhong, Xingshan Zeng, Jiahui Gao, Liangyou Li, Xin Jiang, Lifeng Shang, Ruiming Tang, Qun Liu, and Wei Wang. Learning to edit: Aligning LLMs with knowledge editing. In \textit{Proceedings of the 62nd Annual Meeting of the Association for Computational Linguistics (Volume 1: Long Papers)}, pages 4689–4705. Association for Computational Linguistics, 2024.
    \item \textbf{Yuxin Jiang}, Yufei Wang, Xingshan Zeng, Wanjun Zhong, Liangyou Li, Fei Mi, Lifeng Shang, Xin Jiang, Qun Liu, and Wei Wang. FollowBench: A multi-level fine-grained constraints following benchmark for large language models. In \textit{Proceedings of the 62nd Annual Meeting of the Association for Computational Linguistics (Volume 1: Long Papers)}, pages 4667–4688. Association for Computational Linguistics, 2024.
    \item \textbf{Yuxin Jiang}, Chunkit Chan, Mingyang Chen, and Wei Wang. Lion: Adversarial distillation of proprietary large language models. In \textit{Proceedings of the 2023 Conference on Empirical Methods in Natural Language Processing}, pages 3134–3154. Association for Computational Linguistics, 2023.
    \item \textbf{Yuxin Jiang}, Linhan Zhang, and Wei Wang. Global and local hierarchy-aware contrastive framework for implicit discourse relation recognition. In \textit{Findings of the Association for Computational Linguistics: ACL 2023}, pages 8048–8064. Association for Computational Linguistics, 2023.
    \item \textbf{Yuxin Jiang}, Linhan Zhang, and Wei Wang. Improved universal sentence embeddings with prompt-based contrastive learning and energy-based learning. In \textit{Findings of the Association for Computational Linguistics: EMNLP 2022}, pages 3021–3035. Association for Computational Linguistics, 2022.
\end{itemize}

\bibliographystyle{plain}
\bibliography{citations}

\begin{thebibliography}{100}

\bibitem{agg2021ecqa}
Shourya Aggarwal, Divyanshu Mandowara, Vishwajeet Agrawal, Dinesh Khandelwal, Parag Singla, and Dinesh Garg.
\newblock Explanations for commonsenseqa: New dataset and models.
\newblock In {\em Annual Meeting of the Association for Computational Linguistics}, pages 3050--3065, 2021.

\bibitem{ankner2024perplexed}
Zachary Ankner, Cody Blakeney, Kartik Sreenivasan, Max Marion, Matthew~L Leavitt, and Mansheej Paul.
\newblock Perplexed by perplexity: Perplexity-based data pruning with small reference models.
\newblock {\em arXiv preprint arXiv:2405.20541}, 2024.

\bibitem{arxivdataset}
arXiv.org submitters.
\newblock arxiv dataset, 2023.

\bibitem{askell2021general}
Amanda Askell, Yuntao Bai, Anna Chen, Dawn Drain, Deep Ganguli, Tom Henighan, Andy Jones, Nicholas Joseph, Ben Mann, Nova DasSarma, et~al.
\newblock A general language assistant as a laboratory for alignment.
\newblock {\em arXiv preprint arXiv:2112.00861}, 2021.

\bibitem{austin2021program}
Jacob Austin, Augustus Odena, Maxwell Nye, Maarten Bosma, Henryk Michalewski, David Dohan, Ellen Jiang, Carrie Cai, Michael Terry, Quoc Le, et~al.
\newblock Program synthesis with large language models.
\newblock {\em arXiv preprint arXiv:2108.07732}, 2021.

\bibitem{moh2024ipo}
Mohammad~Gheshlaghi Azar, Zhaohan~Daniel Guo, Bilal Piot, R{\'{e}}mi Munos, Mark Rowland, Michal Valko, and Daniele Calandriello.
\newblock A general theoretical paradigm to understand learning from human preferences.
\newblock In {\em International Conference on Artificial Intelligence and Statistics}, volume 238, pages 4447--4455, 2024.

\bibitem{bach-etal-2022-promptsource}
Stephen Bach, Victor Sanh, Zheng~Xin Yong, Albert Webson, Colin Raffel, Nihal~V. Nayak, Abheesht Sharma, Taewoon Kim, M~Saiful Bari, Thibault Fevry, Zaid Alyafeai, Manan Dey, Andrea Santilli, Zhiqing Sun, Srulik Ben-david, Canwen Xu, Gunjan Chhablani, Han Wang, Jason Fries, Maged Al-shaibani, Shanya Sharma, Urmish Thakker, Khalid Almubarak, Xiangru Tang, Dragomir Radev, Mike Tian-jian Jiang, and Alexander Rush.
\newblock {P}rompt{S}ource: An integrated development environment and repository for natural language prompts.
\newblock In {\em Proceedings of the 60th Annual Meeting of the Association for Computational Linguistics: System Demonstrations}, pages 93--104, Dublin, Ireland, May 2022. Association for Computational Linguistics.

\bibitem{qwen}
Jinze Bai, Shuai Bai, Yunfei Chu, Zeyu Cui, Kai Dang, Xiaodong Deng, Yang Fan, Wenbin Ge, Yu~Han, Fei Huang, Binyuan Hui, Luo Ji, Mei Li, Junyang Lin, Runji Lin, Dayiheng Liu, Gao Liu, Chengqiang Lu, Keming Lu, Jianxin Ma, Rui Men, Xingzhang Ren, Xuancheng Ren, Chuanqi Tan, Sinan Tan, Jianhong Tu, Peng Wang, Shijie Wang, Wei Wang, Shengguang Wu, Benfeng Xu, Jin Xu, An~Yang, Hao Yang, Jian Yang, Shusheng Yang, Yang Yao, Bowen Yu, Hongyi Yuan, Zheng Yuan, Jianwei Zhang, Xingxuan Zhang, Yichang Zhang, Zhenru Zhang, Chang Zhou, Jingren Zhou, Xiaohuan Zhou, and Tianhang Zhu.
\newblock Qwen technical report.
\newblock {\em arXiv preprint arXiv:2309.16609}, 2023.

\bibitem{bai2022training}
Yuntao Bai, Andy Jones, Kamal Ndousse, Amanda Askell, Anna Chen, Nova DasSarma, Dawn Drain, Stanislav Fort, Deep Ganguli, Tom Henighan, et~al.
\newblock Training a helpful and harmless assistant with reinforcement learning from human feedback.
\newblock {\em arXiv preprint arXiv:2204.05862}, 2022.

\bibitem{bai2022constitutional}
Yuntao Bai, Saurav Kadavath, Sandipan Kundu, Amanda Askell, Jackson Kernion, Andy Jones, Anna Chen, Anna Goldie, Azalia Mirhoseini, Cameron McKinnon, et~al.
\newblock Constitutional ai: Harmlessness from ai feedback.
\newblock In {\em arXiv}, 2022.

\bibitem{baichuan2023baichuan2}
Baichuan.
\newblock Baichuan 2: Open large-scale language models.
\newblock {\em arXiv preprint arXiv:2309.10305}, 2023.

\bibitem{bisk2020piqa}
Yonatan Bisk, Rowan Zellers, Ronan~Le Bras, Jianfeng Gao, and Yejin Choi.
\newblock {PIQA:} reasoning about physical commonsense in natural language.
\newblock In {\em The Thirty-Fourth {AAAI} Conference on Artificial Intelligence, {AAAI} 2020, The Thirty-Second Innovative Applications of Artificial Intelligence Conference, {IAAI} 2020, The Tenth {AAAI} Symposium on Educational Advances in Artificial Intelligence, {EAAI} 2020, New York, NY, USA, February 7-12, 2020}, pages 7432--7439. {AAAI} Press, 2020.

\bibitem{bommasani2021opportunities}
Rishi Bommasani, Drew~A Hudson, Ehsan Adeli, Russ Altman, Simran Arora, Sydney von Arx, Michael~S Bernstein, Jeannette Bohg, Antoine Bosselut, Emma Brunskill, et~al.
\newblock On the opportunities and risks of foundation models.
\newblock {\em arXiv preprint arXiv:2108.07258}, 2021.

\bibitem{bradley1952rank}
Ralph~Allan Bradley and Milton~E Terry.
\newblock Rank analysis of incomplete block designs: I. the method of paired comparisons.
\newblock {\em Biometrika}, 39(3/4):324--345, 1952.

\bibitem{broder1997resemblance}
Andrei~Z Broder.
\newblock On the resemblance and containment of documents.
\newblock In {\em Proceedings. Compression and Complexity of SEQUENCES 1997 (Cat. No. 97TB100171)}, pages 21--29. IEEE, 1997.

\bibitem{brown2020language}
Tom Brown, Benjamin Mann, Nick Ryder, Melanie Subbiah, Jared~D Kaplan, Prafulla Dhariwal, Arvind Neelakantan, Pranav Shyam, Girish Sastry, Amanda Askell, Sandhini Agarwal, Ariel Herbert-Voss, Gretchen Krueger, Tom Henighan, Rewon Child, Aditya Ramesh, Daniel Ziegler, Jeffrey Wu, Clemens Winter, Chris Hesse, Mark Chen, Eric Sigler, Mateusz Litwin, Scott Gray, Benjamin Chess, Jack Clark, Christopher Berner, Sam McCandlish, Alec Radford, Ilya Sutskever, and Dario Amodei.
\newblock Language models are few-shot learners.
\newblock In H.~Larochelle, M.~Ranzato, R.~Hadsell, M.F. Balcan, and H.~Lin, editors, {\em Advances in Neural Information Processing Systems}, volume~33, pages 1877--1901. Curran Associates, Inc., 2020.

\bibitem{bukharin2023data}
Alexander Bukharin and Tuo Zhao.
\newblock Data diversity matters for robust instruction tuning.
\newblock {\em arXiv preprint arXiv:2311.14736}, 2023.

\bibitem{cao2024drlc}
Meng Cao, Lei Shu, Lei Yu, Yun Zhu, Nevan Wichers, Yinxiao Liu, and Lei Meng.
\newblock Drlc: Reinforcement learning with dense rewards from llm critic.
\newblock In {\em arXiv}, 2024.

\bibitem{cao2023instruction}
Yihan Cao, Yanbin Kang, and Lichao Sun.
\newblock Instruction mining: High-quality instruction data selection for large language models.
\newblock {\em arXiv preprint arXiv:2307.06290}, 1(3):6, 2023.

\bibitem{cettolo2012wit3}
Mauro Cettolo, Christian Girardi, and Marcello Federico.
\newblock Wit3: Web inventory of transcribed and translated talks.
\newblock In {\em Proceedings of the Conference of European Association for Machine Translation (EAMT)}, pages 261--268, 2012.

\bibitem{chan2024dense}
Alex~J Chan, Hao Sun, Samuel Holt, and Mihaela van~der Schaar.
\newblock Dense reward for free in reinforcement learning from human feedback.
\newblock In {\em arXiv}, 2024.

\bibitem{chen2023maybe}
Hao Chen, Yiming Zhang, Qi~Zhang, Hantao Yang, Xiaomeng Hu, Xuetao Ma, Yifan Yanggong, and Junbo Zhao.
\newblock Maybe only 0.5\% data is needed: A preliminary exploration of low training data instruction tuning.
\newblock {\em arXiv preprint arXiv:2305.09246}, 2023.

\bibitem{chen2022controllable}
Howard Chen, Huihan Li, Danqi Chen, and Karthik Narasimhan.
\newblock Controllable text generation with language constraints.
\newblock {\em arXiv preprint arXiv:2212.10466}, 2022.

\bibitem{chen2023alpagasus}
Lichang Chen, Shiyang Li, Jun Yan, Hai Wang, Kalpa Gunaratna, Vikas Yadav, Zheng Tang, Vijay Srinivasan, Tianyi Zhou, Heng Huang, et~al.
\newblock Alpagasus: Training a better alpaca with fewer data.
\newblock {\em arXiv preprint arXiv:2307.08701}, 2023.

\bibitem{chen2021codex}
Mark Chen, Jerry Tworek, Heewoo Jun, Qiming Yuan, Henrique~Ponde de~Oliveira~Pinto, Jared Kaplan, Harri Edwards, Yuri Burda, Nicholas Joseph, Greg Brockman, Alex Ray, Raul Puri, Gretchen Krueger, Michael Petrov, Heidy Khlaaf, Girish Sastry, Pamela Mishkin, Brooke Chan, Scott Gray, Nick Ryder, Mikhail Pavlov, Alethea Power, Lukasz Kaiser, Mohammad Bavarian, Clemens Winter, Philippe Tillet, Felipe~Petroski Such, Dave Cummings, Matthias Plappert, Fotios Chantzis, Elizabeth Barnes, Ariel Herbert-Voss, William~Hebgen Guss, Alex Nichol, Alex Paino, Nikolas Tezak, Jie Tang, Igor Babuschkin, Suchir Balaji, Shantanu Jain, William Saunders, Christopher Hesse, Andrew~N. Carr, Jan Leike, Josh Achiam, Vedant Misra, Evan Morikawa, Alec Radford, Matthew Knight, Miles Brundage, Mira Murati, Katie Mayer, Peter Welinder, Bob McGrew, Dario Amodei, Sam McCandlish, Ilya Sutskever, and Wojciech Zaremba.
\newblock Evaluating large language models trained on code.
\newblock 2021.

\bibitem{chen-etal-2024-dog}
Yongrui Chen, Haiyun Jiang, Xinting Huang, Shuming Shi, and Guilin Qi.
\newblock {D}o{G}-instruct: Towards premium instruction-tuning data via text-grounded instruction wrapping.
\newblock In Kevin Duh, Helena Gomez, and Steven Bethard, editors, {\em Proceedings of the 2024 Conference of the North American Chapter of the Association for Computational Linguistics: Human Language Technologies (Volume 1: Long Papers)}, pages 4125--4135, Mexico City, Mexico, June 2024. Association for Computational Linguistics.

\bibitem{chen2024rlmec}
Zhipeng Chen, Kun Zhou, Wayne~Xin Zhao, Junchen Wan, Fuzheng Zhang, Di~Zhang, and Ji{-}Rong Wen.
\newblock Improving large language models via fine-grained reinforcement learning with minimum editing constraint.
\newblock In {\em Findings of the Association for Computational Linguistics}, pages 5694--5711, 2024.

\bibitem{chen2024allo}
Zhipeng Chen, Kun Zhou, Wayne~Xin Zhao, Jingyuan Wang, and Ji-Rong Wen.
\newblock Low-redundant optimization for large language model alignment.
\newblock In {\em arXiv}, 2024.

\bibitem{cheng2022replacing}
Pengyu Cheng and Ruineng Li.
\newblock Replacing language model for style transfer.
\newblock {\em arXiv preprint arXiv:2211.07343}, 2022.

\bibitem{cheng-etal-2023-edit}
Siyuan Cheng, Bozhong Tian, Qingbin Liu, Xi~Chen, Yongheng Wang, Huajun Chen, and Ningyu Zhang.
\newblock Can we edit multimodal large language models?
\newblock In Houda Bouamor, Juan Pino, and Kalika Bali, editors, {\em Proceedings of the 2023 Conference on Empirical Methods in Natural Language Processing}, pages 13877--13888, Singapore, December 2023. Association for Computational Linguistics.

\bibitem{vicuna2023}
Wei-Lin Chiang, Zhuohan Li, Zi~Lin, Ying Sheng, Zhanghao Wu, Hao Zhang, Lianmin Zheng, Siyuan Zhuang, Yonghao Zhuang, Joseph~E. Gonzalez, Ion Stoica, and Eric~P. Xing.
\newblock Vicuna: An open-source chatbot impressing gpt-4 with 90\%* chatgpt quality, March 2023.

\bibitem{cho2014learning}
Kyunghyun Cho, Bart Van~Merri{\"e}nboer, Caglar Gulcehre, Dzmitry Bahdanau, Fethi Bougares, Holger Schwenk, and Yoshua Bengio.
\newblock Learning phrase representations using rnn encoder-decoder for statistical machine translation.
\newblock {\em arXiv preprint arXiv:1406.1078}, 2014.

\bibitem{2023palm}
Aakanksha Chowdhery, Sharan Narang, Jacob Devlin, Maarten Bosma, Gaurav Mishra, Adam Roberts, Paul Barham, Hyung~Won Chung, Charles Sutton, Sebastian Gehrmann, Parker Schuh, Kensen Shi, Sasha Tsvyashchenko, Joshua Maynez, Abhishek Rao, Parker Barnes, Yi~Tay, Noam Shazeer, Vinodkumar Prabhakaran, Emily Reif, Nan Du, Ben Hutchinson, Reiner Pope, James Bradbury, Jacob Austin, Michael Isard, Guy Gur{-}Ari, Pengcheng Yin, Toju Duke, Anselm Levskaya, Sanjay Ghemawat, Sunipa Dev, Henryk Michalewski, Xavier Garcia, Vedant Misra, Kevin Robinson, Liam Fedus, Denny Zhou, Daphne Ippolito, David Luan, Hyeontaek Lim, Barret Zoph, Alexander Spiridonov, Ryan Sepassi, David Dohan, Shivani Agrawal, Mark Omernick, Andrew~M. Dai, Thanumalayan~Sankaranarayana Pillai, Marie Pellat, Aitor Lewkowycz, Erica Moreira, Rewon Child, Oleksandr Polozov, Katherine Lee, Zongwei Zhou, Xuezhi Wang, Brennan Saeta, Mark Diaz, Orhan Firat, Michele Catasta, Jason Wei, Kathy Meier{-}Hellstern, Douglas Eck, Jeff Dean, Slav Petrov, and Noah Fiedel.
\newblock Palm: Scaling language modeling with pathways.
\newblock {\em J. Mach. Learn. Res.}, 24:240:1--240:113, 2023.

\bibitem{PAUL2017RLHF}
Paul~F. Christiano, Jan Leike, Tom~B. Brown, Miljan Martic, Shane Legg, and Dario Amodei.
\newblock Deep reinforcement learning from human preferences.
\newblock In {\em Advances in Neural Information Processing Systems}, pages 4299--4307, 2017.

\bibitem{clark2021all}
Elizabeth Clark, Tal August, Sofia Serrano, Nikita Haduong, Suchin Gururangan, and Noah~A Smith.
\newblock All that's' human'is not gold: Evaluating human evaluation of generated text.
\newblock {\em arXiv preprint arXiv:2107.00061}, 2021.

\bibitem{clark2018think}
Peter Clark, Isaac Cowhey, Oren Etzioni, Tushar Khot, Ashish Sabharwal, Carissa Schoenick, and Oyvind Tafjord.
\newblock Think you have solved question answering? try arc, the ai2 reasoning challenge, 2018.

\bibitem{clusmann2023future}
Jan Clusmann, Fiona~R Kolbinger, Hannah~Sophie Muti, Zunamys~I Carrero, Jan-Niklas Eckardt, Narmin~Ghaffari Laleh, Chiara Maria~Lavinia L{\"o}ffler, Sophie-Caroline Schwarzkopf, Michaela Unger, Gregory~P Veldhuizen, et~al.
\newblock The future landscape of large language models in medicine.
\newblock {\em Communications medicine}, 3(1):141, 2023.

\bibitem{cobbe2021gsm8k}
Karl Cobbe, Vineet Kosaraju, Mohammad Bavarian, Mark Chen, Heewoo Jun, Lukasz Kaiser, Matthias Plappert, Jerry Tworek, Jacob Hilton, Reiichiro Nakano, Christopher Hesse, and John Schulman.
\newblock Training verifiers to solve math word problems.
\newblock {\em arXiv preprint arXiv:2110.14168}, 2021.

\bibitem{cohen2024ice}
Roi Cohen, Eden Biran, Ori Yoran, Amir Globerson, and Mor Geva.
\newblock Evaluating the ripple effects of knowledge editing in language models.
\newblock {\em CoRR}, abs/2307.12976, 2023.

\bibitem{together2023redpajama}
Together Computer.
\newblock Redpajama: An open source recipe to reproduce llama training dataset, 2023.

\bibitem{DatabricksBlog2023DollyV2}
Mike Conover, Matt Hayes, Ankit Mathur, Jianwei Xie, Jun Wan, Sam Shah, Ali Ghodsi, Patrick Wendell, Matei Zaharia, and Reynold Xin.
\newblock Free dolly: Introducing the world's first truly open instruction-tuned llm, 2023.

\bibitem{2023opencompass}
OpenCompass Contributors.
\newblock Opencompass: A universal evaluation platform for foundation models.
\newblock \url{https://github.com/open-compass/opencompass}, 2023.

\bibitem{cui2023ultrafeedback}
Ganqu Cui, Lifan Yuan, Ning Ding, Guanming Yao, Wei Zhu, Yuan Ni, Guotong Xie, Zhiyuan Liu, and Maosong Sun.
\newblock Ultrafeedback: Boosting language models with high-quality feedback.
\newblock In {\em arXiv}, 2023.

\bibitem{dai2022kn}
Damai Dai, Li~Dong, Yaru Hao, Zhifang Sui, Baobao Chang, and Furu Wei.
\newblock Knowledge neurons in pretrained transformers.
\newblock In Smaranda Muresan, Preslav Nakov, and Aline Villavicencio, editors, {\em Proceedings of the 60th Annual Meeting of the Association for Computational Linguistics (Volume 1: Long Papers), {ACL} 2022, Dublin, Ireland, May 22-27, 2022}, pages 8493--8502. Association for Computational Linguistics, 2022.

\bibitem{cao2021ke}
Nicola De~Cao, Wilker Aziz, and Ivan Titov.
\newblock Editing factual knowledge in language models.
\newblock In Marie-Francine Moens, Xuanjing Huang, Lucia Specia, and Scott Wen-tau Yih, editors, {\em Proceedings of the 2021 Conference on Empirical Methods in Natural Language Processing}, pages 6491--6506, Online and Punta Cana, Dominican Republic, November 2021. Association for Computational Linguistics.

\bibitem{devlin2019bert}
Jacob Devlin, Ming-Wei Chang, Kenton Lee, and Kristina Toutanova.
\newblock Bert: Pre-training of deep bidirectional transformers for language understanding.
\newblock In {\em Proceedings of the 2019 conference of the North American chapter of the association for computational linguistics: human language technologies, volume 1 (long and short papers)}, pages 4171--4186, 2019.

\bibitem{ding2023ultrachat}
Ning Ding, Yulin Chen, Bokai Xu, Yujia Qin, Shengding Hu, Zhiyuan Liu, Maosong Sun, and Bowen Zhou.
\newblock Enhancing chat language models by scaling high-quality instructional conversations.
\newblock In {\em Conference on Empirical Methods in Natural Language Processing}, pages 3029--3051, 2023.

\bibitem{dong2022survey}
Qingxiu Dong, Lei Li, Damai Dai, Ce~Zheng, Jingyuan Ma, Rui Li, Heming Xia, Jingjing Xu, Zhiyong Wu, Tianyu Liu, et~al.
\newblock A survey on in-context learning.
\newblock {\em arXiv preprint arXiv:2301.00234}, 2022.

\bibitem{dosovitskiy2020image}
Alexey Dosovitskiy, Lucas Beyer, Alexander Kolesnikov, Dirk Weissenborn, Xiaohua Zhai, Thomas Unterthiner, Mostafa Dehghani, Matthias Minderer, Georg Heigold, Sylvain Gelly, et~al.
\newblock An image is worth 16x16 words: Transformers for image recognition at scale.
\newblock {\em arXiv preprint arXiv:2010.11929}, 2020.

\bibitem{du2022glm}
Zhengxiao Du, Yujie Qian, Xiao Liu, Ming Ding, Jiezhong Qiu, Zhilin Yang, and Jie Tang.
\newblock Glm: General language model pretraining with autoregressive blank infilling.
\newblock In {\em Proceedings of the 60th Annual Meeting of the Association for Computational Linguistics (Volume 1: Long Papers)}, pages 320--335, 2022.

\bibitem{llama3modelcard}
Abhimanyu Dubey, Abhinav Jauhri, Abhinav Pandey, Abhishek Kadian, Ahmad Al-Dahle, Aiesha Letman, Akhil Mathur, Alan Schelten, Amy Yang, Angela Fan, et~al.
\newblock The llama 3 herd of models.
\newblock In {\em arXiv}, 2024.

\bibitem{dubois2024length}
Yann Dubois, Bal{\'a}zs Galambosi, Percy Liang, and Tatsunori~B Hashimoto.
\newblock Length-controlled alpacaeval: A simple way to debias automatic evaluators.
\newblock In {\em arXiv}, 2024.

\bibitem{DBLP:conf/acl/FadeevaRSPLMTKP24}
Ekaterina Fadeeva, Aleksandr Rubashevskii, Artem Shelmanov, Sergey Petrakov, Haonan Li, Hamdy Mubarak, Evgenii Tsymbalov, Gleb Kuzmin, Alexander Panchenko, Timothy Baldwin, Preslav Nakov, and Maxim Panov.
\newblock Fact-checking the output of large language models via token-level uncertainty quantification.
\newblock In {\em Findings of the Association for Computational Linguistics}, pages 9367--9385, 2024.

\bibitem{fan-etal-2018-hierarchical}
Angela Fan, Mike Lewis, and Yann Dauphin.
\newblock Hierarchical neural story generation.
\newblock In {\em Proceedings of the 56th Annual Meeting of the Association for Computational Linguistics (Volume 1: Long Papers)}, pages 889--898, Melbourne, Australia, July 2018. Association for Computational Linguistics.

\bibitem{fang2019akd}
Gongfan Fang, Jie Song, Chengchao Shen, Xinchao Wang, Da~Chen, and Mingli Song.
\newblock Data-free adversarial distillation.
\newblock {\em CoRR}, abs/1912.11006, 2019.

\bibitem{DBLP:conf/naacl/FuNJ024}
Jinlan Fu, See{-}Kiong Ng, Zhengbao Jiang, and Pengfei Liu.
\newblock Gptscore: Evaluate as you desire.
\newblock In {\em Conference of the North American Chapter of the Association for Computational Linguistics}, pages 6556--6576, 2024.

\bibitem{furnkranz2010preference}
Johannes F{\"u}rnkranz and Eyke H{\"u}llermeier.
\newblock Preference learning and ranking by pairwise comparison.
\newblock In {\em Preference learning}, pages 65--82. Springer, 2010.

\bibitem{DBLP:journals/corr/abs-2406-20094}
Tao Ge, Xin Chan, Xiaoyang Wang, Dian Yu, Haitao Mi, and Dong Yu.
\newblock Scaling synthetic data creation with 1,000,000,000 personas.
\newblock {\em arXiv preprint arXiv:2406.20094}, 2024.

\bibitem{geng2023koala}
Xinyang Geng, Arnav Gudibande, Hao Liu, Eric Wallace, Pieter Abbeel, Sergey Levine, and Dawn Song.
\newblock Koala: A dialogue model for academic research.
\newblock {\em Blog post, April}, 1, 2023.

\bibitem{geva2021StrategyQA}
Mor Geva, Daniel Khashabi, Elad Segal, Tushar Khot, Dan Roth, and Jonathan Berant.
\newblock Did aristotle use a laptop? {A} question answering benchmark with implicit reasoning strategies.
\newblock {\em Transactions of the Association for Computational Linguistics}, 9:346--361, 2021.

\bibitem{gilardi2023chatgpt}
Fabrizio Gilardi, Meysam Alizadeh, and Ma{\"e}l Kubli.
\newblock Chatgpt outperforms crowd-workers for text-annotation tasks.
\newblock {\em arXiv preprint arXiv:2303.15056}, 2023.

\bibitem{gliwa-etal-2019-samsum}
Bogdan Gliwa, Iwona Mochol, Maciej Biesek, and Aleksander Wawer.
\newblock {SAMS}um corpus: A human-annotated dialogue dataset for abstractive summarization.
\newblock In {\em Proceedings of the 2nd Workshop on New Frontiers in Summarization}, pages 70--79, Hong Kong, China, November 2019. Association for Computational Linguistics.

\bibitem{Google2023}
Google.
\newblock Bard, 2023.

\bibitem{graff2003english}
David Graff, Junbo Kong, Ke~Chen, and Kazuaki Maeda.
\newblock English gigaword.
\newblock {\em Linguistic Data Consortium, Philadelphia}, 4(1):34, 2003.

\bibitem{gu2024survey}
Jiawei Gu, Xuhui Jiang, Zhichao Shi, Hexiang Tan, Xuehao Zhai, Chengjin Xu, Wei Li, Yinghan Shen, Shengjie Ma, Honghao Liu, et~al.
\newblock A survey on llm-as-a-judge.
\newblock {\em arXiv preprint arXiv:2411.15594}, 2024.

\bibitem{DBLP:journals/corr/abs-2305-15717}
Arnav Gudibande, Eric Wallace, Charlie Snell, Xinyang Geng, Hao Liu, Pieter Abbeel, Sergey Levine, and Dawn Song.
\newblock The false promise of imitating proprietary llms.
\newblock {\em CoRR}, abs/2305.15717, 2023.

\bibitem{guo2025deepseek}
Daya Guo, Dejian Yang, Haowei Zhang, Junxiao Song, Ruoyu Zhang, Runxin Xu, Qihao Zhu, Shirong Ma, Peiyi Wang, Xiao Bi, et~al.
\newblock Deepseek-r1: Incentivizing reasoning capability in llms via reinforcement learning.
\newblock {\em arXiv preprint arXiv:2501.12948}, 2025.

\bibitem{guo2024figa}
Geyang Guo, Ranchi Zhao, Tianyi Tang, Xin Zhao, and Ji-Rong Wen.
\newblock Beyond imitation: Leveraging fine-grained quality signals for alignment.
\newblock In {\em International Conference on Learning Representations}, 2024.

\bibitem{hadfield2016cooperative}
Dylan Hadfield-Menell, Stuart~J Russell, Pieter Abbeel, and Anca Dragan.
\newblock Cooperative inverse reinforcement learning.
\newblock {\em Advances in neural information processing systems}, 29, 2016.

\bibitem{hartvigsen2023grace}
Thomas Hartvigsen, Swami Sankaranarayanan, Hamid Palangi, Yoon Kim, and Marzyeh Ghassemi.
\newblock Aging with grace: Lifelong model editing with discrete key-value adaptors.
\newblock In {\em Advances in Neural Information Processing Systems}, 2023.

\bibitem{hase2023does}
Peter Hase, Mohit Bansal, Been Kim, and Asma Ghandeharioun.
\newblock Does localization inform editing? surprising differences in causality-based localization vs. knowledge editing in language models, 2023.

\bibitem{dan2021mmlu}
Dan Hendrycks, Collin Burns, Steven Basart, Andy Zou, Mantas Mazeika, Dawn Song, and Jacob Steinhardt.
\newblock Measuring massive multitask language understanding.
\newblock In {\em 9th International Conference on Learning Representations, {ICLR} 2021, Virtual Event, Austria, May 3-7, 2021}. OpenReview.net, 2021.

\bibitem{hendrycksmath2021}
Dan Hendrycks, Collin Burns, Saurav Kadavath, Akul Arora, Steven Basart, Eric Tang, Dawn Song, and Jacob Steinhardt.
\newblock Measuring mathematical problem solving with the math dataset.
\newblock {\em NeurIPS}, 2021.

\bibitem{DBLP:conf/aaai/HeoLY019}
Byeongho Heo, Minsik Lee, Sangdoo Yun, and Jin~Young Choi.
\newblock Knowledge distillation with adversarial samples supporting decision boundary.
\newblock In {\em The Thirty-Third {AAAI} Conference on Artificial Intelligence, {AAAI} 2019, The Thirty-First Innovative Applications of Artificial Intelligence Conference, {IAAI} 2019, The Ninth {AAAI} Symposium on Educational Advances in Artificial Intelligence, {EAAI} 2019, Honolulu, Hawaii, USA, January 27 - February 1, 2019}, pages 3771--3778. {AAAI} Press, 2019.

\bibitem{hong2024opro}
Jiwoo Hong, Noah Lee, and James Thorne.
\newblock Orpo: Monolithic preference optimization without reference model.
\newblock In {\em arXiv}, 2024.

\bibitem{houlsby2019parameter}
Neil Houlsby, Andrei Giurgiu, Stanislaw Jastrzebski, Bruna Morrone, Quentin De~Laroussilhe, Andrea Gesmundo, Mona Attariyan, and Sylvain Gelly.
\newblock Parameter-efficient transfer learning for nlp.
\newblock In {\em International conference on machine learning}, pages 2790--2799. PMLR, 2019.

\bibitem{hu2022lora}
Edward~J Hu, Yelong Shen, Phillip Wallis, Zeyuan Allen-Zhu, Yuanzhi Li, Shean Wang, Lu~Wang, and Weizhu Chen.
\newblock Lo{RA}: Low-rank adaptation of large language models.
\newblock In {\em International Conference on Learning Representations}, 2022.

\bibitem{huang2023chatgpt}
F~Huang, H~Kwak, and J~An.
\newblock Is chatgpt better than human annotators? potential and limitations of chatgpt in explaining implicit hate speech. arxiv, 2023.

\bibitem{huang2020deep}
Yanhua Huang.
\newblock Deep q-networks.
\newblock {\em Deep reinforcement learning: fundamentals, research and applications}, pages 135--160, 2020.

\bibitem{huang2023c}
Yuzhen Huang, Yuzhuo Bai, Zhihao Zhu, Junlei Zhang, Jinghan Zhang, Tangjun Su, Junteng Liu, Chuancheng Lv, Yikai Zhang, Yao Fu, et~al.
\newblock C-eval: A multi-level multi-discipline chinese evaluation suite for foundation models.
\newblock {\em Advances in Neural Information Processing Systems}, 36:62991--63010, 2023.

\bibitem{ivison2023camelschangingclimateenhancing}
Hamish Ivison, Yizhong Wang, Valentina Pyatkin, Nathan Lambert, Matthew Peters, Pradeep Dasigi, Joel Jang, David Wadden, Noah~A. Smith, Iz~Beltagy, and Hannaneh Hajishirzi.
\newblock Camels in a changing climate: Enhancing lm adaptation with tulu 2, 2023.

\bibitem{ji2024aligner}
Jiaming Ji, Boyuan Chen, Hantao Lou, Donghai Hong, Borong Zhang, Xuehai Pan, Juntao Dai, and Yaodong Yang.
\newblock Aligner: Achieving efficient alignment through weak-to-strong correction.
\newblock In {\em arXiv}, 2024.

\bibitem{jiang2023mistral}
Albert~Q Jiang, Alexandre Sablayrolles, Arthur Mensch, Chris Bamford, Devendra~Singh Chaplot, Diego de~las Casas, Florian Bressand, Gianna Lengyel, Guillaume Lample, Lucile Saulnier, et~al.
\newblock Mistral 7b.
\newblock In {\em arXiv}, 2023.

\bibitem{jin2021medqa}
Di~Jin, Eileen Pan, Nassim Oufattole, Wei-Hung Weng, Hanyi Fang, and Peter Szolovits.
\newblock What disease does this patient have? a large-scale open domain question answering dataset from medical exams.
\newblock {\em Applied Sciences}, 11(14):6421, 2021.

\bibitem{kalpathy2016plus}
Jayashree Kalpathy-Cramer, J~Peter Campbell, Deniz Erdogmus, Peng Tian, Dharanish Kedarisetti, Chace Moleta, James~D Reynolds, Kelly Hutcheson, Michael~J Shapiro, Michael~X Repka, et~al.
\newblock Plus disease in retinopathy of prematurity: improving diagnosis by ranking disease severity and using quantitative image analysis.
\newblock {\em Ophthalmology}, 123(11):2345--2351, 2016.

\bibitem{kaplan2020scaling}
Jared Kaplan, Sam McCandlish, Tom Henighan, Tom~B Brown, Benjamin Chess, Rewon Child, Scott Gray, Alec Radford, Jeffrey Wu, and Dario Amodei.
\newblock Scaling laws for neural language models.
\newblock {\em arXiv preprint arXiv:2001.08361}, 2020.

\bibitem{kaufmann2023survey}
Timo Kaufmann, Paul Weng, Viktor Bengs, and Eyke H{\"u}llermeier.
\newblock A survey of reinforcement learning from human feedback.
\newblock {\em arXiv preprint arXiv:2312.14925}, 10, 2023.

\bibitem{khot2020qasc}
Tushar Khot, Peter Clark, Michal Guerquin, Peter Jansen, and Ashish Sabharwal.
\newblock {QASC:} {A} dataset for question answering via sentence composition.
\newblock In {\em Proceedings of the AAAI Conference on Artificial Intelligence}, pages 8082--8090, 2020.

\bibitem{kingma2014adam}
Diederik~P Kingma and Jimmy Ba.
\newblock Adam: A method for stochastic optimization.
\newblock In {\em arXiv}, 2014.

\bibitem{rik2016mawps}
Rik Koncel{-}Kedziorski, Subhro Roy, Aida Amini, Nate Kushman, and Hannaneh Hajishirzi.
\newblock {MAWPS:} {A} math word problem repository.
\newblock In {\em Conference of the North American Chapter of the Association for Computational Linguistics}, pages 1152--1157, 2016.

\bibitem{kopf2023openassistant}
Andreas K{\"o}pf, Yannic Kilcher, Dimitri Von~R{\"u}tte, Sotiris Anagnostidis, Zhi~Rui Tam, Keith Stevens, Abdullah Barhoum, Duc Nguyen, Oliver Stanley, Rich{\'a}rd Nagyfi, et~al.
\newblock Openassistant conversations-democratizing large language model alignment.
\newblock {\em Advances in Neural Information Processing Systems}, 36:47669--47681, 2023.

\bibitem{kulal2019spoc}
Sumith Kulal, Panupong Pasupat, Kartik Chandra, Mina Lee, Oded Padon, Alex Aiken, and Percy~S Liang.
\newblock Spoc: Search-based pseudocode to code.
\newblock {\em Advances in Neural Information Processing Systems}, 32, 2019.

\bibitem{DBLP:journals/corr/abs-2401-16745}
Wai{-}Chung Kwan, Xingshan Zeng, Yuxin Jiang, Yufei Wang, Liangyou Li, Lifeng Shang, Xin Jiang, Qun Liu, and Kam{-}Fai Wong.
\newblock Mt-eval: {A} multi-turn capabilities evaluation benchmark for large language models.
\newblock {\em CoRR}, abs/2401.16745, 2024.

\bibitem{lai2024large}
Jinqi Lai, Wensheng Gan, Jiayang Wu, Zhenlian Qi, and Philip~S Yu.
\newblock Large language models in law: A survey.
\newblock {\em AI Open}, 2024.

\bibitem{lai2023ds}
Yuhang Lai, Chengxi Li, Yiming Wang, Tianyi Zhang, Ruiqi Zhong, Luke Zettlemoyer, Wen-tau Yih, Daniel Fried, Sida Wang, and Tao Yu.
\newblock Ds-1000: A natural and reliable benchmark for data science code generation.
\newblock In {\em International Conference on Machine Learning}, pages 18319--18345. PMLR, 2023.

\bibitem{DBLP:conf/icml/0001PMMFLBHCRP24}
Harrison Lee, Samrat Phatale, Hassan Mansoor, Thomas Mesnard, Johan Ferret, Kellie Lu, Colton Bishop, Ethan Hall, Victor Carbune, Abhinav Rastogi, and Sushant Prakash.
\newblock {RLAIF} vs. {RLHF:} scaling reinforcement learning from human feedback with {AI} feedback.
\newblock In {\em International Conference on Machine Learning}, 2024.

\bibitem{leike2018scalable}
Jan Leike, David Krueger, Tom Everitt, Miljan Martic, Vishal Maini, and Shane Legg.
\newblock Scalable agent alignment via reward modeling: a research direction.
\newblock {\em arXiv preprint arXiv:1811.07871}, 2018.

\bibitem{levy2017zsre}
Omer Levy, Minjoon Seo, Eunsol Choi, and Luke Zettlemoyer.
\newblock Zero-shot relation extraction via reading comprehension.
\newblock In Roger Levy and Lucia Specia, editors, {\em Proceedings of the 21st Conference on Computational Natural Language Learning (CoNLL 2017), Vancouver, Canada, August 3-4, 2017}, pages 333--342. Association for Computational Linguistics, 2017.

\bibitem{lewis2020rag}
Patrick S.~H. Lewis, Ethan Perez, Aleksandra Piktus, Fabio Petroni, Vladimir Karpukhin, Naman Goyal, Heinrich K{\"{u}}ttler, Mike Lewis, Wen{-}tau Yih, Tim Rockt{\"{a}}schel, Sebastian Riedel, and Douwe Kiela.
\newblock Retrieval-augmented generation for knowledge-intensive {NLP} tasks.
\newblock In Hugo Larochelle, Marc'Aurelio Ranzato, Raia Hadsell, Maria{-}Florina Balcan, and Hsuan{-}Tien Lin, editors, {\em Advances in Neural Information Processing Systems 33: Annual Conference on Neural Information Processing Systems 2020, NeurIPS 2020, December 6-12, 2020, virtual}, 2020.

\bibitem{li2023cmmlu}
Haonan Li, Yixuan Zhang, Fajri Koto, Yifei Yang, Hai Zhao, Yeyun Gong, Nan Duan, and Timothy Baldwin.
\newblock Cmmlu: Measuring massive multitask language understanding in chinese.
\newblock {\em arXiv preprint arXiv:2306.09212}, 2023.

\bibitem{li2024synthetic}
Haoran Li, Qingxiu Dong, Zhengyang Tang, Chaojun Wang, Xingxing Zhang, Haoyang Huang, Shaohan Huang, Xiaolong Huang, Zeqiang Huang, Dongdong Zhang, et~al.
\newblock Synthetic data (almost) from scratch: Generalized instruction tuning for language models.
\newblock {\em arXiv preprint arXiv:2402.13064}, 2024.

\bibitem{li-etal-2024-quantity}
Ming Li, Yong Zhang, Zhitao Li, Jiuhai Chen, Lichang Chen, Ning Cheng, Jianzong Wang, Tianyi Zhou, and Jing Xiao.
\newblock From quantity to quality: Boosting {LLM} performance with self-guided data selection for instruction tuning.
\newblock In Kevin Duh, Helena Gomez, and Steven Bethard, editors, {\em Proceedings of the 2024 Conference of the North American Chapter of the Association for Computational Linguistics: Human Language Technologies (Volume 1: Long Papers)}, pages 7602--7635, Mexico City, Mexico, June 2024. Association for Computational Linguistics.

\bibitem{DBLP:conf/naacl/LiJH21}
Sha Li, Heng Ji, and Jiawei Han.
\newblock Document-level event argument extraction by conditional generation.
\newblock In Kristina Toutanova, Anna Rumshisky, Luke Zettlemoyer, Dilek Hakkani{-}T{\"{u}}r, Iz~Beltagy, Steven Bethard, Ryan Cotterell, Tanmoy Chakraborty, and Yichao Zhou, editors, {\em Proceedings of the 2021 Conference of the North American Chapter of the Association for Computational Linguistics: Human Language Technologies, {NAACL-HLT} 2021, Online, June 6-11, 2021}, pages 894--908. Association for Computational Linguistics, 2021.

\bibitem{li2024live}
Tianle Li, Wei-Lin Chiang, Evan Frick, Lisa Dunlap, Banghua Zhu, Joseph~E Gonzalez, and Ion Stoica.
\newblock From live data to high-quality benchmarks: The arena-hard pipeline, 2024.

\bibitem{li2024selfalignment}
Xian Li, Ping Yu, Chunting Zhou, Timo Schick, Omer Levy, Luke Zettlemoyer, Jason~E Weston, and Mike Lewis.
\newblock Self-alignment with instruction backtranslation.
\newblock In {\em The Twelfth International Conference on Learning Representations}, 2024.

\bibitem{li2021prefix}
Xiang~Lisa Li and Percy Liang.
\newblock Prefix-tuning: Optimizing continuous prompts for generation.
\newblock {\em arXiv preprint arXiv:2101.00190}, 2021.

\bibitem{alpaca_eval}
Xuechen Li, Tianyi Zhang, Yann Dubois, Rohan Taori, Ishaan Gulrajani, Carlos Guestrin, Percy Liang, and Tatsunori~B. Hashimoto.
\newblock Alpacaeval: An automatic evaluator of instruction-following models.
\newblock \url{https://github.com/tatsu-lab/alpaca_eval}, 2023.

\bibitem{li2023large}
Yinheng Li, Shaofei Wang, Han Ding, and Hang Chen.
\newblock Large language models in finance: A survey.
\newblock In {\em Proceedings of the fourth ACM international conference on AI in finance}, pages 374--382, 2023.

\bibitem{lin2004rouge}
Chin-Yew Lin.
\newblock Rouge: A package for automatic evaluation of summaries.
\newblock In {\em Text summarization branches out}, pages 74--81, 2004.

\bibitem{lison-tiedemann-2016-opensubtitles2016}
Pierre Lison and J{\"o}rg Tiedemann.
\newblock {O}pen{S}ubtitles2016: Extracting large parallel corpora from movie and {TV} subtitles.
\newblock In {\em Proceedings of the Tenth International Conference on Language Resources and Evaluation ({LREC}'16)}, pages 923--929, Portoro{\v{z}}, Slovenia, May 2016. European Language Resources Association (ELRA).

\bibitem{liu2023m3ke}
Chuang Liu, Renren Jin, Yuqi Ren, Linhao Yu, Tianyu Dong, Xiaohan Peng, Shuting Zhang, Jianxiang Peng, Peiyi Zhang, Qingqing Lyu, et~al.
\newblock M3ke: A massive multi-level multi-subject knowledge evaluation benchmark for chinese large language models.
\newblock {\em arXiv preprint arXiv:2305.10263}, 2023.

\bibitem{liu2023your}
Jiawei Liu, Chunqiu~Steven Xia, Yuyao Wang, and Lingming Zhang.
\newblock Is your code generated by chatgpt really correct? rigorous evaluation of large language models for code generation.
\newblock {\em Advances in Neural Information Processing Systems}, 36:21558--21572, 2023.

\bibitem{liu2023statistical}
Tianqi Liu, Yao Zhao, Rishabh Joshi, Misha Khalman, Mohammad Saleh, Peter~J Liu, and Jialu Liu.
\newblock Statistical rejection sampling improves preference optimization.
\newblock In {\em International Conference on Learning Representations}, 2024.

\bibitem{liu2023agentbench}
Xiao Liu, Hao Yu, Hanchen Zhang, Yifan Xu, Xuanyu Lei, Hanyu Lai, Yu~Gu, Hangliang Ding, Kaiwen Men, Kejuan Yang, et~al.
\newblock Agentbench: Evaluating llms as agents.
\newblock {\em arXiv preprint arXiv:2308.03688}, 2023.

\bibitem{liu2019roberta}
Yinhan Liu, Myle Ott, Naman Goyal, Jingfei Du, Mandar Joshi, Danqi Chen, Omer Levy, Mike Lewis, Luke Zettlemoyer, and Veselin Stoyanov.
\newblock Roberta: A robustly optimized bert pretraining approach.
\newblock {\em arXiv preprint arXiv:1907.11692}, 2019.

\bibitem{liu2021swin}
Ze~Liu, Yutong Lin, Yue Cao, Han Hu, Yixuan Wei, Zheng Zhang, Stephen Lin, and Baining Guo.
\newblock Swin transformer: Hierarchical vision transformer using shifted windows.
\newblock In {\em Proceedings of the IEEE/CVF international conference on computer vision}, pages 10012--10022, 2021.

\bibitem{lu2023tabmwp}
Pan Lu, Liang Qiu, Kai{-}Wei Chang, Ying~Nian Wu, Song{-}Chun Zhu, Tanmay Rajpurohit, Peter Clark, and Ashwin Kalyan.
\newblock Dynamic prompt learning via policy gradient for semi-structured mathematical reasoning.
\newblock In {\em International Conference on Learning Representations}, 2023.

\bibitem{mckenzie2023inverse}
Ian~R McKenzie, Alexander Lyzhov, Michael Pieler, Alicia Parrish, Aaron Mueller, Ameya Prabhu, Euan McLean, Aaron Kirtland, Alexis Ross, Alisa Liu, et~al.
\newblock Inverse scaling: When bigger isn't better.
\newblock {\em arXiv preprint arXiv:2306.09479}, 2023.

\bibitem{meng2022rome}
Kevin Meng, David Bau, Alex Andonian, and Yonatan Belinkov.
\newblock Locating and editing factual associations in {GPT}.
\newblock In Sanmi Koyejo, S.~Mohamed, A.~Agarwal, Danielle Belgrave, K.~Cho, and A.~Oh, editors, {\em Advances in Neural Information Processing Systems 35: Annual Conference on Neural Information Processing Systems 2022, NeurIPS 2022, New Orleans, LA, USA, November 28 - December 9, 2022}, 2022.

\bibitem{meng2023memit}
Kevin Meng, Arnab~Sen Sharma, Alex~J. Andonian, Yonatan Belinkov, and David Bau.
\newblock Mass-editing memory in a transformer.
\newblock In {\em The Eleventh International Conference on Learning Representations, {ICLR} 2023, Kigali, Rwanda, May 1-5, 2023}. OpenReview.net, 2023.

\bibitem{meng2024simpo}
Yu~Meng, Mengzhou Xia, and Danqi Chen.
\newblock Simpo: Simple preference optimization with a reference-free reward.
\newblock In {\em arXiv}, 2024.

\bibitem{DBLP:conf/nips/MicaelliS19}
Paul Micaelli and Amos~J. Storkey.
\newblock Zero-shot knowledge transfer via adversarial belief matching.
\newblock In Hanna~M. Wallach, Hugo Larochelle, Alina Beygelzimer, Florence d'Alch{\'{e}}{-}Buc, Emily~B. Fox, and Roman Garnett, editors, {\em Advances in Neural Information Processing Systems 32: Annual Conference on Neural Information Processing Systems 2019, NeurIPS 2019, December 8-14, 2019, Vancouver, BC, Canada}, pages 9547--9557, 2019.

\bibitem{miha2018obqa}
Todor Mihaylov, Peter Clark, Tushar Khot, and Ashish Sabharwal.
\newblock Can a suit of armor conduct electricity? {A} new dataset for open book question answering.
\newblock In {\em Conference on Empirical Methods in Natural Language Processing}, pages 2381--2391, 2018.

\bibitem{miller-1992-wordnet}
George~A. Miller.
\newblock {W}ord{N}et: A lexical database for {E}nglish.
\newblock In {\em Speech and Natural Language: Proceedings of a Workshop Held at Harriman, New York, {F}ebruary 23-26, 1992}, 1992.

\bibitem{mishra-etal-2022-cross}
Swaroop Mishra, Daniel Khashabi, Chitta Baral, and Hannaneh Hajishirzi.
\newblock Cross-mishra-etal-2022-cross.
\newblock In Smaranda Muresan, Preslav Nakov, and Aline Villavicencio, editors, {\em Proceedings of the 60th Annual Meeting of the Association for Computational Linguistics (Volume 1: Long Papers)}, pages 3470--3487, Dublin, Ireland, May 2022. Association for Computational Linguistics.

\bibitem{mitchell2022mend}
Eric Mitchell, Charles Lin, Antoine Bosselut, Chelsea Finn, and Christopher~D. Manning.
\newblock Fast model editing at scale.
\newblock In {\em The Tenth International Conference on Learning Representations, {ICLR} 2022, Virtual Event, April 25-29, 2022}. OpenReview.net, 2022.

\bibitem{mitchell2022serac}
Eric Mitchell, Charles Lin, Antoine Bosselut, Christopher~D. Manning, and Chelsea Finn.
\newblock Memory-based model editing at scale.
\newblock In Kamalika Chaudhuri, Stefanie Jegelka, Le~Song, Csaba Szepesv{\'{a}}ri, Gang Niu, and Sivan Sabato, editors, {\em International Conference on Machine Learning, {ICML} 2022, 17-23 July 2022, Baltimore, Maryland, {USA}}, volume 162 of {\em Proceedings of Machine Learning Research}, pages 15817--15831. {PMLR}, 2022.

\bibitem{mostafazadeh-etal-2016-corpus}
Nasrin Mostafazadeh, Nathanael Chambers, Xiaodong He, Devi Parikh, Dhruv Batra, Lucy Vanderwende, Pushmeet Kohli, and James Allen.
\newblock A corpus and cloze evaluation for deeper understanding of commonsense stories.
\newblock In {\em Proceedings of the 2016 Conference of the North {A}merican Chapter of the Association for Computational Linguistics: Human Language Technologies}, pages 839--849, San Diego, California, June 2016. Association for Computational Linguistics.

\bibitem{2023orca}
Subhabrata Mukherjee, Arindam Mitra, Ganesh Jawahar, Sahaj Agarwal, Hamid Palangi, and Ahmed~Hassan Awadallah.
\newblock Orca: Progressive learning from complex explanation traces of {GPT-4}.
\newblock {\em CoRR}, abs/2306.02707, 2023.

\bibitem{nallapati2016abstractive}
Ramesh Nallapati, Bowen Zhou, Caglar Gulcehre, Bing Xiang, et~al.
\newblock Abstractive text summarization using sequence-to-sequence rnns and beyond.
\newblock {\em arXiv preprint arXiv:1602.06023}, 2016.

\bibitem{narayan2018xsum}
Shashi Narayan, Shay~B. Cohen, and Mirella Lapata.
\newblock Don't give me the details, just the summary! topic-aware convolutional neural networks for extreme summarization.
\newblock In Ellen Riloff, David Chiang, Julia Hockenmaier, and Jun'ichi Tsujii, editors, {\em Proceedings of the 2018 Conference on Empirical Methods in Natural Language Processing, Brussels, Belgium, October 31 - November 4, 2018}, pages 1797--1807. Association for Computational Linguistics, 2018.

\bibitem{ng2000algorithms}
Andrew~Y Ng, Stuart Russell, et~al.
\newblock Algorithms for inverse reinforcement learning.
\newblock In {\em Icml}, volume~1, page~2, 2000.

\bibitem{nguyen2024betteralignmentinstructionbackandforth}
Thao Nguyen, Jeffrey Li, Sewoong Oh, Ludwig Schmidt, Jason Weston, Luke Zettlemoyer, and Xian Li.
\newblock Better alignment with instruction back-and-forth translation, 2024.

\bibitem{novikova-etal-2017-e2e}
Jekaterina Novikova, Ond{\v{r}}ej Du{\v{s}}ek, and Verena Rieser.
\newblock The {E}2{E} dataset: New challenges for end-to-end generation.
\newblock In {\em Proceedings of the 18th Annual {SIG}dial Meeting on Discourse and Dialogue}, pages 201--206, Saarbr{\"u}cken, Germany, August 2017. Association for Computational Linguistics.

\bibitem{2023gpt4}
OpenAI.
\newblock {GPT-4} technical report.
\newblock {\em CoRR}, abs/2303.08774, 2023.

\bibitem{openai2022chatgpt}
TB~OpenAI.
\newblock Chatgpt: Optimizing language models for dialogue.
\newblock {\em OpenAI}, 2022.

\bibitem{ouyang2022training}
Long Ouyang, Jeffrey Wu, Xu~Jiang, Diogo Almeida, Carroll Wainwright, Pamela Mishkin, Chong Zhang, Sandhini Agarwal, Katarina Slama, Alex Ray, et~al.
\newblock Training language models to follow instructions with human feedback.
\newblock {\em Advances in Neural Information Processing Systems}, 35:27730--27744, 2022.

\bibitem{ozturk2005preference}
Meltem {\"O}zt{\"u}rk, Alexis Tsouki{\`a}s, and Philippe Vincke.
\newblock Preference modelling.
\newblock {\em Multiple criteria decision analysis: State of the art surveys}, 78:27--59, 2005.

\bibitem{pace2024west}
Aliz{\'e}e Pace, Jonathan Mallinson, Eric Malmi, Sebastian Krause, and Aliaksei Severyn.
\newblock West-of-n: Synthetic preference generation for improved reward modeling.
\newblock In {\em ICLR 2024 Workshop on Navigating and Addressing Data Problems for Foundation Models}, 2024.

\bibitem{pan2022effects}
Alexander Pan, Kush Bhatia, and Jacob Steinhardt.
\newblock The effects of reward misspecification: Mapping and mitigating misaligned models.
\newblock {\em arXiv preprint arXiv:2201.03544}, 2022.

\bibitem{papineni2002bleu}
Kishore Papineni, Salim Roukos, Todd Ward, and Wei-Jing Zhu.
\newblock Bleu: a method for automatic evaluation of machine translation.
\newblock In {\em Proceedings of the 40th annual meeting of the Association for Computational Linguistics}, pages 311--318, 2002.

\bibitem{park2024rdpo}
Ryan Park, Rafael Rafailov, Stefano Ermon, and Chelsea Finn.
\newblock Disentangling length from quality in direct preference optimization.
\newblock In {\em arXiv}, 2024.

\bibitem{paster2024openwebmath}
Keiran Paster, Marco~Dos Santos, Zhangir Azerbayev, and Jimmy Ba.
\newblock Openwebmath: An open dataset of high-quality mathematical web text.
\newblock In {\em The Twelfth International Conference on Learning Representations}, 2024.

\bibitem{pauls2011faster}
Adam Pauls and Dan Klein.
\newblock Faster and smaller n-gram language models.
\newblock In {\em Proceedings of the 49th annual meeting of the Association for Computational Linguistics: Human Language Technologies}, pages 258--267, 2011.

\bibitem{plackett1975analysis}
Robin~L Plackett.
\newblock The analysis of permutations.
\newblock {\em Journal of the Royal Statistical Society Series C: Applied Statistics}, 24(2):193--202, 1975.

\bibitem{rafailov2024scaling}
Rafael Rafailov, Yaswanth Chittepu, Ryan Park, Harshit~Sushil Sikchi, Joey Hejna, Brad Knox, Chelsea Finn, and Scott Niekum.
\newblock Scaling laws for reward model overoptimization in direct alignment algorithms.
\newblock {\em Advances in Neural Information Processing Systems}, 37:126207--126242, 2024.

\bibitem{DBLP:journals/corr/abs-2404-12358}
Rafael Rafailov, Joey Hejna, Ryan Park, and Chelsea Finn.
\newblock From \emph{r} to q\({}^{\mbox{*}}\): Your language model is secretly a q-function.
\newblock In {\em arXiv}, 2024.

\bibitem{rafailov2024dpo}
Rafael Rafailov, Archit Sharma, Eric Mitchell, Christopher~D Manning, Stefano Ermon, and Chelsea Finn.
\newblock Direct preference optimization: Your language model is secretly a reward model.
\newblock In {\em Advances in Neural Information Processing Systems}, 2023.

\bibitem{ranjan2024comprehensive}
Rajesh Ranjan, Shailja Gupta, and Surya~Narayan Singh.
\newblock A comprehensive survey of bias in llms: Current landscape and future directions.
\newblock {\em arXiv preprint arXiv:2409.16430}, 2024.

\bibitem{reimers2019sentencebert}
Nils Reimers and Iryna Gurevych.
\newblock Sentence-bert: Sentence embeddings using siamese bert-networks.
\newblock In {\em Proceedings of the 2019 Conference on Empirical Methods in Natural Language Processing}. Association for Computational Linguistics, 11 2019.

\bibitem{sakaguchi2019winogrande}
Keisuke Sakaguchi, Ronan~Le Bras, Chandra Bhagavatula, and Yejin Choi.
\newblock Winogrande: An adversarial winograd schema challenge at scale.
\newblock {\em arXiv preprint arXiv:1907.10641}, 2019.

\bibitem{DBLP:conf/iclr/SanhWRBSACSRDBX22}
Victor Sanh, Albert Webson, Colin Raffel, Stephen~H. Bach, Lintang Sutawika, Zaid Alyafeai, Antoine Chaffin, Arnaud Stiegler, Arun Raja, Manan Dey, M~Saiful Bari, Canwen Xu, Urmish Thakker, Shanya~Sharma Sharma, Eliza Szczechla, Taewoon Kim, Gunjan Chhablani, Nihal~V. Nayak, Debajyoti Datta, Jonathan Chang, Mike~Tian{-}Jian Jiang, Han Wang, Matteo Manica, Sheng Shen, Zheng~Xin Yong, Harshit Pandey, Rachel Bawden, Thomas Wang, Trishala Neeraj, Jos Rozen, Abheesht Sharma, Andrea Santilli, Thibault F{\'{e}}vry, Jason~Alan Fries, Ryan Teehan, Teven~Le Scao, Stella Biderman, Leo Gao, Thomas Wolf, and Alexander~M. Rush.
\newblock Multitask prompted training enables zero-shot task generalization.
\newblock In {\em The Tenth International Conference on Learning Representations, {ICLR} 2022, Virtual Event, April 25-29, 2022}. OpenReview.net, 2022.

\bibitem{schulman2015high}
John Schulman, Philipp Moritz, Sergey Levine, Michael Jordan, and Pieter Abbeel.
\newblock High-dimensional continuous control using generalized advantage estimation.
\newblock {\em arXiv preprint arXiv:1506.02438}, 2015.

\bibitem{schulman2017proximal}
John Schulman, Filip Wolski, Prafulla Dhariwal, Alec Radford, and Oleg Klimov.
\newblock Proximal policy optimization algorithms.
\newblock {\em arXiv preprint arXiv:1707.06347}, 2017.

\bibitem{shanahan2023role}
Murray Shanahan, Kyle McDonell, and Laria Reynolds.
\newblock Role-play with large language models.
\newblock {\em arXiv preprint arXiv:2305.16367}, 2023.

\bibitem{shao2024deepseekmath}
Zhihong Shao, Peiyi Wang, Qihao Zhu, Runxin Xu, Junxiao Song, Xiao Bi, Haowei Zhang, Mingchuan Zhang, YK~Li, Y~Wu, et~al.
\newblock Deepseekmath: Pushing the limits of mathematical reasoning in open language models.
\newblock {\em arXiv preprint arXiv:2402.03300}, 2024.

\bibitem{beijing_academy_of_artificial_intelligence}
Xiaofeng Shi, Lulu Zhao, Hua Zhou, and Donglin Hao.
\newblock Industrycorpus2, 2024.

\bibitem{DBLP:conf/iclr/SinitsinPPPB20}
Anton Sinitsin, Vsevolod Plokhotnyuk, Dmitry~V. Pyrkin, Sergei Popov, and Artem Babenko.
\newblock Editable neural networks.
\newblock In {\em 8th International Conference on Learning Representations, {ICLR} 2020, Addis Ababa, Ethiopia, April 26-30, 2020}. OpenReview.net, 2020.

\bibitem{smith2020controlling}
Eric~Michael Smith, Diana Gonzalez-Rico, Emily Dinan, and Y-Lan Boureau.
\newblock Controlling style in generated dialogue.
\newblock {\em arXiv preprint arXiv:2009.10855}, 2020.

\bibitem{srivastava2022beyond}
Aarohi Srivastava, Abhinav Rastogi, Abhishek Rao, Abu Awal~Md Shoeb, Abubakar Abid, Adam Fisch, Adam~R Brown, Adam Santoro, Aditya Gupta, Adri{\`a} Garriga-Alonso, et~al.
\newblock Beyond the imitation game: Quantifying and extrapolating the capabilities of language models.
\newblock {\em arXiv preprint arXiv:2206.04615}, 2022.

\bibitem{suzgun2022bbh}
Mirac Suzgun, Nathan Scales, Nathanael Sch{\"{a}}rli, Sebastian Gehrmann, Yi~Tay, Hyung~Won Chung, Aakanksha Chowdhery, Quoc~V. Le, Ed~H. Chi, Denny Zhou, and Jason Wei.
\newblock Challenging big-bench tasks and whether chain-of-thought can solve them.
\newblock {\em CoRR}, abs/2210.09261, 2022.

\bibitem{talmor2019commonsenseqa}
Alon Talmor, Jonathan Herzig, Nicholas Lourie, and Jonathan Berant.
\newblock {C}ommonsense{QA}: A question answering challenge targeting commonsense knowledge.
\newblock In Jill Burstein, Christy Doran, and Thamar Solorio, editors, {\em Proceedings of the 2019 Conference of the North {A}merican Chapter of the Association for Computational Linguistics: Human Language Technologies, Volume 1 (Long and Short Papers)}, pages 4149--4158, Minneapolis, Minnesota, June 2019. Association for Computational Linguistics.

\bibitem{tang2023struc}
Xiangru Tang, Yiming Zong, Yilun Zhao, Arman Cohan, and Mark Gerstein.
\newblock Struc-bench: Are large language models really good at generating complex structured data?
\newblock {\em arXiv preprint arXiv:2309.08963}, 2023.

\bibitem{alpaca}
Rohan Taori, Ishaan Gulrajani, Tianyi Zhang, Yann Dubois, Xuechen Li, Carlos Guestrin, Percy Liang, and Tatsunori~B. Hashimoto.
\newblock Stanford alpaca: An instruction-following llama model.
\newblock \url{https://github.com/tatsu-lab/stanford_alpaca}, 2023.

\bibitem{qwen2.5}
Qwen Team.
\newblock Qwen2.5: A party of foundation models, September 2024.

\bibitem{OpenHermes}
Teknium.
\newblock Openhermes 2.5: An open dataset of synthetic data for generalist llm assistants, 2023.

\bibitem{tiedemann2012parallel}
J{\"o}rg Tiedemann.
\newblock Parallel data, tools and interfaces in opus.
\newblock In {\em Lrec}, volume 2012, pages 2214--2218. Citeseer, 2012.

\bibitem{tjong-kim-sang-de-meulder-2003-introduction}
Erik~F. Tjong Kim~Sang and Fien De~Meulder.
\newblock Introduction to the {C}o{NLL}-2003 shared task: Language-independent named entity recognition.
\newblock In {\em Proceedings of the Seventh Conference on Natural Language Learning at {HLT}-{NAACL} 2003}, pages 142--147, 2003.

\bibitem{torabi2018behavioral}
Faraz Torabi, Garrett Warnell, and Peter Stone.
\newblock Behavioral cloning from observation.
\newblock {\em arXiv preprint arXiv:1805.01954}, 2018.

\bibitem{2023llama}
Hugo Touvron, Thibaut Lavril, Gautier Izacard, Xavier Martinet, Marie{-}Anne Lachaux, Timoth{\'{e}}e Lacroix, Baptiste Rozi{\`{e}}re, Naman Goyal, Eric Hambro, Faisal Azhar, Aur{\'{e}}lien Rodriguez, Armand Joulin, Edouard Grave, and Guillaume Lample.
\newblock Llama: Open and efficient foundation language models.
\newblock {\em CoRR}, abs/2302.13971, 2023.

\bibitem{touvron2023llama}
Hugo Touvron, Louis Martin, Kevin Stone, Peter Albert, Amjad Almahairi, Yasmine Babaei, Nikolay Bashlykov, Soumya Batra, Prajjwal Bhargava, Shruti Bhosale, Dan Bikel, Lukas Blecher, Cristian~Canton Ferrer, Moya Chen, Guillem Cucurull, David Esiobu, Jude Fernandes, Jeremy Fu, Wenyin Fu, Brian Fuller, Cynthia Gao, Vedanuj Goswami, Naman Goyal, Anthony Hartshorn, Saghar Hosseini, Rui Hou, Hakan Inan, Marcin Kardas, Viktor Kerkez, Madian Khabsa, Isabel Kloumann, Artem Korenev, Punit~Singh Koura, Marie-Anne Lachaux, Thibaut Lavril, Jenya Lee, Diana Liskovich, Yinghai Lu, Yuning Mao, Xavier Martinet, Todor Mihaylov, Pushkar Mishra, Igor Molybog, Yixin Nie, Andrew Poulton, Jeremy Reizenstein, Rashi Rungta, Kalyan Saladi, Alan Schelten, Ruan Silva, Eric~Michael Smith, Ranjan Subramanian, Xiaoqing~Ellen Tan, Binh Tang, Ross Taylor, Adina Williams, Jian~Xiang Kuan, Puxin Xu, Zheng Yan, Iliyan Zarov, Yuchen Zhang, Angela Fan, Melanie Kambadur, Sharan Narang, Aurelien Rodriguez, Robert Stojnic, Sergey Edunov, and Thomas
  Scialom.
\newblock Llama 2: Open foundation and fine-tuned chat models, 2023.

\bibitem{tsai-etal-2021-style}
Alicia Tsai, Shereen Oraby, Vittorio Perera, Jiun-Yu Kao, Yuheng Du, Anjali Narayan-Chen, Tagyoung Chung, and Dilek Hakkani-Tur.
\newblock Style control for schema-guided natural language generation.
\newblock In {\em Proceedings of the 3rd Workshop on Natural Language Processing for Conversational AI}, pages 228--242, Online, November 2021. Association for Computational Linguistics.

\bibitem{tunstall2023zephyr}
Lewis Tunstall, Edward Beeching, Nathan Lambert, Nazneen Rajani, Kashif Rasul, Younes Belkada, Shengyi Huang, Leandro von Werra, Cl{\'e}mentine Fourrier, Nathan Habib, et~al.
\newblock Zephyr: Direct distillation of lm alignment.
\newblock In {\em arXiv}, 2023.

\bibitem{vrandevcic2014wikidata}
Denny Vrande{\v{c}}i{\'c} and Markus Kr{\"o}tzsch.
\newblock Wikidata: a free collaborative knowledgebase.
\newblock {\em Communications of the ACM}, 57(10):78--85, 2014.

\bibitem{DBLP:journals/corr/abs-2309-08952}
Jiaan Wang, Yunlong Liang, Zengkui Sun, Yuxuan Cao, and Jiarong Xu.
\newblock Cross-lingual knowledge editing in large language models.
\newblock {\em CoRR}, abs/2309.08952, 2023.

\bibitem{wang2024survey}
Jiahao Wang, Bolin Zhang, Qianlong Du, Jiajun Zhang, and Dianhui Chu.
\newblock A survey on data selection for llm instruction tuning.
\newblock {\em arXiv preprint arXiv:2402.05123}, 2024.

\bibitem{DBLP:journals/corr/abs-2305-17926}
Peiyi Wang, Lei Li, Liang Chen, Dawei Zhu, Binghuai Lin, Yunbo Cao, Qi~Liu, Tianyu Liu, and Zhifang Sui.
\newblock Large language models are not fair evaluators.
\newblock {\em CoRR}, abs/2305.17926, 2023.

\bibitem{wang2023easyedit}
Peng Wang, Ningyu Zhang, Xin Xie, Yunzhi Yao, Bozhong Tian, Mengru Wang, Zekun Xi, Siyuan Cheng, Kangwei Liu, Guozhou Zheng, and Huajun Chen.
\newblock Easyedit: An easy-to-use knowledge editing framework for large language models.
\newblock {\em CoRR}, abs/2308.07269, 2023.

\bibitem{wang2024essence}
Xinpeng Wang, Shitong Duan, Xiaoyuan Yi, Jing Yao, Shanlin Zhou, Zhihua Wei, Peng Zhang, Dongkuan Xu, Maosong Sun, and Xing Xie.
\newblock On the essence and prospect: An investigation of alignment approaches for big models.
\newblock {\em arXiv preprint arXiv:2403.04204}, 2024.

\bibitem{wang-etal-2023-self-instruct}
Yizhong Wang, Yeganeh Kordi, Swaroop Mishra, Alisa Liu, Noah~A. Smith, Daniel Khashabi, and Hannaneh Hajishirzi.
\newblock Self-instruct: Aligning language models with self-generated instructions.
\newblock In {\em Proceedings of the 61st Annual Meeting of the Association for Computational Linguistics (Volume 1: Long Papers)}, pages 13484--13508, Toronto, Canada, July 2023. Association for Computational Linguistics.

\bibitem{wang-etal-2022-super}
Yizhong Wang, Swaroop Mishra, Pegah Alipoormolabashi, Yeganeh Kordi, Amirreza Mirzaei, Atharva Naik, Arjun Ashok, Arut~Selvan Dhanasekaran, Anjana Arunkumar, David Stap, Eshaan Pathak, Giannis Karamanolakis, Haizhi Lai, Ishan Purohit, Ishani Mondal, Jacob Anderson, Kirby Kuznia, Krima Doshi, Kuntal~Kumar Pal, Maitreya Patel, Mehrad Moradshahi, Mihir Parmar, Mirali Purohit, Neeraj Varshney, Phani~Rohitha Kaza, Pulkit Verma, Ravsehaj~Singh Puri, Rushang Karia, Savan Doshi, Shailaja~Keyur Sampat, Siddhartha Mishra, Sujan Reddy~A, Sumanta Patro, Tanay Dixit, and Xudong Shen.
\newblock Super-{N}atural{I}nstructions: Generalization via declarative instructions on 1600+ {NLP} tasks.
\newblock In Yoav Goldberg, Zornitsa Kozareva, and Yue Zhang, editors, {\em Proceedings of the 2022 Conference on Empirical Methods in Natural Language Processing}, pages 5085--5109, Abu Dhabi, United Arab Emirates, December 2022. Association for Computational Linguistics.

\bibitem{wang2023interactive}
Zekun Wang, Ge~Zhang, Kexin Yang, Ning Shi, Wangchunshu Zhou, Shaochun Hao, Guangzheng Xiong, Yizhi Li, Mong~Yuan Sim, Xiuying Chen, et~al.
\newblock Interactive natural language processing.
\newblock {\em arXiv preprint arXiv:2305.13246}, 2023.

\bibitem{wang2024comprehensive}
Zhichao Wang, Bin Bi, Shiva~Kumar Pentyala, Kiran Ramnath, Sougata Chaudhuri, Shubham Mehrotra, Xiang-Bo Mao, Sitaram Asur, et~al.
\newblock A comprehensive survey of llm alignment techniques: Rlhf, rlaif, ppo, dpo and more.
\newblock In {\em arXiv}, 2024.

\bibitem{wei2022finetuned}
Jason Wei, Maarten Bosma, Vincent Zhao, Kelvin Guu, Adams~Wei Yu, Brian Lester, Nan Du, Andrew~M. Dai, and Quoc~V Le.
\newblock Finetuned language models are zero-shot learners.
\newblock In {\em International Conference on Learning Representations}, 2022.

\bibitem{wei2022emergent}
Jason Wei, Yi~Tay, Rishi Bommasani, Colin Raffel, Barret Zoph, Sebastian Borgeaud, Dani Yogatama, Maarten Bosma, Denny Zhou, Donald Metzler, et~al.
\newblock Emergent abilities of large language models.
\newblock {\em arXiv preprint arXiv:2206.07682}, 2022.

\bibitem{wei2022chain}
Jason Wei, Xuezhi Wang, Dale Schuurmans, Maarten Bosma, Fei Xia, Ed~Chi, Quoc~V Le, Denny Zhou, et~al.
\newblock Chain-of-thought prompting elicits reasoning in large language models.
\newblock {\em Advances in neural information processing systems}, 35:24824--24837, 2022.

\bibitem{weidinger2021ethical}
Laura Weidinger, John Mellor, Maribeth Rauh, Conor Griffin, Jonathan Uesato, Po-Sen Huang, Myra Cheng, Mia Glaese, Borja Balle, Atoosa Kasirzadeh, et~al.
\newblock Ethical and social risks of harm from language models.
\newblock {\em arXiv preprint arXiv:2112.04359}, 2021.

\bibitem{wiener1960some}
Norbert Wiener.
\newblock Some moral and technical consequences of automation: As machines learn they may develop unforeseen strategies at rates that baffle their programmers.
\newblock {\em Science}, 131(3410):1355--1358, 1960.

\bibitem{wirth2017survey}
Christian Wirth, Riad Akrour, Gerhard Neumann, and Johannes F{\"u}rnkranz.
\newblock A survey of preference-based reinforcement learning methods.
\newblock {\em Journal of Machine Learning Research}, 18(136):1--46, 2017.

\bibitem{wu2023style}
Minghao Wu and Alham~Fikri Aji.
\newblock Style over substance: Evaluation biases for large language models.
\newblock {\em arXiv preprint arXiv:2307.03025}, 2023.

\bibitem{xia2024less}
Mengzhou Xia, Sadhika Malladi, Suchin Gururangan, Sanjeev Arora, and Danqi Chen.
\newblock Less: Selecting influential data for targeted instruction tuning.
\newblock {\em arXiv preprint arXiv:2402.04333}, 2024.

\bibitem{DBLP:conf/eacl/XiaoW21}
Yijun Xiao and William~Yang Wang.
\newblock On hallucination and predictive uncertainty in conditional language generation.
\newblock In {\em Conference of the European Chapter of the Association for Computational Linguistics}, pages 2734--2744, 2021.

\bibitem{xie2024finben}
Qianqian Xie, Weiguang Han, Zhengyu Chen, Ruoyu Xiang, Xiao Zhang, Yueru He, Mengxi Xiao, Dong Li, Yongfu Dai, Duanyu Feng, Yijing Xu, Haoqiang Kang, Ziyan Kuang, Chenhan Yuan, Kailai Yang, Zheheng Luo, Tianlin Zhang, Zhiwei Liu, GUOJUN XIONG, Zhiyang Deng, Yuechen Jiang, Zhiyuan Yao, Haohang Li, Yangyang Yu, Gang Hu, Huang Jiajia, Xiao-Yang Liu, Alejandro Lopez-Lira, Benyou Wang, Yanzhao Lai, Hao Wang, Min Peng, Sophia Ananiadou, and Jimin Huang.
\newblock Finben: An holistic financial benchmark for large language models.
\newblock In {\em The Thirty-eight Conference on Neural Information Processing Systems Datasets and Benchmarks Track}, 2024.

\bibitem{xu2024wizardlm}
Can Xu, Qingfeng Sun, Kai Zheng, Xiubo Geng, Pu~Zhao, Jiazhan Feng, Chongyang Tao, Qingwei Lin, and Daxin Jiang.
\newblock Wizard{LM}: Empowering large pre-trained language models to follow complex instructions.
\newblock In {\em The Twelfth International Conference on Learning Representations}, 2024.

\bibitem{xu2022rag}
Jing Xu, Arthur Szlam, and Jason Weston.
\newblock Beyond goldfish memory: Long-term open-domain conversation.
\newblock In Smaranda Muresan, Preslav Nakov, and Aline Villavicencio, editors, {\em Proceedings of the 60th Annual Meeting of the Association for Computational Linguistics (Volume 1: Long Papers), {ACL} 2022, Dublin, Ireland, May 22-27, 2022}, pages 5180--5197. Association for Computational Linguistics, 2022.

\bibitem{xu2024magpie}
Zhangchen Xu, Fengqing Jiang, Luyao Niu, Yuntian Deng, Radha Poovendran, Yejin Choi, and Bill~Yuchen Lin.
\newblock Magpie: Alignment data synthesis from scratch by prompting aligned llms with nothing.
\newblock {\em arXiv preprint arXiv:2406.08464}, 2024.

\bibitem{yang2023foundation}
Sherry Yang, Ofir Nachum, Yilun Du, Jason Wei, Pieter Abbeel, and Dale Schuurmans.
\newblock Foundation models for decision making: Problems, methods, and opportunities.
\newblock {\em arXiv preprint arXiv:2303.04129}, 2023.

\bibitem{yao2022react}
Shunyu Yao, Jeffrey Zhao, Dian Yu, Nan Du, Izhak Shafran, Karthik Narasimhan, and Yuan Cao.
\newblock React: Synergizing reasoning and acting in language models.
\newblock {\em arXiv preprint arXiv:2210.03629}, 2022.

\bibitem{yu2023metamath}
Longhui Yu, Weisen Jiang, Han Shi, Jincheng YU, Zhengying Liu, Yu~Zhang, James Kwok, Zhenguo Li, Adrian Weller, and Weiyang Liu.
\newblock Metamath: Bootstrap your own mathematical questions for large language models.
\newblock In {\em International Conference on Learning Representations}, 2024.

\bibitem{yu2024mates}
Zichun Yu, Spandan Das, and Chenyan Xiong.
\newblock Mates: Model-aware data selection for efficient pretraining with data influence models.
\newblock {\em Advances in Neural Information Processing Systems}, 37:108735--108759, 2024.

\bibitem{DBLP:conf/nips/YuanNL21}
Weizhe Yuan, Graham Neubig, and Pengfei Liu.
\newblock Bartscore: Evaluating generated text as text generation.
\newblock In {\em Advances in Neural Information Processing Systems}, pages 27263--27277, 2021.

\bibitem{yudkowsky2016ai}
Eliezer Yudkowsky.
\newblock The ai alignment problem: why it is hard, and where to start.
\newblock {\em Symbolic Systems Distinguished Speaker}, 4(1), 2016.

\bibitem{yue2024mammoth2}
Xiang Yue, Tuney Zheng, Ge~Zhang, and Wenhu Chen.
\newblock Mammoth2: Scaling instructions from the web.
\newblock {\em Advances in Neural Information Processing Systems}, 2024.

\bibitem{yujian2007normalized}
Li~Yujian and Liu Bo.
\newblock A normalized levenshtein distance metric.
\newblock {\em IEEE transactions on pattern analysis and machine intelligence}, 29(6):1091--1095, 2007.

\bibitem{zhang2022survey}
Hanqing Zhang, Haolin Song, Shaoyu Li, Ming Zhou, and Dawei Song.
\newblock A survey of controllable text generation using transformer-based pre-trained language models.
\newblock {\em ACM Computing Surveys}, 2022.

\bibitem{zhang2024knowedit}
Ningyu Zhang, Yunzhi Yao, Bozhong Tian, Peng Wang, Shumin Deng, Mengru Wang, Zekun Xi, Shengyu Mao, Jintian Zhang, Yuansheng Ni, Siyuan Cheng, Ziwen Xu, Xin Xu, Jia{-}Chen Gu, Yong Jiang, Pengjun Xie, Fei Huang, Lei Liang, Zhiqiang Zhang, Xiaowei Zhu, Jun Zhou, and Huajun Chen.
\newblock A comprehensive study of knowledge editing for large language models.
\newblock {\em CoRR}, abs/2401.01286, 2024.

\bibitem{zhang2023instruction}
Shengyu Zhang, Linfeng Dong, Xiaoya Li, Sen Zhang, Xiaofei Sun, Shuhe Wang, Jiwei Li, Runyi Hu, Tianwei Zhang, Fei Wu, et~al.
\newblock Instruction tuning for large language models: A survey.
\newblock {\em arXiv preprint arXiv:2308.10792}, 2023.

\bibitem{zhang2018generating}
Yizhe Zhang, Michel Galley, Jianfeng Gao, Zhe Gan, Xiujun Li, Chris Brockett, and Bill Dolan.
\newblock Generating informative and diverse conversational responses via adversarial information maximization.
\newblock {\em Advances in Neural Information Processing Systems}, 31, 2018.

\bibitem{zhao2024wildchat}
Wenting Zhao, Xiang Ren, Jack Hessel, Claire Cardie, Yejin Choi, and Yuntian Deng.
\newblock Wildchat: 1m chat{GPT} interaction logs in the wild.
\newblock In {\em The Twelfth International Conference on Learning Representations}, 2024.

\bibitem{zhao2023large}
Yilun Zhao, Haowei Zhang, Shengyun Si, Linyong Nan, Xiangru Tang, and Arman Cohan.
\newblock Large language models are effective table-to-text generators, evaluators, and feedback providers.
\newblock {\em arXiv preprint arXiv:2305.14987}, 2023.

\bibitem{zheng2023judging}
Lianmin Zheng, Wei-Lin Chiang, Ying Sheng, Siyuan Zhuang, Zhanghao Wu, Yonghao Zhuang, Zi~Lin, Zhuohan Li, Dacheng Li, Eric Xing, Hao Zhang, Joseph~E. Gonzalez, and Ion Stoica.
\newblock Judging {LLM}-as-a-judge with {MT}-bench and chatbot arena.
\newblock In {\em Thirty-seventh Conference on Neural Information Processing Systems Datasets and Benchmarks Track}, 2023.

\bibitem{zhong2023agieval}
Wanjun Zhong, Ruixiang Cui, Yiduo Guo, Yaobo Liang, Shuai Lu, Yanlin Wang, Amin Saied, Weizhu Chen, and Nan Duan.
\newblock Agieval: {A} human-centric benchmark for evaluating foundation models.
\newblock {\em CoRR}, abs/2304.06364, 2023.

\bibitem{zhong2023mquake}
Zexuan Zhong, Zhengxuan Wu, Christopher~D. Manning, Christopher Potts, and Danqi Chen.
\newblock Mquake: Assessing knowledge editing in language models via multi-hop questions.
\newblock In Houda Bouamor, Juan Pino, and Kalika Bali, editors, {\em Proceedings of the 2023 Conference on Empirical Methods in Natural Language Processing, {EMNLP} 2023, Singapore, December 6-10, 2023}, pages 15686--15702. Association for Computational Linguistics, 2023.

\bibitem{zhou2023ifeval}
Jeffrey Zhou, Tianjian Lu, Swaroop Mishra, Siddhartha Brahma, Sujoy Basu, Yi~Luan, Denny Zhou, and Le~Hou.
\newblock Instruction-following evaluation for large language models, 2023.

\end{thebibliography}

\appendix

\chapter{Appendix for Chapter 3}

\section{Data Statistics}
\label{appendix:data stat}
Table \ref{c3-tab: agieval_sat} and Table \ref{c3-tab: bbheval_sat} show the data statistics of AGIEval and BIG-Bench Hard, respectively.

\begin{table}[!h]
\footnotesize
\centering
\begin{tabular}{l|c|c}
\toprule
\textbf{Task}          & \textbf{\# Examples} & \textbf{\# Choices} \\ \midrule
AQuA-RAT               & 254                  & 5                   \\
LogiQA                 & 651                  & 4                   \\
LSAT-AR                & 230                  & 5                   \\
LSAT-LR                & 510                  & 5                   \\
LSAT-RC                & 269                  & 5                   \\
SAT-Math               & 220                  & 4                   \\
SAT-English            & 206                  & 4                   \\
SAT-English (w/o Psg.) & 206                  & 4           
\\ \bottomrule
\end{tabular}
\caption{Statistics of AGIEval dataset.}
\label{c3-tab: agieval_sat}
\end{table}

\begin{table}[!h]
\scriptsize
\centering
\begin{tabular}{l|c|c}
\toprule
\textbf{Task}                         & \textbf{\# Examples} & \textbf{\# Choices} \\ \midrule
Boolean Expressions                   & 250                  & 2                   \\
Causal Judgement                      & 187                  & 2                   \\
Date Understanding                    & 250                  & 6                   \\
Disambiguation QA                     & 250                  & 4                   \\
Formal Fallacies                      & 250                  & 2                   \\
Geometric Shapes                      & 250                  & 11                  \\
Hyperbaton                            & 250                  & 2                   \\
Logical Deduction (5 objects)         & 250                  & 5                   \\
Logical Deduction (7 objects)         & 250                  & 7                   \\
Logical Deduction (3 objects)         & 250                  & 3                   \\
Movie Recommendation                  & 250                  & 5                   \\
Navigate                              & 250                  & 2                   \\
Penguins in a Table                   & 146                  & 5                   \\
Reasoning about Colored Objects       & 250                  & 18                  \\
Ruin Names                            & 250                  & 11                   \\
Salient Translation Error Detection   & 250                  & 6                   \\
Snarks                                & 178                  & 2                   \\
Sports Understanding                  & 250                  & 2                   \\
Temporal Sequences                    & 250                  & 4                   \\
Tracking Shuffled Objects (5 objects) & 250                  & 5                   \\
Tracking Shuffled Objects (7 objects) & 250                  & 7                   \\
Tracking Shuffled Objects (3 objects) & 250                  & 3                   \\
Web of Lies                           & 250                  & 2 \\ \bottomrule                 
\end{tabular}
\caption{Statistics of BIG-Bench Hard dataset.}
\label{c3-tab: bbheval_sat}
\end{table}

\section{Baselines}
\label{appendix:baseline}
\begin{itemize}
\item \textbf{LLaMA} \cite{touvron2023llama} is a collection of foundation language models ranging from 7B to 65B parameters. It is trained on trillions of tokens from publicly available datasets and is demonstrated to outperform larger-size LLMs such as GPT-3 (175B) across a multitude of benchmarks. We use the official code from LLaMA \footnote{\url{https://github.com/facebookresearch/llama}}.
\item \textbf{Alpaca} \cite{alpaca} is a project initiated by Stanford University with the objective of developing and disseminating an open-source model that adeptly follows instructions. It is based on LLaMA and fine-tuned on 52K instruction-following examples generated by querying OpenAI’s text-davinci-003 model. On the self-instruct evaluation set, Alpaca mirrors text-davinci-003, but is notably more compact and cost-effective to reproduce. We use the official code from Alpaca \footnote{\url{https://github.com/tatsu-lab/stanford_alpaca}}.
\item \textbf{WizardLM} \cite{xu2024wizardlm} employs LLMs instead of humans to automatically mass-produce open-domain instructions of various difficulty levels, to improve the performance of LLMs. 
It uses an Evol-Instruct method to bootstrap the 52k instruction-following examples of Alapca into a larger set of 250k more intricate instructions. Out of this larger set, 70k examples were selected to fine-tune LLaMA. We use WizardLM-7B-V1.0 from the official code \footnote{\url{https://github.com/nlpxucan/WizardLM}}.
\item \textbf{Vicuna} \cite{zheng2023judging}, a superior open-source chatbot, excels in generating fluid and captivating responses to user queries. It is based on LLaMA and fine-tuned on 70K user-shared conversations collected from ShareGPT, a platform designed for sharing interactions with ChatGPT. Its impressive capabilities make it one of the leading open instruction-following models today. Vicuna achieves competitive performance against proprietary models such as ChatGPT and Bard \cite{Google2023}. We use Vicuna-7B-V1.1 and Vicuna-13B-V1.1 from FastChat \footnote{\url{https://github.com/lm-sys/FastChat}}.
\item \textbf{ChatGPT} \cite{openai2022chatgpt}, a product of OpenAI, is an advanced AI chatbot renowned for its ability to interact with users in an authentically human and engaging manner. The chatbot is built on powerful LLMs such as GPT-3.5 and GPT-4, which are trained on a vast corpus of internet text data.
ChatGPT undergoes fine-tuning via both supervised and reinforcement learning techniques, with the human trainers providing necessary feedback and direction.
\end{itemize}

\section{Implementation Details}
\paragraph{Training Hyperparameters.}
The training process is conducted on 8 A100 GPUs.
During each iteration of adversarial knowledge distillation, the hyperparameters for training are shown in Table \ref{c3-tab: hyperparameters}.

\begin{table}[h]
\small
\centering
{\begin{tabular}{l|c|c}
\toprule
\textbf{Hyperparameter} & \textbf{Lion-7B} & \textbf{Lion-13B} \\ \midrule
Batch size              & 128              & 128               \\
Learning rate           & 2e-5             & 2e-5              \\
Epoches                 & 3                & 3                 \\
Max length              & 1024             & 1024              \\
Optimizer               & AdamW            & AdamW             \\
Scheduler               & cosine           & cosine            \\
Weight decay            & 0                & 0                 \\
Warmup ratio            & 0.03             & 0.03           \\ \bottomrule
\end{tabular}}
\caption{\label{c3-tab: hyperparameters}
Training hyperparameters.}
\end{table}

\label{c3-sec: gpt_parameter}
\paragraph{Querying the gpt-3.5-turbo API.}
We use different sets of hyperparameters when querying the gpt-3.5-turbo API for different roles (Teacher, Referee, Generator). These hyperparameters are found to work well and we listed them in Table \ref{c3-tab: api}.

\begin{table}[!h]
\small
\centering
\begin{tabular}{l|c|c|c|c}
\toprule
\textbf{Role}  & 
\textbf{temperature}  &
\textbf{top\_p}  &
\textbf{beam\_size (n)}  &
\textbf{max\_tokens}
\\
\midrule
Teacher   & 0.7 & 1.0  & 1 & 1024 \\
Referee   & 0.2 & 1.0  & 1 & 512 \\
Generator   & 1.0 & 1.0  & 1 & 512 \\
\bottomrule
\end{tabular}
\caption{\label{c3-tab: api}
Hyperparameters for querying OpenAI gpt-3.5-turbo API under different roles.}
\end{table}

\section{Prompt Templates for Our Adversarial Distillation Framework}
\label{c3-sec: prompt template}
Fine-tuning an LLM (i.e. ChatGPT) is costly and intricate, human-tailored prompt templates are utilized to solve various tasks.
The prompt template of the \textbf{Teacher} for generating responses is shown in Table \ref{c3-tab: response}.
The prompt template of the \textbf{Referee} for comparing the quality of two responses generated by two AI assistants is shown in Table \ref{c3-tab: discriminator_gpt-3.5-turbo}.
The prompt templates of the \textbf{Generator} for generating new hard instructions and new easy instructions are shown in Table \ref{c3-tab: generator_hard} and Table \ref{c3-tab: generator_no_hard}, respectively.

\begin{table*}[!h]
\small
\centering
\begin{tabular}{l|l}
\toprule
system content &
\parbox[c]{13cm}{
\texttt{You are a helpful assistant that generates a response to a given task instruction.}
}\\
\toprule
{user content} &
\parbox[c]{13cm}{
\texttt{\#\#\# Instruction:} \\ 
\texttt{\{instruction\}} \\ \\
\texttt{\#\#\# Response:} \\
} \\
\bottomrule
\end{tabular}
\caption{Prompt template of gpt-3.5-turbo for generating responses. Note that the original instruction in Alpaca is composed of an instruction
prompt and an instance input. For example, the instruction prompt is ``write an abstract about the following method'', and the instance input is ``knowledge distillation''.
For a better adaption to real-world scenarios, we concatenate the instruction prompt and the instruction prompt into one instruction using a line break.}
\label{c3-tab: response}
\end{table*}

\begin{table*}[!h]
\small
\centering
\begin{tabular}{l|l}
\toprule
system content &
\parbox[c]{13cm}{
\texttt{You are a helpful and precise assistant for checking the quality of the answer.}
}\\
\toprule
user content &
\parbox[c]{13cm}{
$\left[ \texttt{Instruction} \right]$\\ 
\texttt{\{instruction\}} \\ \\ 
$\left[ \texttt{The Start of Assistant 1's Answer} \right]$\\ 
\texttt{\{answer\_1\}}\\ 
$\left[ \texttt{The End of Assistant 1's Answer} \right]$\\ \\
$\left[ \texttt{The Start of Assistant 2's Answer} \right]$\\ 
\texttt{\{answer\_2\}} \\
$\left[ \texttt{The End of Assistant 2's Answer} \right]$\\ \\
$\left[ \texttt{System} \right]$\\ 
\texttt{We would like to request your feedback on the performance of two AI assistants in response to the user instruction and input displayed above.} \\ \\
\texttt{Please rate the helpfulness, relevance, accuracy, and level of detail of their responses. Each assistant receives an overall score on a scale of 1 to 10, where a higher score indicates better overall performance.} \\ \\
\texttt{Please first provide a comprehensive explanation of your evaluation, avoiding any potential bias and ensuring that the order in which the responses were presented does not affect your judgment. Then, output two lines indicating the scores for Assistant 1 and 2, respectively.} \\ \\
\texttt{Output with the following format:} \\
\texttt{Evaluation evidence: <your evaluation explanation here>} \\
\texttt{Score of the Assistant 1: <score>} \\
\texttt{Score of the Assistant 2: <score>}
}
\\
\bottomrule
\end{tabular}
\caption{Prompt template of gpt-3.5-turbo for comparing the quality of two responses generated by two AI assistants.}
\label{c3-tab: discriminator_gpt-3.5-turbo}
\end{table*}

\begin{table*}[!h]
\small
\centering
\begin{tabular}{l|l}
\toprule
system content &
\parbox[c]{13cm}{
\texttt{You are a helpful assistant.}
}\\
\toprule
user content &
\parbox[c]{13cm}{
\texttt{I want you to act as an Instruction Creator.} \\ 
\texttt{Your goal is to draw inspiration from the \#Given Instruction\# to create a brand new instruction.} \\ 
\texttt{This new instruction should belong to the same domain and the same task type as the \#Given Instruction\#.} \\ 
\texttt{The LENGTH and difficulty level of the \#Created Instruction\# should be similar to that of the \#Given Instruction\#.} \\ 
\texttt{The \#Created Instruction\# must be reasonable and must be understood and responded to by humans.} \\
\texttt{'\#Given Instruction\#', '\#Created Instruction\#', 'given instruction' and 'created instruction' are not allowed to appear in \#Created Instruction\#.} \\ \\
\texttt{\#Given Instruction\#:} \\
\texttt{\{instruction\}} \\ \\
\texttt{\#Created Instruction\#:} \\
} \\
\bottomrule
\end{tabular}
\caption{Prompt template of gpt-3.5-turbo for generating new hard instructions.}
\label{c3-tab: generator_hard}
\end{table*}

\begin{table*}[!h]
\small
\centering
\begin{tabular}{l|l}
\toprule
system content &
\parbox[c]{13cm}{
\texttt{You are a helpful assistant.}
}\\
\toprule
user content &
\parbox[c]{13cm}{
\texttt{I want you to act as an Instruction Creator.} \\ 
\texttt{Your goal is to draw inspiration from the \#Given Instruction\# to create a brand new instruction.} \\ 
\texttt{This new instruction should belong to the same domain as the \#Given Instruction\# but be even more rare.} \\ 
\texttt{The LENGTH and difficulty level of the \#Created Instruction\# should be similar to that of the \#Given Instruction\#.} \\ 
\texttt{The \#Created Instruction\# must be reasonable and must be understood and responded to by humans.} \\
\texttt{'\#Given Instruction\#', '\#Created Instruction\#', 'given instruction' and 'created instruction' are not allowed to appear in \#Created Instruction\#.} \\ \\
\texttt{\#Given Instruction\#:} \\
\texttt{\{instruction\}} \\ \\
\texttt{\#Created Instruction\#:} \\
} \\
\bottomrule
\end{tabular}
\caption{Prompt template of gpt-3.5-turbo for generating new easy instructions.}
\label{c3-tab: generator_no_hard}
\end{table*}

\chapter{Appendix for Chapter 4}
\section{Implementation Details}
\label{c4-sec: appendix_implementation}
Our implementation is based on the alignment-handbook repo\footnote{\url{https://github.com/huggingface/alignment-handbook}}.
The training procedure was executed on 4 NVIDIA A800 GPUs, each equipped with 80GB of memory.
The duration required to train a single instance of the model, specifically the Llama3-8B-base, was approximately 9 hours.
The specific hyperparameters used during training are detailed in Table \ref{c4-tab: hyperparameters}.
Notably, all models were trained using the same set of hyperparameters, except for the maximum sequence length, which was set to 2048 for the 14B LLMs to mitigate computational bottlenecks.

\begin{table}[!h]
\centering
{\begin{tabular}{l|c}
\toprule
\textbf{Hyperparameter} & \textbf{Value} \\ \midrule
Batch size              & 128                            \\
Learning rate           & 2e-5                         \\
Epoches                 & 4                              \\
Max length              & 4096 (2048 for 14B LLMs)                          \\
Optimizer               & AdamW                     \\
Scheduler               & cosine             \\
Weight decay            & 0                       \\
Warmup ratio            & 0.1                  \\ \bottomrule
\end{tabular}}
\caption{\label{c4-tab: hyperparameters}
Training hyperparameters for Llama3-8B-base and Qwen2.5-1.5/3/7/14B-base.}
\end{table}

\section{Evaluation Details}
\label{c4-sec: evaluation_details}
Table \ref{c4-tab: evaluation_details} lists the evaluation details for AlpacaEval 2~\cite{alpaca_eval}, Arena-Hard~\cite{li2024live}, MT-Bench~\cite{zheng2023judging}, and IFEval~\cite{zhou2023ifeval}.
AlpacaEval 2 comprises 805 questions from 5 datasets, and MT-Bench spans 8 categories with a total of 80 questions.
Arena-Hard is an enhanced version of MT-Bench, featuring 500 well-defined technical problem-solving queries.
IFEval consists of 541 samples, each containing 1 to 3 verifiable constraints.
Evaluation metrics are reported in accordance with each benchmark's protocol.

\begin{table*}[!h]
\centering
\scriptsize
\begin{tabular}{lccccc}
\toprule
\textbf{Benchmark} & \textbf{\# Exs.} & \textbf{Baseline Model} & \textbf{Judge Model} & \textbf{Scoring Type} & \textbf{Metric} \\
\midrule
AlpacaEval 2 & 805 & GPT-4 Turbo & GPT-4 Turbo & Pairwise comparison & Length-controlled win rate \\
Arena-Hard   & 500 & GPT-4-0314 & GPT-4 Turbo & Pairwise comparison & Win rate \\
MT-Bench     & 80  & -          & GPT-4/GPT-4 Turbo & Single-answer grading & Rating of 1-10 \\
IFEval & 541 & - & - & Rule-based verification & Accuracy \\
\bottomrule
\end{tabular}
\caption{Evaluation details for AlpacaEval 2~\cite{alpaca_eval}, Arena-Hard~\cite{li2024live}, MT-Bench~\cite{zheng2023judging}, and IFEval~\cite{zhou2023ifeval}. The baseline model refers to the model compared against.}
\label{c4-tab: evaluation_details}
\end{table*}

\section{Dataset Analysis}
\label{c4-sec: data analysis}
\begin{figure}[!h]
\centering
\includegraphics[width=0.7\linewidth]{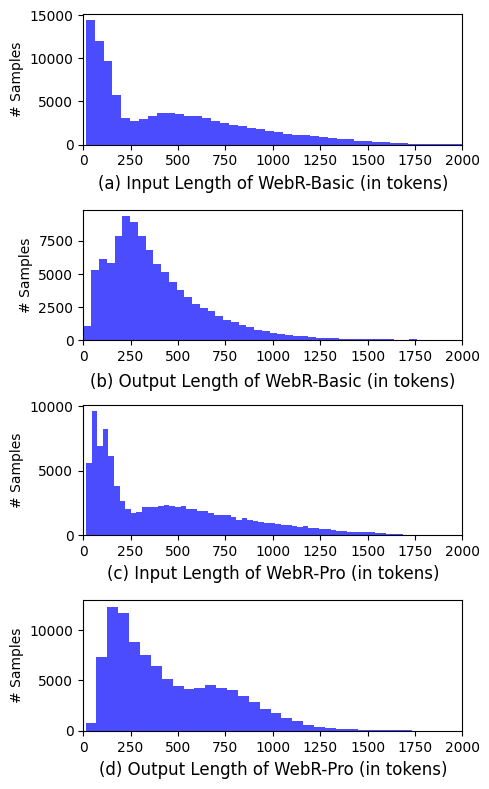}
\caption{
Lengths of instructions and responses in WebR-Basic and WebR-Pro. 
}
\label{c4-fig: token_len}
\end{figure}

Statistics including token lengths of instructions and responses are illustrated in Figure \ref{c4-fig: token_len}.
Tokens are counted using the \texttt{tiktoken} library\footnote{\url{https://github.com/openai/tiktoken}}.
For WebR-Basic, the average token lengths of instructions and responses are 441.41 and 381.28, respectively.
For WebR, the average token lengths of instructions and responses are 439.88 and 457.34, respectively.

\section{Prompt Template}
\begin{figure*}[!ht]
\begin{center}
\includegraphics[width=\linewidth]{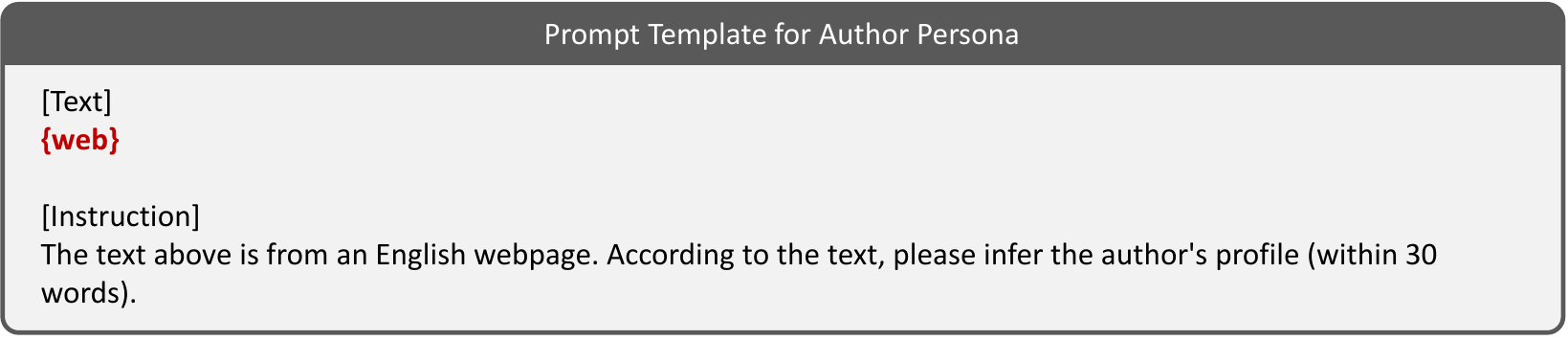}
\end{center}
\caption{Prompt template for generating author persona.}
\label{c4-fig: persona_prompt}
\end{figure*}

\begin{figure*}[!ht]
\begin{center}
\includegraphics[width=\linewidth]{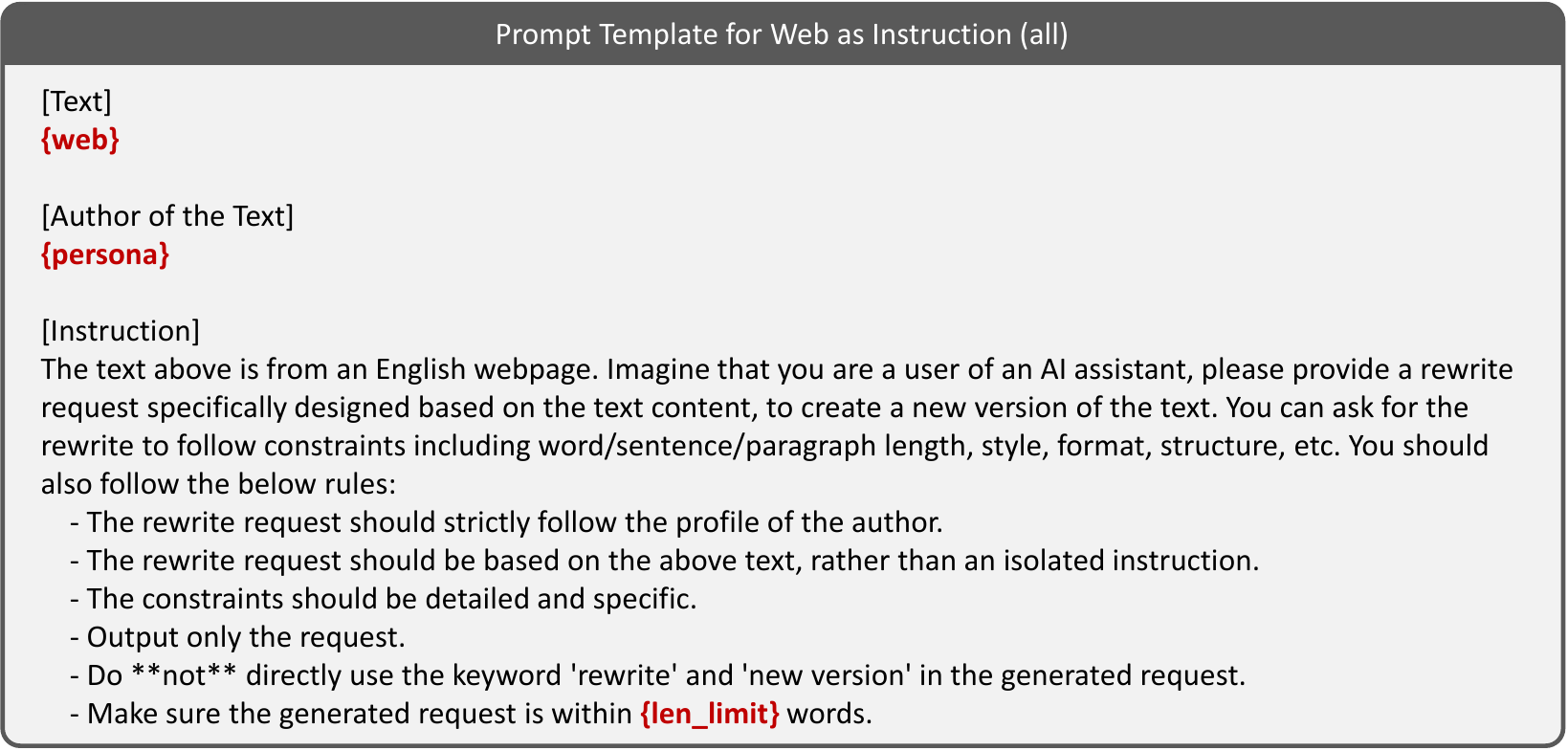}
\end{center}
\caption{Prompt template for \textit{Web as Instruction} (generating the rewrite request based on the whole web content).}
\label{c4-fig: l2l_all_prompt}
\end{figure*}

\begin{figure*}[!ht]
\begin{center}
\includegraphics[width=\linewidth]{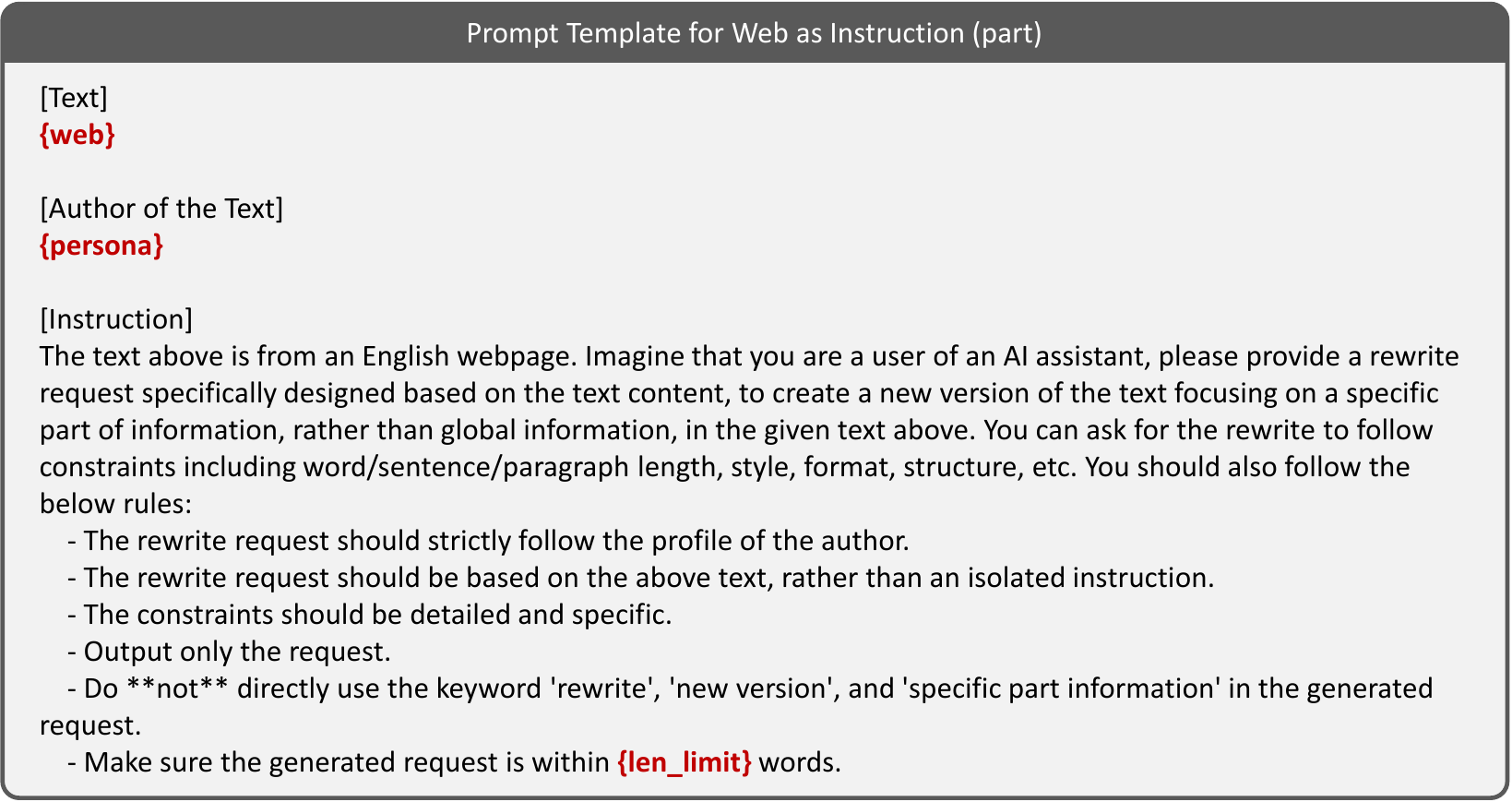}
\end{center}
\caption{Prompt template for \textit{Web as Instruction} (generating the rewrite request based on the specific part of the web content).}
\label{c4-fig: l2l_part_prompt}
\end{figure*}

\begin{figure*}[!ht]
\begin{center}
\includegraphics[width=\linewidth]{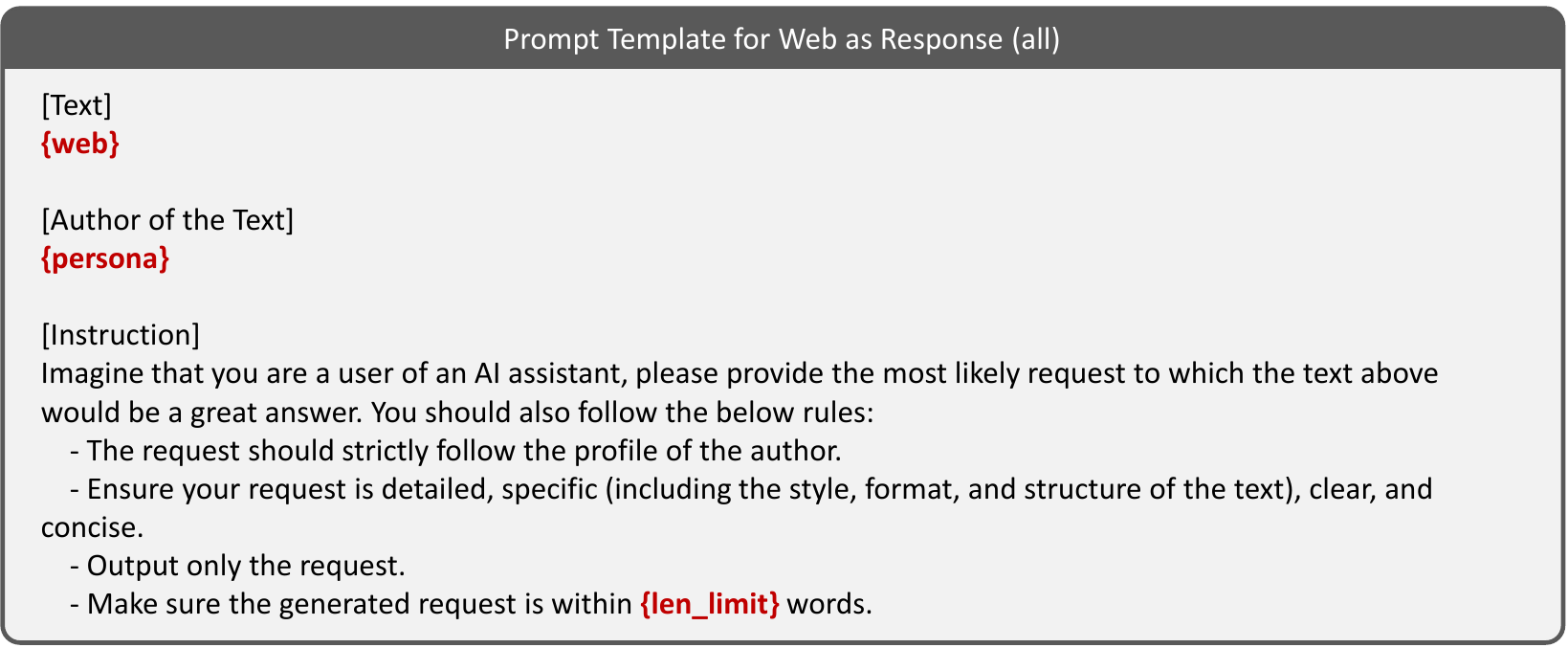}
\end{center}
\caption{Prompt template for \textit{Web as Response} (generating the latent instruction based on the whole web content).}
\label{c4-fig: s2l_all_prompt}
\end{figure*}

\begin{figure*}[!ht]
\begin{center}
\includegraphics[width=\linewidth]{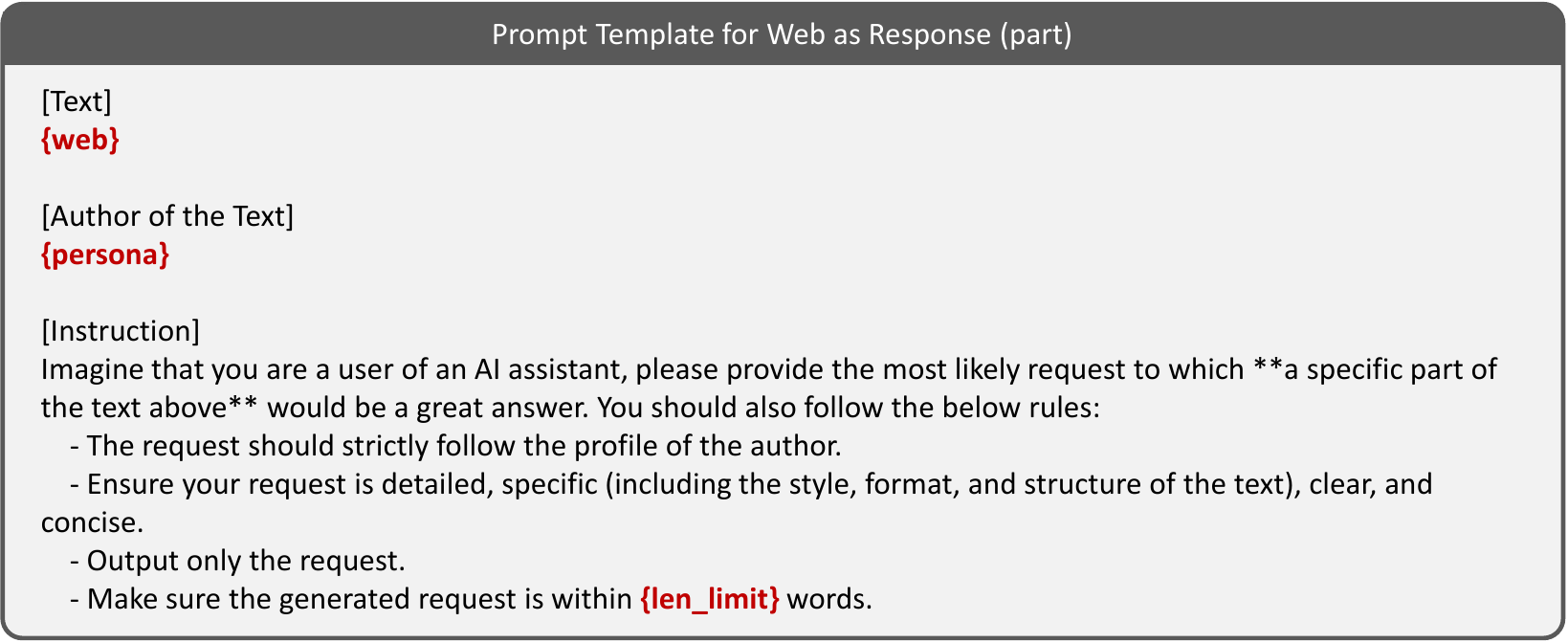}
\end{center}
\caption{Prompt template for \textit{Web as Response} (generating the latent instruction based on the specific part of the web content).}
\label{c4-fig: s2l_part_prompt}
\end{figure*}

\begin{figure*}[!ht]
\begin{center}
\includegraphics[width=\linewidth]{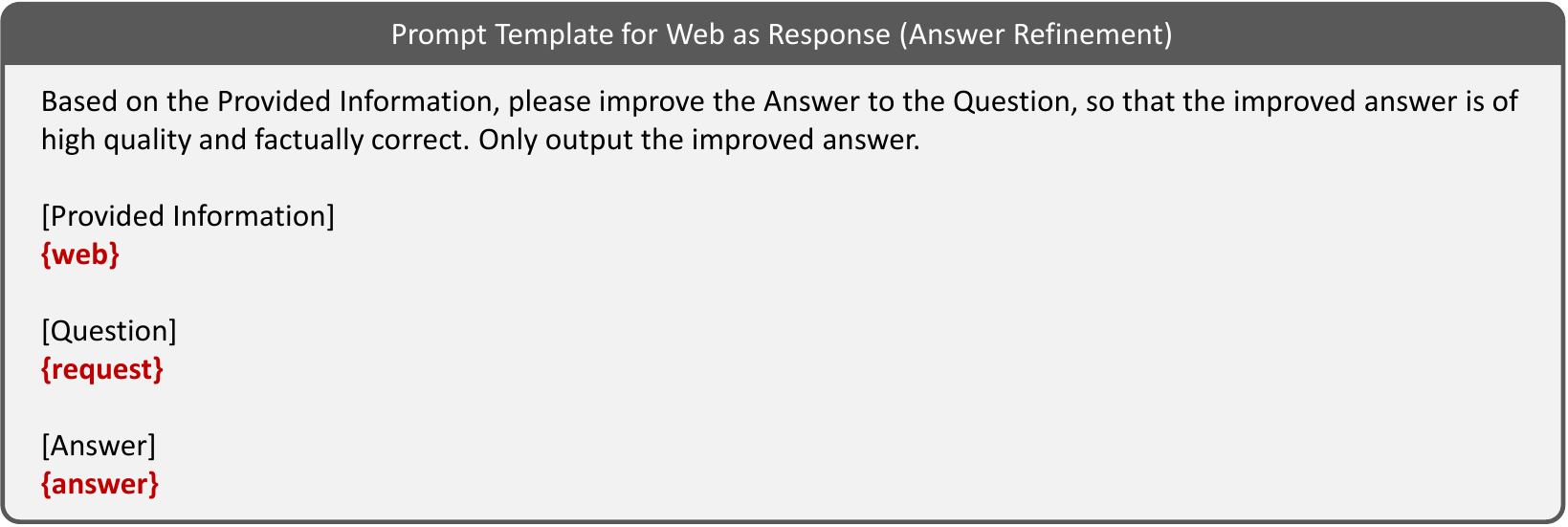}
\end{center}
\caption{Prompt template for \textit{Web as Response} (answer refinement).}
\label{c4-fig: refine_prompt}
\end{figure*}

Figure \ref{c4-fig: persona_prompt} shows the prompt template for generating the author persona according to the web content.
Figure \ref{c4-fig: l2l_all_prompt} shows the prompt template for generating the rewrite request based on the whole web content.
Figure \ref{c4-fig: l2l_part_prompt} shows the prompt template for generating the rewrite request based on a specific part of the web content.
Figure \ref{c4-fig: s2l_all_prompt} shows the prompt template for generating the latent instruction corresponding to the whole web content.
Figure \ref{c4-fig: s2l_part_prompt} shows the prompt template for generating the latent instruction corresponding to a specific part of the web content.
Figure \ref{c4-fig: refine_prompt} shows the prompt template for generating a refined response based on the raw web and the instruction.

\chapter{Appendix for Chapter 5}

\section{Details of Training Data Construction}
\label{c5-sec: appendix_data_construction}

\subsection{Synthetics of Out-of-scope Examples}
\label{c5-sec: appendix_out_scope}
\begin{figure*}[!h]
\centering
\includegraphics[width=\linewidth]{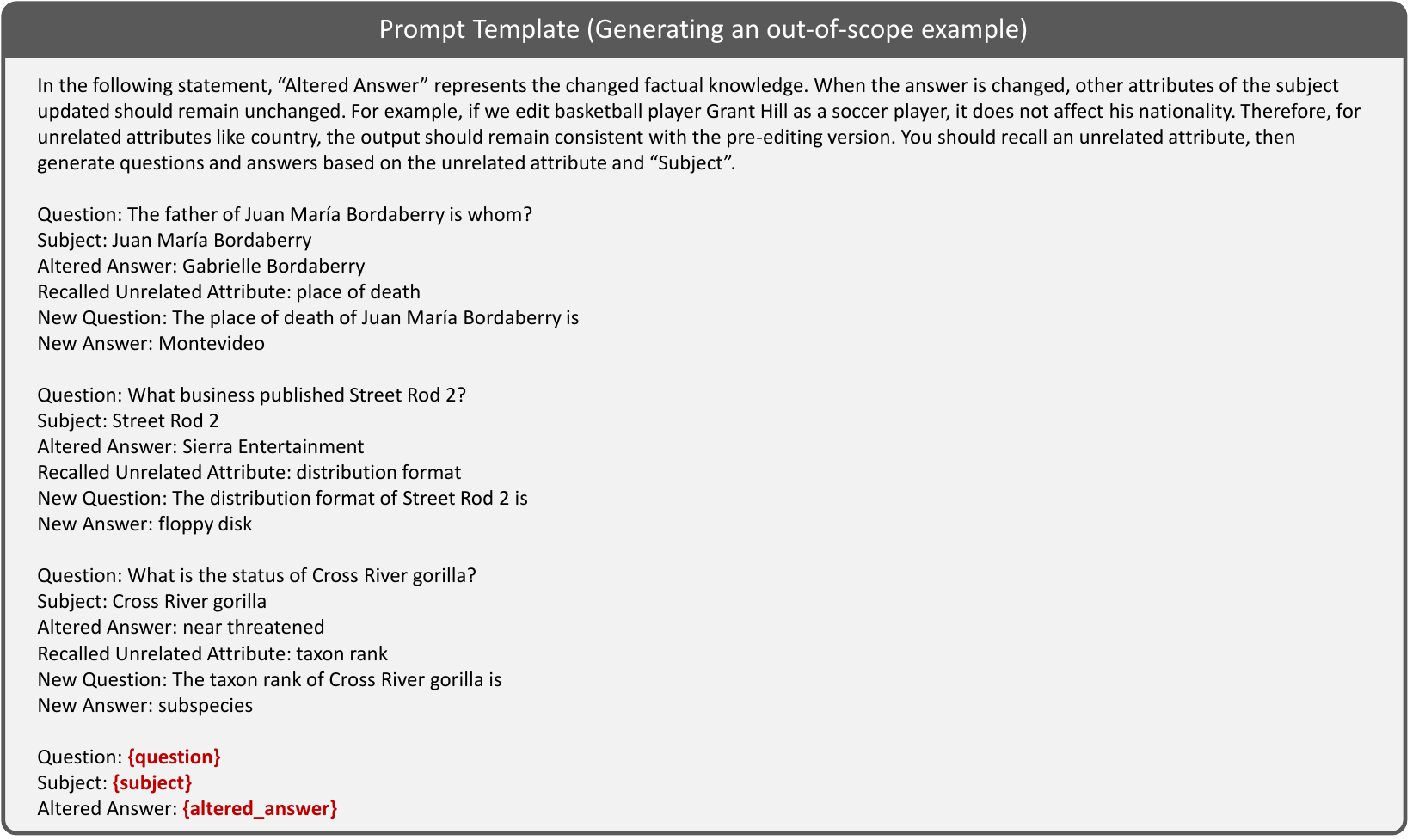}
\caption{
Prompt template for generating an out-of-scope example.
}
\label{c5-fig: example_prompt}
\end{figure*}

As shown in Figure \ref{c5-fig: example_prompt}, we employ a few-shot manual demonstration as a prompt to guide GPT-4 in producing the desired query and answer.

\subsection{Synthetics of Free-text In-scope Question-answering Pairs}
\label{c5-sec: appendix_free_text}
\begin{figure*}[!h]
\centering
\includegraphics[width=\linewidth]{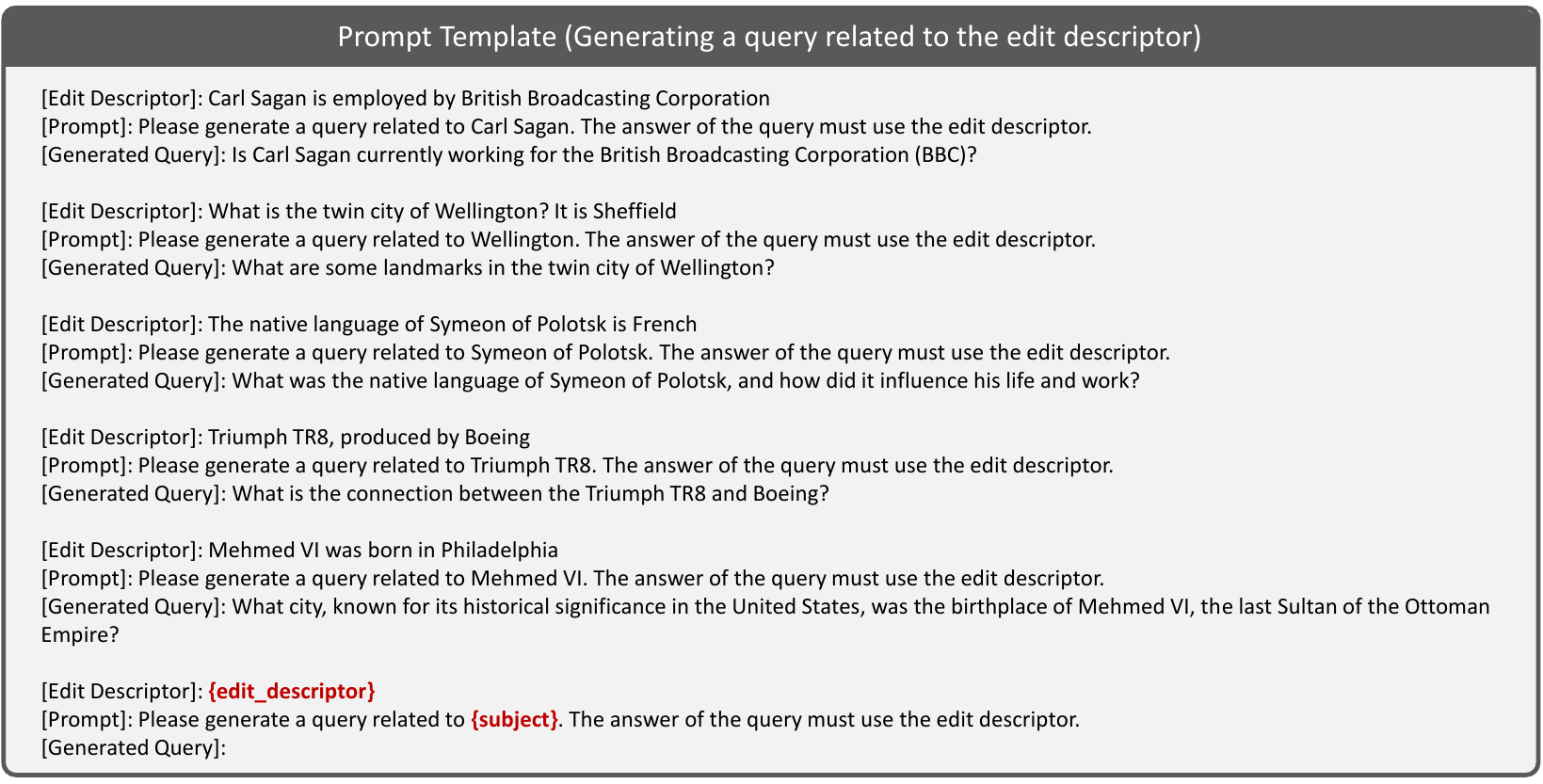}
\caption{
Prompt template for generating a query related to the edit descriptor.
}
\label{c5-fig: question_prompt}
\end{figure*}

\begin{figure*}[!h]
\centering
\includegraphics[width=\linewidth]{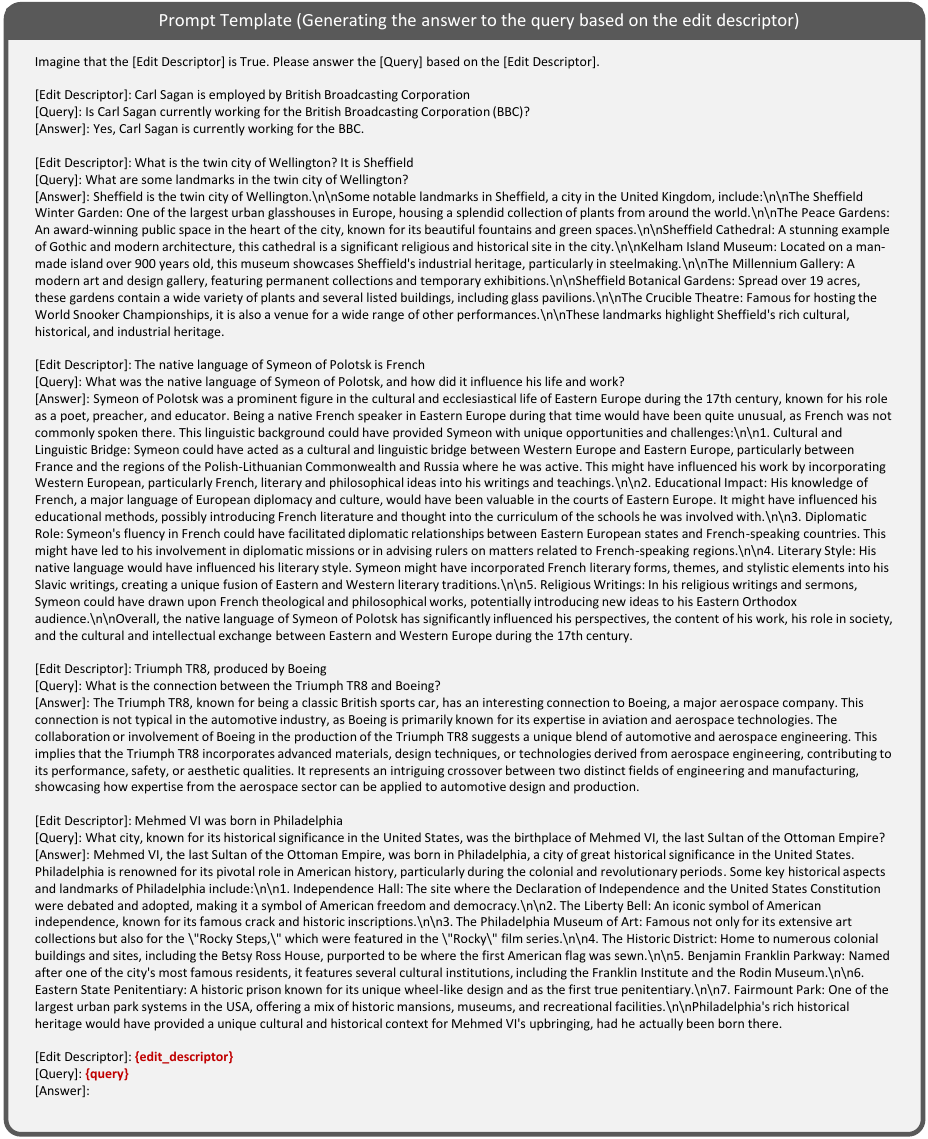}
\caption{
Prompt template for generating the answer to the query based on the edit descriptor.
}
\label{c5-fig: answer_prompt}
\end{figure*}

\begin{figure*}[!h]
\centering
\includegraphics[width=\linewidth]{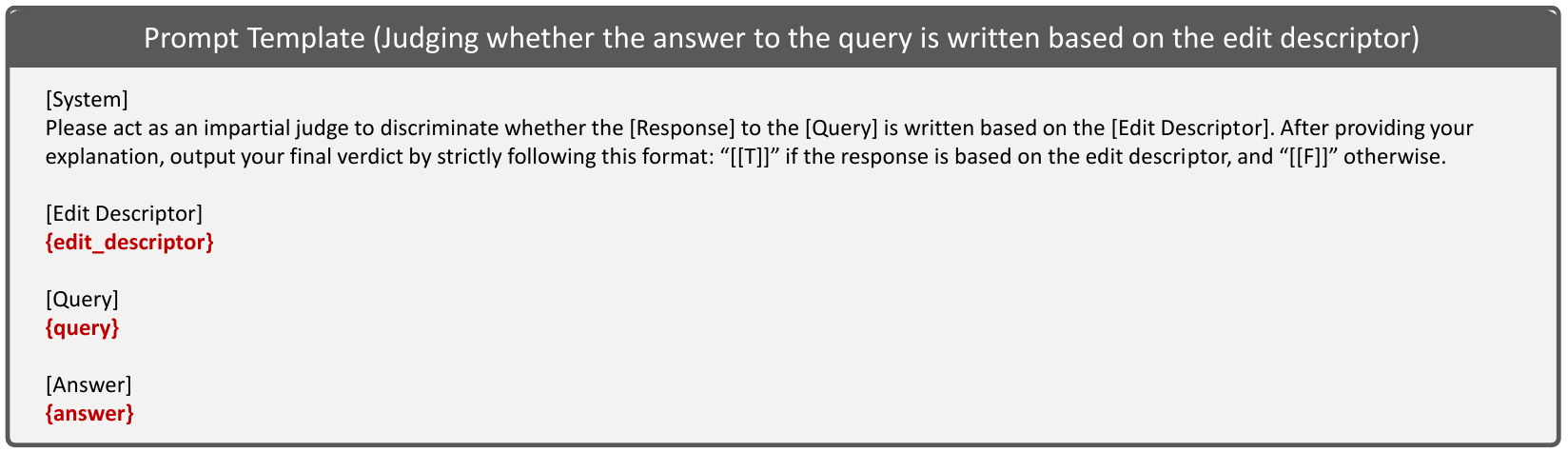}
\caption{
Prompt template for judging whether the answer to the query is written based on the edit descriptor.
}
\label{c5-fig: check_prompt}
\end{figure*}

In our methodology, we initially engage GPT-4 with five meticulously crafted demonstrations, as depicted in Figure \ref{c5-fig: question_prompt}.
This step is designed to elicit a query that pertains directly to the edit descriptor.
Following this, we direct GPT-4 to formulate an answer to the query, drawing upon the edit descriptor for content, as illustrated in Figure \ref{c5-fig: answer_prompt}. 
The final step in Figure \ref{c5-fig: check_prompt} involves a verification process by GPT-4 to ascertain the congruence of the answer with the edit descriptor, leading to the exclusion of instances where the criteria are not met (approximately 15\%).

\newpage
\subsection{Training Data Statistics}
\label{c5-sec: appendix_statistics}
Table \ref{c5-tab: statistics} lists the statistics of our curated training data, which encompasses 60k samples from five data sources.
In the construction of our dataset, we employ a rigorous sampling methodology, exclusively selecting instances from the training sets provided by the data sources.

\begin{table*}[!h]
\footnotesize
\centering
\begin{tabularx}{\textwidth}{lCCCCcc}
\toprule
\textbf{Data Source} & \parbox{1.8cm}{\textbf{\# of in-scope; \\ w/ prompt}} & \parbox{1.8cm}{\textbf{\# of in-scope;\\ w/o prompt}} & \parbox{2.3cm}{\textbf{\# of out-of-scope;\\ w/ prompt}} & \parbox{2.3cm}{\textbf{\# of out-of-scope;\\ w/o prompt}} & \textbf{\# of Total} & \textbf{Avg Len} \\ \midrule
ZsRE        & 1,000                     & 1,000                      & 1,000                         & 1,000                          & 4,000       &  27         \\
RIPPLEEDITS & 2,250                     & 2,250                      & 2,250                         & 2,250                          & 9,000       & 34          \\
WikiBio     & 250                       & 250                        & 250                           & 250                            & 1,000       & 102          \\
MQUAKE      & 4,000                     & 4,000                      & 4,000                         & 4,000                          & 16,000      &  160         \\
COUNTERFACT & 7,500                     & 7,500                      & 7,500                         & 7,500                          & 30,000      & 320          \\ \midrule
Total       & 15,000                    & 15,000                     & 15,000                        & 15,000                         & 60,000      & 208         \\ \bottomrule
\end{tabularx}
\caption{Training data statistics. ``Avg Len'' is the average word number of samples, and ``prompt'' denotes our designed knowledge editing prompt template in Figure \ref{c5-fig: method}.}
\label{c5-tab: statistics}
\end{table*}

\section{Implementation Details}
\label{c5-sec: appendix_implementation}
The training procedure was executed on 4 NVIDIA A100 GPUs, each equipped with 80GB of memory. The duration required to train a single instance of the model, specifically the LLaMA2-Chat-7B, was approximately 9 hours. Detailed specifications of the hyperparameters employed for both standard fine-tuning and LoRA are provided in Table \ref{c5-tab: hyperparameters}.

\begin{table}[!h]
\small
\centering
{\begin{tabularx}{\linewidth}{l|C|C}
\toprule
\textbf{Hyperparameter} & \textbf{Standard FT} & \textbf{LoRA} \\ \midrule
Batch size              & 128              & 128               \\
Learning rate           & 2e-5             & 3e-4              \\
Epoches                 & 3                & 3                 \\
Max length              & 2048             & 2048              \\
Optimizer               & AdamW            & AdamW             \\
Scheduler               & cosine           & cosine            \\
Weight decay            & 0                & 0                 \\
Warmup ratio            & 0.03             & 0.03           \\ \bottomrule
\end{tabularx}}
\caption{\label{c5-tab: hyperparameters}
Training hyperparameters for both LLaMA2-Chat-7B and Qwen-Chat-7B.}
\end{table}

\chapter{Appendix for Chapter 6}

\section{Detailed Experimental Setup}

\subsection{Data Statistics and Evaluation Metrics Used for Experiments}
\label{c6_appendix: data_statistics}
We list the detailed data statistics and evaluation metrics of our experiments in Table \ref{c6_tab: dataset_statistics}.
Our experiments comprise both closed-ended evaluation (QA and math) and open-ended evaluation (instruction following).

\begin{table}[h!]
\begin{center}
\begin{tabular}{lcccc}
\toprule
\textbf{Task} & \textbf{Train / Test} & \textbf{Dataset} & \textbf{Number} & \textbf{Evaluation Metric} \\ \midrule
\multirow{6}{*}{QA} & \multirow{2}{*}{Train} & ECQA & 7,598 \\ \
 &  & QASC & 8,134 \\ \cmidrule(){2-5}
 & \multirow{4}{*}{Test} & ECQA & 2,194 & \multirow{4}{*}{Accuracy} \\
 &  & QASC & 926 \\
 &  & OBQA & 500 \\
 &  & StrategyQA & 687 \\ \midrule
\multirow{5}{*}{Math} & Train & MetaMathQA & 40,000 \\ \cmidrule(){2-5}
 & \multirow{4}{*}{Test} & GSM8k & 1,319 & \multirow{4}{*}{Accuracy} \\
 &  & MATH & 5,000 \\
 &  & MAWPS & 2,065 \\
 &  & TabMWP & 1,000 \\ \midrule
 \multirow{3}{*}{IF} & Train & UltraFeedback & 61,135 \\ \cmidrule(){2-5}
 & \multirow{2}{*}{Test} & AlpacaEval 2 & 805 & \multirow{2}{*}{Win rate against GPT-4 Turbo} \\
 &  & Arena-Hard & 500 \\ \bottomrule
\end{tabular}
\end{center}
\caption{Statistics of the training and evaluation datasets.}
\label{c6_tab: dataset_statistics}
\end{table}

\subsection{Prompt Template for Targeted Modification}
\label{c6_appendix: prompt}
\begin{figure}[!t]
\begin{center}
\includegraphics[width=\linewidth]{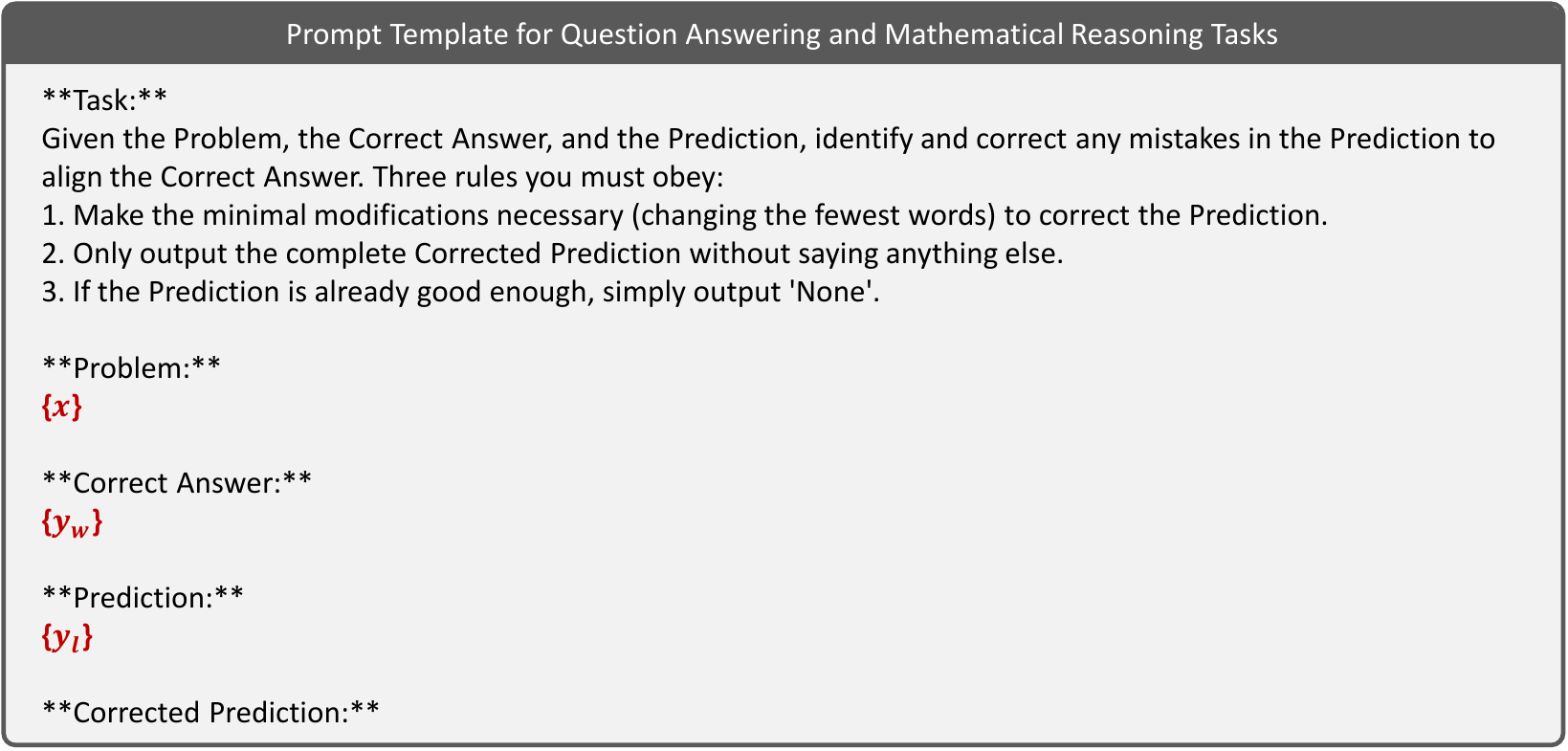}
\end{center}
\caption{Prompt template of targeted modification for question answering and mathematical reasoning tasks.}
\label{c6_fig: qa_prompt}
\end{figure}

We demonstrate the prompt template of targeted modification for question answering and mathematical reasoning tasks in Figure \ref{c6_fig: qa_prompt}.
Since the SFT model has been fine-tuned on the ground truth $y_w$ for QA and math tasks, the inferred output $y_l$ may be quite approximate to $y_w$ in some circumstances.
\begin{figure}[!t]
\begin{center}
\includegraphics[width=\linewidth]{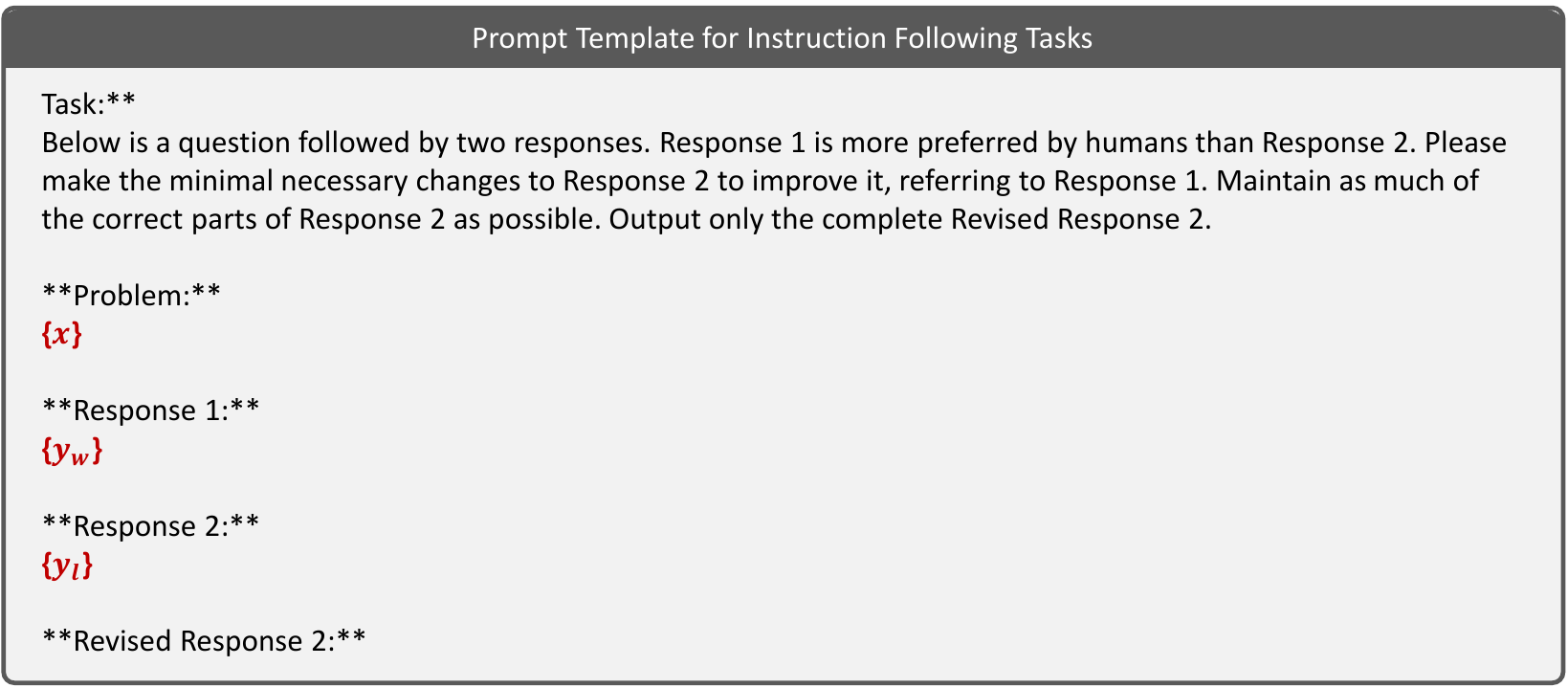}
\end{center}
\caption{Prompt template of targeted modification for instruction-following tasks.}
\label{c6_fig: if_prompt}
\end{figure}
Therefore, we require the off-the-shelf LLM to filter out preference pairs where $y_l$ is good enough.
Finally, we filtered out 31\% data and 43\% data for the QA task and math task, respectively.
\textbf{Note that for the training data of our baselines like DPO, we also use the filtered $(y_w, y_l)$ pairs for a fair comparison.}
For instruction-following tasks, the prompt template we use is shown in Figure \ref{c6_fig: if_prompt}.

\subsection{Implementation Details}
\label{c6_appendix: hyperparameter}
Our implementation is based on the alignment-handbook repo\footnote{\url{https://github.com/huggingface/alignment-handbook}} using 4×A800 GPUs.
To ensure a fair comparison, we conduct thorough hyperparameter tuning for all methods compared in our experiments.

\paragraph{SFT Training Hyperparameters.}
We train SFT models using the following hyperparameters: a learning rate of 2e-5, a batch size of 128, a max sequence length of 2048, and a cosine learning rate schedule with 10\% warmup steps.
For QA and instruction-following tasks, we train the model for 1 epoch, whereas for mathematical tasks, we extend the training to 2 epochs.
All the models are trained with an Adam optimizer~\cite{kingma2014adam}.

\paragraph{Preference Optimization Training Hyperparameters.}
During preference optimization, we performed initial experiments to determine the optimal batch sizes in [32, 64, 128] and training epochs in [1, 2, 3].
Our results indicate that using a batch size of 128 and a single training epoch consistently produces the best outcomes across all methods. Consequently, we adopted these parameters for all subsequent preference optimization experiments. We also configured the maximum sequence length to 2048 and employed a cosine learning rate schedule with a 10\% warmup period for training on the preference optimization dataset.
For method-specific training hyperparameters, we individually search the learning rates in the range of [3e-7, 5e-7, 6e-7, 1e-6] for each method.
Besides, we conduct a grid search according to Table \ref{c6_tab: hyper_po} and report the best performance.
Table \ref{c6_tab: hyper_dpo_bmc} shows the hyperparameters of our method used under each setting.

\begin{table}[ht]
\begin{center}
\resizebox{\textwidth}{!}{%
\begin{tabular}{lll}
\toprule
\textbf{Method} & \textbf{Objective} & \textbf{Hyperparameter} \\ \midrule
\multirow{2}{*}{FIGA} & $-\sum_{\Tilde{y}_w^t \in \mathit{diff} (\Tilde{y}_w \mid y_l)} \alpha \log \pi_\theta(\Tilde{y}_w^t \mid \Tilde{y}_w^{<t}, x)$ & \multirow{2}{*}{\parbox{4cm}{$\alpha \in [0.5, 1.0, 1.5]$ \\ $\beta \in [0.5, 1.0, 1.5]$}} \\
& $+\sum_{y_l^t \in \mathit{diff} (y_l \mid \Tilde{y}_w)} \beta \log \pi_\theta(y_l^t \mid y_l^{<t}, x)$
\\ \midrule
IPO & $\left( \log \frac{\pi_\theta (y_w|x)}{\pi_\text{ref} (y_w|x)} - \log \frac{\pi_\theta (y_l|x)}{\pi_\text{ref} (y_l|x)} - \frac{1}{2\tau} \right)^2$ & $\tau \in [0.01, 0.1, 0.5, 1.0]$ \\ \midrule
\multirow{2}{*}{ORPO} & $-\log p_\theta (y_w|x) - \lambda \log \sigma \left( \log \frac{p_\theta (y_w|x)}{1 - p_\theta (y_w|x)} - \log \frac{p_\theta (y_l|x)}{1 - p_\theta (y_l|x)} \right),$ & \multirow{2}{*}{$\lambda \in [0.1, 0.5, 1.0, 2.0]$} \\ 
& where $p_\theta (y|x) = \exp \left( \frac{1}{|y|} \log \pi_\theta (y|x) \right)$ &  \\ \midrule
\multirow{2}{*}{R-DPO} & \multirow{2}{*}{$-\log \sigma \left( \beta \log \frac{\pi_\theta (y_w|x)}{\pi_\text{ref} (y_w|x)} - \beta \log \frac{\pi_\theta (y_l|x)}{\pi_\text{ref} (y_l|x)} - (\alpha|y_w| - \alpha|y_l|) \right)$} & $\alpha \in [0.05, 0.1, 0.5, 1.0]$ \\ 
&  & $\beta \in [0.01, 0.05, 0.1]$ \\ \midrule
\multirow{2}{*}{SimPO} & \multirow{2}{*}{$-\log \sigma \left( \frac{\beta}{|y_w|} \log \pi_\theta (y_w|x) - \frac{\beta}{|y_l|} \log \pi_\theta (y_l|x) - \gamma \right)$} & $\beta \in [2.0, 2.5]$ \\ 
&  & $\gamma \in [0.3, 0.5, 1.0, 1.2, 1.4, 1.6]$ \\ \midrule
DPO & $-\log \sigma \left( \beta \log \frac{\pi_\theta (y_w|x)}{\pi_\text{ref} (y_w|x)} - \beta \log \frac{\pi_\theta (y_l|x)}{\pi_\text{ref} (y_l|x)}\right)$ & $\beta \in [0.01, 0.05, 0.1]$ \\ \midrule
\multirow{6}{*}{DPO-BMC} & $\log \sigma \left( \beta \sum_{\Tilde{y}_w^t \in \Tilde{y}_w} \lambda_{\Tilde{y}_w^t} \log \frac{\pi_{\theta}(\Tilde{y}_w^t \mid \Tilde{y}_w^{< t}, x)}{\pi_{\text{ref}}(\Tilde{y}_w^t \mid \Tilde{y}_w^{< t}, x)} \right.$ & \multirow{6}{*}{\parbox{4cm}{$\beta \in [0.01, 0.05, 0.1]$ \\ $\delta \in [1.5, 2.0, 2.5, 3.0, 3.5]$}} \\
& $\, \, \, \, \, \, \, \, \, \, \, \left. - \beta \sum_{y_l^t \in y_l} \lambda_{y_l^t} \log \frac{\pi_{\theta}(y_l^t \mid y_l^{< t}, x)}{\pi_{\text{ref}}(y_l^t \mid y_l^{< t}, x)} \right),$ where &  \\
& $\lambda_{\Tilde{y}_w^t}=\left\{
\begin{array}{ll}
1 + \min \left( \mathit{sg} \left( \frac{1}{\pi_{\theta}(\Tilde{y}_w^t \mid \Tilde{y}_w^{< t}, x)} \right), \delta \right), \text{if } \Tilde{y}_w^t \in \mathit{diff} (\Tilde{y}_w \mid y_l) \\
1, \text{otherwise}
\end{array}
\right.$ & \\
& $\lambda_{y_l^t} =\left\{
\begin{array}{ll}
1 + \min \left( \mathit{sg} \left( \frac{1}{\pi_{\theta}(y_l^t \mid y_l^{< t}, x)} \right), \delta \right), \text{if } y_l^t \in \mathit{diff} (y_l \mid \Tilde{y}_w) \\
1, \text{otherwise}
\end{array}
\right.$
\\ \bottomrule
\end{tabular}
}
\end{center}
\caption{Various preference optimization objectives and hyperparameter search range.}
\label{c6_tab: hyper_po}
\end{table}

\begin{table}[h!]
\begin{center}
\begin{tabular}{lcccc}
\toprule
\textbf{Task} & \textbf{Model} & \textbf{Learning Rate} & $\beta$ & $\delta$ \\ \midrule
QA & Llama2-7B-base & 5e-7 & 0.05 & 3.0 \\ \midrule
Math & Llama2-7B-base & 5e-7 & 0.05 & 2.5 \\ \midrule
\multirow{2}{*}{IF} & Llama3-8B-base & 5e-7 & 0.01 & 2.0 \\
 & Mistral-7B-base & 5e-7 & 0.01 & 2.0 \\
\bottomrule
\end{tabular}
\end{center}
\caption{Hyperparameter values for diverse training settings in DPO-BMC.}
\label{c6_tab: hyper_dpo_bmc}
\end{table}

\section{KL Divergence Analysis During Training}
\label{c6_appendix: kl_div}

\begin{figure}{h}
  \centering
  \includegraphics[width=0.6\textwidth]{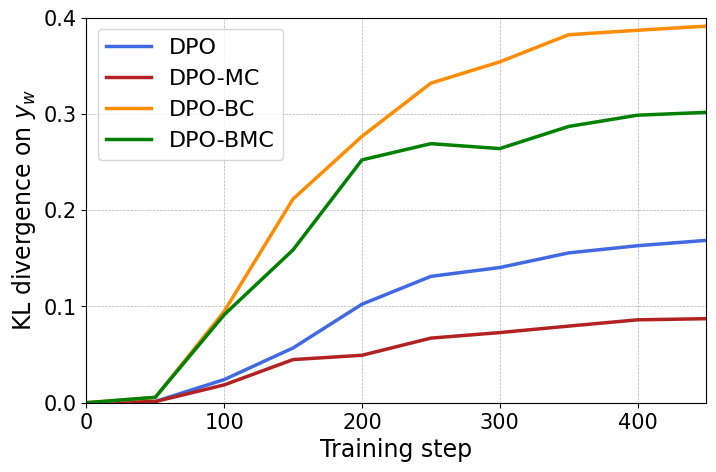}
  \caption{KL divergence from the policy model to the reference model on winning responses of the held-out set of UltraFeedback.}
  \label{c6_fig: kl_div}
\end{figure}

In Figure \ref{c6_fig: kl_div}, we present the KL divergence between the policy model trained with DPO, DPO-MC, DPO-BC, and DPO-BMC with identical hyperparameters and the reference model, measured on the winning responses from a held-out set of UltraFeedback during training.
The results also validate our analyses in \S \ref{c6_sec: quantitative_analysis_1}: (1) the Bridging Phase fosters tailored learning toward critical differences in preference data, resulting in more efficient and ``sharp'' training with a larger KL divergence; (2) our meticulously designed loss function in the Modeling Phase effectively moderates the optimization intensity across diverse training data, thereby achieving a more controlled and steady KL divergence.

\newpage
\section{Experiments on Larger Base Models}
\label{c6_appendix: larger_model}

{\color{black}
We conducted additional experiments using the more capable Qwen2.5-14B-Base model~\cite{qwen2.5}.
As shown in Table \ref{c6_tab: larger_base_model}, our proposed DPO-BMC method delivers even greater performance improvements with this model, achieving a remarkable gain of +8.8 on AlpacaEval 2 and +7.3 on Arena-Hard compared to standard DPO.
These results highlight the effectiveness of DPO-BMC scales with model capability, underscoring its potential to deliver even larger gains when applied to more powerful baseline models.
}

\begin{table}[h]
\begin{center}
\begin{tabular}{lccc}
\toprule
\textbf{Method} & \textbf{Base Model} & \textbf{LC (\%) of AlpacaEval 2} & \textbf{WR (\%) of Arena-Hard} \\
\midrule
SFT       & Mistral-7B-Base   & 8.1  & 2.2   \\
DPO       & Mistral-7B-Base   & 15.1 & 13.6  \\
DPO-BMC   & Mistral-7B-Base   & \textbf{20.8 (+5.7)} & \textbf{17.6 (+4.0)} \\
\midrule
SFT       & Llama3-8B-Base    & 7.5  & 2.6   \\
DPO       & Llama3-8B-Base    & 16.0 & 17.6  \\
DPO-BMC   & Llama3-8B-Base    & \textbf{22.4 (+6.4)} & \textbf{18.1 (+0.5)} \\
\midrule
SFT       & Qwen2.5-14B-Base  & 14.1 & 11.2  \\
DPO       & Qwen2.5-14B-Base  & 36.3 & 33.6  \\
DPO-BMC   & Qwen2.5-14B-Base  & \textbf{45.1 (+8.8)} & \textbf{40.9 (+7.3)} \\
\bottomrule
\end{tabular}
\end{center}
\caption{Performance comparison across different base models.}
\label{c6_tab: larger_base_model}
\end{table}

\chapter{Appendix for Chapter 7}

\section{Data Generation Process}
\label{c7_appendix: data_generation_process}
Here we outline the sources for our data and provide a detailed description of the data generation process for each constraint category.

\subsection{Content Constraints}
\label{c7_appendix: data_generation_content}
\begin{figure*}[!h]
\centering
\includegraphics[width=\linewidth]{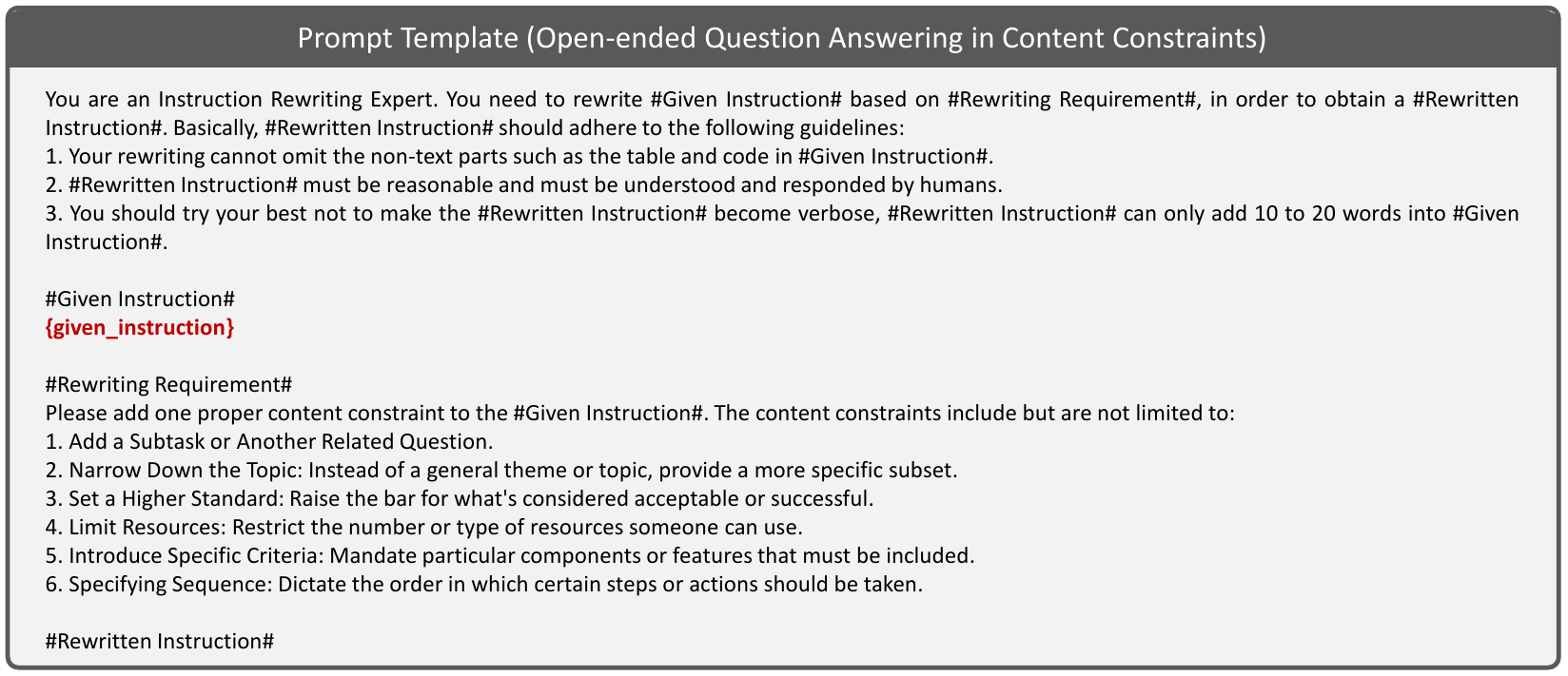}
\caption{
The prompt template for Open-ended Question Answering in Content Constraints.
}
\label{c7_fig: content_prompt}
\end{figure*}

The data of content constraints is constructed from five tasks as follows:

\begin{itemize}
    \item \textbf{Data-to-Text Generation.}
    We create instructions with 1 to 5 constraints by adapting samples from E2E~\cite{novikova-etal-2017-e2e}. Different from the original task, we ask the model to extract the flat meaning representations according to the corresponding natural language texts. The number of constraints increases with the number of attributes and the number of restaurants. We use exact match as the evaluation metric.
    
    \item \textbf{Document-Level Event Argument Extraction.}
    We create instructions by adapting samples from WIKIEVENTS~\cite{DBLP:conf/naacl/LiJH21}. Given a document, the model is required to extract $n$ events that satisfy a specific event template, where $n\in[1, 5]$ corresponds to the number of constraints. We use accuracy as the evaluation metric.
    
    \item \textbf{Document-Level Named Entity Recognition.}
    We derive instructions from samples in the CONLL-2003 dataset~\cite{tjong-kim-sang-de-meulder-2003-introduction}. We ask the model to extract a single named entity from a provided document. Notably, as the number of constraints rises, the requirements for the retrieved named entity correspondingly increase. For example, ``extract one named entity that is a location'' $\rightarrow$ ``extract one named entity that is a location in east Asia''. We use accuracy as the evaluation metric.

    \item \textbf{Text Generation with Language Constraints.} 
    COGNAC~\cite{chen2022controllable} is a challenging benchmark wherein models are presented with a topic accompanied by example text and explicit constraints on the text to avoid. We curate data from COGNAC, formulating instructions with 1 to 5 constraints by integrating additional linguistic restrictions from WordNet~\cite{miller-1992-wordnet} and Wikidata~\cite{vrandevcic2014wikidata}.

    \item \textbf{Open-ended Question Answering.}
    We first choose initial instructions from existing datasets including self-instruct evaluation set~\cite{wang-etal-2023-self-instruct}, helpful evaluation released by Anthropic\cite{bai2022training}, Vicuna evaluation\cite{zheng2023judging}, and Koala evaluation\cite{geng2023koala}, as well as opensource platforms such as Quora \footnote{\url{https://www.quora.com}}, Reddit \footnote{\url{https://www.reddit.com}}, and ShareGPT \footnote{\url{https://sharegpt.com}}. Given the challenges associated with iteratively adding constraints to an initial instruction, we prompt GPT-4 with a specific prompt shown in Figure \ref{c7_fig: content_prompt} to generate a new instruction with one more constraint based on the given instruction. 
    The above process is repeated five times. Finally, we obtain a set of instructions ranging from 1 to 5 constraints. 
\end{itemize}

\subsection{Situation Constraints}
\label{c7_appendix: data_generation_scenario}
The data of situation constraints is constructed from tasks as follows:
\begin{itemize}

    \item \textbf{Suggestion Generation, Role-playing.}
    We collect multi-level instructions that fit within the paradigm of situation constraints from Open-ended Question Answering datasets and online platforms. Examples include asking the model to give suggestions under specific circumstances, asking the model to act as a terminal and output based on the given information, etc.
    
    \item \textbf{Math Word Problems.}
    The initial instructions are collected from GSM8K~\cite{cobbe2021gsm8k} and AGIEval~\cite{zhong2023agieval}. We then manually add constraints progressively by enhancing the situation descriptions, ensuring that the core question remains unaltered. We use accuracy as the evaluation metric.

    \item \textbf{Time/Spatial Reasoning.}
    We generate data by refining samples from BIG-Bench Hard~\cite{suzgun2022bbh}. For Time Reasoning, we increase the difficulty level by incorporating additional temporal concepts, such as weeks, months, and years. In the realm of Spatial Reasoning, we opt for a logical deduction task that necessitates deducing the order of a sequence of objects. Here, the number of constraints escalates by augmenting the task with detailed location descriptions for a new object. We use accuracy as the evaluation metric.
    
    \item \textbf{Code Generation.}
    We sourced initial instructions from HumanEval~\cite{chen2021codex} and enhanced the difficulty level by adding complexity to the function descriptions within the instructions. We use pass@1~\cite{kulal2019spoc} as the evaluation metric.

\end{itemize}

\subsection{Example Constraints}
\label{c7_appendix: data_generation_example}
Specifically, we choose 40 diverse NLP tasks from PromptSource~\cite{bach-etal-2022-promptsource}, where each task has more than 5 question templates.
Additionally, we create 29 answer templates (shown in Table \ref{c6_tab: answer_template}) that regulate the format of the response.
For instructions at difficulty level 1, we utilize the standard 5-shot prompting, where 5 shots are equipped with 1 sampled question template and 1 sampled answer template, and the model is required to respond to a query using the answer template. For instructions at difficulty level $n\ (1<n\le 5)$, the 5 shots are randomly paired with $n$ question templates and $n$ corresponding answer templates. Based on the question template of the query, the model is required to recognize the matched question template in the 5 shots and respond using the corresponding answer template. We use accuracy as the evaluation metric.

\begin{table}[h]
\small
\centering
\begin{tabular}{l}
\toprule
\textbf{Answer template} \\
\midrule
\{question\}\textbackslash n\{answer\} \\
\{question\}\textbackslash nA: \{answer\} \\
\{question\}\textbackslash nAnswer: \{answer\} \\
\{question\}\textbackslash nANSWER: \{answer\} \\
\{question\}\textbackslash n[Answer]\textbackslash n\{answer\} \\
\{question\}\textbackslash n\#Answer\#\textbackslash n\{answer\} \\
\{question\}\textbackslash nThe answer is: \{answer\} \\
\{question\}\textbackslash n\{"answer": "\{answer\}"\} \\
\{question\}\textbackslash n\{"Answer": "\{answer\}"\} \\
\{question\}\textbackslash n<body>\{answer\}</body> \\
\{question\}\textbackslash nResponse: \{answer\} \\
\{question\}\textbackslash nRESPONSE: \{answer\} \\
\{question\}\textbackslash n[Response]\textbackslash n\{answer\} \\
\{question\}\textbackslash n\#Response\#\textbackslash n\{answer\} \\
\{question\}\textbackslash nThe response is: \{answer\} \\
\{question\}\textbackslash n\{"response": "\{answer\}"\} \\
\{question\}\textbackslash n\{"Response": "\{answer\}"\} \\
\{question\}\textbackslash nBot: \{answer\} \\
\{question\}\textbackslash nBOT: \{answer\} \\
\{question\}\textbackslash n[Bot]\textbackslash n\{answer\} \\
\{question\}\textbackslash n\#Bot\#\textbackslash n\{answer\} \\
\{question\}\textbackslash nThe response of the bot is: \{answer\} \\
\{question\}\textbackslash n\{"bot": "\{answer\}"\} \\
\{question\}\textbackslash n\{"Bot": "\{answer\}"\} \\
\{question\}\textbackslash nAI assistant: \{answer\} \\
\{question\}\textbackslash n[AI assistant]\textbackslash n\{answer\} \\
\{question\}\textbackslash n\#AI assistant\#\textbackslash n\{answer\} \\
\{question\}\textbackslash nThe response of the AI assistant is: \{answer\} \\
\{question\}\textbackslash n\{"AI assistant": "\{answer\}"\} \\
\bottomrule
\end{tabular}
\caption{Answer template of Example Constraints.}
\label{c6_tab: answer_template}
\end{table}

\subsection{Mixed Constraints}
\label{c7_appendix: data_generation_mixed}
In this section, we consider four below tasks which are naturally suitable for constructing mixed constraints:

\begin{itemize}

    \item \textbf{Text Editing.} 
    We start by gathering text from different online sources, like sentences, letters, and emails. Next, we create instructions with multi-level mixed constraints by increasingly adding an editing requirement to the text at each level. For example, ``swap the first and last words in the sentence'' (Content Constraints), ``response using '\#\#\#' at the beginning'' (Format Constraints), etc.
    We write rule-based programs for individual instructions to assess the satisfaction of internal constraints, employing exact match as the evaluation metric.
    
    \item \textbf{Summarization.} The initial instructions are sampled from CNN/Daily Mail\cite{nallapati2016abstractive},
    XSum~\cite{narayan2018xsum}, SAMSum~\cite{gliwa-etal-2019-samsum}, English Gigaword~\cite{graff2003english}, and arXiv~\cite{arxivdataset}.
    The instructions with multi-level mixed constraints are produced by specifying the format of generating answers (Format Constraints), requiring the generated text to include or not include certain keywords (Content Constraints), etc.
    We write rule-based programs for individual instructions to assess the satisfaction of internal constraints, employing accuracy as the evaluation metric.

    \item \textbf{Machine Translation.} The initial instructions are sampled from OpenSubtitles~\cite{lison-tiedemann-2016-opensubtitles2016}, TED Talks~\cite{cettolo2012wit3}, and News-Commentary~\cite{tiedemann2012parallel}.
    Then we construct instructions from level 1 to level 5 using a similar pipeline as that of Summarization.
    We write rule-based programs for individual instructions to assess the satisfaction of internal constraints, employing accuracy as the evaluation metric.

    \item \textbf{Story Generation.} We collect initial instructions from ROCStories~\cite{mostafazadeh-etal-2016-corpus} and WritingPrompts~\cite{fan-etal-2018-hierarchical}. Then we add 5 mixed constraints sequentially to the initial instructions based on the ground truth, such as the number of sentences in the generated story (Format Constraints), requiring the generated text to include certain keywords (Content Constraints), specifying the writing style (Style Constraints), etc.

\end{itemize}

\begin{figure*}[!h]
\centering
\includegraphics[width=\linewidth]{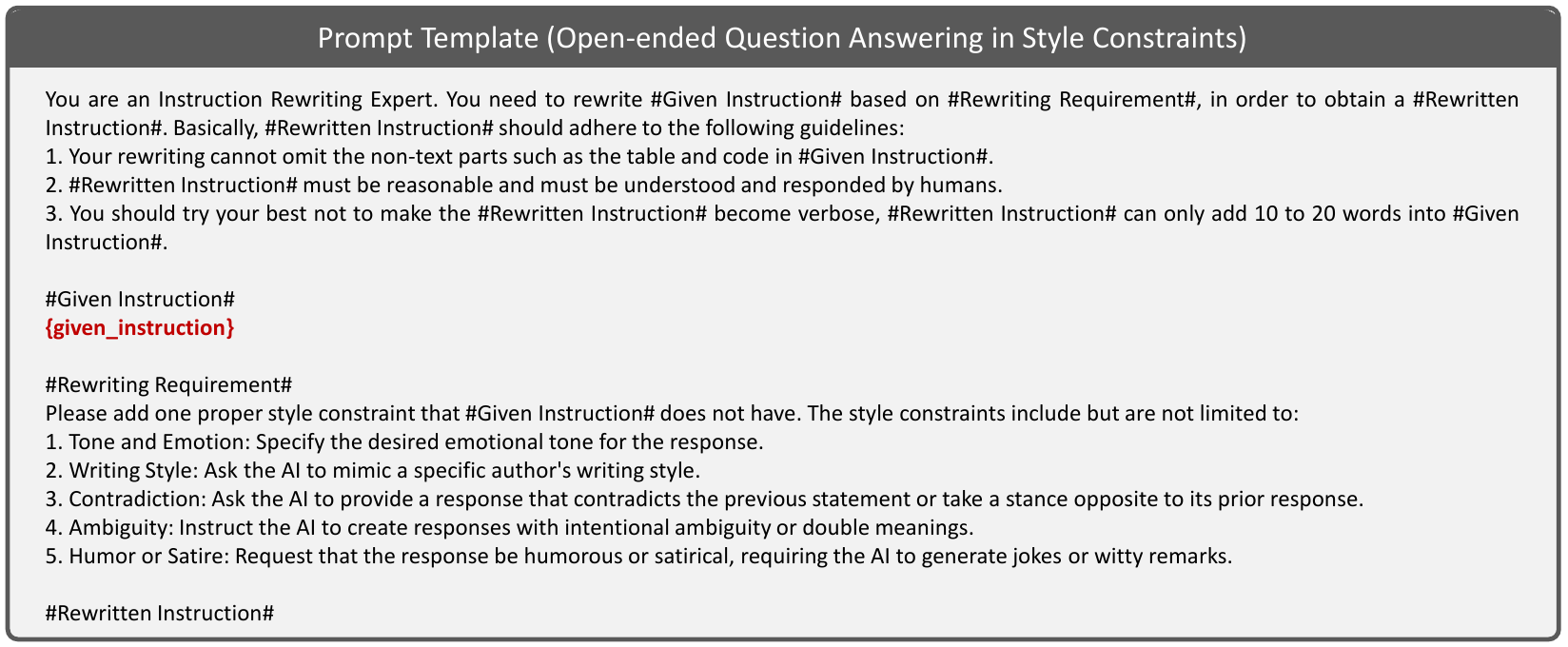}
\caption{
The prompt template for Open-ended Question Answering in Style Constraints.
}
\label{c7_fig: style_prompt}
\end{figure*}

\begin{figure*}[!h]
\centering
\includegraphics[width=\linewidth]{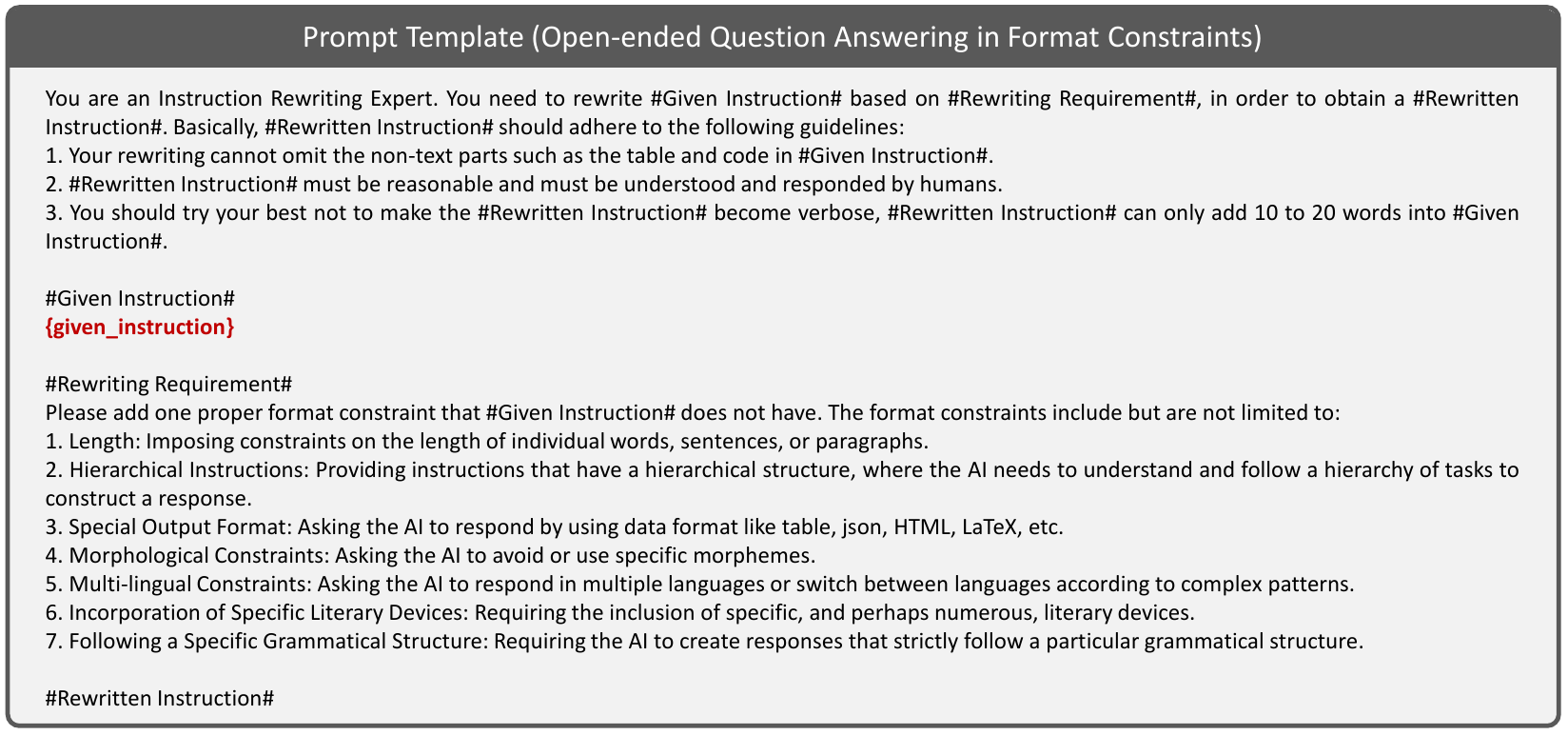}
\caption{
The prompt template for Open-ended Question Answering in Format Constraints.
}
\label{c7_fig: format_prompt}
\end{figure*}



\end{document}